\pdfoutput=1 
\nonstopmode

\documentclass[10pt,twocolumn,letterpaper]{article}

 \usepackage[pagenumbers]{cvpr} %

\usepackage[utf8]{inputenc} %
\usepackage[T1]{fontenc}    %
\usepackage{url}            %
\usepackage{booktabs}       %
\usepackage{amsfonts}       %
\usepackage{nicefrac}       %
\usepackage{microtype}      %
\usepackage{xcolor}         %
\usepackage{algorithm,algpseudocode}
\usepackage{amsmath}
\usepackage{tikz}
\usepackage[export]{adjustbox}
\usepackage{graphicx}
\usepackage{subcaption}
\usetikzlibrary{spy}
\usepackage{svg}
\usepackage{pgfplots}
\pgfplotsset{compat=1.18} 
\usepackage{multirow}
\usepackage{tikz}
\usepackage{tabularx}
\usepackage[outdir=./]{epstopdf}
\usepackage[htt]{hyphenat}  %
\usepackage{textgreek}

    \newlength{\zz}
    \newlength{\xx}
    \newlength{\bb}
    \newlength{\cc}
\definecolor{cvprblue}{rgb}{0.21,0.49,0.74}
\usepackage[pagebackref,breaklinks,colorlinks,citecolor=cvprblue]{hyperref}

\def\paperID{3798} %
\def\confName{CVPR}
\def\confYear{2024}

\title{Differential Diffusion: Giving Each Pixel Its Strength}

\author{Eran Levin\\
Tel Aviv University\\
{\tt\small eran.levin@cs.tau.ac.il}
\and
Ohad Fried\\
Reichman University\\
{\tt\small ofried@runi.ac.il}
}
\begin{document}
\twocolumn[{
    \renewcommand\twocolumn[1][]{#1}
    \maketitle
    \newcommand{\vv}{5pt}
\newcommand{\af}{1.81cm}
\newcommand{\aff}{1.81cm}

\begin{center}
\begin{tabular}{ccc}
\multicolumn{2}{c}{\textbf{Image + Map}} & \textbf{Output} \\[-0.2cm]
\includegraphics[width=\af,frame,valign=T]{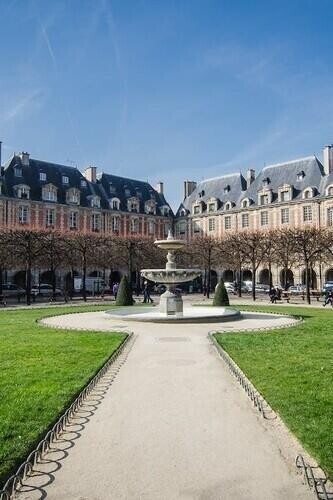} &
\includegraphics[width=\af,frame,valign=T]{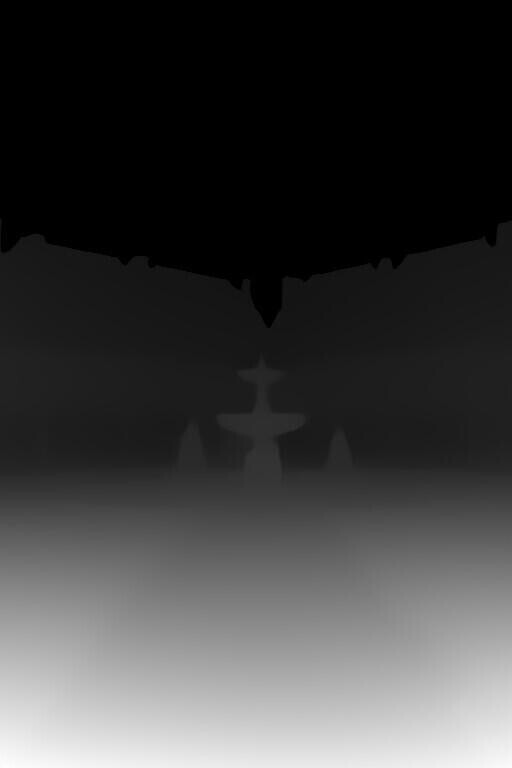} &
\includegraphics[width=\aff,frame,valign=T]{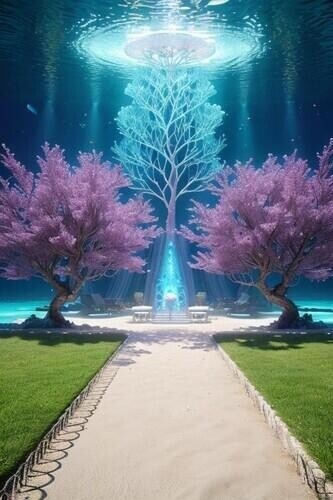} 
\end{tabular}
\hfill
\begin{tabular}{ccc}
\multicolumn{2}{c}{\textbf{Image + Map}} & \textbf{Output} \\[-0.2cm]
\includegraphics[width=\af,frame,valign=T]{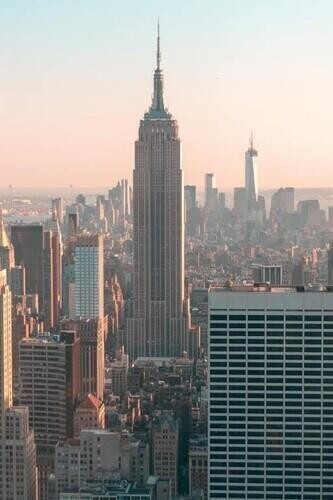} &
\includegraphics[width=\af,frame,valign=T]{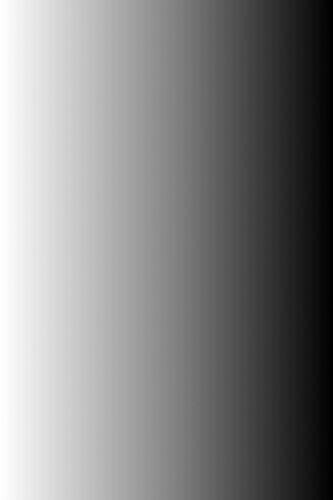} &
\includegraphics[width=\aff,frame,valign=T]{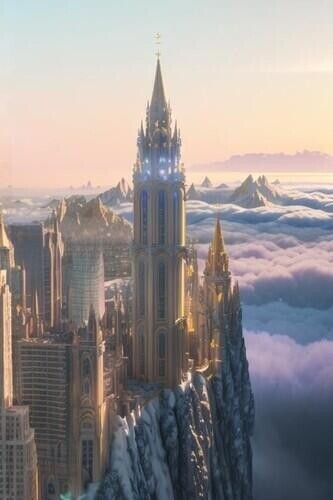} 
\end{tabular}
\hfill
\begin{tabular}{ccc}
\multicolumn{2}{c}{\textbf{Image + Map}} & \textbf{Output} \\[-0.2cm]
\includegraphics[width=\af,frame,valign=T]{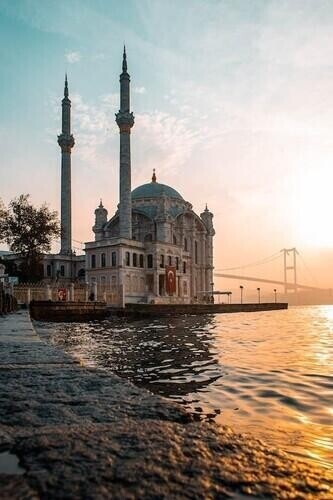} &
\includegraphics[width=\af,frame,valign=T]{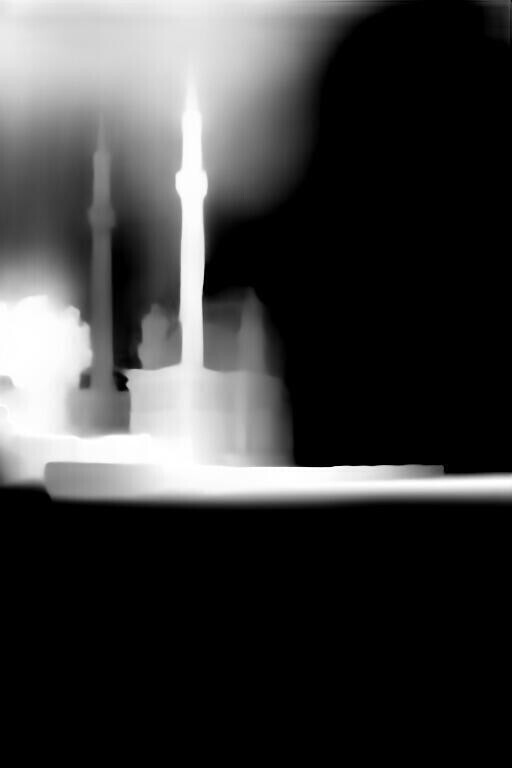} &
\includegraphics[width=\aff,frame,valign=T]{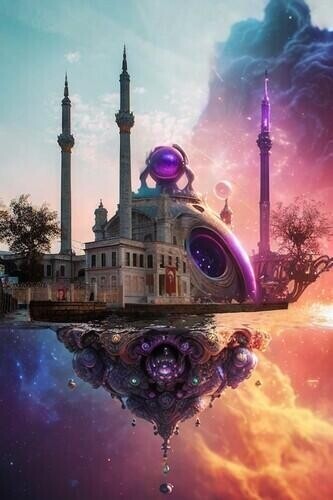} 
\end{tabular}

\vspace{-0.1cm}
\begin{tabular}{ccc}
\includegraphics[width=\af,frame,valign=T]{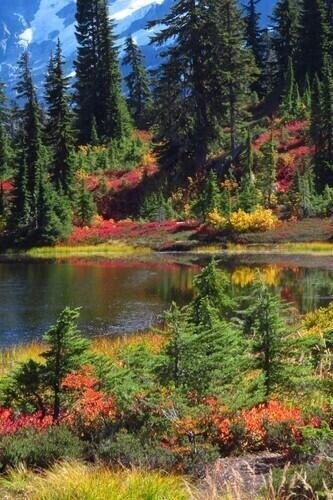} &
\includegraphics[width=\af,frame,valign=T]{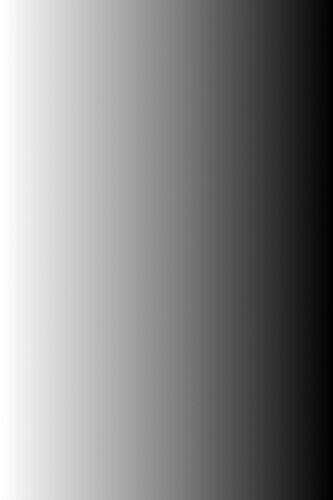} &
\includegraphics[width=\aff,frame,valign=T]{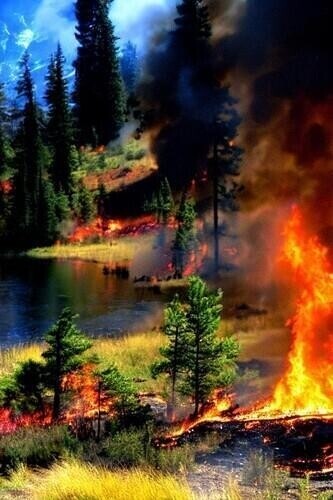} 
\end{tabular}
\hfill
\begin{tabular}{ccc}
\includegraphics[width=\af,frame,valign=T]{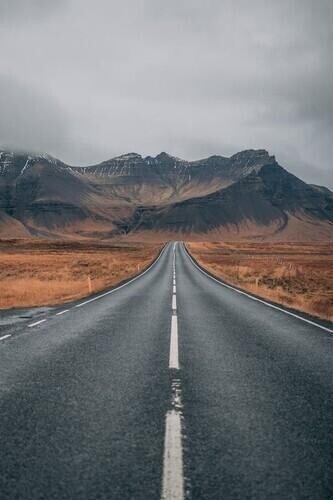} &
\includegraphics[width=\af,frame,valign=T]{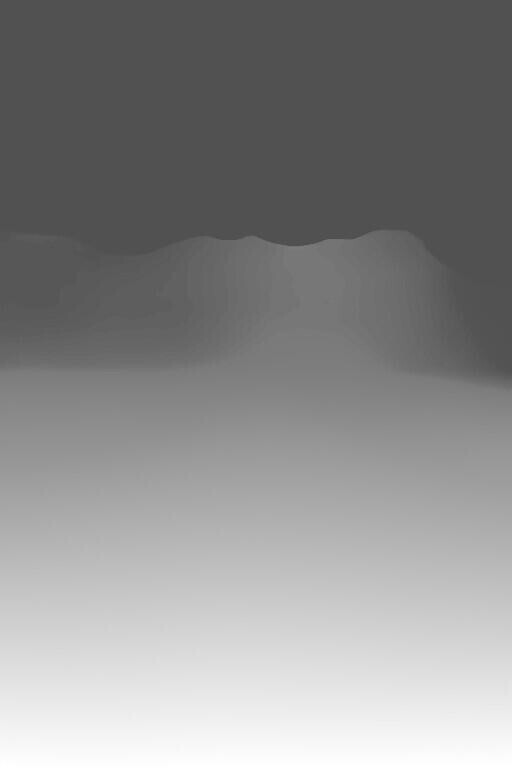} &
\includegraphics[width=\aff,frame,valign=T]{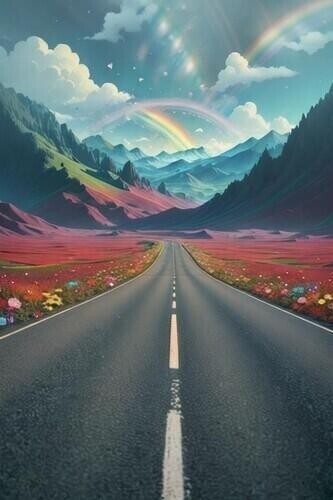} 
\end{tabular}
\hfill
\begin{tabular}{ccc}
\includegraphics[width=\af,frame,valign=T]{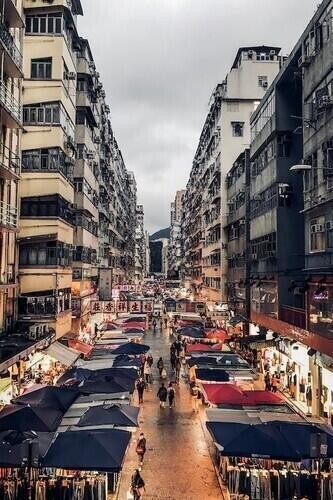} &
\includegraphics[width=\af,frame,valign=T]{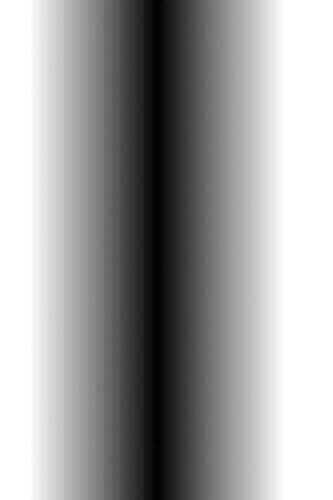} & 
\includegraphics[width=\aff,frame,valign=T]{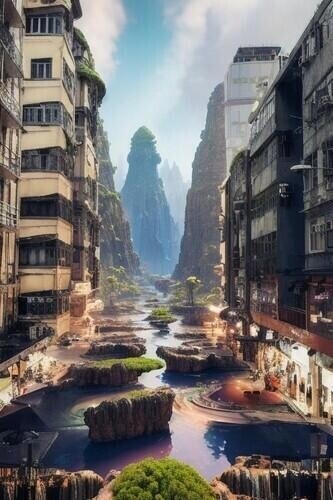} 
\end{tabular}

    \captionsetup{type=figure}
    \caption{Our method changes different regions of an image in different amounts, according to a given map and text prompt. 
    This controllability allows us to reproduce gradual processes (e.g., fire, bottom left) and to seamlessly blend between varying edit strengths.
    Edits, left to right: “tree of life under the sea”, “palace above the clouds”, “3d depth outer space nebulae background”, “fire”, “whimsical illustration of a rainbow”, “fantasy art” (full prompts in the supplemental).  
    }
    \label{fig:teaser}
\end{center}

}]
\newcommand{\ignorethis}[1]{}
\newcommand{\redund}[1]{#1}

\newcommand{\Identity   }     {\mat{I}}
\newcommand{\Zero       }     {\mathbf{0}}
\newcommand{\Reals      }     {{\textrm{I\kern-0.18em R}}}
\newcommand{\isdefined  }     {\mbox{\hspace{0.5ex}:=\hspace{0.5ex}}}
\newcommand{\texthalf   }     {\ensuremath{\textstyle\frac{1}{2}}}
\newcommand{\half       }     {\ensuremath{\frac{1}{2}}}
\newcommand{\third      }     {\ensuremath{\frac{1}{3}}}
\newcommand{\fourth     }     {\ensuremath{\frac{1}{4}}}

\newcommand{\Lone} {\ensuremath{L_1}}
\newcommand{\Ltwo} {\ensuremath{L_2}}

\newcommand{\mat        } [1] {{\text{\boldmath $\mathbit{#1}$}}}
\newcommand{\Approx     } [1] {\widetilde{#1}}
\newcommand{\change     } [1] {\mbox{{\footnotesize $\Delta$} \kern-3pt}#1}

\newcommand{\Order      } [1] {O(#1)}
\newcommand{\set        } [1] {{\lbrace #1 \rbrace}}
\newcommand{\floor      } [1] {{\lfloor #1 \rfloor}}
\newcommand{\ceil       } [1] {{\lceil  #1 \rceil }}
\newcommand{\inverse    } [1] {{#1}^{-1}}
\newcommand{\transpose  } [1] {{#1}^\mathrm{T}}
\newcommand{\invtransp  } [1] {{#1}^{-\mathrm{T}}}
\newcommand{\relu       } [1] {{\lbrack #1 \rbrack_+}}

\newcommand{\abs        } [1] {{| #1 |}}
\newcommand{\Abs        } [1] {{\left| #1 \right|}}
\newcommand{\norm       } [1] {{\| #1 \|}}
\newcommand{\Norm       } [1] {{\left\| #1 \right\|}}
\newcommand{\pnorm      } [2] {\norm{#1}_{#2}}
\newcommand{\Pnorm      } [2] {\Norm{#1}_{#2}}
\newcommand{\inner      } [2] {{\langle {#1} \, | \, {#2} \rangle}}
\newcommand{\Inner      } [2] {{\left\langle \begin{array}{@{}c|c@{}}
                               \displaystyle {#1} & \displaystyle {#2}
                               \end{array} \right\rangle}}

\newcommand{\twopartdef}[4]
{
  \left\{
  \begin{array}{ll}
    #1 & \mbox{if } #2 \\
    #3 & \mbox{if } #4
  \end{array}
  \right.
}

\newcommand{\fourpartdef}[8]
{
  \left\{
  \begin{array}{ll}
    #1 & \mbox{if } #2 \\
    #3 & \mbox{if } #4 \\
    #5 & \mbox{if } #6 \\
    #7 & \mbox{if } #8
  \end{array}
  \right.
}

\newcommand{\len}[1]{\text{len}(#1)}

\newlength{\w}
\newlength{\h}
\newlength{\x}

\definecolor{darkred}{rgb}{0.7,0.1,0.1}
\definecolor{darkgreen}{rgb}{0.1,0.6,0.1}
\definecolor{cyan}{rgb}{0.7,0.0,0.7}
\definecolor{otherblue}{rgb}{0.1,0.4,0.8}
\definecolor{maroon}{rgb}{0.76,.13,.28}
\definecolor{burntorange}{rgb}{0.81,.33,0}

\ifdefined\ShowNotes
  \newcommand{\colornote}[3]{{\color{#1}\textbf{#2} #3\normalfont}}
\else
  \newcommand{\colornote}[3]{}
\fi

\newcommand {\reqs}[1]{\colornote{red}{\tiny #1}}

\newcommand {\new}[1]{{\color{red}{#1}}}

\newcommand*\rot[1]{\rotatebox{90}{#1}}

\newcommand {\newstuff}[1]{#1}

\newcommand{\woBGmask}{{w/o~bg~\&~mask}}
\newcommand{\woMask}{{w/o~mask}}

\providecommand{\keywords}[1]
{
  \textbf{\textit{Keywords---}} #1
}

\newcommand{\GAN}{\textit{GAN}}
\newcommand{\data}{\mathit{data}}
\newcommand{\unionGAN}{\textsc{UnionGAN}\xspace}
\newcommand {\ganArrow}[2]{\ensuremath{\GAN_{{#1} \rightarrow {#2}}}}
\newcommand {\gan}[1]{\ensuremath{\GAN_{#1}}}
\newcommand{\DALLE}{{DALL$\cdot$E}}

\newcommand{\xhat}{\widehat{x}}

\makeatletter
\newcommand{\ogeneric}[2][0.7]{%
  \vphantom{\oplus}\mathpalette\o@generic{{#1}{#2}}%
}
\newcommand{\o@generic}[2]{\o@@generic#1#2}
\newcommand{\o@@generic}[3]{%
  \begingroup
  \sbox\z@{$\m@th#1\oplus$}%
  \dimen@=\dimexpr\ht\z@+\dp\z@\relax
  \savebox\tw@[\totalheight]{$\m@th#1\bigcirc$}%
  \makebox[\wd\z@]{%
    \ooalign{%
      $#1\vcenter{\hbox{\resizebox{\dimen@}{!}{\usebox\tw@}}}$\cr
      \hidewidth
      $#1\vcenter{\hbox{\resizebox{#2\dimen@}{!}{$#1\vphantom{\oplus}{#3}$}}}$%
      \hidewidth
      \cr
    }%
  }%
  \endgroup
}
\makeatother

\newcommand{\ewgt}{\mathrel{\ogeneric[0.6]{\geq}}}
\newcommand{\ewlt}{\mathrel{\ogeneric[0.6]{\leq}}}

\newcommand{\ewgteq}{\mathrel{\ogeneric[0.6]{>}}}
\newcommand{\ewlteq}{\mathrel{\ogeneric[0.6]{<}}}

\renewcommand{\cref}[1]{\PackageError{cref}{\string\cref\space is disabled}{Please use \string\Cref\space instead.}}
    \newlength{\ww}

\newcommand{\algcomment}[1]{%
    \vspace{-\baselineskip}%
    \noindent%
    {\footnotesize #1\par}%
    \vspace{\baselineskip}%
    }

\begin{abstract}
Diffusion models have revolutionized image generation and editing,
producing state-of-the-art results in conditioned and unconditioned image synthesis.
While current techniques enable user control over the degree of change in an image edit, the controllability is limited to global changes over an entire edited region. 
This paper introduces a novel framework 
that enables 
customization of the amount of change \emph{per pixel} or \emph{per image region}.
Our framework can be integrated into any existing diffusion model, enhancing it with this capability.
Such granular control on the quantity of change opens up a diverse array of new editing capabilities, such as control of the extent to which individual objects are modified, or 
the ability to introduce gradual spatial changes.
Furthermore, we showcase the framework's effectiveness in soft-inpainting---the completion of 
portions of an image while subtly adjusting the surrounding areas to ensure seamless integration.
Additionally, we introduce a new tool for exploring the effects of different change quantities. 
Our framework operates solely during inference, requiring no model training or fine-tuning. We demonstrate our method with the current open state-of-the-art models, and validate it via both quantitative and qualitative comparisons, and a user study.
Our code will be available at: \url{https://github.com/exx8/differential-diffusion}.

\end{abstract}

\section{Introduction}

Recently, diffusion models have taken the lead in image generation~\cite{dhariwal2021diffusion}, offering a robust method to generate and edit high quality images~\cite{saharia2022palette,rombach2022highresolution}. 
Typical editing methods allow specifying only one change quantity (which is usually termed ``strength''), changing the entire image uniformly.
Inpainting methods, at most, allow partitioning the picture into an unchanged and a changed region according to a single selected strength. 
In this paper, we introduce a major advancement of editing with diffusion models—a framework that allows the user to change an arbitrary number of regions of the picture by different strengths efficiently and simultaneously,
thus generalizing some of the edit tasks present today.
This is primarily accomplished by the insight that, by selectively modifying various regions at different timesteps during the diffusion's inference process, we can control the fidelity to the original image on a spatial basis. Our framework does \emph{not} require any optimization process such as fine-tuning or training.
It enables greater flexibility and finer-grained editing of the image, allowing use cases that have not been attainable by any previous methods, such as editing a region with a continuous range of strength values.

Consider for example, introducing a wildfire into a wooded area of a photo. As fire is a continuous phenomenon, 
we would not want to make abrupt and complete transformations such as replacing all the trees with burnt stumps.
Instead, we would like to introduce different amounts of fire into different regions on the photo, in a controllable manner (\Cref{fig:teaser} bottom-right).

We will show that even for simple edits (with as little as three regions), controlling the amount of change of each region opens up a wide range of unique edits (e.g., \Cref{video_game}). 
In addition, we demonstrate that 
we can apply soft-inpainting that exhibits superior blending and visual quality compared to conventional blending methods.
Furthermore, we demonstrate our method's capability to generate a tool we call ``Strength Fan'', which enables the analysis of the effects of different strength levels for a given prompt and input image.

\subsection{Contributions}
We introduce several key observations (\Cref{observations}) that allow us to create an efficient inference process (\Cref{algo-analysis}).
We demonstrate several applications of our method (\Cref{sec:Applications}), and evaluate it against baselines and other methods (\Cref{baselines-comparison,evaluation}).
Our main contributions are:
\begin{itemize}
    \item We define a new concept termed ``change map'' that generalizes the ``mask'' concept of image editing, and create a framework that implements it (\Cref{method}).
    \item We extend the framework to apply better soft-inpainting than was previously possible (\Cref{soft-inpainting}).
    \item We introduce a tool to visualize the effects of different strength values (\Cref{str_fan}).
\item We devise a new evaluation procedure~(\Cref{eval:reconstruction}) and metrics (\Cref{eval:metric}) to compare different techniques according to their adherence to a change map.    
\end{itemize}
Our framework does not require training or fine-tuning, and it only adds minimal memory overhead to the inference process (\Cref{subsec:memory}).

\section{Related Work}

We first review methods that generate complete images from text (\Cref{subsec:text-based-synt}), and then describe editing methods using either text (\Cref{subsec:text-based-editing}) or masks (\Cref{subsec:stroke-based-editing}) as input.

\subsection{Text-based Image Synthesis} 
\label{subsec:text-based-synt}
In recent years, research on text-based image generation
has become increasingly prevalent.
Early models such as DRAW~\cite{gregor2015draw} and alignDRAW~\cite{mansimov2016generating}, while impressive, tend to produce blurry outputs. 
Many papers~\cite{reed2016generative,zhang2022crossmodal,pmlr-v70-arjovsky17a,xu2017attngan,zhang2017stackgan,sauer2023stylegant,liu2021fusedream} propose GAN-based~\cite{goodfellow2014generative} solutions,
which produce higher quality output, but they often lack coherence---they sometimes miss hierarchical structures, which lead complex objects to be blurry~\cite{reed2016generative}.
\noindent Recently, diffusion models~\cite{sohldickstein2015deep,ho2020denoising} have emerged as a leading solution for image generation~\cite{dhariwal2021diffusion,nichol2021improved},
and for text-to-image synthesis~\cite{zhang2023texttoimage}. 
PromptPaint~\cite{chung2023promptpaint}, MultiDiffusion~\cite{bar2023multidiffusion}, and SpaText~\cite{Avrahami_2023_CVPR} offer methods to synthesize images based on user-provided prompts and shapes.
Recently, a number of models that support text-based image synthesis were released~\cite{rombach2022highresolution,razzhigaev2023kandinsky,DeepFloydIF,podell2023sdxl}.
Our method is an enhancement of editing methods rather than a method for producing whole new images.

\subsection{Text-based Editing}
\label{subsec:text-based-editing}
Several approaches take an image as input and use text to guide the editing process. 
DiffusionCLIP~\cite{kim2022diffusionclip} uses domain-specific diffusion models, DDIM inversion, and fine-tuning, for image editing.  
``More Control for Free!''~\cite{liu2022control} suggests a framework for semantic image generation that allows guiding it simultaneously with a text and an image. 
InstructPix2Pix~\cite{brooks2023instructpix2pix} enables the user to edit pictures with human-friendly instructions. Prompt-to-prompt~\cite{hertz2022prompt} presents “real image editing” which utilizes inversion, extracting text-guided masks from attention layers.
DiffEdit~\cite{couairon2022diffedit} introduces a diffusion-based text-guided photo-editing method, which extracts masks guided by a reference prompt through the comparison of noise outputs, followed by a modified DDIM denoising. 
Other method exists, addressing various challenges in the realm of video and photo content editing~\cite{ceylan2023pix2video,bartal2022text2live,kawar2023imagic,zhang2022sine,nichol2022glide,xie2022smartbrush,wang2023instructedit}. 
While the methods above yield impressive outcomes, we claim that masks offer irreplaceable controllability that surpasses textual prompts alone. This distinction is amplified by our paper, which introduces strength spatial control. 

\subsection{Mask-based Editing}
\label{subsec:stroke-based-editing}
Lately, multiple methods which support mask-based editing emerged~\cite{Lugmayr_2022_CVPR,yu2023inpaint,Suvorov_2022_WACV,Kim_2022_CVPR,9730792,ackermann2022highresolution,andonian2023paint}. Next, we describe diffusion models that support mask-based editing, followed by techniques for mask-based editing that are applicable to diffusion models in general.
\subsubsection{Diffusion Models with Native Support}
\noindent Palette~\cite{saharia2022palette} introduces a diffusion-based model that solves four tasks including inpainting---the ability to discard and re-synthesize parts of an image in a photo-realistic way. In contrast to our solution, it always discards the selected content completely and does not grant the users the capability to guide content completion through text-based instructions.
Recently, Stable Diffusion~\cite{rombach2022highresolution}---a deep learning diffusion model which supports inpainting and I2I, gained attention from the research community, industry, and the public,
in part because the researchers chose to freely release the model's checkpoints\footnote{We hope that other researchers (and companies) will choose to do the same. Our code is available at \href{https://github.com/exx8/differential-diffusion}{https://github.com/exx8/differential-diffusion}.
}.
In recent months, more diffusion models that support inpainting and I2I have been released, such as Kandinsky~\cite{razzhigaev2023kandinsky}, DeepFloyd IF~\cite{DeepFloydIF}, and Stable Diffusion XL~\cite{podell2023sdxl}. All are compatible with our method (\Cref{extension-for-different-diffusion-models-main}). 
All these models offer guided inpainting, taking as input a binary edit mask, a guiding text, and a strength parameter. However, they do not allow specifying different strengths in the same transformation. All but DeepFloyd IF require special fine-tuning for the inpaint task as opposed to our solution which does not require fine-tuning or training. 
\subsubsection{Other Approaches}
 Blended Diffusion~\cite{Avrahami_2022} allows the user to edit specific regions of a picture according to a text.
Blended Latent Diffusion~\cite{Avrahami_2023} expands this approach to latent-based diffusion models.
All these solutions
offer minimal control on the strength applied during the edit. They allow only manipulations that are equivalent to editing by a change map with two or fewer distinct values.
SDEdit~\cite{meng2022sdedit} is a diffusion-based method for editing images. The paper introduces the “realism-faithfulness trade-off”, which is now referred to as “strength”.
It demonstrates that diffusion models can be used for editing images by starting the diffusion process at a later stage than usual with matching noise.
In addition, SDEdit offers stroke-based picture editing.
Unlike our method, it lacks text-guided capabilities, and the strength is uniform for the entire edited region.
Our method generalizes most of the approaches described by enabling local strength control, converting it from a scalar into a matrix, while allowing text-guided editing.

\section{Method}
\label{method}
Given an image, a mono-channel change map representing the desired change amount of each pixel, and a text prompt, 
our goal is to edit the image
to produce a high-quality result that satisfies the desired change and adheres to the prompt.

\subsection{Preliminaries}
\label{sec:Overview}
Diffusion models are deep learning models that have been inspired by thermodynamics \cite{sohldickstein2015deep}.
In computer vision context, they are usually trained to gradually denoise images, that have been corrupted by a random Gaussian noise.
As explained in Denoising Diffusion Probabilistic Models~\cite{ho2020denoising}, typically, the image-to-image translation process 
(\textbf{The Inference Process}) 
begins with an image with added Gaussian noise,
then in an iterative process, the noise is gradually removed.
This inference process creates a series of images (\textbf{Intermediate Images}), where each is the result of the denoising operation of the previous one (\textbf{The Inference Chain}).
The \textbf{Prompt} is a text input that controls the content of the generated picture.
The \textbf{Strength} is a quantity that determines the amount of change applied by the edit. 
For this paper, a \textbf{Mask} is defined as a map comprised of only two distinct values, as used in traditional inpainting and localized editing.
Extending the concepts of strengths and masks, in this paper we introduce the \textbf{Change Map}, a matrix of the same dimensions as the original input image, 
describing the strength of the edit to be applied at each location.

In Latent Diffusion Models \cite{rombach2021highresolution}, the diffusion process occurs in a dimensionally reduced latent space, which the original image is translated to by a \textbf{Latent Encoder} at the beginning of the inference process.
Therefore, in these models, the intermediate images are represented as their latent version.
At the end of the process, a \textbf{Latent Decoder} translates the latent output into an image. 

In this paper, we present \emph{Differential Diffusion} --- an inference-time enhancement of image-to-image diffusion models that adds the ability to 
control the amount of change applied to each image region according to a change map.
Our method decomposes the map into a series of nested masks that are applied iteratively, such that each region begins the inference at a different timestep according to the masks.
\newcommand{\algrule}[1][.2pt]{\par\vskip.5\baselineskip\hrule height #1\par\vskip.5\baselineskip}

\begin{algorithm}[t]
\begin{flushleft}
\caption{Differential Image to Image Diffusion}
\label{algo}

\hspace*{\algorithmicindent} \textbf{Input} $x$ (image to edit), $k$ (number of steps), $\mu$ (change map with values between 0 and 1), $p$ (prompt) \\
\hspace*{\algorithmicindent} \textbf{Output} $\xhat$
\begin{algorithmic}[1]
\Procedure{inference}{$x$, $k$, $\mu$, $p$}

        \State $z_{init}$ = ldm\_encode($x$)
        \label{algo-encode}
        \State $\mu_s$ = down\_sample($\mu$)
        \label{algo-downsample}
        \State $z'_k$ = add\_noise($z_{init}$, $k$) \label{algo-add-noise}

        \State $z_k$ = denoise($z'_k$, $p$, $k$)
        \For{t = $k-1$ to $0$} \label{algo-loop-begins}
            
                \State $z_t'$ = add\_noise($z_{init}$, $t$)
                \State $mask$ = $\mu_s \ewlteq \frac{k-t}{k}$ \label{no-future-hinting-change}
                \State $z_t^{mix}$ = $z_{t+1} \odot mask +  z_t' \odot (1 - mask)$ \label{injection}
                \State $z_t$ = denoise($z_t^{mix}$, $p$, $t$)

        \EndFor
    \State $\xhat$ = ldm\_decode($z_0$)
     \State return $\xhat$

\EndProcedure
\algrule

\noindent We denote $\ewlteq$, $\odot$ as element-wise less-than and element-wise multiplication, respectively.
$\ewlteq$ returns a tensor of 1s and 0s.
\end{algorithmic}
\end{flushleft}
\end{algorithm}

\begin{figure}
  \centering
  \includegraphics[width=\columnwidth]{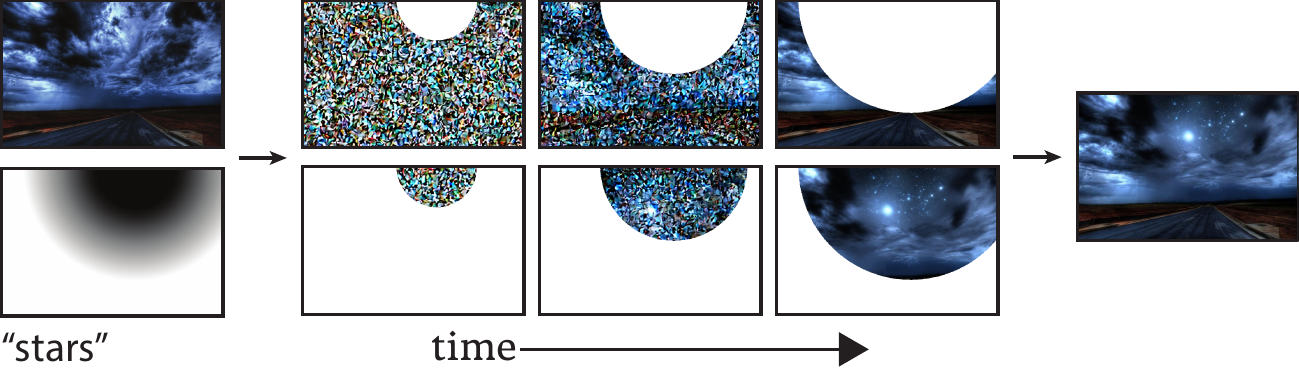}
  \caption{\textbf{Breakdown of \Cref{algo} \Cref{injection} over time.}
  Top: 
  $z_t' \odot (1 - mask)$, the regions
copied from a noised version of the input.
  Bottom:
  $z_{t+1} \odot mask$, the residue regions copied from the previous U-Net output.
  Observe how the change map determines the inference process---the darker the region, the earlier it is copied from the residue.
}
  \label{tensor-print}
\end{figure}

\subsection{Observations}
\label{observations}
Our method is based on the following observations:
\begin{enumerate}
    \item \textbf{The Suffix Principle}: For every complete image-to-image inference chain $\Sigma$, every suffix $\sigma$ is also an inference chain. Let $N$,$n$ be the number of timesteps of $\Sigma$,$\sigma$, the noise levels in $\sigma$'s intermediate images match those of an inference chain, with a strength of $\frac{n}{N}$.
    \item \textbf{Overridability}:
    Regions in the intermediate images can be overridden by external content with the same distribution, and influence the generated image without breaking the inference process. 
    \item \label{latent-locallity} \textbf{Locality}: All the latent encoders that were examined~\cite{rombach2022highresolution,podell2023sdxl,razzhigaev2023kandinsky} generally encode pixels to the same relative positions, which is the ratio between their positions to the dimensions of the picture.
\end{enumerate}

\subsection{Algorithm}
\label{algo-analysis}
In our framework, we change the inference algorithm of the diffusion process~(\Cref{algo}).
First, the original image is encoded to the latent space ($z_{init}$), and the map is downsampled to the latent space spatial dimensions. Due to \textit{locality}, the down-sampled map ($\mu_s$) aligns with the positions of the latent pixels in the latent tensor.
The denoising loop is changed for each time step $t$ as follows.
We first noise the encoded original image according to the current timestep ($z'_t$).
 Then, we calculate a mask of all points which are lower than the threshold $\frac{k-t}{k}$ in the change map. 
 $\frac{k-t}{k}$ is the value of the complement of the strength. Therefore, the values of the mask determine the last timestep where each region is overridden, controlling the amount of change of the region, due to \textit{the suffix principle}.
 
The masks are nested~(\Cref{tensor-print}), therefore some regions are copied from the noised original image multiple times, in contrast to copying each region once according to its strength. Because the diffusion model has not been trained on intermediate images with holes, this mimics the distribution which it has been trained on, and gives it an advance knowledge of the content of lighter regions.
 We conducted an experiment that affirms these advantages in \Cref{future_hinting_examples}---we changed \Cref{no-future-hinting-change} in \Cref{algo} to $mask =\frac{k-t}{k}\ewgt \mu_s \ewgteq \frac{k-(t+1)}{k}$, which means for each timestep, only the regions that match the exact timestep were copied.
Next, we copy all the selected regions in the mask from the previous timestep. The rest is copied from the noised version of the original image (\Cref{fig:diff-inference}).
This is possible due to \textit{Overridability}.
Finally, the U-Net denoises the result ($z_t^{mix}$). 
After the loop, $z_0$ is decoded to the pixel space, yielding the result ($\xhat$).
\subsubsection{Optimization: Skipping}
We observed that the inference process can be optimized for change maps devoid of small values. In this scenario, we can significantly enhance efficiency by skipping steps related to these small values and avoiding their calculation, as their values will be overridden by the injections of subsequent steps.
It can be implemented by adding
before \Cref{algo-add-noise} in \Cref{algo}: ${L=\left\lfloor \left(1-\min(\mu_{s})\right)\cdot k\right\rfloor }$
and replace the references for k to L in  \Crefrange{algo-add-noise}{algo-loop-begins} only. Refer to the supplementary materials for the explicit algorithm.

\subsubsection{Technical Details}
\label{prompt-discovery} 
\textbf{Model:} We have used the checkpoint \texttt{512-base-ema.ckpt} of Stable Diffusion 2.1, unless otherwise stated. 
\textbf{Prompts:} 
For most experiments, we used a simple description of the edit. For others, we found it beneficial to expand the prompt by taking the input image, reversing it into a prompt via a Clip interrogator that uses both BLIP~\cite{li2022blip} and CLIP~\cite{radford2021learning}, and then adding the desired edit to the prompt.
\textbf{Change Maps:}
Our method does not assume anything about the source of the change maps. We show results using various sources: (1) Segment-Anything~\cite{kirillov2023segment} and Language-Segment-Anything, mainly for discrete change maps. (2) MiDaS~\cite{Ranftl2022} mainly for continuous change maps; from a single depth map, many change maps can be created by simple histogram transformations. (3) Manually drawn change maps. \textbf{Other settings:}
Unless stated otherwise, we used 100 inference steps for each experiment.

\subsection{Extension For Different Diffusion Models}
\label{extension-for-different-diffusion-models-main}
We generalize our framework for Stable Diffusion XL~\cite{podell2023sdxl}, Kandinsky~\cite{razzhigaev2023kandinsky}, and DeepFloyd IF~\cite{DeepFloydIF}. See \Cref{fig-extension-model} and the supplementary materials for more information.
\newcommand{\www}{0.1135\textwidth}

\begin{figure}[t]
  \centering
  \begin{tabular}{cccc}
    \textbf{Input Image} & \textbf{Change Map} & \textbf{Ours} & \textbf{No Nesting} \\

    \includegraphics[width=\www,frame]{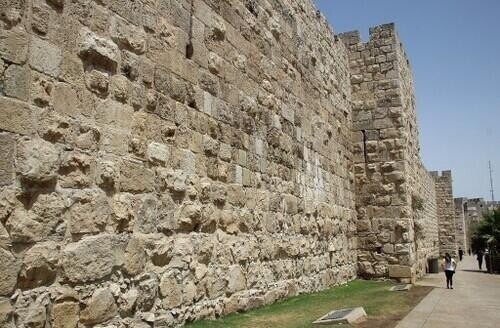} &
    \includegraphics[width=\www,frame]{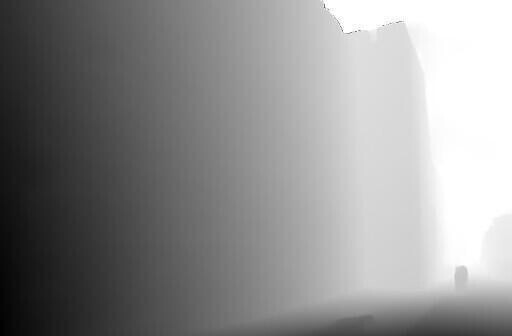} &
    \includegraphics[width=\www,frame]{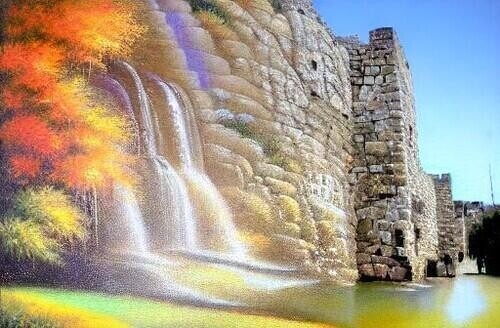} &
    \includegraphics[width=\www,frame]{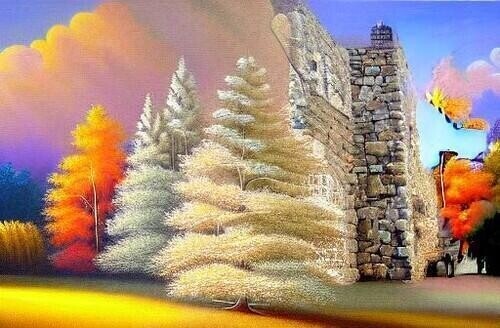} \\
    
    \includegraphics[width=\www,frame]{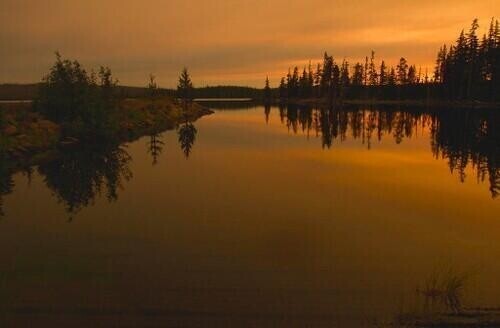} &
    \includegraphics[width=\www,frame]{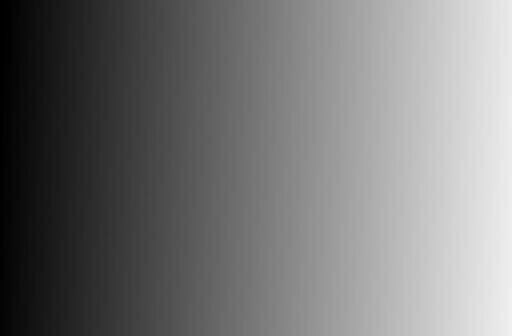} &
    \includegraphics[width=\www,frame]{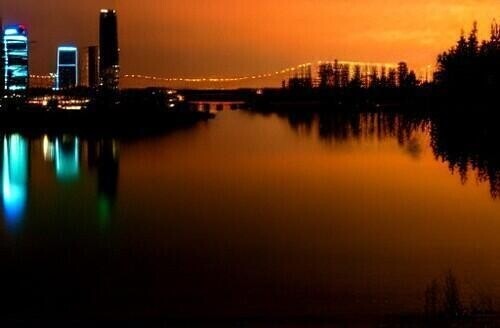} &
    \includegraphics[width=\www,frame]{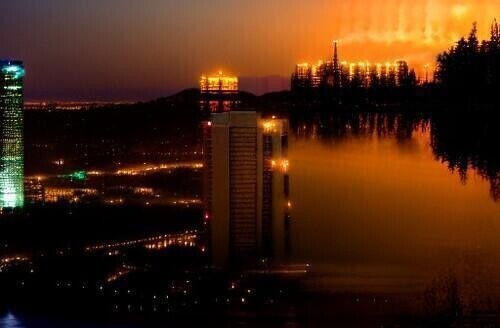} \\
  \end{tabular}
  
  \caption{\textbf{Ablation of nested masks.} 
  Our result is more complex, blends better with the scene, and less blurry. Note the difference in transitions (1st row: the sharp transition in the wall) and placements (2nd row: the building is inside the lake).
The seed is fixed for each row.
Prompts: “a fine art painting”, “a city skyline…”.
  }
  \label{future_hinting_examples}
\end{figure}

\begin{figure}
  \centering
  \includegraphics[width=\columnwidth]{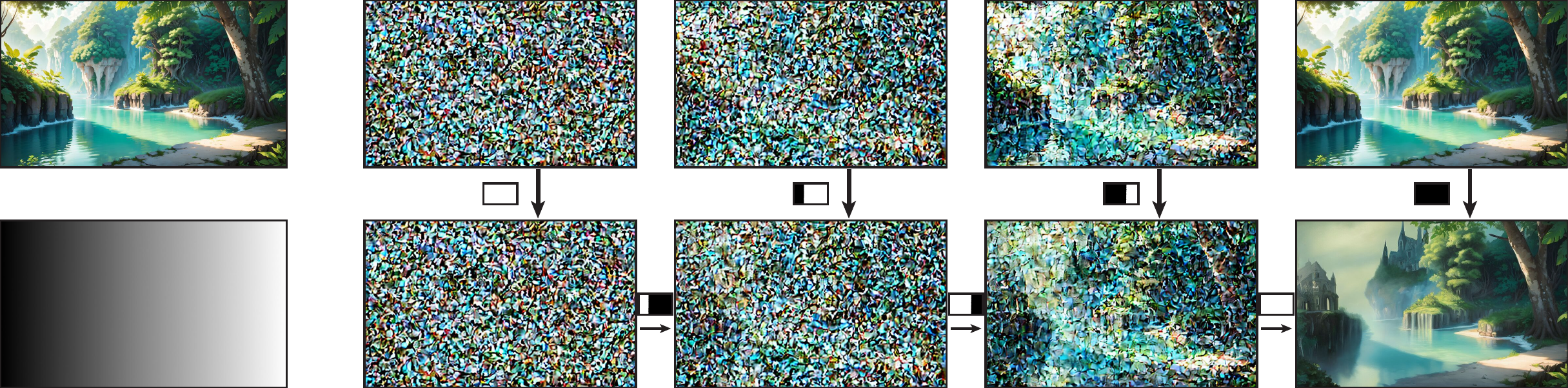}
  \caption{\textbf{illustration to the inference process.}
  Top: 
  $z_t'$ - the original image noised to the current timestep.
  bottom:
the intermediate images that the diffusion model denoises. 
The masks near the arrow represent the regions that were copied from each picture. %
Follow the arrows to discern the influence of the origins on the output image, and observe the correlation with the decomposed masks and the change map.
 The prompt is ``Gothic painting''.
 }
 \label{fig:diff-inference}
\end{figure}

\section{Applications}
\label{sec:Applications}
\label{sec:Multi-Strength Editing}

Our method is the first to allow users to edit images with an arbitrary number of strengths (\Cref{fig:teaser}). This introduces fine grained control over the impact of the edit prompts on different image regions (\Cref{video_game}). It also allows continuous edit transitions (e.g., \Cref{comparison_gallery,future_hinting_examples,fig-extension-model}), a feature that has been found to be visually appealing in our user study (\Cref{sec:user-study}).
Besides our main application of guiding an edit with a change map, our method also supports soft-inpainting, and allows for a novel visualization tool we call a strength fan, as we detail next.

\subsection{Soft-Inpainting}
\label{soft-inpainting}
Inpainting with no softening can sometimes appear unnaturally superimposed due to noticeable differences in style and lighting compared to the surrounding background. ``Soft-Inpainting'' is the process of completing parts of a picture while gently modifying the surrounding regions to guarantee smooth blending.
We extend our framework to support Soft-Inpainting, by allowing the user to input a binary \textit{mask} and a scalar value to determine transition softness. We Gaussian blur the mask according to the softness parameter, turning it into a \textit{change map}, that is then processed by our standard framework. 
Our method achieves superior soft-inpaint editing compared to previous methods (\Cref{fig:monks-short}).

\subsection{Strength Fan}
\label{str_fan}
When editing images with diffusion models, finding the perfect balance between preserving certain elements and altering others can be challenging. Often, choosing the right strength for an edit can be non-intuitive, especially as the optimal setting varies across input prompts and images. To simplify this process, we propose a new visualization tool called ``Strength Fan''.  This fan is a modified image created by dividing it into columns, with each column undergoing editing at a different strength level. 
This allows users to observe multiple strength settings simultaneously, thereby simplifying the task of comparing and tuning edit strengths 
(\Cref{str_fan_short}).
Our framework is uniquely suited to produce strength fans, as we only need to invoke our inference chain \emph{once}, with a change map of rectangles corresponding to the various strengths under examination.

\section{Evaluation}
\newcommand{\wwww}{0.305\columnwidth}

\begin{figure}[t]
  \centering
  \begin{tabular}{cccc}
    \textbf{Input Image} & \textbf{Change Map} & \textbf{Output} & \\

    \includegraphics[width=\wwww,frame]{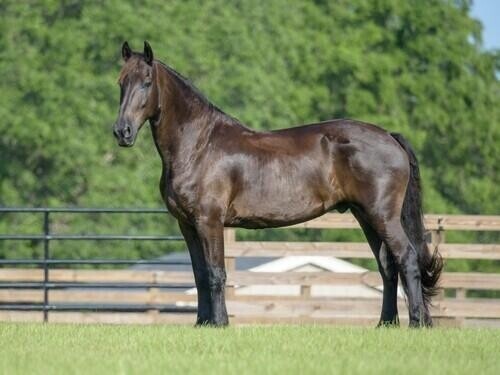} &
    \includegraphics[width=\wwww,frame]{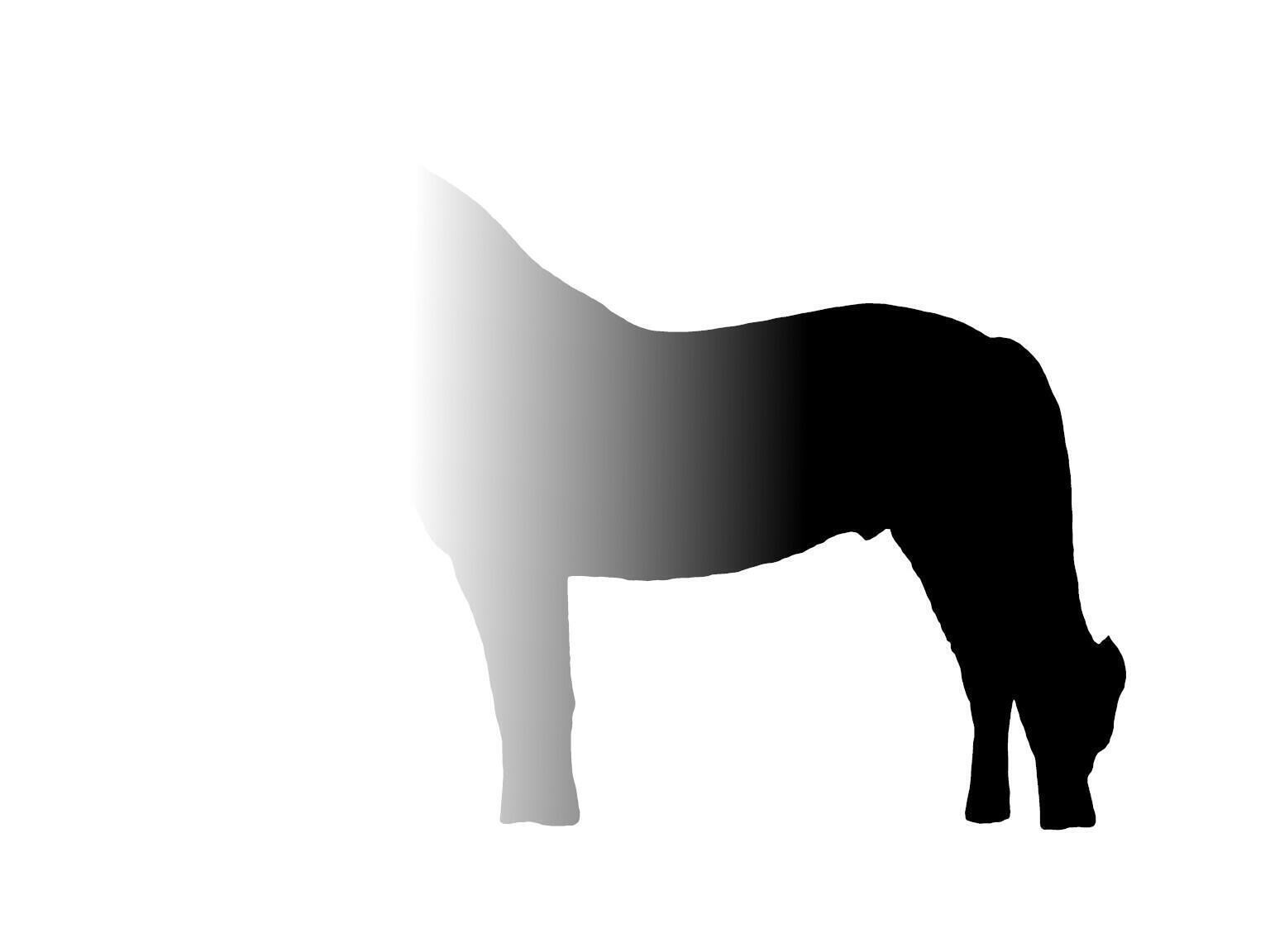} &
    \includegraphics[width=\wwww,frame]{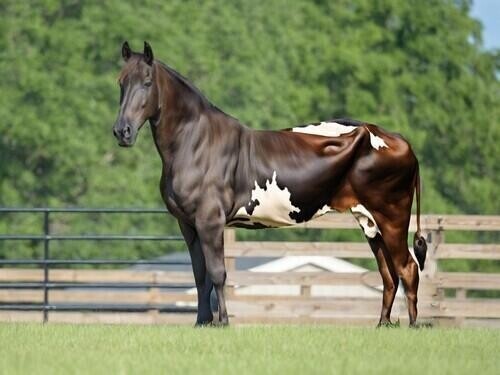} &
    \raisebox{1.8cm}{\rotatebox{270}{SDXL~\cite{podell2023sdxl}}}   \\
    
    \includegraphics[width=\wwww,frame]{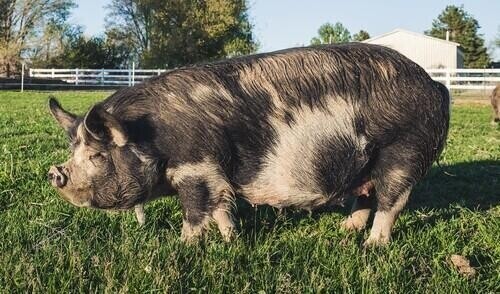} &
    \includegraphics[width=\wwww,frame]{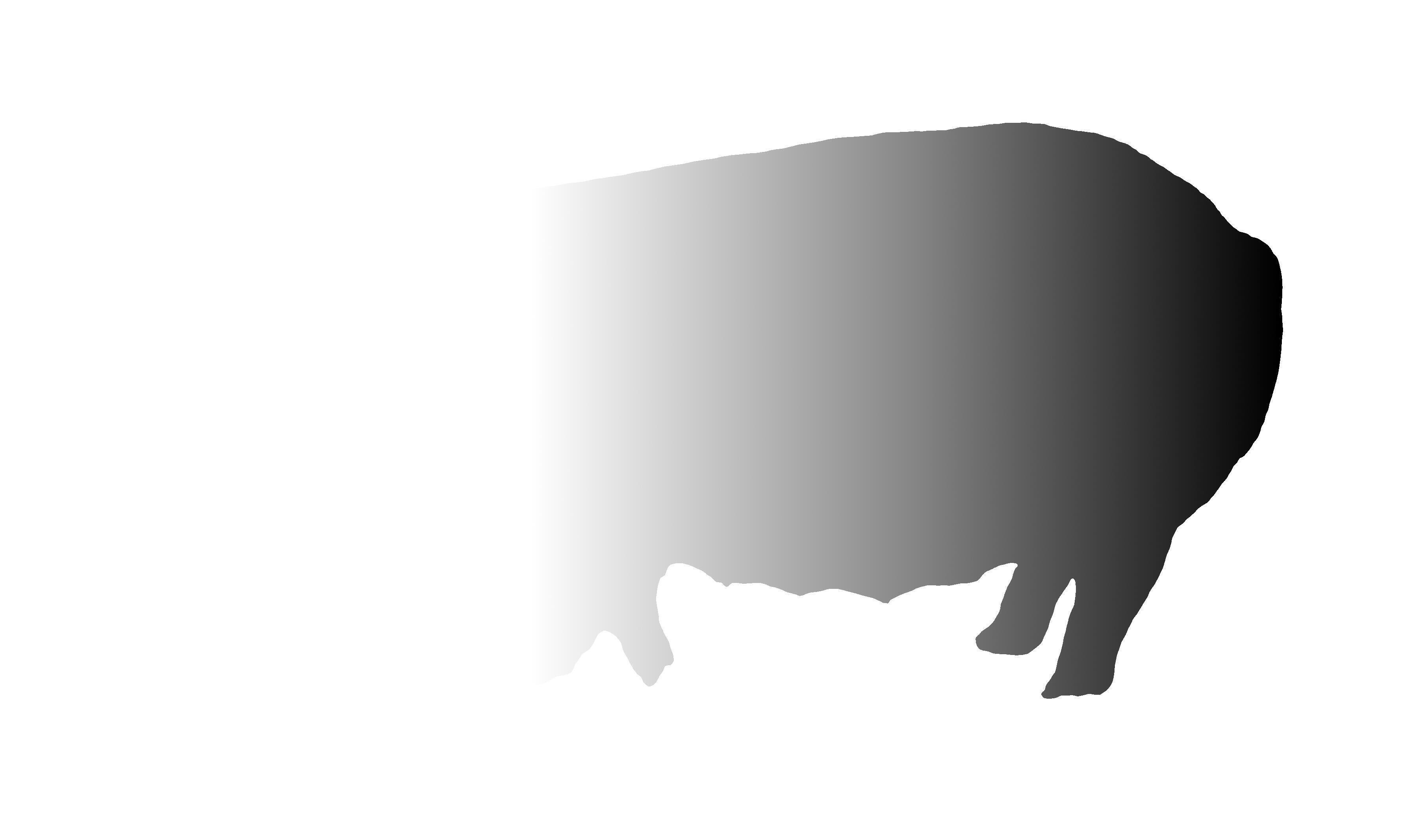} &
    \includegraphics[width=\wwww,frame]{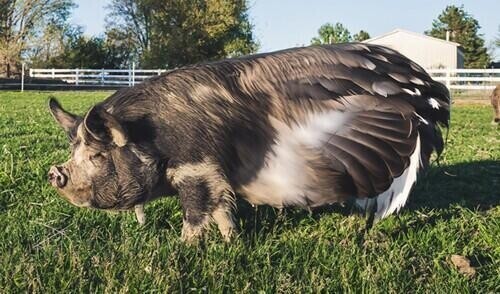} &
    \raisebox{1.2cm}{\rotatebox{270}{IF~\cite{DeepFloydIF}}}  \\

    \includegraphics[width=\wwww,frame]{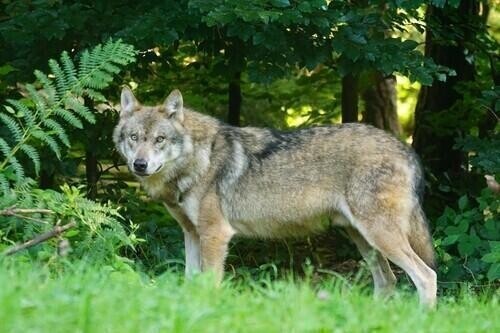} &
    \includegraphics[width=\wwww,frame]{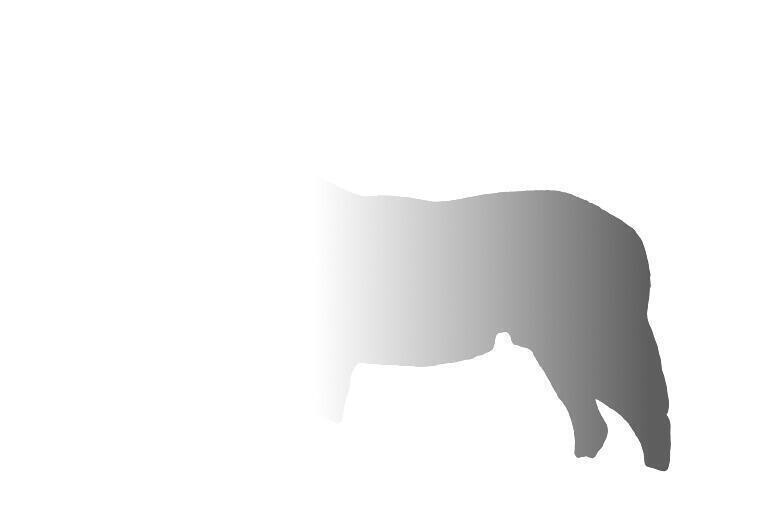} &
    \includegraphics[width=\wwww,frame]{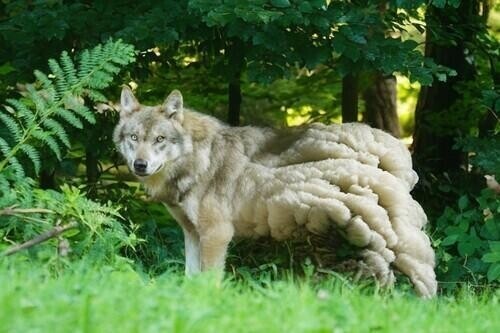} &
    \raisebox{1.6cm}{\rotatebox{270}{Kand.~\cite{razzhigaev2023kandinsky}}} 
  \end{tabular}
  
  \caption{\textbf{Our method with different diffusion models.} We applied our framework to several diffusion models: SDXL~\cite{podell2023sdxl}, DeepFloyd IF~\cite{DeepFloydIF}, and Kandinsky~\cite{razzhigaev2023kandinsky}, demonstrating its generality. Prompts: “cow”, “feathers”, “sheepskin”.}
  \label{fig-extension-model}
\end{figure}

We compare our method to baselines (\Cref{baselines-comparison}) and to other methods
(\Cref{evaluation}), evaluate our method via a user study~(\Cref{sec:user-study}), and report memory consumption and running time~(\Cref{subsec:memory}).

\subsection{Comparison to Baselines}
\label{baselines-comparison}
We compare our inference algorithm to four baseline alternatives,
founded on different ideas. Visual results are in \Cref{fig:alternatives}.

We start with several iterative alternatives.
\textbf{Composition}:\label{composition} Since most existing diffusion methods accept a \emph{single} strength parameter (e.g., SDEdit~\cite{meng2022sdedit}), a straightforward method to support a varying change map is to decompose the map into K different masks (we use $K=100$), each with a single value, and applying them iteratively. 
The image degrades rapidly (after applying five masks), leaving a meaningless image at the end. The deterioration is mostly caused by the recurrent transition via the latent encoder of Stable-Diffusion.
\textbf{Tiling}: We try to circumvent the previous limitation by avoiding the cumulative degradation. As before, we decompose the change map into 100 masks, and then apply a series of inpaints of the masks successively. But instead of using all the pixels of the output image, after each inpainting we copy the pixels that lie outside the current mask  from the previous step (for the first step we copy them from the original image). This guarantees that each region will only traverse once through the latent space. Here, the model has difficulties inpainting narrow masks, which is inherent in tiling and every process that decomposes the map into fine-grained masks, leading to pixels being replaced with gray stripes. \textbf{Five Tiles}:
We attempted to overcome the limitations of tiling by increasing the size of the tiles. In this alternative, we grouped the map values to five distinct bins.
We then apply tiling as before, this time with only 5 masks. This method generates reasonable visual content by avoiding the use of narrow mask segments. However, the method usually fails to semantically change the regions except for the darkest tile, thus not fulfilling the desired task. All iterative alternatives exhibit inefficient running times, sometimes tens of times longer than our method, depending on the number of applied masks. 

\textbf{Masked Noise}: It can be hypothesized that spatially adjusting the added noise's magnitude can produce similar results to our method. We show it cannot. In this alternative, we multiplied the added noise by the change map. The model soon converges to a random single-color image. This can be anticipated, as the model was originally trained to handle a specific noise distribution, which this technique alters. 

\subsection{Comparison to Other Methods}
\label{evaluation}
\newcommand{\ad}{0.23\columnwidth}

\begin{figure}[t]
\centering
\begin{tabular}{ccccc}
 \textbf{Input Image} & \textbf{Inpaint Mask} & \textbf{No Softening} & \textbf{\textalpha-Compose} \\

\includegraphics[width=\ad,frame]{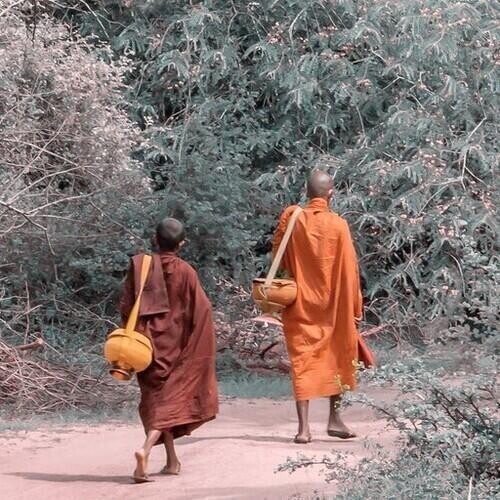} &
\includegraphics[width=\ad,frame]{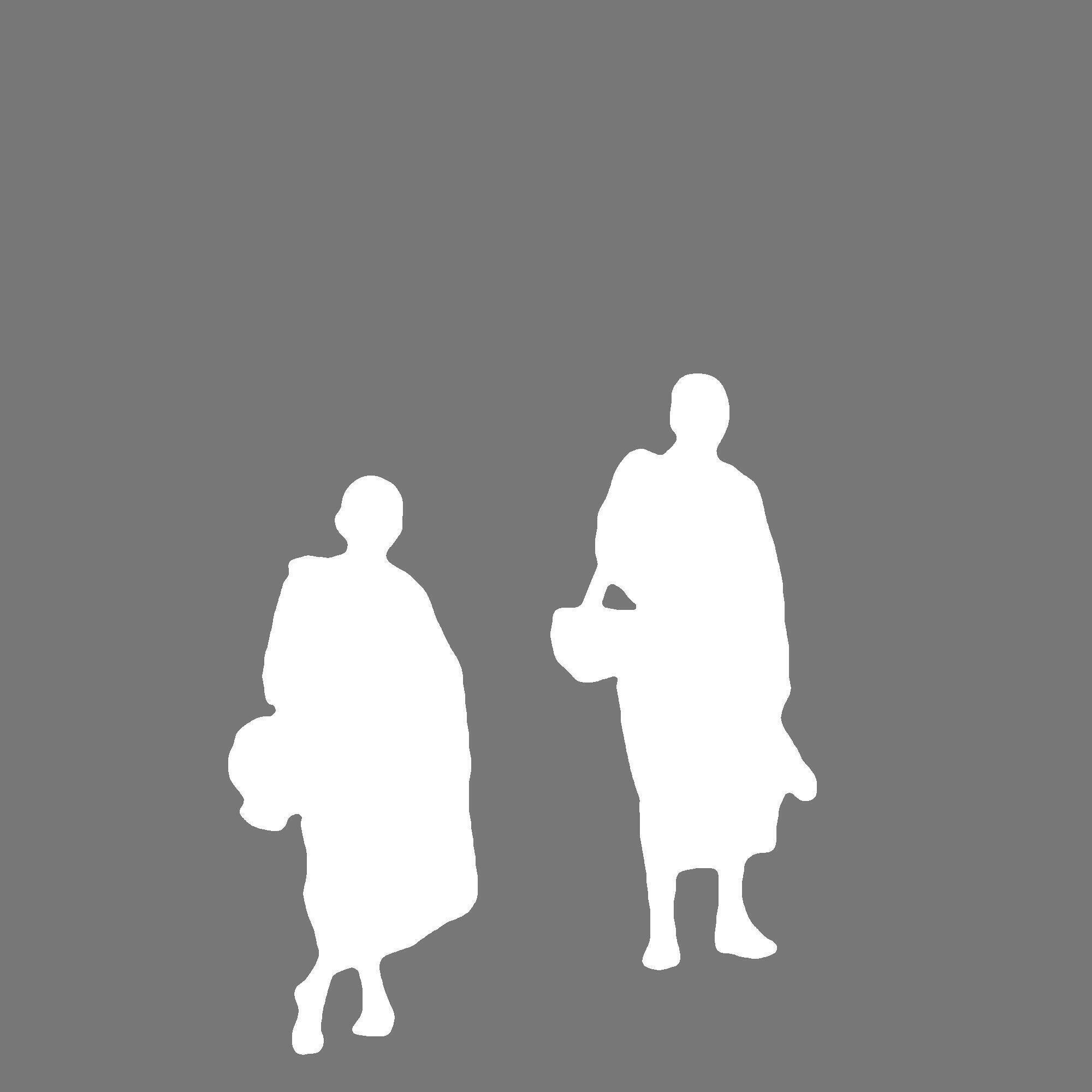} &
\includegraphics[width=\ad,frame]{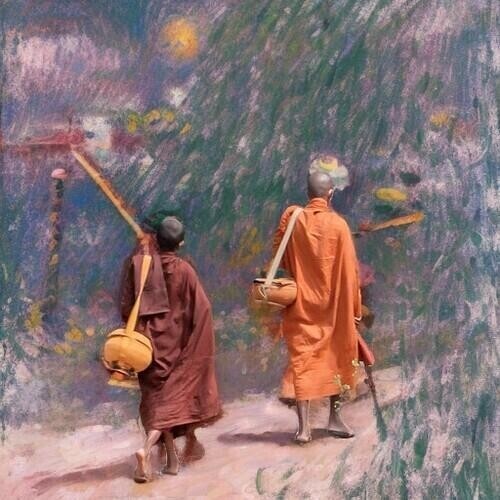} &
\includegraphics[width=\ad,frame]{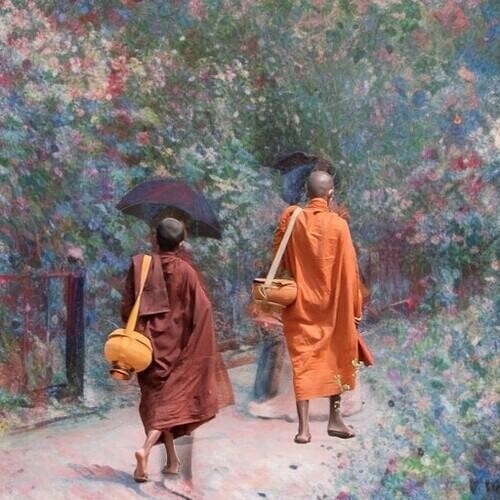} \\
  \textbf{Poisson} & \textbf{Laplace} & \textbf{Standard Soft} & \textbf{Ours} \\
\includegraphics[width=\ad,frame]{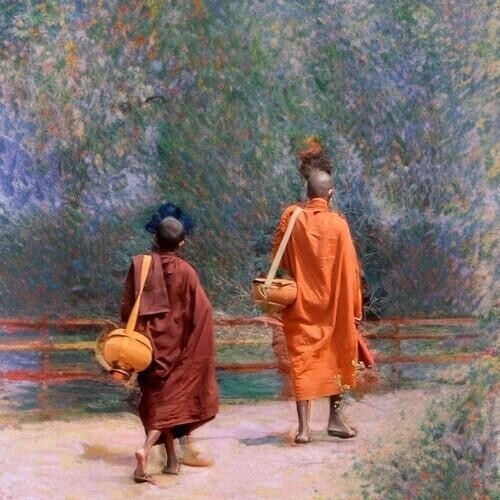} &
\includegraphics[width=\ad,frame]{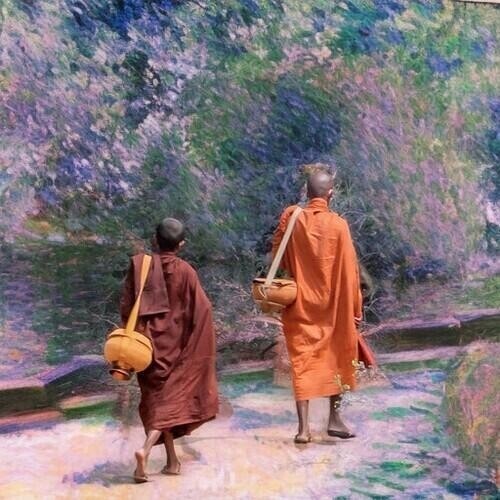} &
\includegraphics[width=\ad,frame]{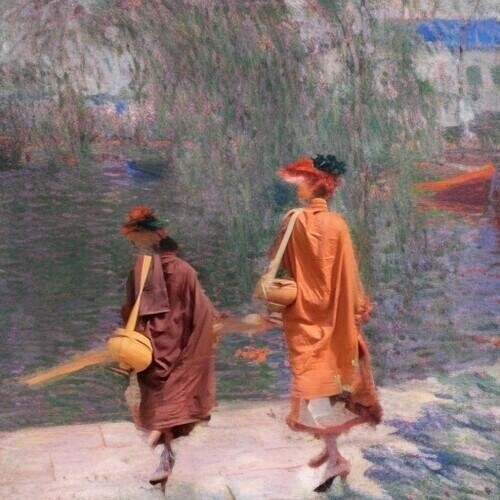} &
\includegraphics[width=\ad,frame]{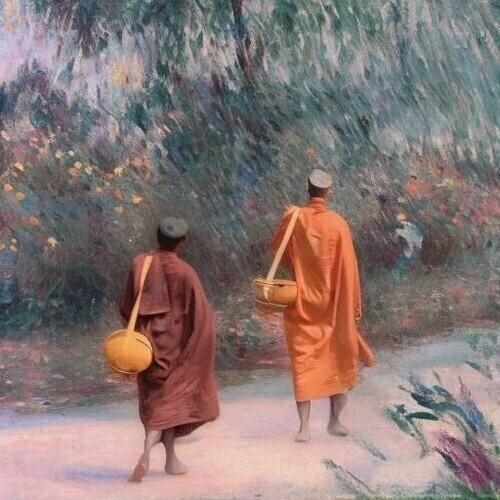} 
\end{tabular}
\caption{\textbf{Soft-inpainting.}
We compare our approach to no softening, \textalpha-compositing, Poission-based~\cite{10.1145/882262.882269} and Laplace-based~\cite{10.1145/245.247} compositing, and standard soft-inpainting (as implemented in Stable Diffusion web UI~\cite{AUTOMATIC1111_Stable_Diffusion_Web_2022}). 
For \textalpha-compositing, Poisson-based and Laplace-based methods, we blend the original image with a regular inpaint result using a Gaussian blurred version of the inpaint mask. 
In all other methods, artifacts appear in the transition area, and the unchanged region looks pasted. For standard softening, even the inner parts of the figures are corrupted.
Our method produces a more natural blend.
Prompt: ``Impressionist''.
}

\label{fig:monks-short}
\end{figure}

We demonstrate the effectiveness of our change maps over text-guided-only methods: InstructPix2Pix~\cite{brooks2023instructpix2pix} and DiffEdit~\cite{couairon2022diffedit} in \Cref{video_game},
and over mask-based methods: Stable Diffusion 2's Text-Guided Inpainting~\cite{sd2}, and Blended Latent Diffusion~\cite{Avrahami_2023} in \Cref{comparison_gallery}.

For quantitative comparison, we begin by demonstrating how to measure edit strength spatially given input-output pairs (\Cref{eval:reconstruction}). We then use this technique to establish metrics for assessing change map adherence (\Cref{eval:metric}), and use them to compare our method to Stable Diffusion 2 Inpaint and Blended Latent Diffusion (\Cref{metric-results}).

\subsubsection{Edit Strength Measurement}
\newcommand{\ab}{0.234\columnwidth}
\newcommand{\ac}{5pt}

\begin{figure}[t]
\centering
\begin{tabular}{cc@{\hskip 0.2cm}cc} 
\vspace{-0.1cm}  \includegraphics[width=\ab,height=\ac,frame]{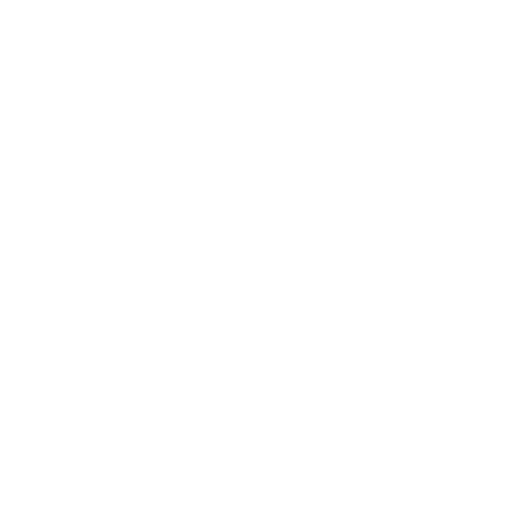}  &
\includegraphics[width=\ab,height=\ac,frame]{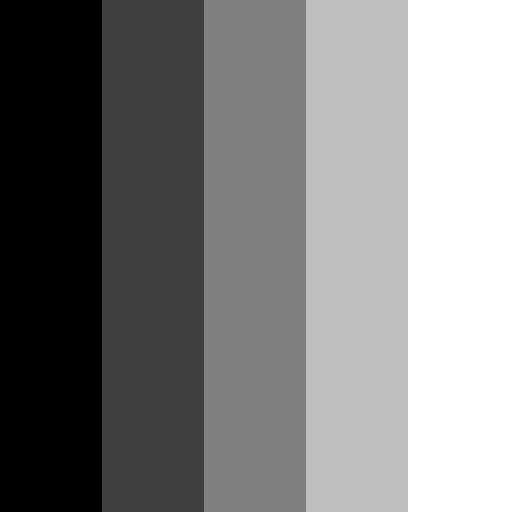} &
\includegraphics[width=\ab,height=\ac,frame]{imgs/str_fan/fans/null.jpg}  &
\includegraphics[width=\ab,height=\ac,frame]{imgs/str_fan/fans/stripe_5.jpg}
\\
\includegraphics[width=\ab,frame]{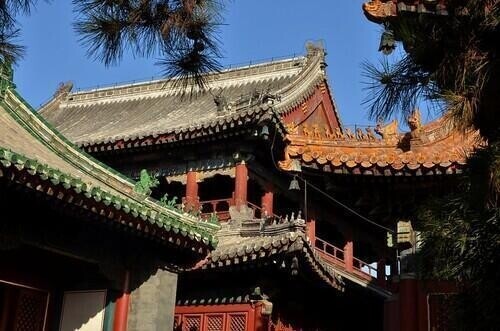} &
\includegraphics[width=\ab,frame]{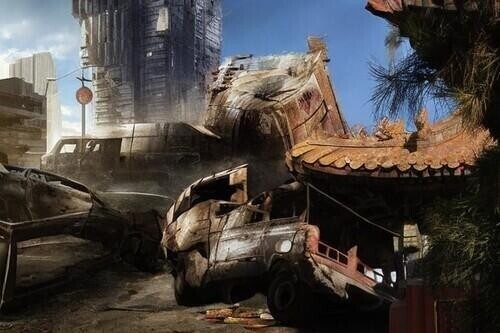}
&
\includegraphics[width=\ab,frame]{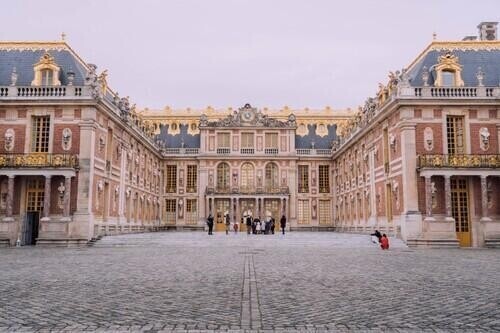}  &
\includegraphics[width=\ab,frame]{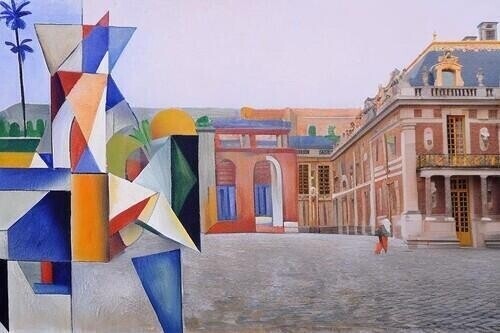}
\end{tabular}
\caption{\textbf{Strength Fans.} Our Strength Fans allow users to visually compare and tune edit strengths.
Prompts: ``post-apocalyptic'', ``Cubist painting''. Extended version in the supplemental materials.}
\label{str_fan_short}
\end{figure}

\label{eval:reconstruction}
We propose the following procedure for spatially measuring edit strength, providing a means to evaluate the applied map during an edit.
Given a change map $M$, and an input-output image pair, we first compute the LPIPS~\cite{zhang2018unreasonable} perceptual similarity map~\cite{zhang2018perceptualsimilarity} between input and output.
For robustness,
we repeat this procedure with 1,000 different input-output pairs 
and average the results. This process is referred to as ``biased measurement''.
For every change map $M$, the edit strength measurement map~($E_M$) for a method is defined to be the biased measurement of $M$ subtracted by the biased measurement of a black map (full change), 
which eliminates
spatial bias in the perceptual similarity map.

\subsubsection{Adherence Metrics: CAM and DAM}
\label{eval:metric}
Our objective is to measure the extent to which each method adheres to change maps and to establish metrics for assessing adherence.
As far as we know, we are the first to quantify change map adherence. 
We suggest two metrics:
Correlation Adherence
Metric~(CAM) and Distance Adherence Metric~(DAM).
In general, CAM tends to focus on high-level features; for example, CAM will usually assign a poor similarity score when comparing maps of different shapes.
DAM, on the other hand, focuses on lower-level features; for instance, DAM will usually assign a lower similarity score to maps that differ by regions in which their brightness has been changed.
As a result, each metric offers a distinctive view of adherence quality. Refer to the supplementary material for an example of their differences.

Let $M$ be a change map. CAM is defined as: $CAM(M,E_M)=\rho(M,E_M)$, where $\rho$ is the Pearson correlation coefficient calculated element-wise. 
DAM is defined as: $ DAM(M,E_M) = \underset{(a,b)\in\mathbb{R}^2}{\min} \left\Vert M - aE_M + b \right\Vert_F 
$, where $\left\Vert \cdot\right\Vert _{F}$ is the Frobenius norm.
We introduced the parameters $a$ and $b$
because LPIPS has a magnitude and additive bias depending on the change map and method used.
Pearson correlation coefficients are invariant under positive multiplication and addition.
\newcommand{\ff}{0.06}
\begin{figure}[t]
  \centering
  \setlength{\tabcolsep}{1pt}
  \renewcommand{\arraystretch}{2}
  \setlength{\ww}{0.067\textwidth}
    
\includegraphics[width=\columnwidth]{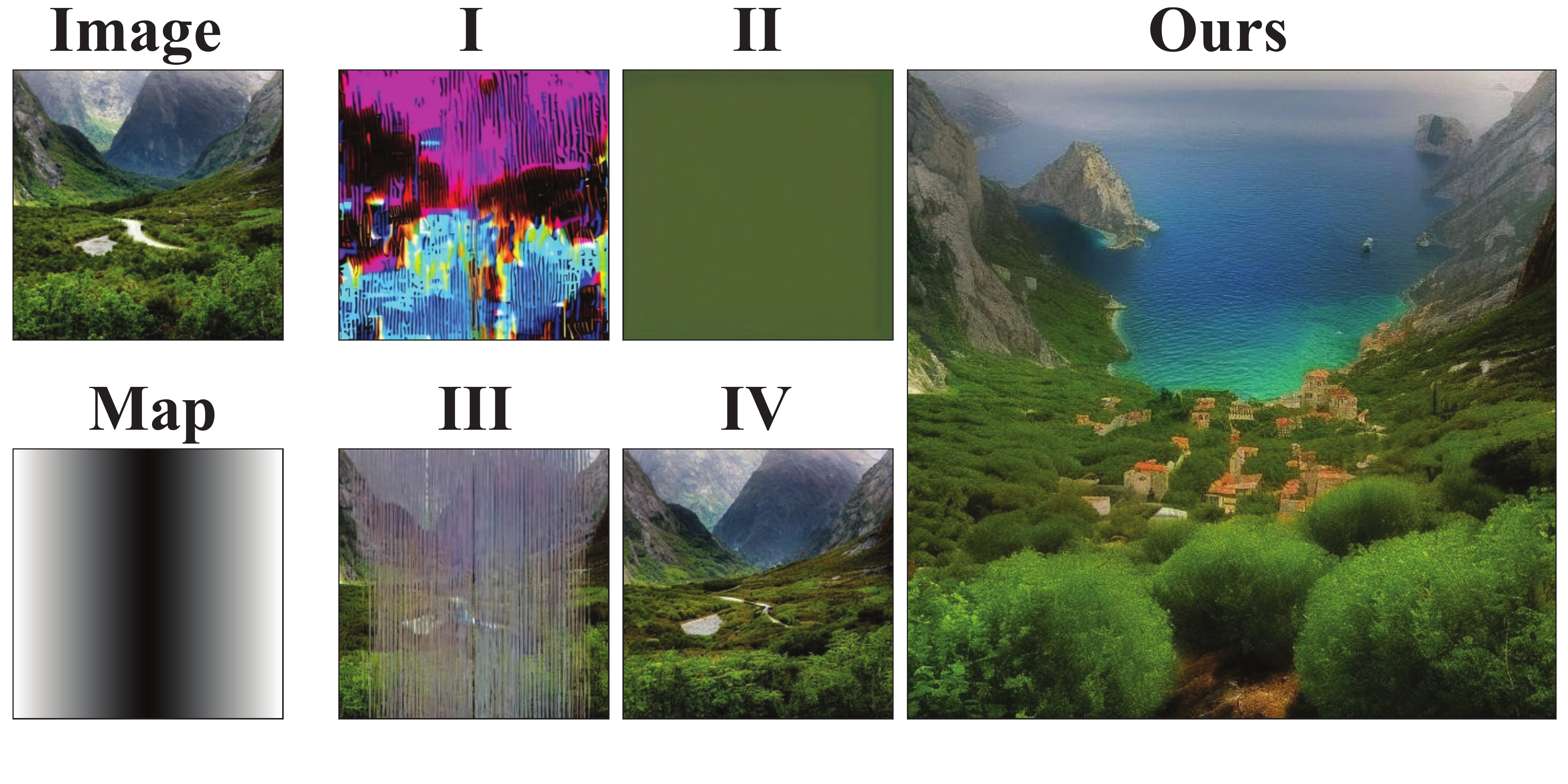}
  
  \caption{\textbf{Comparison to baselines.}
  All baselines fail, even for this simple unstructured scene and a smooth map. ``Composition'' (I) and ``Masked Noise'' (II) both fail to create a meaningful image, while ``Tiling'' (III) does not produce an edit related to the prompt and corrupts the image. ``Five Tiles'' (IV) stands out as the most successful among the baselines. Albeit, the edit is primarily noticeable in the darkest tile.
  Prompt: ``Mediterranean Sea''.
  }
 \label{fig:alternatives} 
\end{figure}

\subsubsection{Experimental Setup and Results}
\label{metric-results}
We sample 1,000 images randomly from ImageNet's validation split~\cite{ILSVRC15}, 
and use them to measure 3 different maps, with an empty prompt.
As we are the first method to allow change maps with an arbitrary number of strengths,
other methods support only two strengths. 
For a fair comparison, we tested other methods using the best performing binarized version of the maps (by sweeping over threshold values and reporting the best result). 
Our results produce the most accurate maps (\Cref{fig:reconstruction}), and our method obtains the best numerical scores (\Cref{cam_dam_table}).

\subsection{User Study}
\label{sec:user-study}
We conducted a user study that included STEM students and volunteers from social media platforms to assess the usability of our method. Each participant received 30 questions at random from a pool of 69 questions. Each question was answered on average by 8 participants. The participants were able to skip questions if they were undecided.
The study had three parts:
\textbf{1. Image-to-Map Matching:} Participants identified the applied map out of three options for ten input-output image pairs. This part tests the ability of untrained users to estimate the impact of  different change maps on the output and their intuitiveness.
\textbf{2. Method Comparison:} Participants ranked three methods (our method, blended-latent-diffusion, Stable-Diffusion's inpaint) based on adherence to change maps and visual quality for given input images and maps.
\textbf{3. Text-Guided Editing Assessment:} Participants chose between two images, one edited with text guidance and the other without, for triplets consisting of input images, change maps, and prompts.
This part tests whether the changes which are applied by our framework adhere to the textual prompt. Throughout all stages of the study, both the images and prompts were selected at random from the InstructPix2Pix~\cite{brooks2023instructpix2pix} dataset. Images smaller than $256\times256$ pixels were excluded, as they are unsuitable for all the techniques under examination.
 For the second and third parts, the change maps were automatically created using MiDaS~\cite{Ranftl2022}.
For evaluation, we determined results based on the majority vote.  As shown in~\Cref{tbl:user-study1}, users mostly chose the applied change maps, 
and also chose the text-guided images on most of the edits.
The user mostly preferred our framework over alternative approaches, as shown in~\Cref{tbl:user-study2}.
Examples of the questions can be found in the supplementary materials.

\begin{table}[tbp]
    \setlength{\tabcolsep}{3.5pt}
    \begin{tabular}{llcccc}
        \toprule

               &                  & \multicolumn{3}{c}{Pattern} \\
        \cmidrule{3-5}
        Metric & Method           & Gradient & Shapes & Triangles \\
        \midrule
        \multirow{3}{*}{CAM ($\uparrow$)} & Ours & \textbf{0.97} & \textbf{0.81} & \textbf{0.93} \\
                             & BLD~\cite{Avrahami_2023} & 0.92 & 0.68 & 0.83 \\
                             & SD Inpaint~\cite{sd2} & 0.93 & 0.65 & 0.82 \\
        \midrule
        \multirow{3}{*}{DAM ($\downarrow$)} & Ours & \textbf{19.41} & \textbf{52.2} & \textbf{35.75} \\
                             & BLD~\cite{Avrahami_2023} & 29.05 & 65.65 & 53.95 \\
                             & SD Inpaint~\cite{sd2} & 28.69 & 67.84 & 54.44 \\
        \bottomrule
    \end{tabular}
        \caption{Comparison between different methods' performance across different patterns by utilizing CAM and DAM metrics.}
\label{cam_dam_table}
\end{table}

\begin{table}[ht]
    \setlength{\tabcolsep}{6pt}
    \centering
    \begin{tabular}{lcc}
        \toprule
        Criteria & Match & P-Value \\
        \midrule
        Map Matching & 80.43\% & $1.31\times10^{-5}$\\
        Text-Guidance & 92.11\% &$3.64\times 10^{-4}$\\
        \bottomrule

    \end{tabular}
    \caption{
    Users are able to identify which change map was used for a given edit, and to determine that a text-guided edit is indeed closer to the text than a non-guided edit. This shows that the use of text-guided change maps leads to perceptible results.
    }
    \label{tbl:user-study1}
\end{table}

\begin{table}[t]
    \setlength{\tabcolsep}{4.5pt}
    \centering
    \begin{tabular}{lcccc}
        \toprule
        Criteria & BLD~\cite{Avrahami_2023} & SD2.1~\cite{sd2} & Ours & P-Value \\
        \midrule
        Map adherence & 10\% & 32\% & \textbf{58\%} & 0.0164 \\
        Visual Quality & 10\% & 35\% & \textbf{55\%} & $0.037$ \\
        \bottomrule

    \end{tabular}
\caption{ 
The preferred method  by  users according to map adherence and visual quality. The results are statistically significant. 
}
\label{userstudy-table}
 \label{tbl:user-study2}

\end{table}

\begin{figure}[t]
\captionsetup[subfigure]{labelformat=empty}

  \centering
\begin{subfigure}{0.32\columnwidth}
\includegraphics[width=0.99\linewidth,frame]{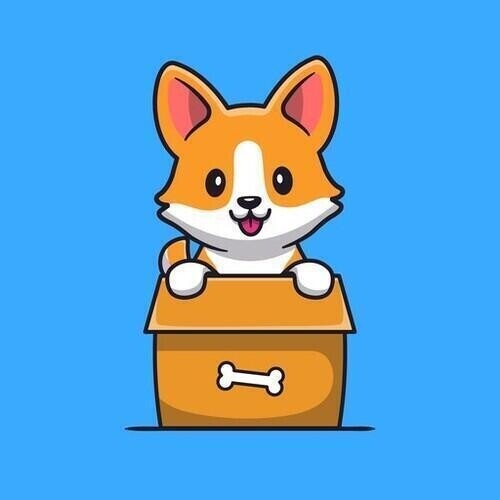}
\caption{Input Image}
\end{subfigure}
\hfill
\begin{subfigure}{0.32\columnwidth}
\includegraphics[width=0.99\linewidth,frame]{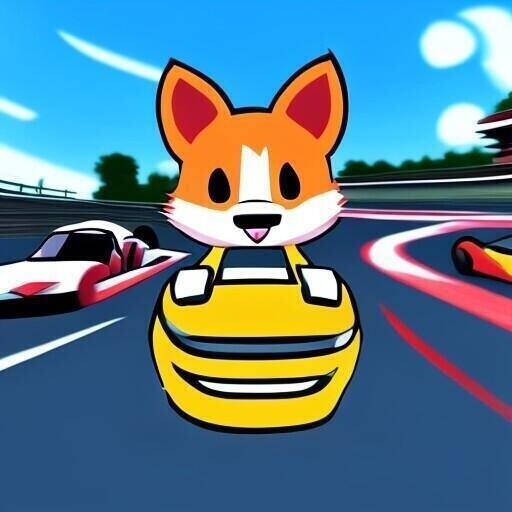}
\caption{InstructPix2Pix~\cite{brooks2023instructpix2pix}}
\end{subfigure}
\hfill
\begin{subfigure}{0.32\columnwidth}
\includegraphics[width=0.99\linewidth,frame]{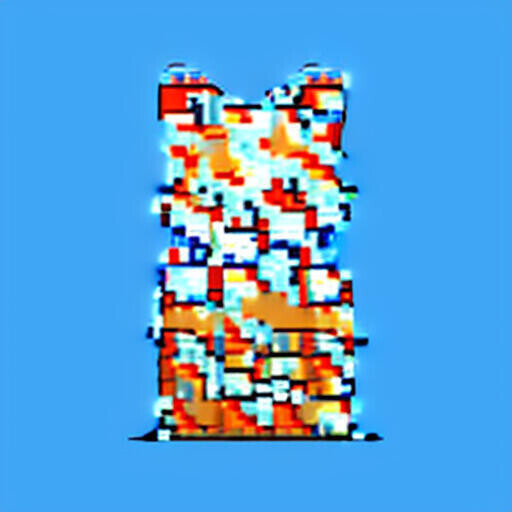}
\caption{DiffEdit~\cite{couairon2022diffedit}}
\end{subfigure}

\vspace{0.2cm}
\begin{subfigure}{0.32\columnwidth}
\includegraphics[width=0.99\linewidth,frame]{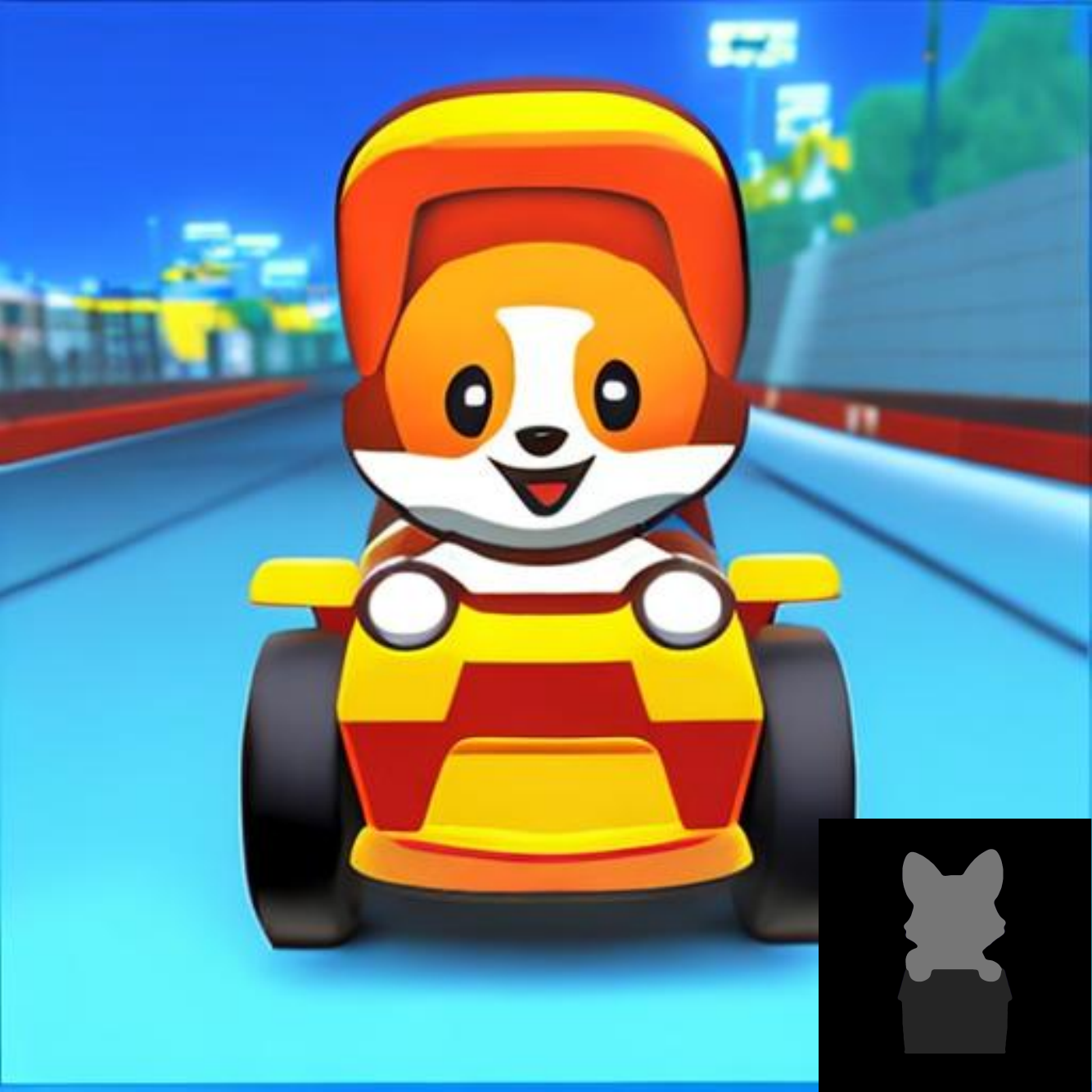}
\caption{Ours 1}
\end{subfigure}
\hfill
\begin{subfigure}{0.32\columnwidth}
\includegraphics[width=0.99\linewidth,frame]{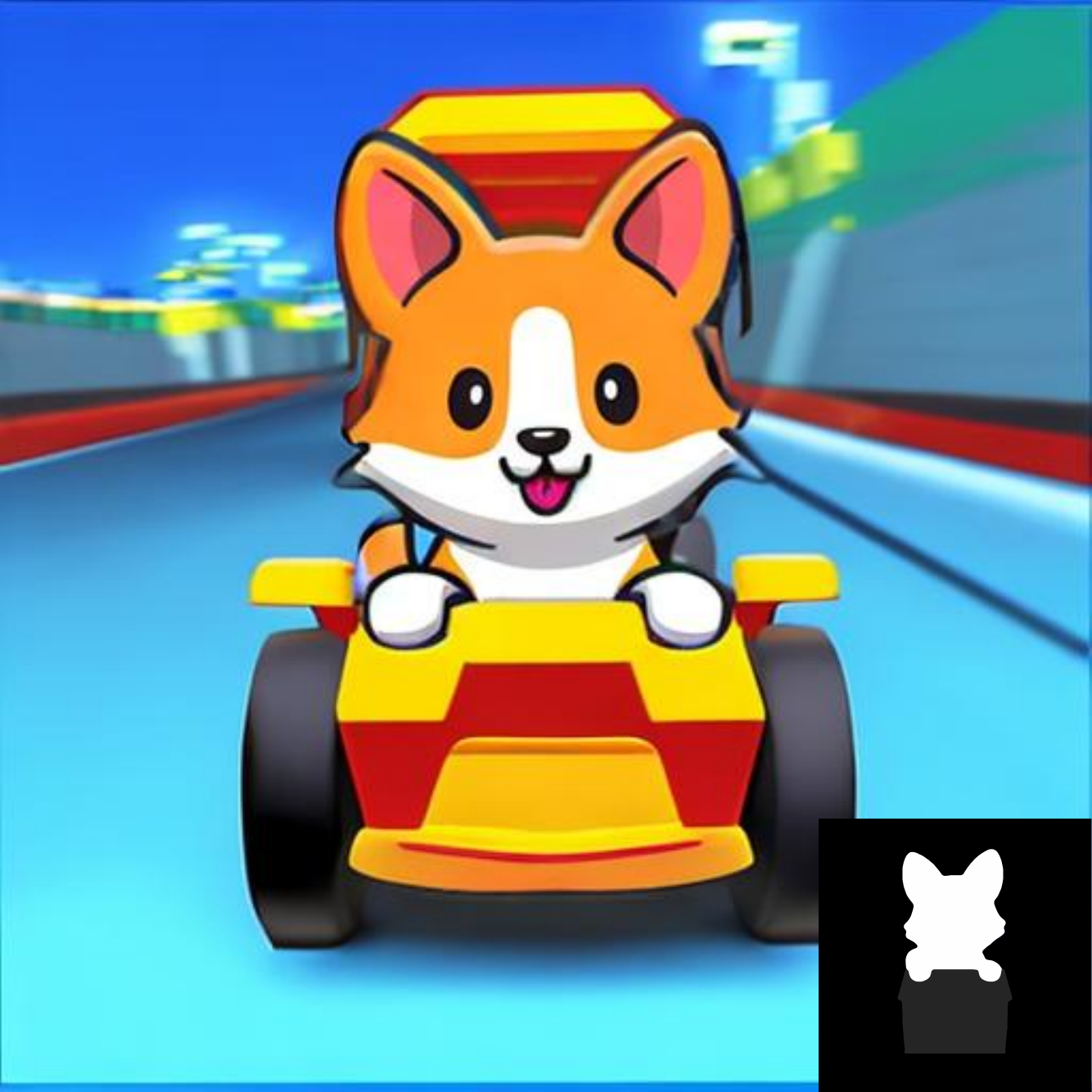}
\caption{Ours 2} 
\end{subfigure}
\hfill
\begin{subfigure}{0.32\columnwidth}
\includegraphics[width=0.99\linewidth,frame]{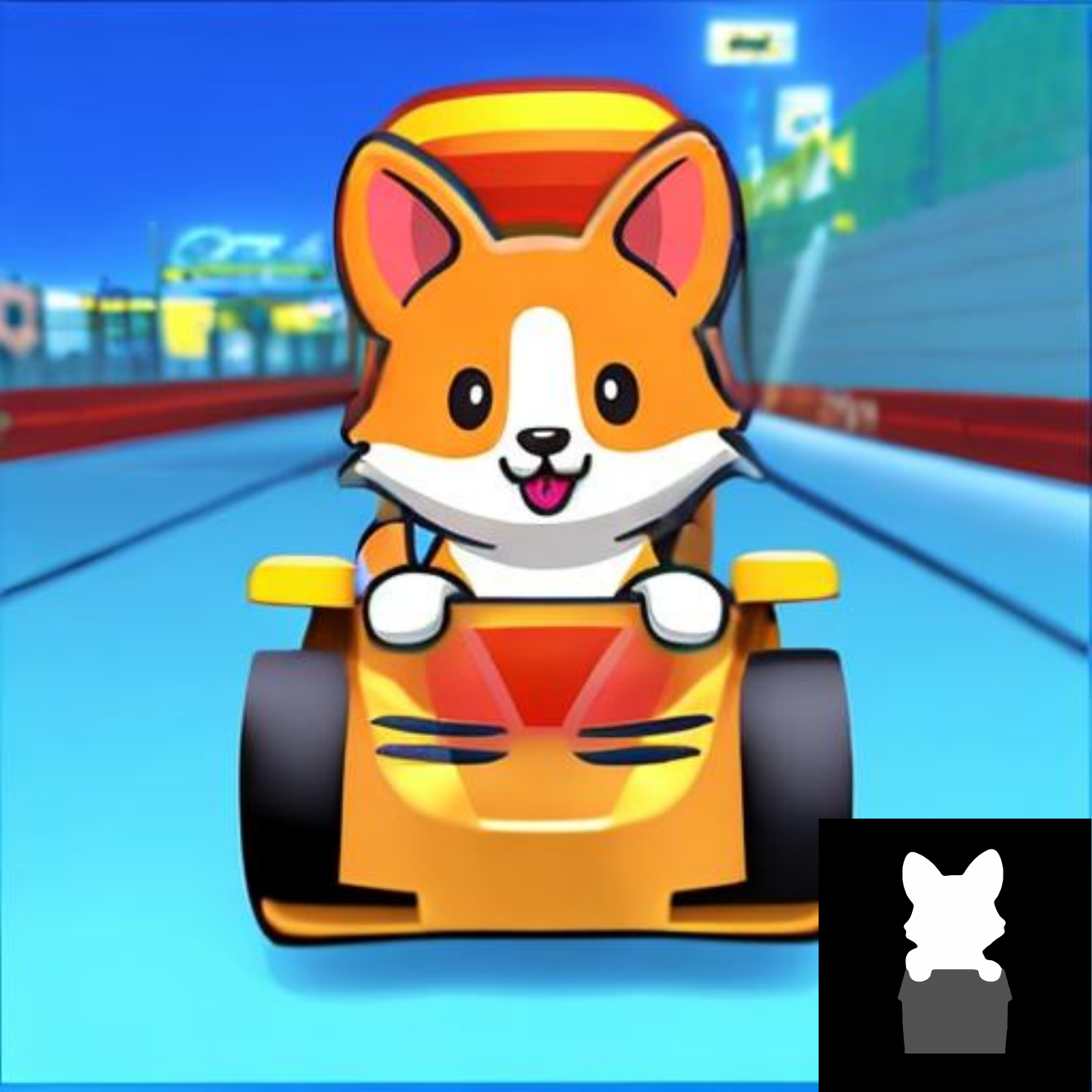}
\caption{Ours 3} 
\end{subfigure}
  \caption{\textbf{Comparison to InstructPix2Pix~\cite{brooks2023instructpix2pix} and DiffEdit~\cite{couairon2022diffedit}.} The prompts are ``[in a] race car video game'', and the same seed.  
  Our change map controllability allows the user to experiment with a variety of different edits.
The user can specify how faithful the dog will be to the original image, and how much the box and background will change. 
DiffEdit's reference prompt is “dog”; the technique mistakenly include the surrounding box and struggle to generate a high-quality result. InstructPix2Pix changes the entire image uniformly. The supplementary materials contain more examples. Change maps in bottom-right inset.
}
\label{video_game}
\end{figure}

\subsection{Memory Consumption \& Inference Time}
\label{subsec:memory}
We measured the inference memory consumption of Stable Diffusion's img2img~\cite{rombach2021highresolution} with and without our framework. The overhead of using our framework is less than 3MB (0.07\%).
In addition, we measured our model with and without skipping. For high-fidelity maps, skipping can reduce 89\% of the inference time. See the supplemental materials for more details.

\newlength{\uu}
\setlength{\uu}{0.23\columnwidth}

\begin{figure}[t]
  \centering
\begin{tabular}{cccccc}
 \textbf{Input} & \textbf{Ours} & \textbf{BLD~\cite{Avrahami_2023}} & \textbf{SD 2.1~\cite{sd2}} \\

\includegraphics[width=\uu,frame]{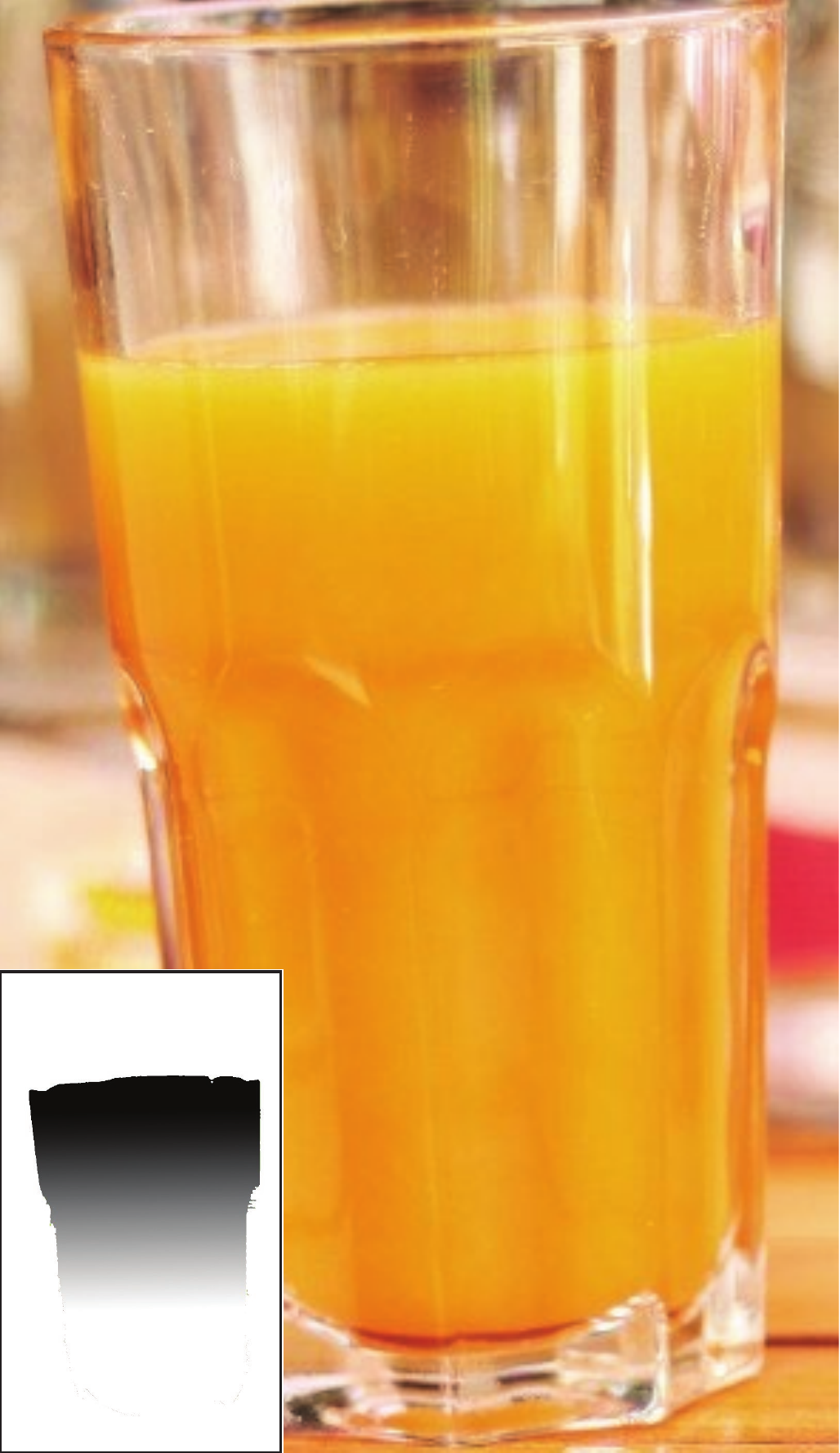}  & \includegraphics[width=\uu,frame]{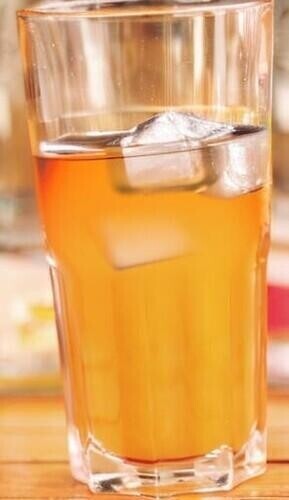} & \includegraphics[width=\uu,frame]{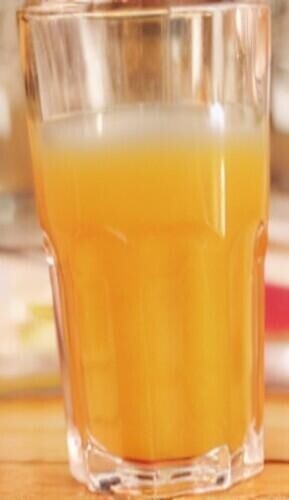} 
& \includegraphics[width=\uu,frame]{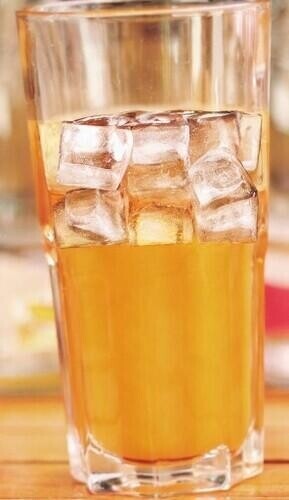} & \\
\includegraphics[width=\uu,frame]{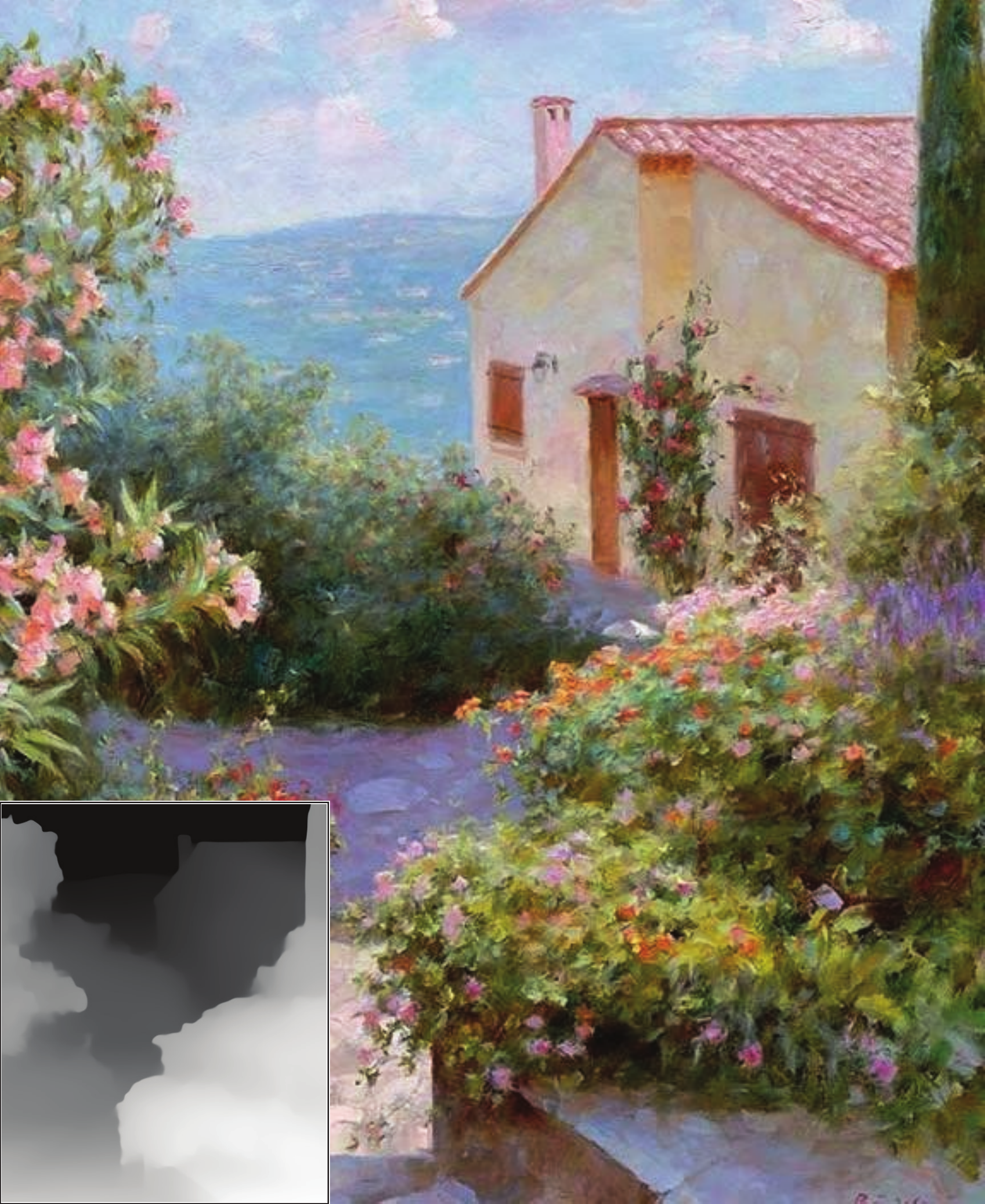}& \includegraphics[width=\uu,frame]{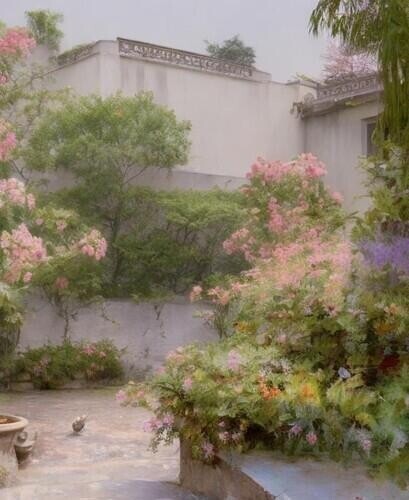} & \includegraphics[width=\uu,frame]{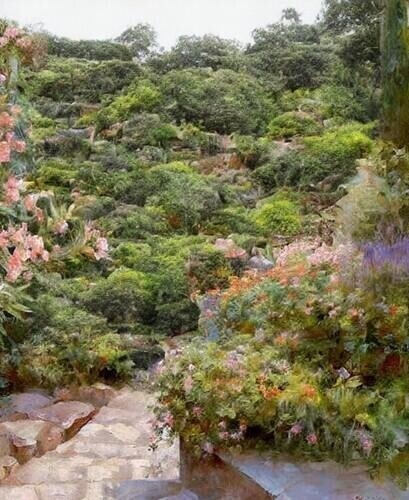} 
& \includegraphics[width=\uu,frame]{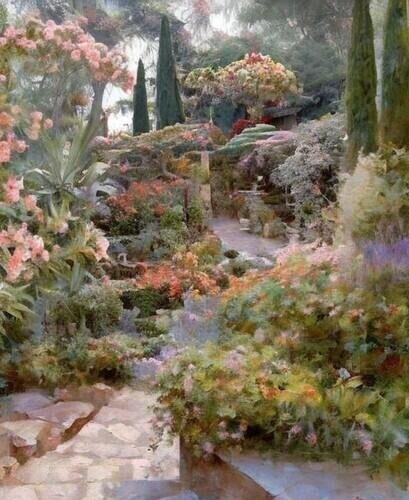} & \\
\includegraphics[width=\uu,frame]{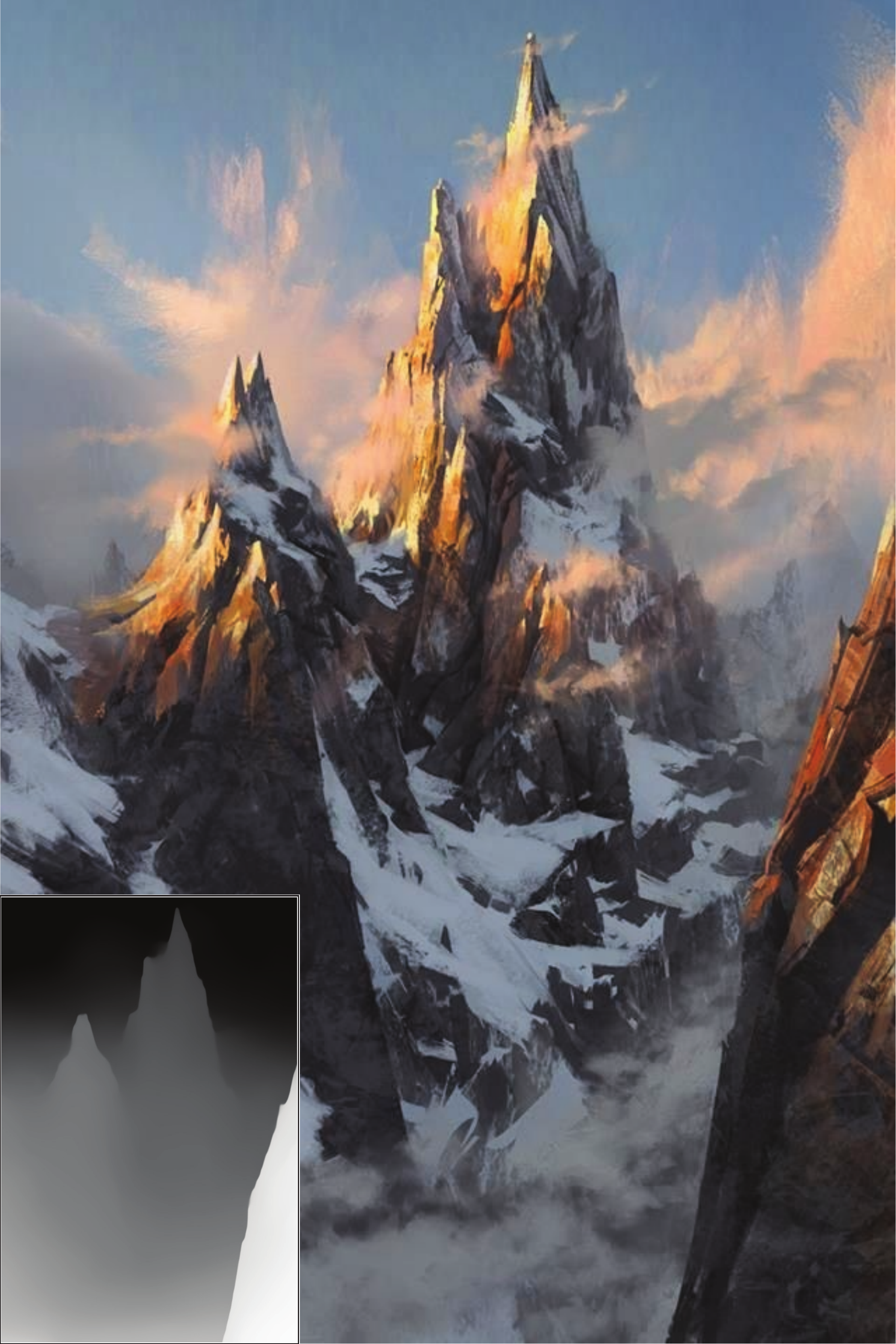} & \includegraphics[width=\uu,frame]{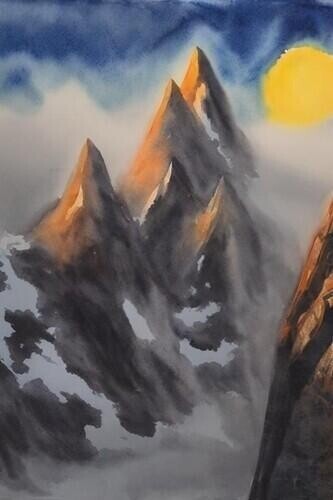} & \includegraphics[width=\uu,frame]{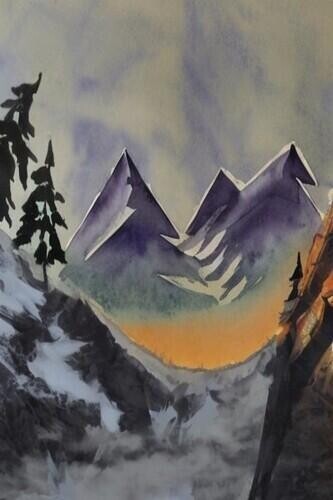} 
& \includegraphics[width=\uu,frame]{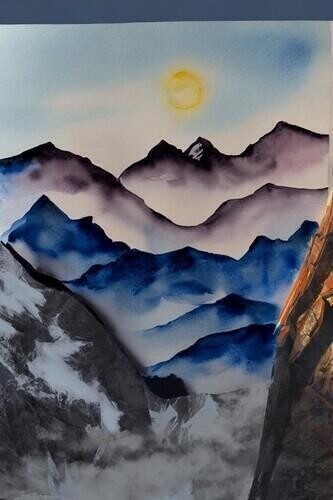} & \\
 \\
\end{tabular}
  \caption{
  \textbf{Comparison to Blended Latent Diffusion~\cite{Avrahami_2023} and Stable Diffusion 2.1's Inpaint~\cite{sd2}.} 1st row: ``glasses with ice cubes'', all other methods fail to blend the ice with the juice, as such a blend should be gradual.
  2nd row: ``Photograph titled Villa Garden by the photographer Bi Wei Liang Tronolone''. Because masks contain only two distinct values, it is not possible to differ the rear, the front, and the house. Therefore, the house is absent in the other methods.
  3rd row: ``Night court mountains, watercolor painting'', to achieve the watercolor style, a high strength is required for the skies. For other methods this destroys the structure of the mountain completely, and the intended similarity between the edited picture to the original is lost. Change maps are presented in the bottom-left inset.
 }
  \label{comparison_gallery}
\end{figure}

\section{Limitations and Future Work}
One limitation of our method is in users' ability to anticipate the result of a given edit.
Some users already struggle to predict the effects of strengths in existing diffusion models, 
and our method further complicates the input-output relation by
allowing users to select multiple values for each map. While this is far from solved, we alleviate the issue by introducing our strength fans.

In the future, our algorithm can be further optimized by calculating all $z'_t$s and mask decompositions in advance and in parallel, as they are independent of the diffusion process.
Another research direction is on different methods to automatically produce change maps.
In this paper, the user creates the change map, or uses depth or segmentation maps. We believe that other algorithms for map generation can extend our framework to solve various editing tasks.
Lastly, we have shown Strength Fan as a tool for exploring the editing space. We believe that many other tools can be developed by using different map patterns and with automatically generated change maps.
\newcommand{\hh}{0.08\textwidth}

\begin{figure}[t]
  \centering
  \renewcommand{\arraystretch}{1}
  
  \begin{tabular}{ccccc}
    & \textbf{Pattern} & \textbf{Ours} & \textbf{BLD~\cite{Avrahami_2023}} & \textbf{SD2.1~\cite{sd2}} \\

    \raisebox{0.10cm}{\rotatebox{90}{Gradient}} &
    \includegraphics[width=\hh,frame]{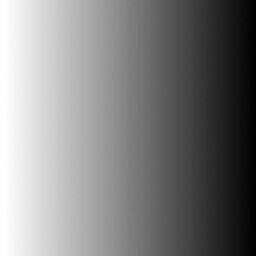} &
    \includegraphics[width=\hh,frame]{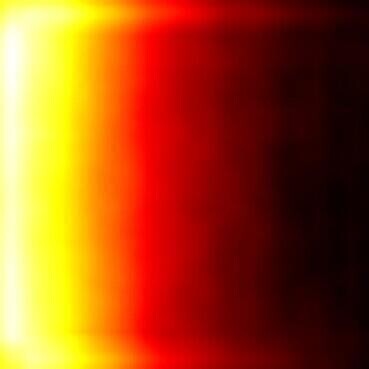} &
    \includegraphics[width=\hh,frame]{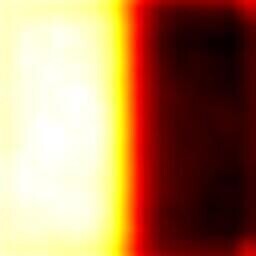} &
    \includegraphics[width=\hh,frame]{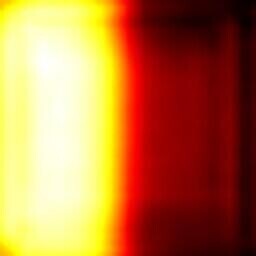} \\

    \raisebox{0.22cm}{\rotatebox{90}{Shapes}} &
    \includegraphics[width=\hh,frame]{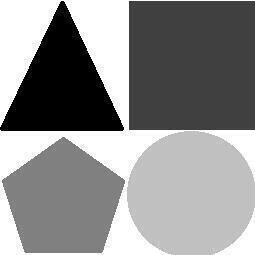} &
    \includegraphics[width=\hh,frame]{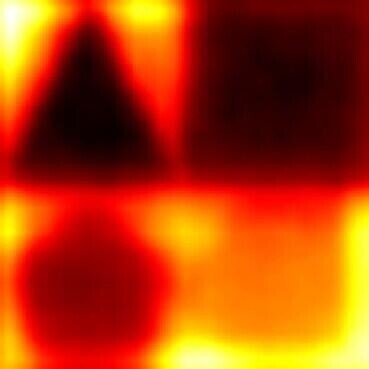} &
    \includegraphics[width=\hh,frame]{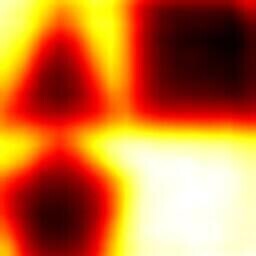} &
    \includegraphics[width=\hh,frame]{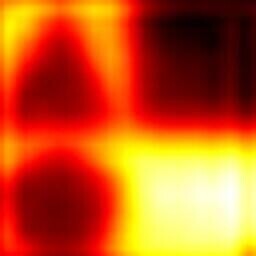} \\

    \raisebox{0.01cm}{\rotatebox{90}{Triangles}} &
    \includegraphics[width=\hh,frame]{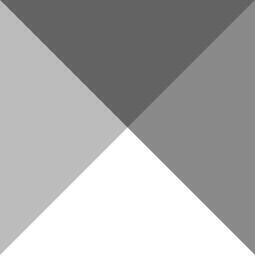} &
    \includegraphics[width=\hh,frame]{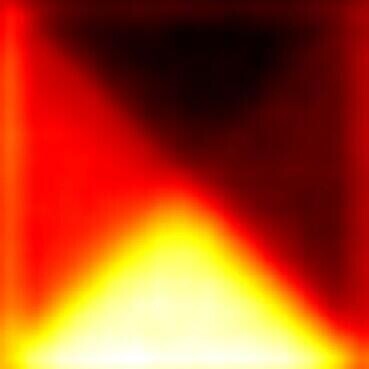} &
    \includegraphics[width=\hh,frame]{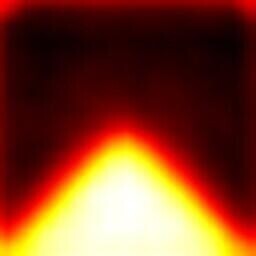} &
    \includegraphics[width=\hh,frame]{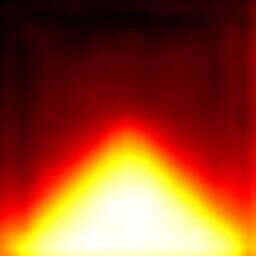} \\
    
  \end{tabular}
\caption{\textbf{Edit strength measurement maps.} We evaluate our method, Blended Latent Diffusion~\cite{Avrahami_2023}, and Stable Diffusion 2.1's inpaint~\cite{sd2}. Comparing the measured map to the original, we see that the other methods are able to express only two strengths, even when the original change map contains more.}
\label{fig:reconstruction}
\end{figure}

\section{Conclusion}
In this paper, we presented a new framework that enables region-wise control over the strength of the image-to-image translation process. Our framework does not require optimization or training, has minimal overhead, and can be used with diverse images and prompts.
Our work expands the scalar strength parameter into a more flexible 2D array, and we hope other researchers will advance this further.

\section*{Acknowledgments}
We thank Almog Friedlander, Yuval Ran-Milo, Shahar Sarfaty and Eyal Michaeli for proofreading the paper.
This work was supported in part by the Israel Science Foundation
(grant No.\ 1574/21), and by the Tel Aviv University Center for AI and Data Science (TAD).

{
    \small
    \bibliographystyle{ieeenat_fullname}
    \bibliography{ms}
}
\clearpage

\maketitlesupplementary
We present additional results (\Cref{fig:str_fan_general,fig:monks,fig:peacock,orange-cubes-seed,venice_inverse,fig:eagle-president,fig:alternatives_fire,single-reconstruction,lego_airplane}), extend the discussion on experiments from the main paper~(\Cref{fig:user-study-screenshots,fig:runtime-min-value}), provide additional experiments that show the usability of our framework (\Cref{cam-dam-couterexample}), and specify additional technical details for reproducibility (\Cref{prompt}).

We provide further details on the ``skipping'' optimization in \Cref{skipping-running-time}.
We provide an extended discussion on strength fans in \Cref{sec:str_fan}.
We provide additional comparisons for soft-inpainting in \Cref{fig:monks}. 
In figure 10 in the main paper we show a comparison between our method and other mask-based methods. 
In \Cref{orange-cubes-seed} we show that these results have not been caused by outlier seeds. We show that for different seeds, our method produces better results. 
 We show more examples of the control that change maps introduce in \Cref{tapestry_horses,fig:eagle-president,venice_inverse}.

In Section 5.1 of the main paper we discuss the limitations of traversing multiple times across the latent space. We show the degradation in the image quality in \Cref{latent-deg-large}.
We provide additional comparisons to the baselines described in this section in \Cref{fig:alternatives_fire}.
In Section 5.2 of the main paper we use 1,000 pairs of images to measure the applied change map. We show an ablation to this amplification in \Cref{single-reconstruction}, where we compare it to a single pair measurement, and also show the spatial bias of the LPIPS similarity maps~\cite{zhang2018unreasonable}.

We show that our algorithm is compatible with many samplers in \Cref{samplers}.
We show more examples of edits in \Cref{lego_airplane}.
In the main paper, we have outlined the structure of the user study; in \Cref{fig:user-study-screenshots} we provide representative questions from the study.

\section{Running Time of Skipping}
\label{skipping-running-time}
We measured our model with and without skipping, using gradient maps with different minimum values. We average results over 100 different inference processes. 
In the main paper, we detail the modification of the original algorithm to integrate the skipping optimization.
We write the explicit skipping algorithm in \Cref{s_algo-with-skipping}, and we visualize the runtime as a function of the minimum value in the change map in~\Cref{fig:runtime-min-value}. 
A clear linear correlation exists between these two variables.
Our measurements (\Cref{tbl:skipping-runtime}) demonstrate the utility of the skipping optimization, by saving up to 89 percent of the running time.

\section{Strength Fan}
\label{sec:str_fan}
The main paper demonstrates the ability of our framework to create a “strength fan”, a tool to estimate the effects of different strength values for given prompts and seeds. We believe that with the adoption of the tool, tuning the strength parameter when using diffusion models will become more prevalent. We envision that this tool will be integrated into existing software solutions, where users will specify the number of strengths they wish to explore simultaneously. 
We show several strength fans in \Cref{fig:str_fan_general}. 
Some users may opt to iteratively tune the strength by using fans with different magnitudes and offsets, as we demonstrate in \Cref{fig:str_fan_tuning}.

\section{CAM versus DAM properties}
In the main paper, CAM and DAM consistently ranked methods in the same order, leading to the tempting assumption that the metrics give similar information. However, this assumption is incorrect. In~\Cref{cam-dam-couterexample}  we present a counterexample: comparing map 1 (the circle) to the two other maps using CAM and DAM. As observed, while CAM indicates a stronger similarity to the halved circle, DAM indicates a higher similarity with the inverted circle.  
In the case of the inverted map, CAM yields the minimal similarity score possible, while DAM provides the maximal similarity score. %
The differences between the metrics are also notable when comparing the triangle map to the shaded circle one. By just changing the shape, the CAM drops to almost 0, while DAM retains a similar score. 

\section{Extensions For Various Diffusion Models}
\label{extension-for-different-diffusion-models}
In the main paper, we mostly present our framework applied to Stable Diffusion 2.1~\cite{rombach2022highresolution}.
We generalized our framework to other diffusion models.
\subsection{Stable Diffusion XL}
The latest version of Stable Diffusion, XL, introduces several modifications, most of which are unrelated to our method. However, it offers a mode of ensemble of experts for inferring high-quality images. In this ensemble, a “base model” is responsible for handling higher timesteps, while a “refiner model” takes care of the rest. The split ratio $s$ is a user-defined parameter. Therefore, our inference process is similarly split. 
In our algorithm, for $t$ smaller than $sk$, the denoising (in line 10) is performed by the base model. Otherwise, denoising is conducted by the refiner.

\subsection{Kandinsky}
Kandinsky, despite its differences from Stable Diffusion's models, seamlessly fits into our algorithm without requiring any special adaptations.

\subsection{DeepFloyd IF}
We adapt DeepFloyd's~\cite{DeepFloydIF} cascading paradigm, by applying our algorithm for the first two (of three) stages of inference.
We suggest preserving the original inference process for the last stage, as this stage primarily handles super-resolution.

\section{Automatic Edits Creation}
In the main paper we mostly assume that the change map inputs to the framework are handily chosen and crafted by the users. In this section, we show that some procedures can be employed to automate those choices. In~\Cref{sec:auto-change-maps} we use the input images and the prompts from the InstructPix2Pix dataset~\cite{brooks2023instructpix2pix}. For change maps, we first explore three simple fixed change maps. Surprisingly, even simple patterns seem to produce interesting results.
We then explore depth maps generated by MiDaS~\cite{Ranftl2022} with no further intervention in~\Cref{midas-no-intervention}.
The continuous nature of these change maps produces successful and unique edits, despite being chosen arbitrarily.

\newcommand{\ii}{0.1055\textwidth}

\begin{figure*}[t]
\centering
\begin{tabular}{ccccccccccc}
\textbf{\footnotesize{Input Image}} & 
\textbf{\footnotesize{Inpaint Mask}} & 
\textbf{\footnotesize{No Softening}} & 
\textbf{\footnotesize{Blurred Mask}} & 
\textbf{\footnotesize{\textalpha-Compose}} & 
\textbf{\footnotesize{Poisson}} & 
\textbf{\footnotesize{Laplace}} & 
\textbf{\footnotesize{Standard Soft}} & 
\textbf{\footnotesize{Ours}} \\
 
\includegraphics[width=\ii,frame]{imgs/comparison/monks/original_image.jpg} &
\includegraphics[width=\ii,frame]{imgs/comparison/monks/mask_monks_original.jpg} &
\includegraphics[width=\ii,frame]{imgs/comparison/monks/sharp_mask.jpg} &
\includegraphics[width=\ii,frame]{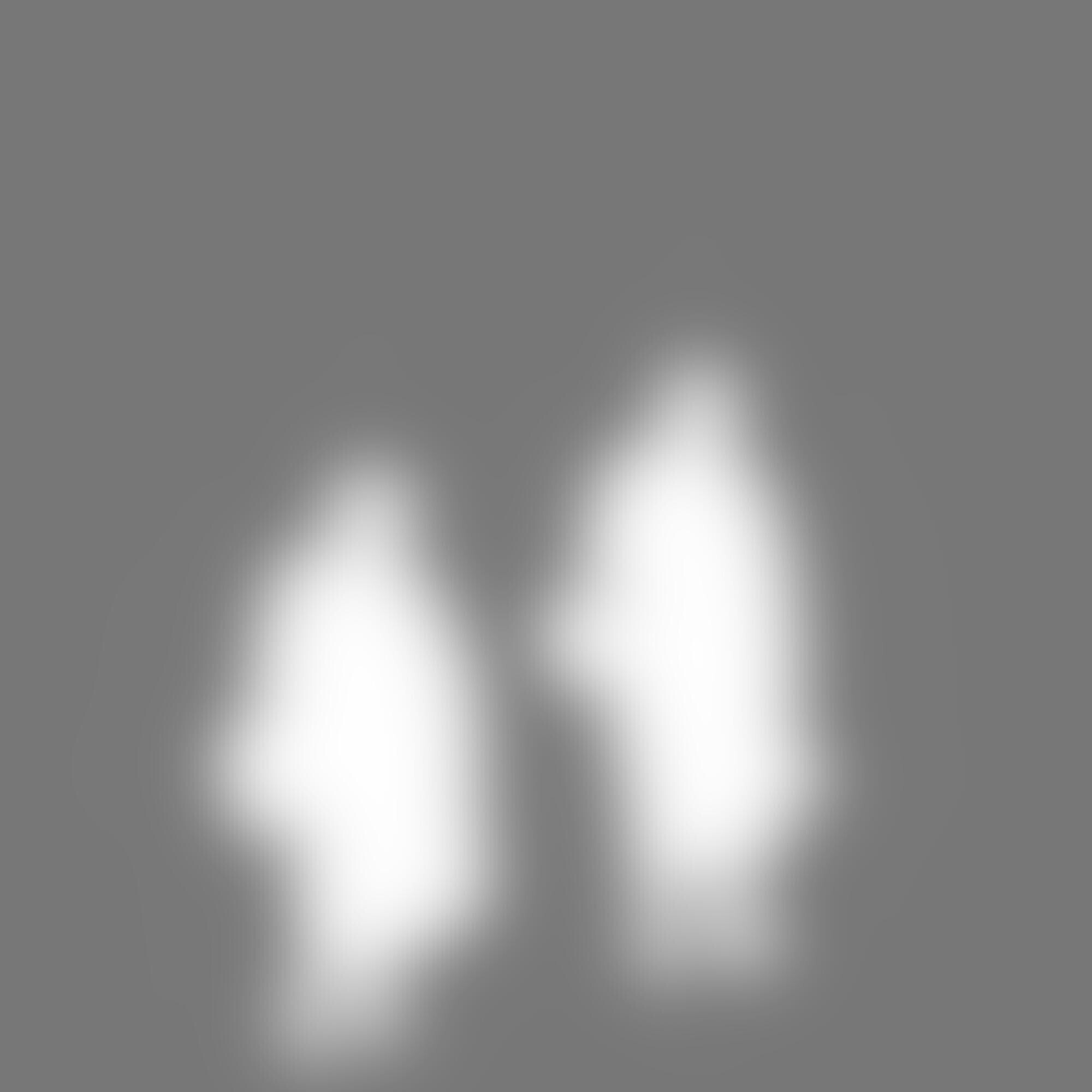} &  
\includegraphics[width=\ii,frame]{imgs/comparison/monks/monk_alpha_blend.jpg} &
\includegraphics[width=\ii,frame]{imgs/comparison/monks/poission.jpg} &
\includegraphics[width=\ii,frame]{imgs/comparison/monks/monk_laplace.jpg} &
\includegraphics[width=\ii,frame]{imgs/comparison/monks/blured_mask.jpg} &
\includegraphics[width=\ii,frame]{imgs/comparison/monks/augmented_39_of_monks_blur.jpg} \\
\includegraphics[width=\ii,frame]{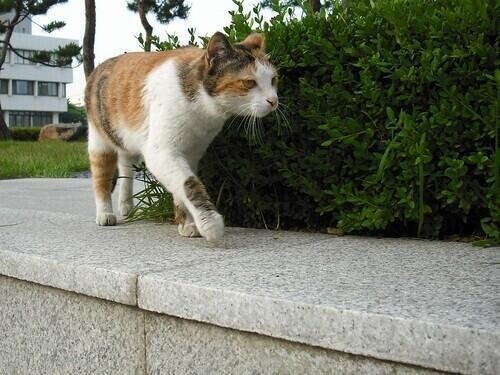} &
\includegraphics[width=\ii,frame]{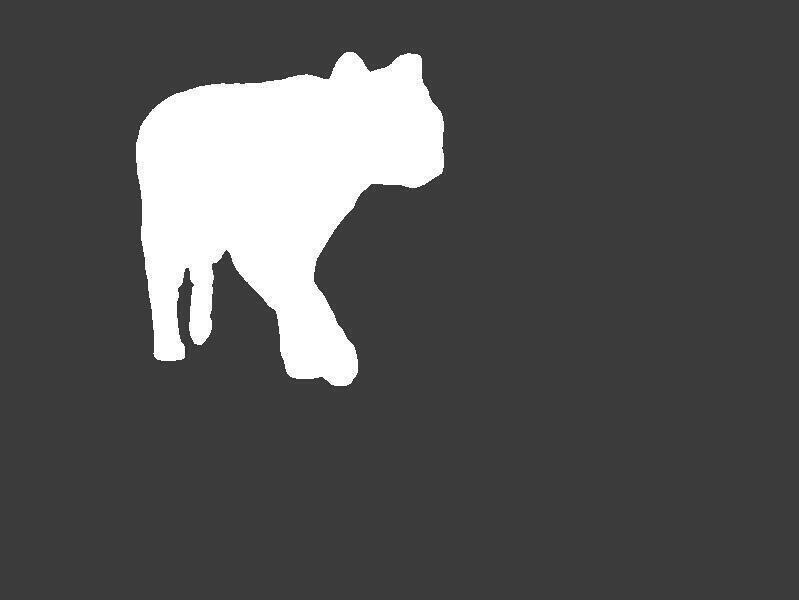} &
\includegraphics[width=\ii,frame]{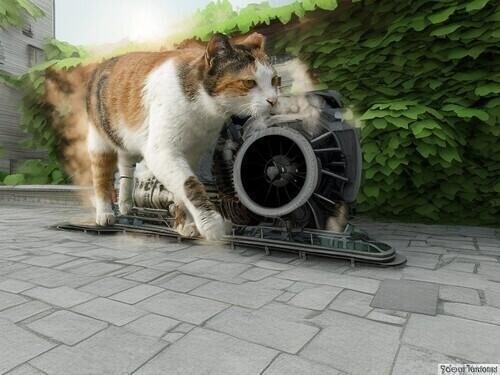} &
\includegraphics[width=\ii,frame]{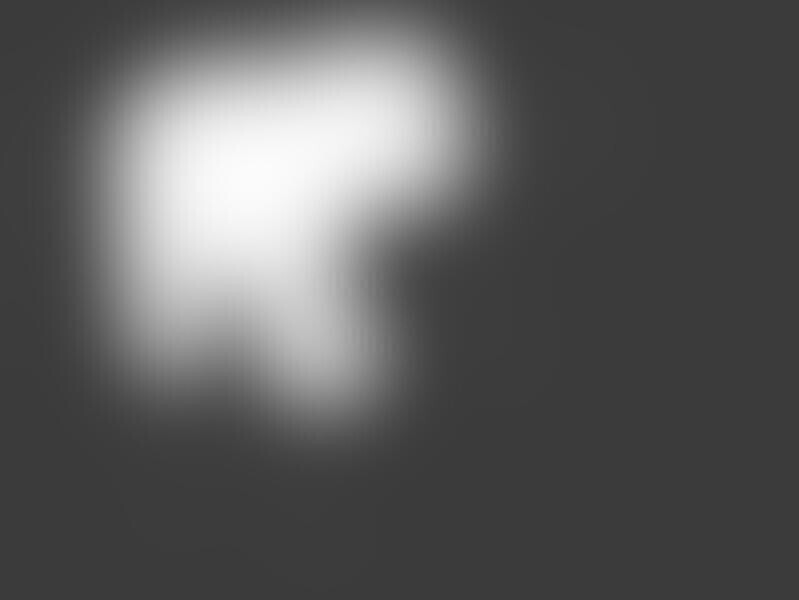} & 
\includegraphics[width=\ii,frame]{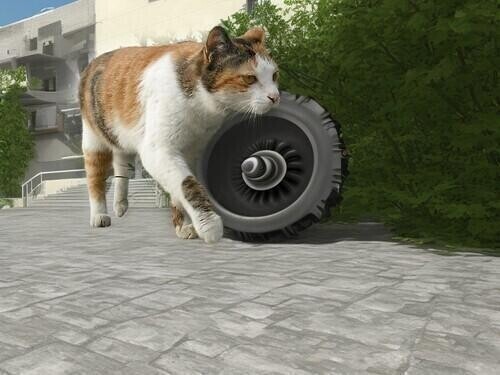} &
\includegraphics[width=\ii,frame]{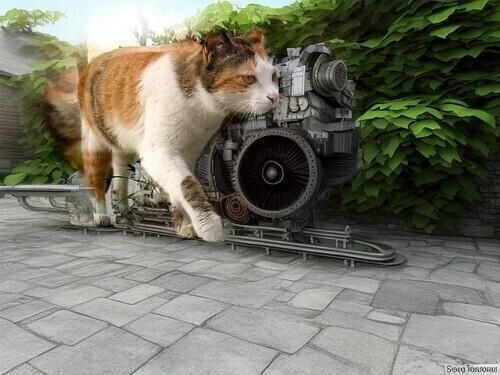} &
\includegraphics[width=\ii,frame]{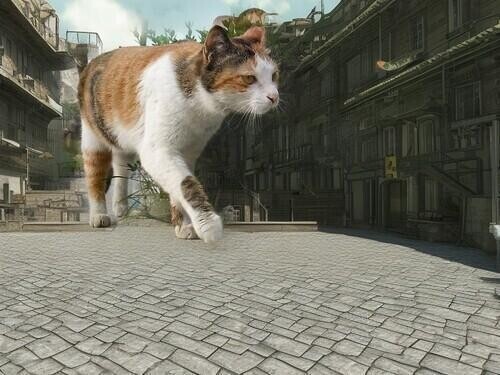} &
\includegraphics[width=\ii,frame]{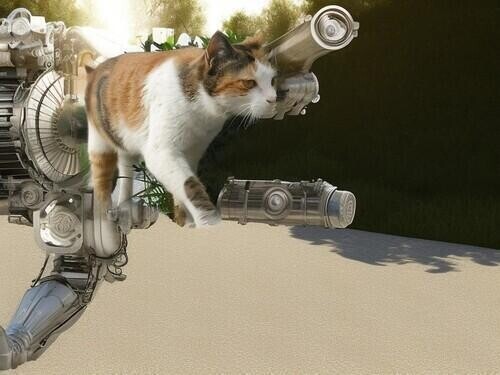} &
\includegraphics[width=\ii,frame]{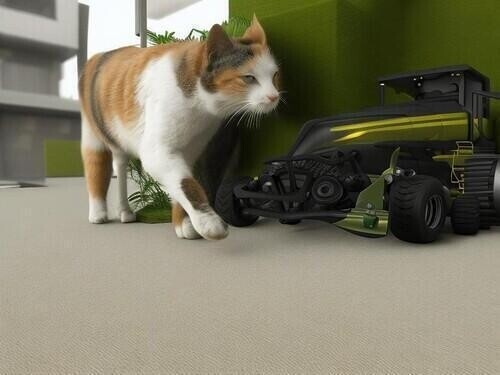} \\
\includegraphics[width=\ii,frame]{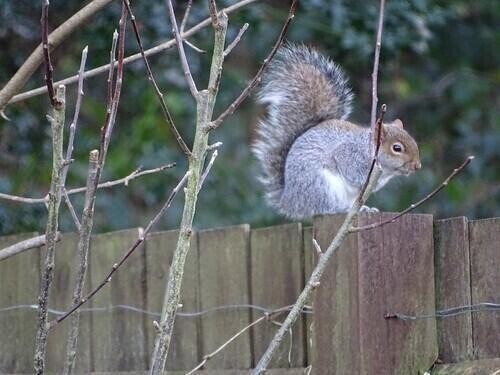} &
\includegraphics[width=\ii,frame]{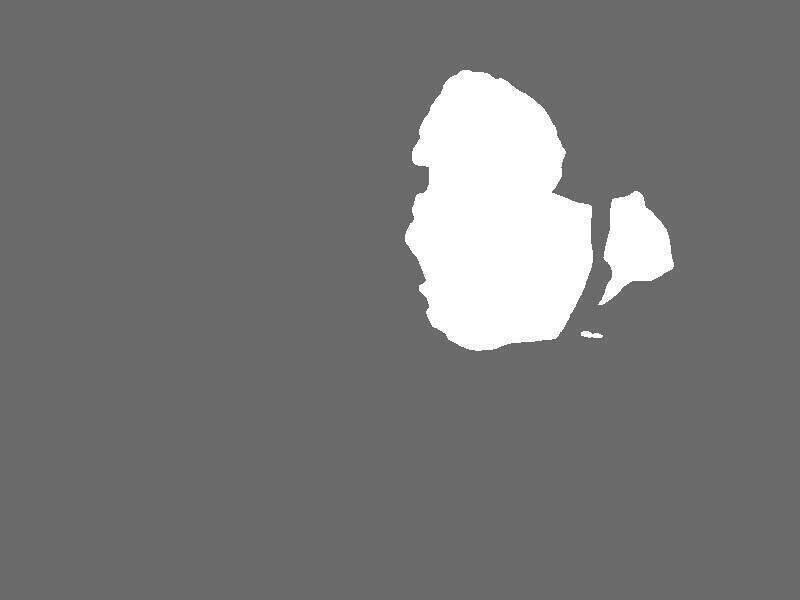} &
\includegraphics[width=\ii,frame]{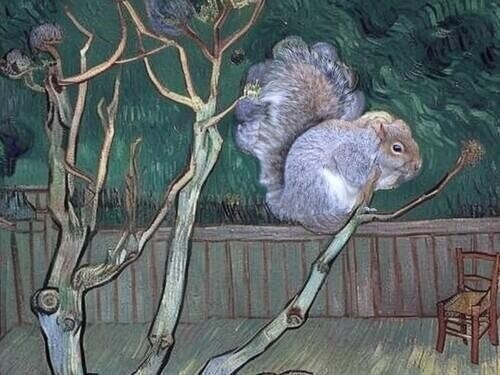} &
\includegraphics[width=\ii,frame]{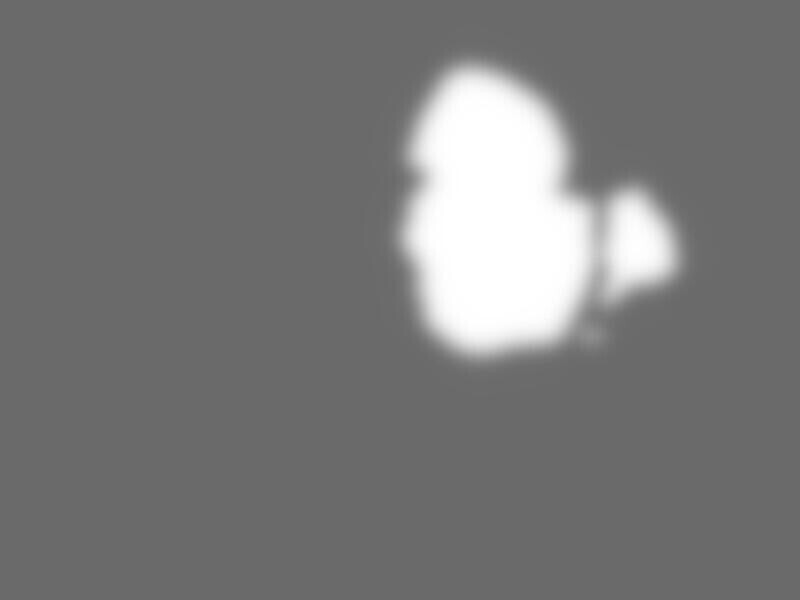} &
\includegraphics[width=\ii,frame]{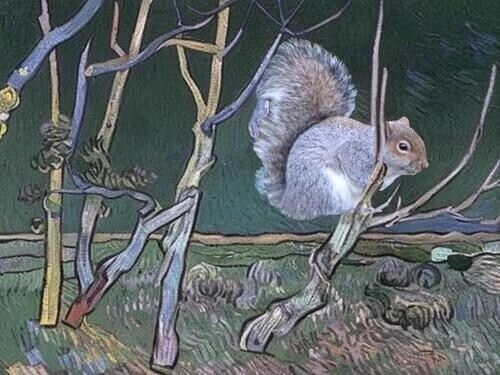} &
\includegraphics[width=\ii,frame]{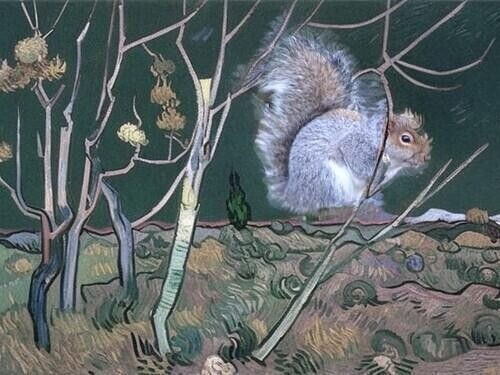} &
\includegraphics[width=\ii,frame]{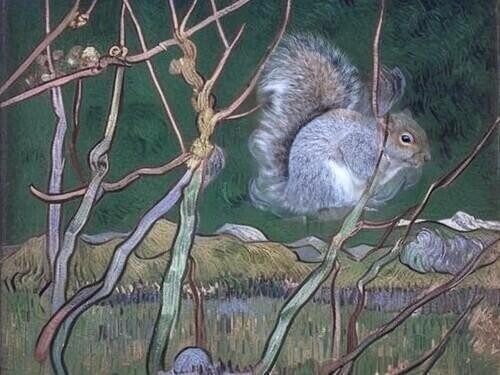} &
\includegraphics[width=\ii,frame]{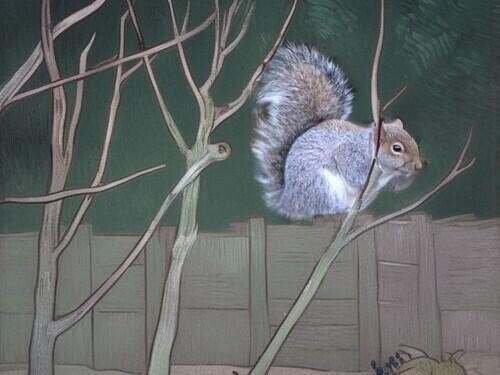} &
\includegraphics[width=\ii,frame]{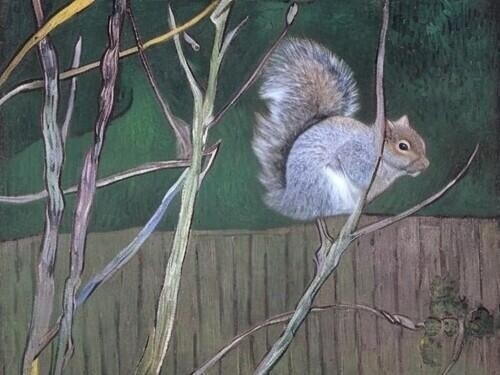} \\
\end{tabular}
\caption{Extended version of Figure 6 from the main paper.  
We compare our approach to no softening, \textalpha-compositing, Poission-based~\cite{10.1145/882262.882269} and Laplace-based~\cite{10.1145/245.247} compositing, and standard soft-inpainting (as implemented in Stable Diffusion web UI~\cite{AUTOMATIC1111_Stable_Diffusion_Web_2022}). 
For \textalpha-compositing, Poisson-based and Laplace-based methods, we blend the original image with a regular inpaint result using a Gaussian blurred version of the inpaint mask. 
Our method produces a much more natural blend.
The prompts are  ``Impressionist'',  ``Real Engine'', ``Van Gogh''.
The blurring radii are: 64px, 40px, 10px.}

\label{fig:monks}
\end{figure*}

    \newlength{\dd}
    \setlength{\dd}{2.8cm}
    \newlength{\ee}
    \setlength{\ee}{1.71428571\dd}
    
\begin{figure*}
    \centering
    \begin{tabular}{cccccc}

    Input Image & Change Map & \multicolumn{2}{c}{Ours with different seeds} & \multicolumn{2}{c}{BLD~\cite{Avrahami_2023} with different seeds} \\[0.1cm]
    
    \includegraphics[height=\ee,width=\dd,frame]{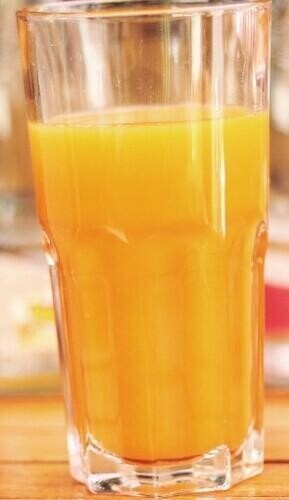} &
    \includegraphics[height=\ee,width=\dd,frame]{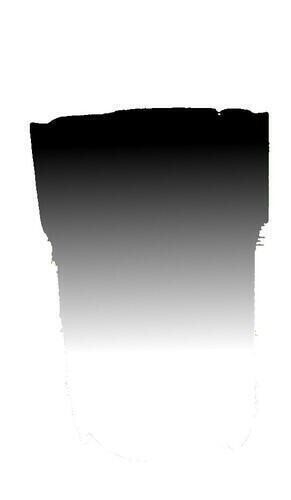} &
    \includegraphics[height=\ee,width=\dd,frame]{imgs/gallery/orange_cubes/augmented_39_of_23699976063_d423553477_k_B.jpg} &

    \includegraphics[height=\ee,width=\dd,frame]{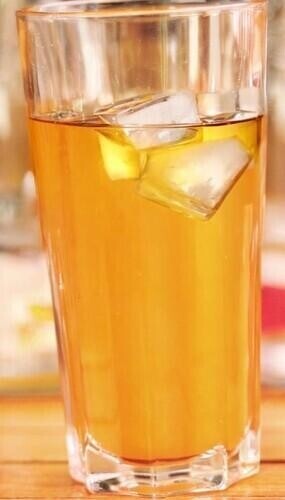} &
    \includegraphics[height=\ee,width=\dd,frame]{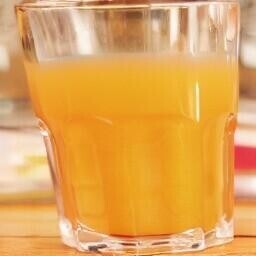} &
    \includegraphics[height=\ee,width=\dd,frame]{imgs/gallery/orange_cubes/omri_0002.jpg} \\[0.4cm]

    \multicolumn{6}{c}{Stable Diffusion 2's Text-Guided Inpainting~\cite{sd2} with different seeds} \\[0.1cm]

    \includegraphics[height=\ee,width=\dd,frame,valign=b]{imgs/gallery/orange_cubes/inpaint/inpaint_augmented_49_of_23699976063_d423553477_k_B.jpg} &
    \includegraphics[height=\ee,width=\dd,frame,valign=b]{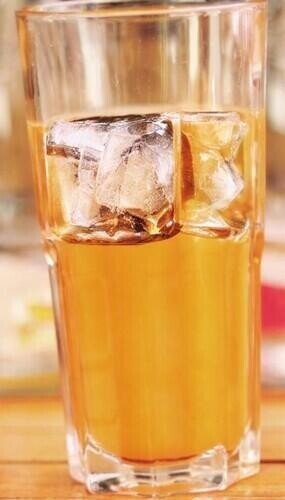} &
    \includegraphics[height=\ee,width=\dd,frame,valign=b]{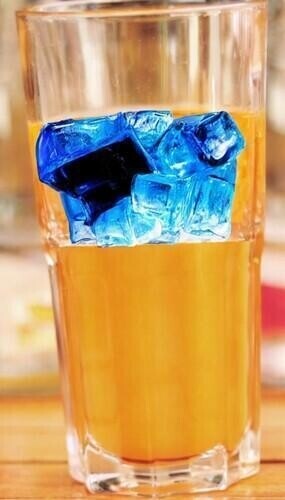} &
    \includegraphics[height=\ee,width=\dd,frame,valign=b]{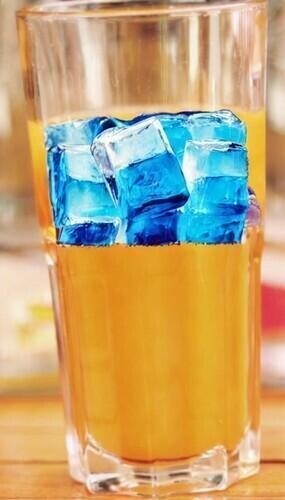} &
    \includegraphics[height=\ee,width=\dd,frame,valign=b]{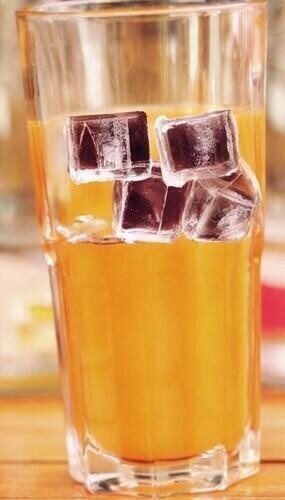} &
    \includegraphics[height=\ee,width=\dd,frame,valign=b]{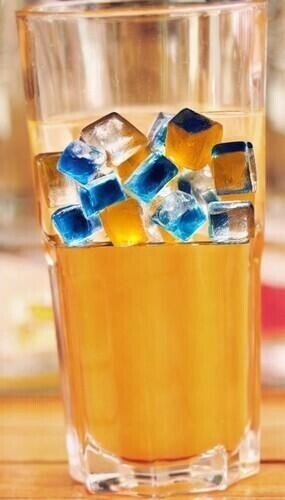} \\[0.2cm]
  
  \end{tabular}
  \caption{\textbf{Comparison to other methods with different seeds.} The prompt for all results is ``glass with ice cubes''. Blended Latent Diffusion fails to generate ice-cubes, and Stable Diffusion cannot properly blend between the added cubes and the juice.}
\label{orange-cubes-seed}
\end{figure*}

\begin{figure*}
  \centering
  \begin{tabular}{c@{\hskip 0.3cm}cc@{\hskip 0.3cm}cc}
  \textbf{Input Image} & \textbf{Change Map} & \textbf{Output} & \textbf{Change Map}& \textbf{Output}\\[0.1cm]
    \includegraphics[width=0.189\textwidth,frame]{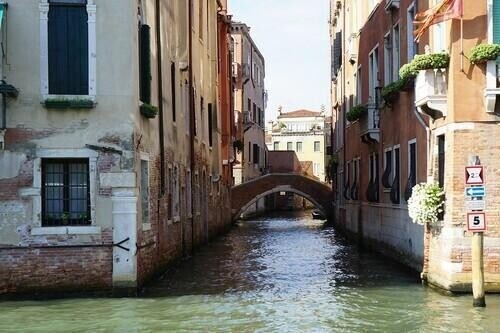} &
    \includegraphics[width=0.189\textwidth,frame]{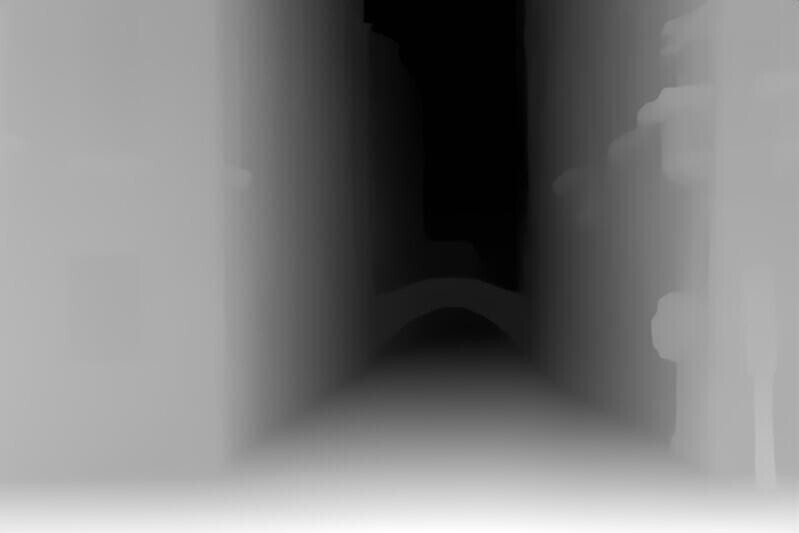} &
    \includegraphics[width=0.189\textwidth,frame]{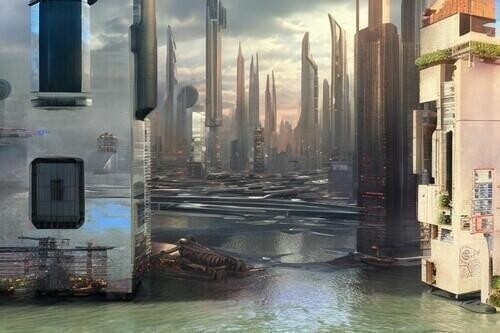} &
    \includegraphics[width=0.189\textwidth,frame]{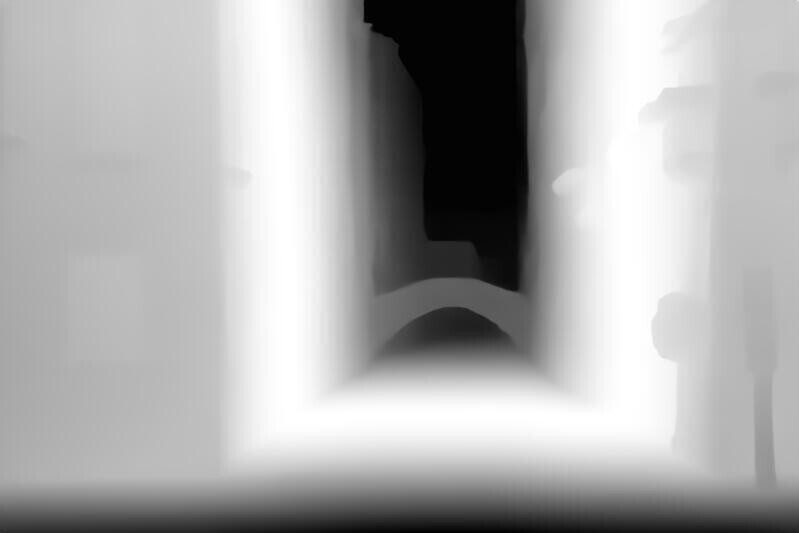} &
    \includegraphics[width=0.189\textwidth,frame]{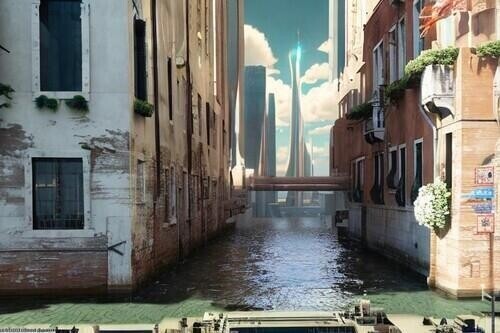} \\[0.2cm]
    &
    \includegraphics[width=0.189\textwidth,frame]{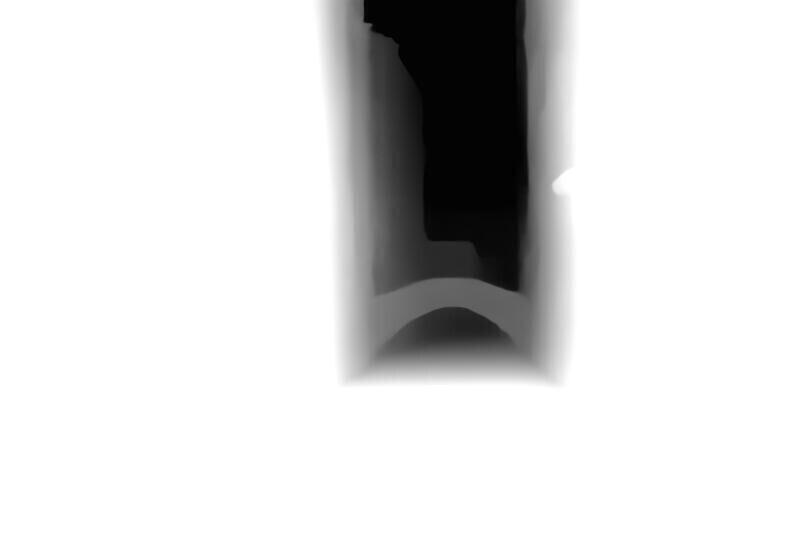} &
    \includegraphics[width=0.189\textwidth,frame]{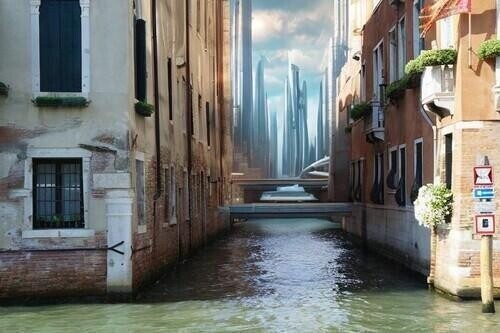} &
    \includegraphics[width=0.189\textwidth,frame]{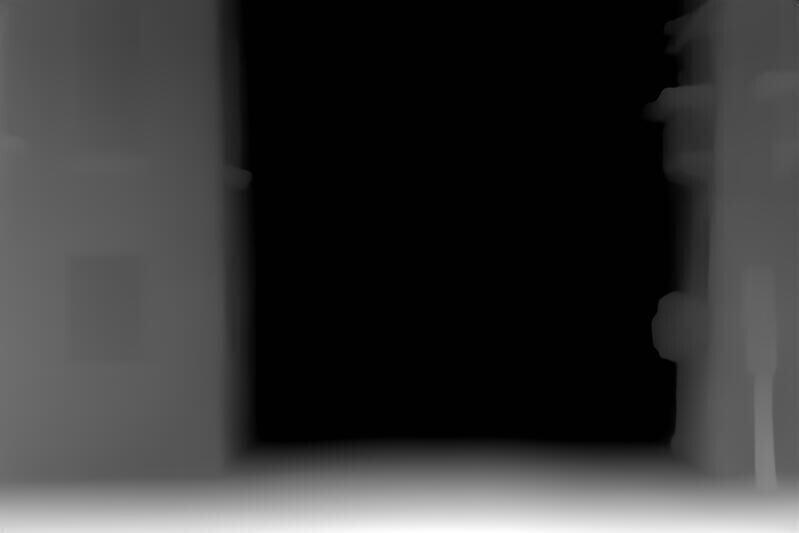} &
    \includegraphics[width=0.189\textwidth,frame]{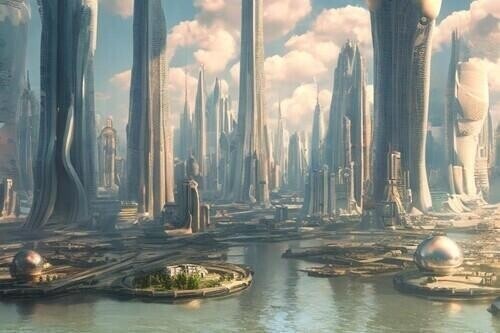} 
  \end{tabular}
  \caption{\textbf{Various change maps created from the same depth map.}
  As described in the main paper, a MiDaS~\cite{Ranftl2022,Ranftl2021} depth map can be used to create many different change maps via simple image transformations. Prompt: ``a futuristic city with tall buildings and a lot of traffic in the foreground and a cloudy sky in the background, a detailed matte painting, afrofuturism, Beeple, cyberpunk city''.}
  \label{venice_inverse}
\end{figure*}

\begin{figure*}[t]
\captionsetup[subfigure]{labelformat=empty}

  \centering
\begin{subfigure}{0.24\textwidth}
\includegraphics[width=\linewidth,frame]{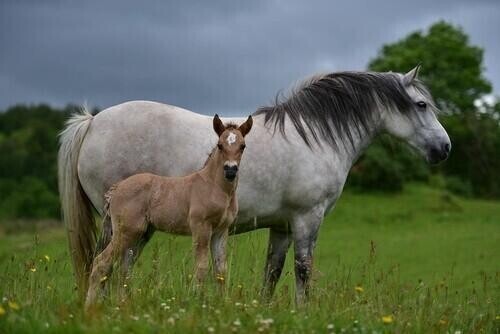}
\caption{Input Image}
\end{subfigure}
\vspace{0.5em}

\begin{subfigure}{0.24\textwidth}
\includegraphics[width=\linewidth,frame]{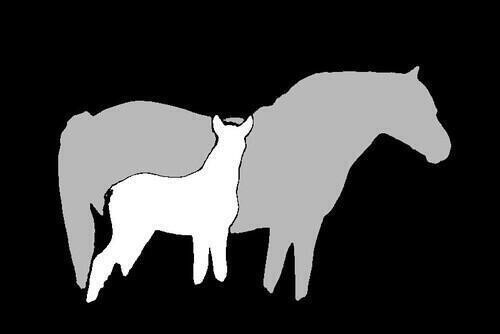}
\caption{Change Map 1}
\end{subfigure}
\begin{subfigure}{0.24\textwidth}
\includegraphics[width=\linewidth,frame]{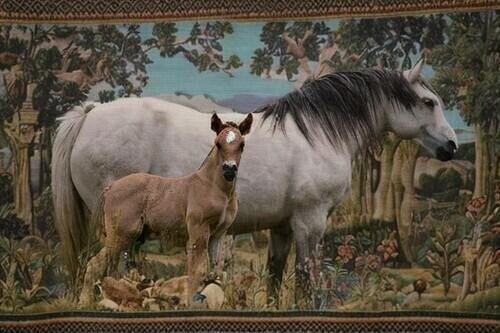}
\caption{Output 1}
\end{subfigure}
\hfill
\begin{subfigure}{0.24\textwidth}
\includegraphics[width=\linewidth,frame]{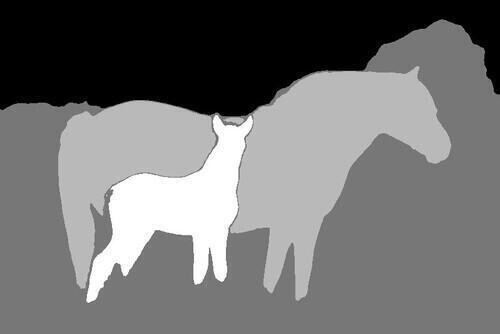}
\caption{Change Map 2}
\end{subfigure}
\begin{subfigure}{0.24\textwidth}
\includegraphics[width=\linewidth,frame]{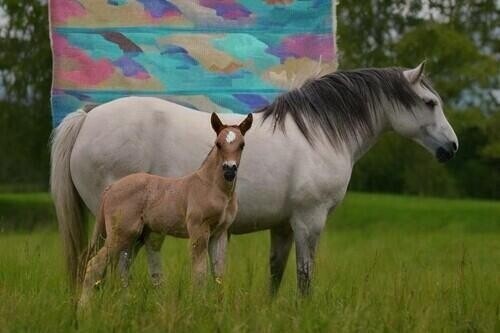}
\caption{Output 2}
\end{subfigure}
\vspace{0.5em}

\begin{subfigure}{0.24\textwidth}
\includegraphics[width=\linewidth,frame]{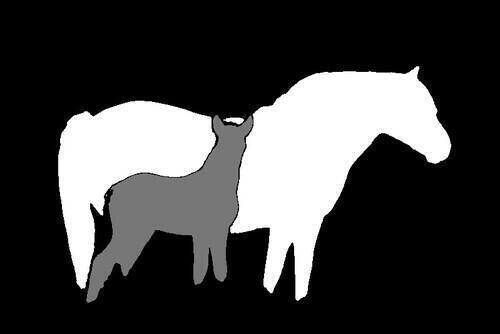}
\caption{Change Map 3} 
\end{subfigure}
\begin{subfigure}{0.24\textwidth}
\includegraphics[width=\linewidth,frame]{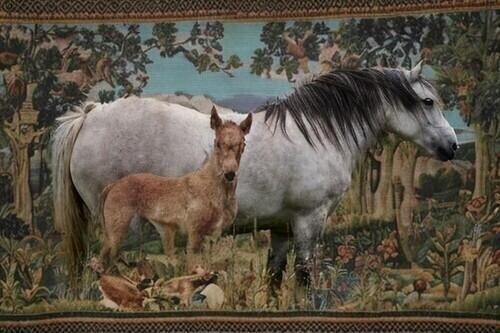}
\caption{Output 3} 
\end{subfigure}
\hfill
\begin{subfigure}{0.24\textwidth}
\includegraphics[width=\linewidth,frame]{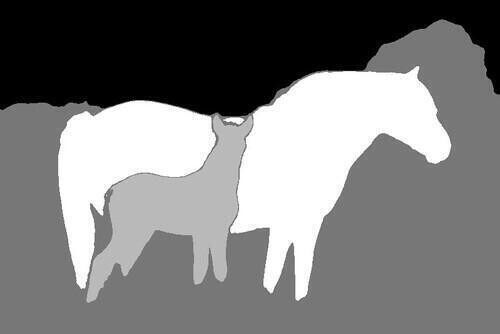}
\caption{Change Map 4} 
\end{subfigure}
\begin{subfigure}{0.24\textwidth}
\includegraphics[width=\linewidth,frame]{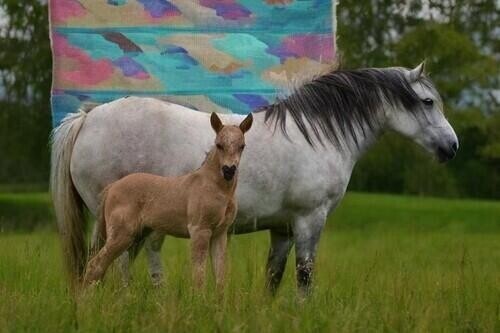}
\caption{Output 4} 
\end{subfigure}
\vspace{0.5em}

\begin{subfigure}{0.24\textwidth}
\includegraphics[width=\linewidth,frame]{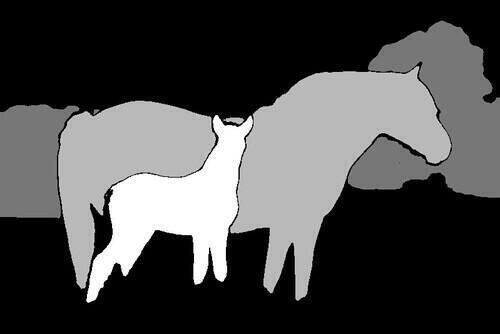}
\caption{Change Map 5} 
\end{subfigure}
\begin{subfigure}{0.24\textwidth}
\includegraphics[width=\linewidth,frame]{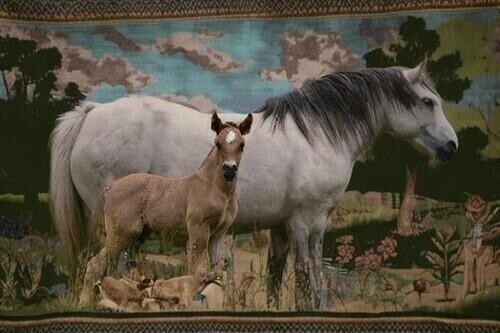}
\caption{Output 5} 
\end{subfigure}
\hfill
\begin{subfigure}{0.24\textwidth}
\includegraphics[width=\linewidth,frame]{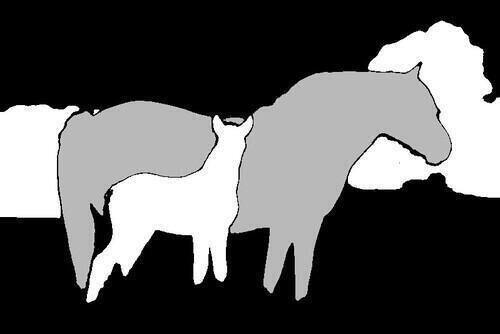}
\caption{Change Map 6} 
\end{subfigure}
\begin{subfigure}{0.24\textwidth}
\includegraphics[width=\linewidth,frame]{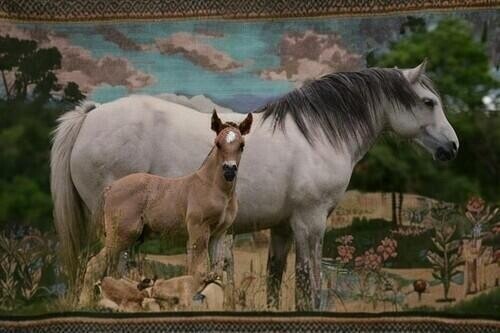}
\caption{Output 6} 
\end{subfigure}
\vspace{0.5em}

\begin{subfigure}{0.24\textwidth}
\includegraphics[width=\linewidth,frame]{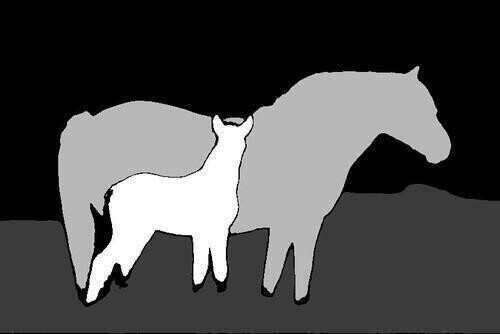}
\caption{Change Map 7} 
\end{subfigure}
\begin{subfigure}{0.24\textwidth}
\includegraphics[width=\linewidth,frame]{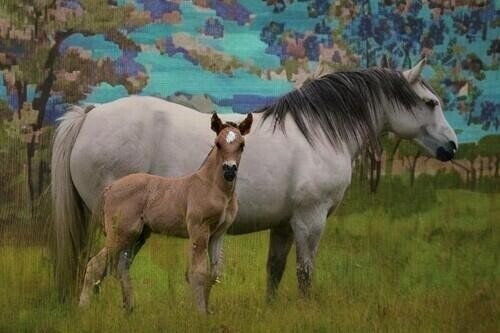}
\caption{Output 7} 
\end{subfigure}
\hfill
\begin{subfigure}{0.24\textwidth}
\includegraphics[width=\linewidth,frame]{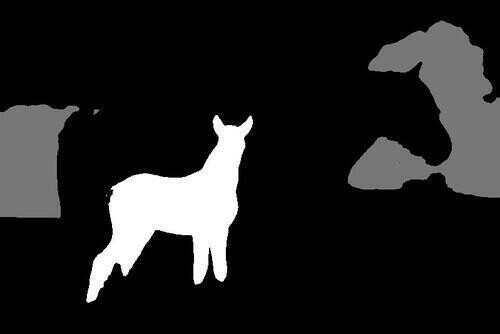}
\caption{Change Map 8} 
\end{subfigure}
\begin{subfigure}{0.24\textwidth}
\includegraphics[width=\linewidth,frame]{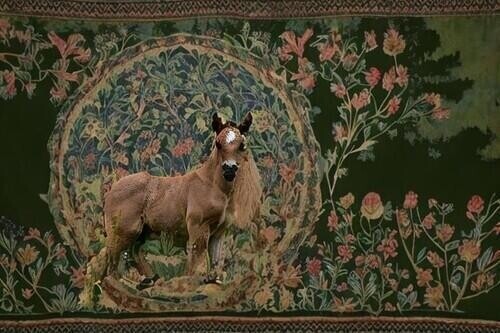}
\caption{Output 8} 
\end{subfigure}
\vspace{0.1em}
  \caption{\textbf{Change map control.} Altering the strength of different regions produces a wide range of edits. The same prompt “tapestry” and seed were used for all examples. 
Notice how the outputs adhere to the change maps.}

\label{tapestry_horses}
\end{figure*}

\begin{figure*}[t]
  \setlength{\ww}{0.135\textwidth}
  
  \centering
  
\begin{subfigure}{\ww}
\includegraphics[width=\linewidth,frame]{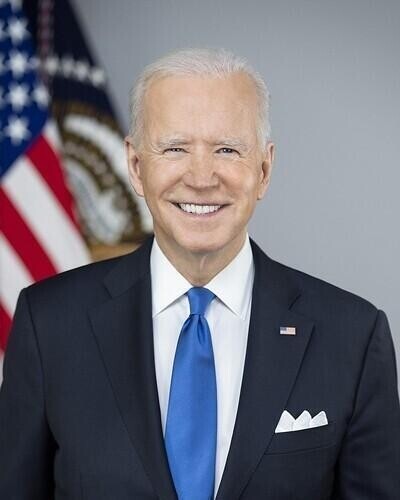}
\caption{Input Image}
\end{subfigure}\hspace{0.5mm}
\begin{subfigure}{\ww}
\includegraphics[width=\linewidth,frame]{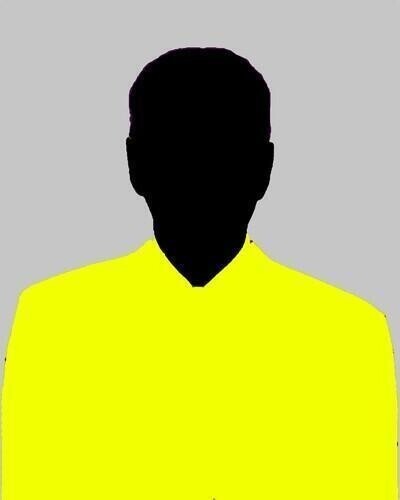}
\caption{Input Map}
\end{subfigure}\hspace{1.5mm}
\begin{subfigure}{\ww}
\includegraphics[width=\linewidth,frame]{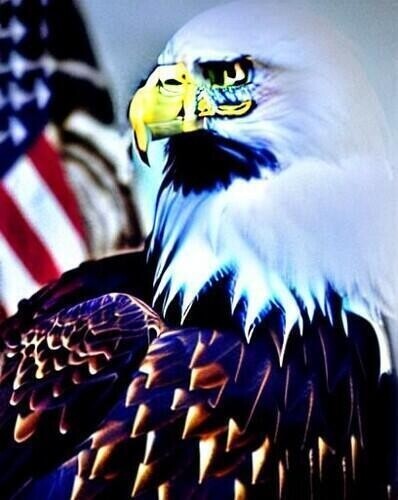}
\caption{0.8}
\end{subfigure}\hspace{0.5mm}
\begin{subfigure}{\ww}
\includegraphics[width=\linewidth,frame]{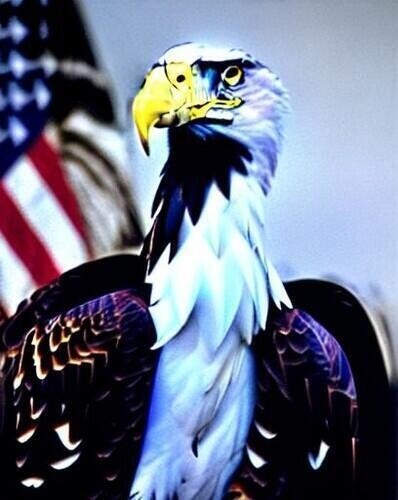}
\caption{0.76}
\end{subfigure}\hspace{0.5mm}
\begin{subfigure}{\ww}
\includegraphics[width=\linewidth,frame]{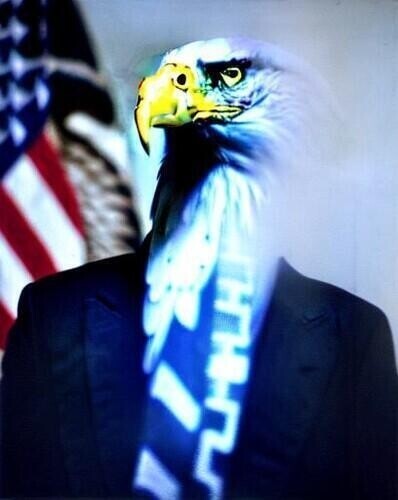}
\caption{0.72}
\end{subfigure}\hspace{0.5mm}
\begin{subfigure}{\ww}
\includegraphics[width=\linewidth,frame]{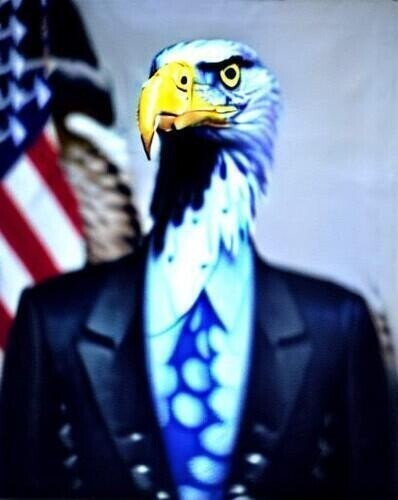}
\caption{0.68}
\end{subfigure}\hspace{0.5mm}
\begin{subfigure}{\ww}
\includegraphics[width=\linewidth,frame]{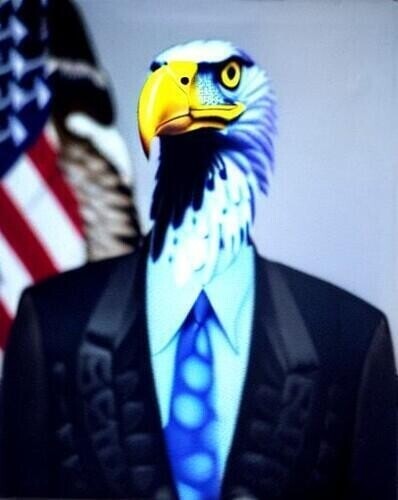}
\caption{0.64}
\end{subfigure}%
  \caption{\textbf{Changing the strength of a single region.} The yellow region of the input map is assigned the value listed below the outputs c--g. The prompt is ``eagle'', expanded as described in Section 3.3.2 of the main paper.}
  \label{fig:eagle-president}
\end{figure*}

\input{figs/latent_degradation_zoom_2}
\newcommand{\qq}{0.134\textwidth}
\begin{figure*}[t]
  \centering
  \setlength{\tabcolsep}{1pt}
  \renewcommand{\arraystretch}{2}
    
 \begin{tabular}{cc@{\hskip 0.15cm}ccccc}
    Input Image
    & Change Map & Ours & Composition 
    & Tiling
    & Five Tiles
    & Masked Noise
    \\
    
    \includegraphics[width=\qq,height=\qq,frame]{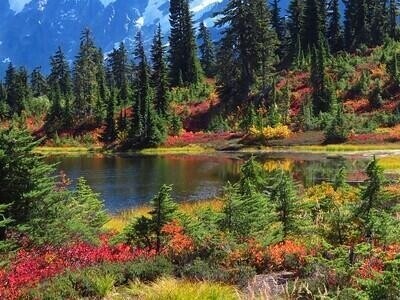} &
    \includegraphics[width=\qq,height=\qq,frame]{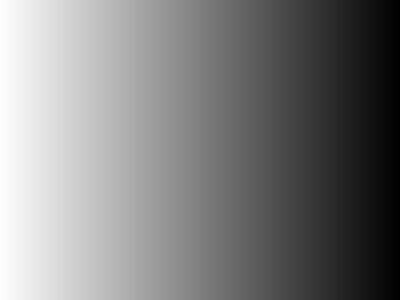} &
    \includegraphics[width=\qq,height=\qq,frame]{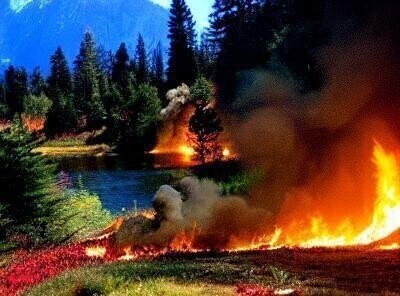} &
    \includegraphics[width=\qq,height=\qq,frame]{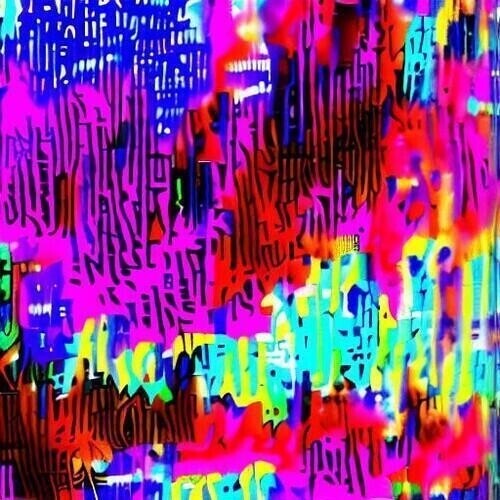} &
    \includegraphics[width=\qq,height=\qq,frame]{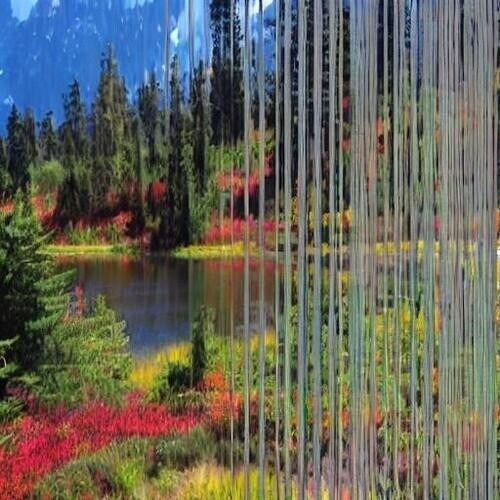} &
    \includegraphics[width=\qq,height=\qq,frame]{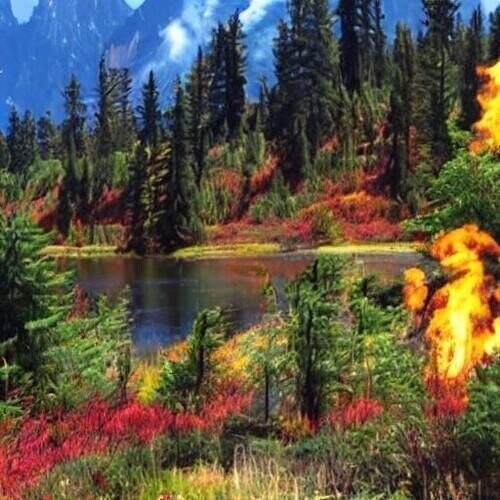} &
    \includegraphics[width=\qq,height=\qq,frame]{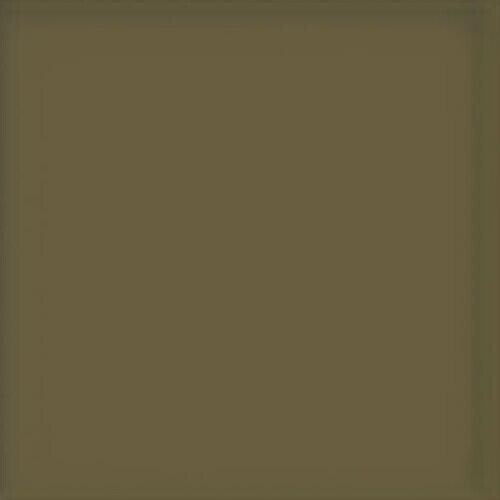} \\
    
    \includegraphics[width=\qq,height=\qq,frame]{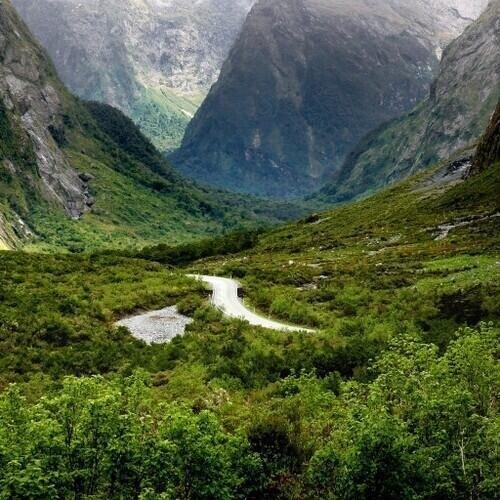} &
    \includegraphics[width=\qq,height=\qq,frame]{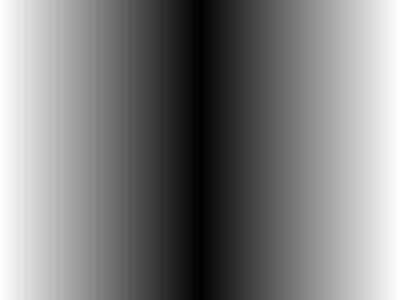} &
    \includegraphics[width=\qq,height=\qq,frame]{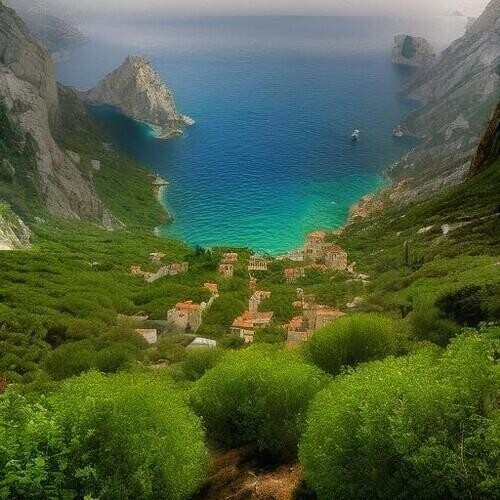} &
    \includegraphics[width=\qq,height=\qq,frame]{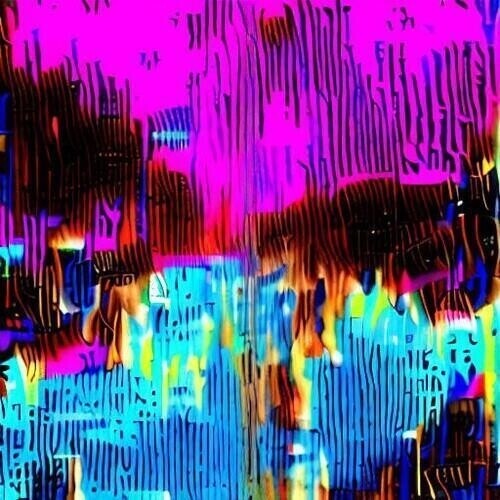} &
    \includegraphics[width=\qq,height=\qq,frame]{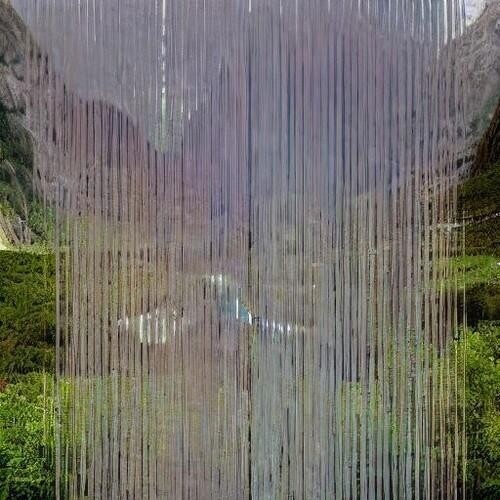} &
    \includegraphics[width=\qq,height=\qq,frame]{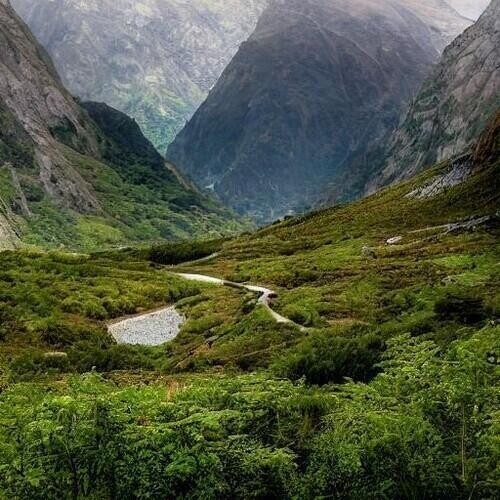} &
    \includegraphics[width=\qq,height=\qq,frame]{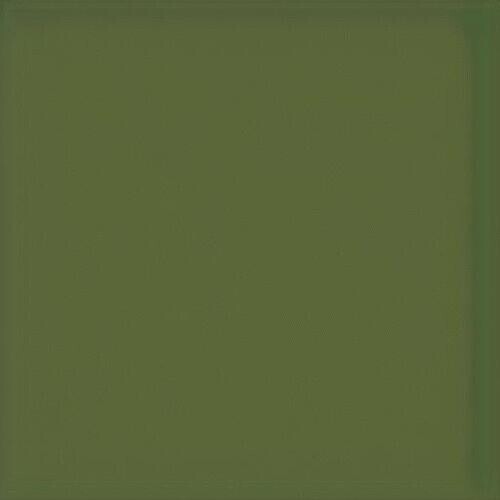} \\
  \end{tabular}
  \caption{
  \textbf{Comparison to the baslines.}
  Even for simple unstructured scenes and smooth maps, all baselines fail. ``Composition'' and ``Masked Noise'' both fail to create a meaningful image, while ``Tiling'' does not produce an edit related to the prompt and corrupts the image. ``Five Tiles'' stands out as the most successful among the baselines. Albeit, the edit is primarily noticeable in the darkest tile.
  Prompt: ``everything is burning, fire'', ``Mediterranean Sea''.
 }
 \label{fig:alternatives_fire} 
\end{figure*}

 \begin{figure*}
  \centering
  \begin{subfigure}{0.3\columnwidth}
    \includegraphics[width=\linewidth,frame]{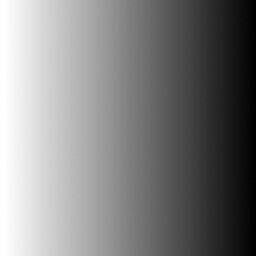}
    \caption{Input Change Map}
  \end{subfigure}\hspace{0.02\columnwidth}%
  \begin{subfigure}{0.3\columnwidth}
    \includegraphics[width=\linewidth,frame]{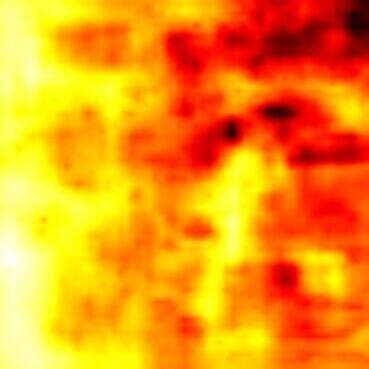}
    \caption{1 pair}
  \end{subfigure}\hspace{0.02\columnwidth}%
  \begin{subfigure}{0.3\columnwidth}
    \includegraphics[width=\linewidth,frame]{imgs/eval/full_reconstruction/gradient_reconstruction.jpg}
    \caption{1,000 pairs}
  \end{subfigure}
  \\
  \hspace{0.02\columnwidth}%
  \begin{subfigure}{0.3\columnwidth}
    \includegraphics[width=\linewidth,frame]{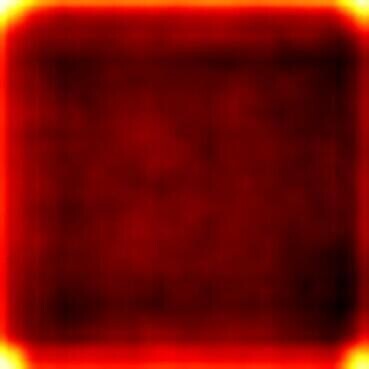}
    \caption{LPIPS spatial bias}
  \end{subfigure}

  \caption{\textbf{Intermediate results of change map measurement.} Edit strength measurement of a change map based on a single pair vs.\ 1,000 input-output pairs. 
  Using only one pair produces a measurement which is somewhat similar to the true map, but extremely noisy.
  The LPIPS spatial bias is displayed on the 2nd row. 
  }
  \label{single-reconstruction}
\end{figure*}

\begin{figure*}[t]
\captionsetup[subfigure]{labelformat=empty}

\centering

\begin{subfigure}{0.19\textwidth}
\includegraphics[width=\linewidth,frame]{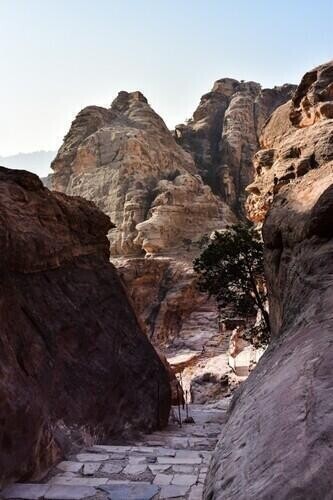}
\caption{Input Image}
\end{subfigure}
\hfill
\begin{subfigure}{0.19\textwidth}
\includegraphics[width=\linewidth,frame]{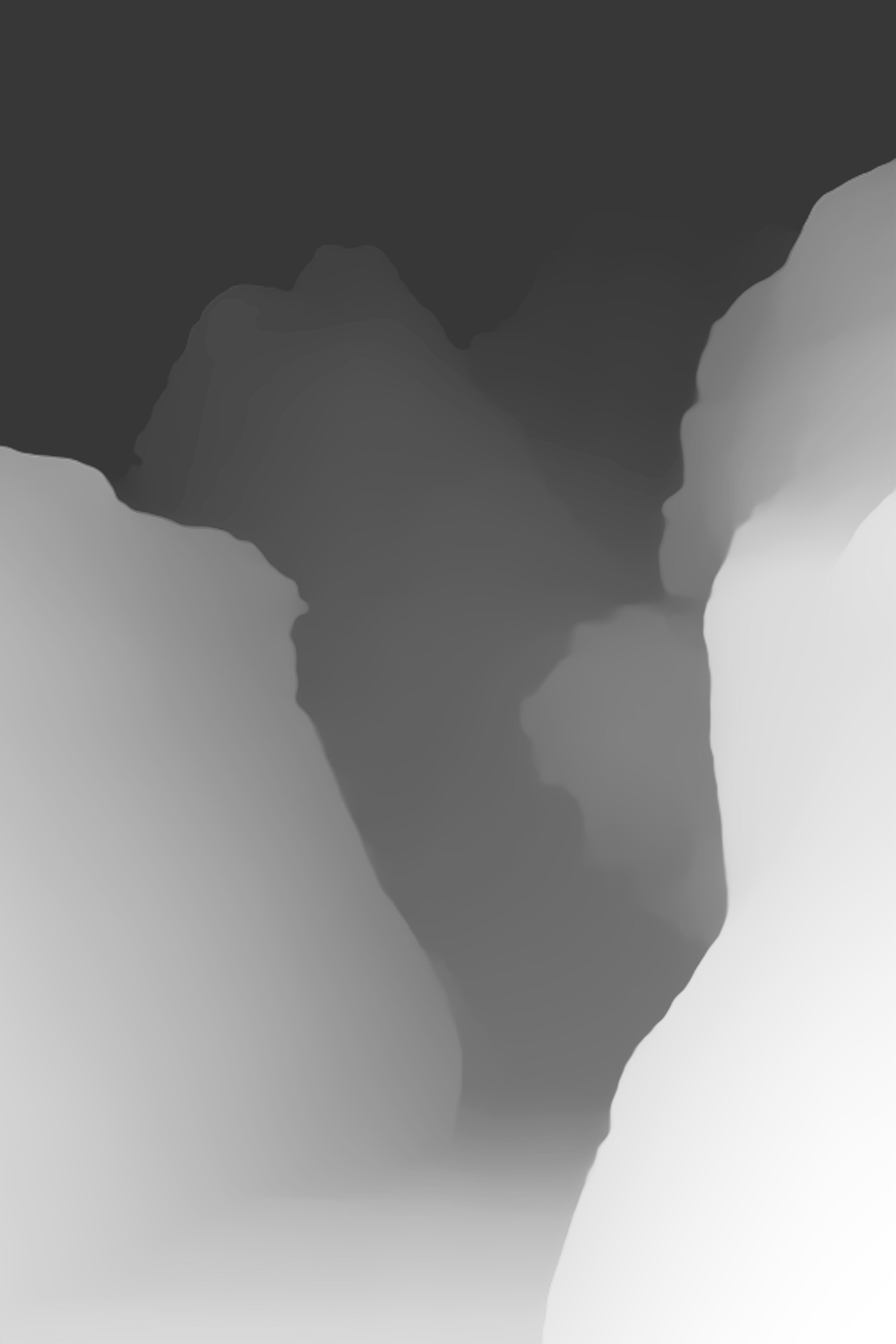}
\caption{Input Change Map}
\end{subfigure}
\hfill
\begin{subfigure}{0.19\textwidth}
\includegraphics[width=\linewidth,frame]{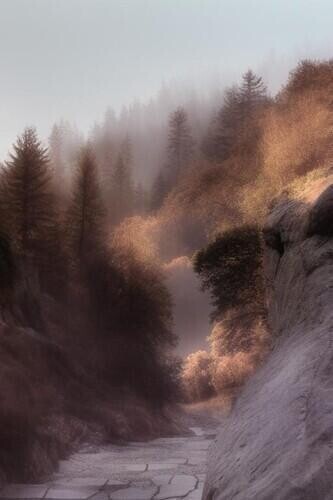}
\caption{DDPM~\cite{ho2020denoising}}
\end{subfigure}
\hfill
\begin{subfigure}{0.19\textwidth}
\includegraphics[width=\linewidth,frame]{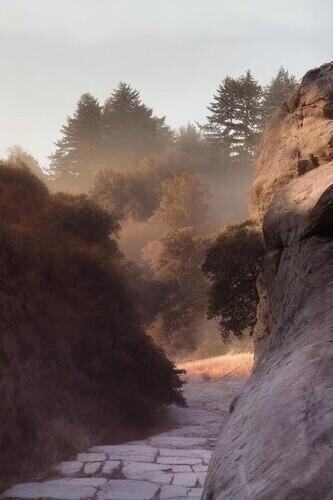}
\caption{DDIM~\cite{song2022denoising}}
\end{subfigure}
\hfill
\begin{subfigure}{0.19\textwidth}
\includegraphics[width=\linewidth,frame]{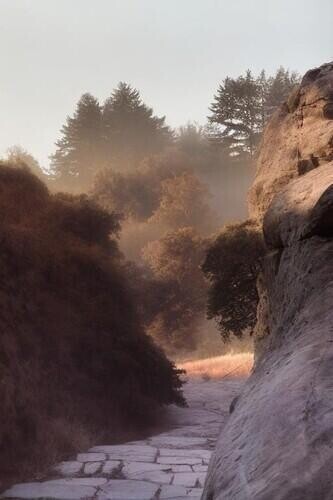}
\caption{DEIS~\cite{zhang2023fast}}
\end{subfigure}
\vspace{0.2cm}

\begin{subfigure}{0.19\textwidth}
\includegraphics[width=\linewidth,frame]{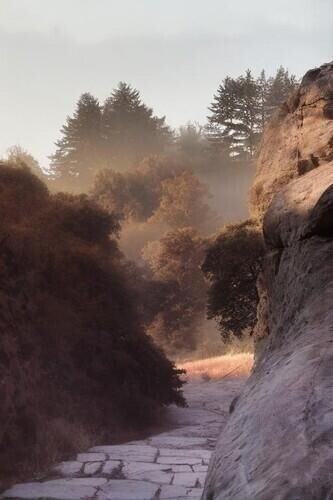}
\caption{DPM Solver Multi Step~\cite{lu2022dpmsolver}}
\end{subfigure}
\hfill
\begin{subfigure}{0.19\textwidth}
\includegraphics[width=\linewidth,frame]{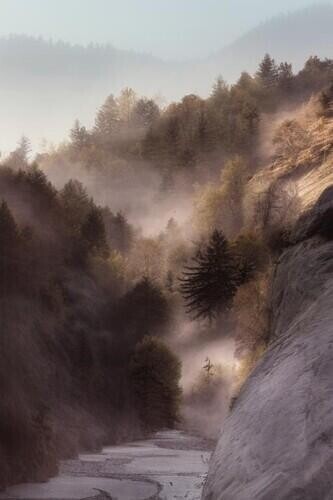}
\caption{DPM Solver SDE~\cite{lu2022dpmsolver}}
\end{subfigure}
\hfill
\begin{subfigure}{0.19\textwidth}
\includegraphics[width=\linewidth,frame]{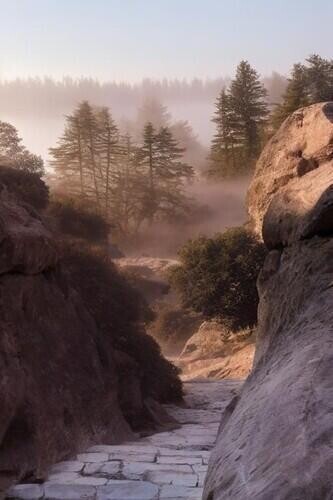}
\caption{DPM Solver Single Step~\cite{lu2022dpmsolver}}
\end{subfigure}
\hfill
\begin{subfigure}{0.19\textwidth}
\includegraphics[width=\linewidth,frame]{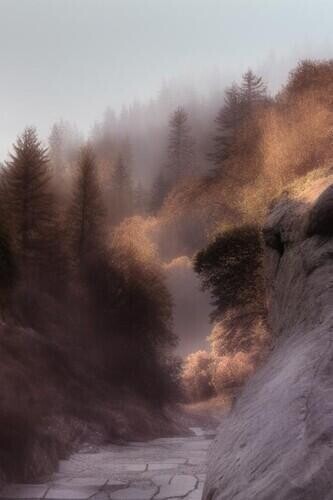}
\caption{Euler Ancestral Discrete~\cite{karras2022elucidating}}
\end{subfigure}
\hfill
\begin{subfigure}{0.19\textwidth}
\includegraphics[width=\linewidth,frame]{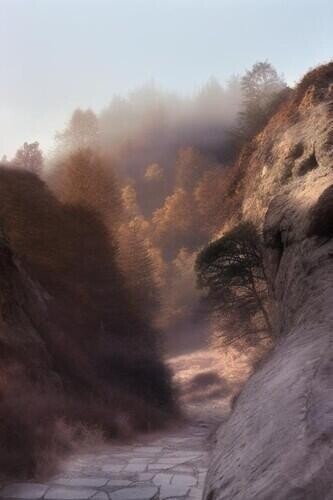}
\caption{Euler Discrete~\cite{karras2022elucidating}}
\end{subfigure}
\vspace{0.2cm}

\begin{subfigure}{0.19\textwidth}
\includegraphics[width=\linewidth,frame]{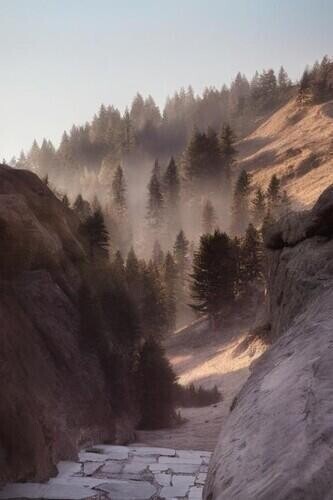}
\caption{Heun Discrete~\cite{karras2022elucidating}}
\end{subfigure}
\hfill
\begin{subfigure}{0.19\textwidth}
\includegraphics[width=\linewidth,frame]{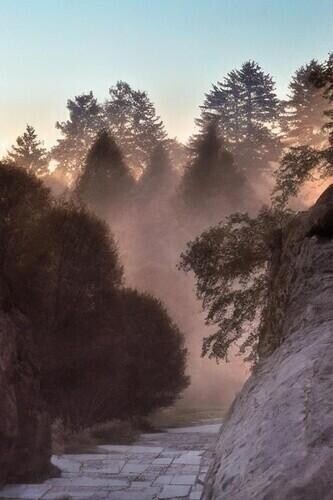}
\caption{KDPM2 Ances.\ Discrete~\cite{karras2022elucidating}}
\end{subfigure}
\hfill
\begin{subfigure}{0.19\textwidth}
\includegraphics[width=\linewidth,frame]{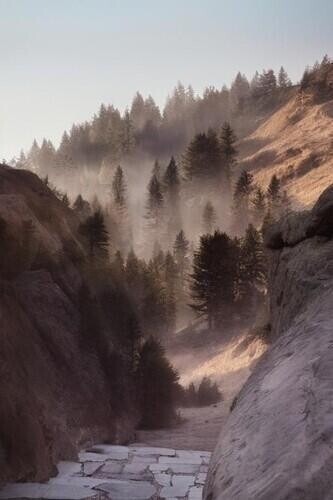}
\caption{KDPM2 Discrete~\cite{karras2022elucidating}}
\end{subfigure}
\hfill
\begin{subfigure}{0.19\textwidth}
\includegraphics[width=\linewidth,frame]{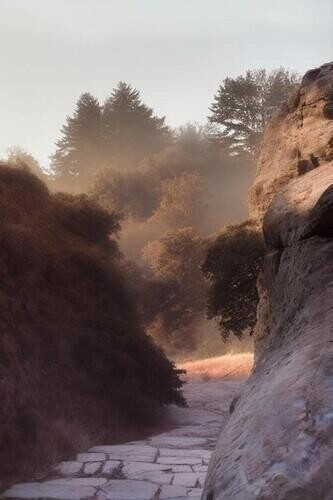}
\caption{LMS Discrete~\cite{karras2022elucidating}}
\end{subfigure}
\hfill
\begin{subfigure}{0.19\textwidth}
\includegraphics[width=\linewidth,frame]{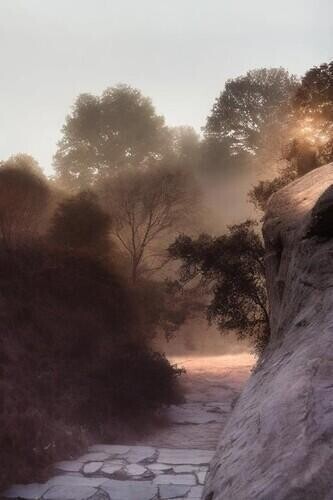}
\caption{PNDM~\cite{liu2022pseudo}}
\end{subfigure}
\vspace{0.01cm} 

  \caption{\textbf{Various sampling techniques.} Our method is compatible with various samplers that support image-to-image translations. All samplers that are examined maintain adherence to the original input, to the change map, and to the prompt ``Subtle morning mist''.}
  \label{samplers}
\end{figure*}

\begin{figure*}[t]

  \centering
  \begin{tabular}{cccp{0.5cm}ccc}

\textbf{Input Image} & \textbf{Input Change Map} & \textbf{Output Image} & &
\textbf{Input Image} & \textbf{Input Change Map} & \textbf{Output Image} \\
\includegraphics[width=\xx,height=\zz,frame]{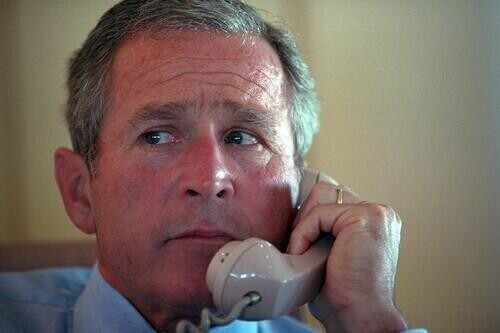} &
\includegraphics[width=\xx,height=\zz,frame]{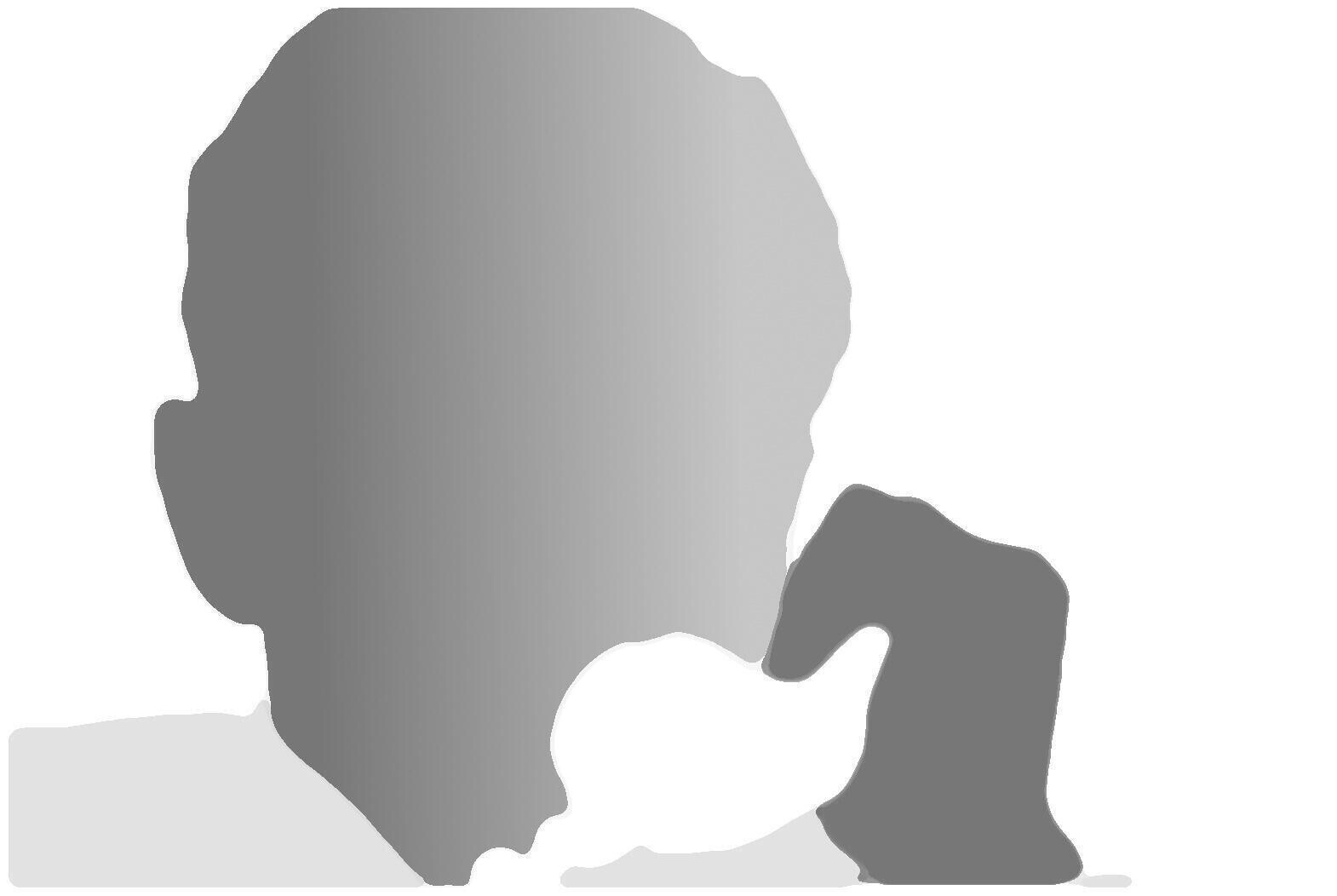} &
\includegraphics[width=\xx,height=\zz,frame]{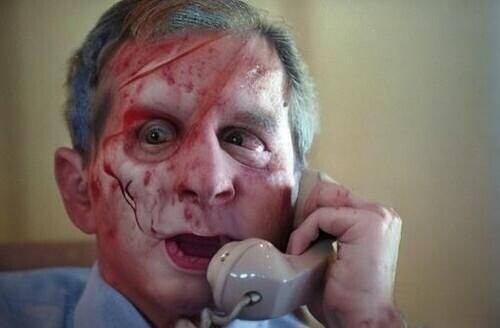} & &
\includegraphics[width=\xx,height=\zz,frame]{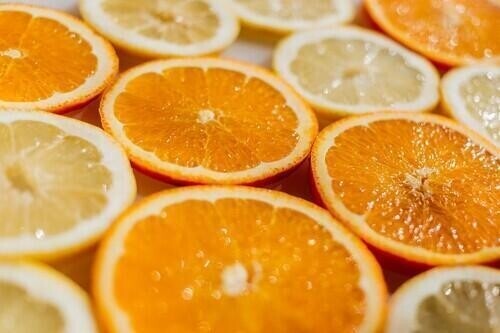} & \includegraphics[width=\xx,height=\zz,frame]{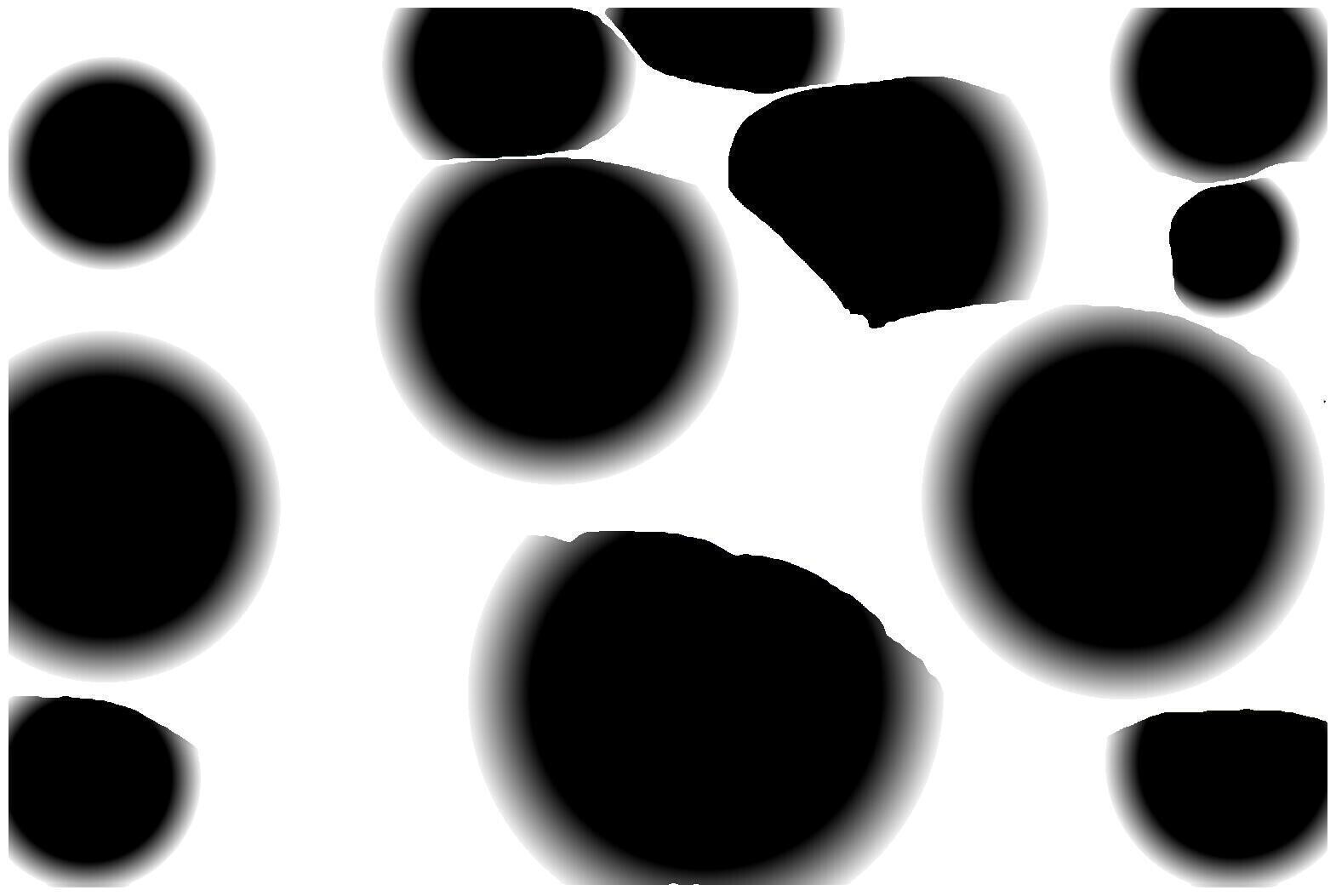} & \includegraphics[width=\xx,height=\zz,frame]{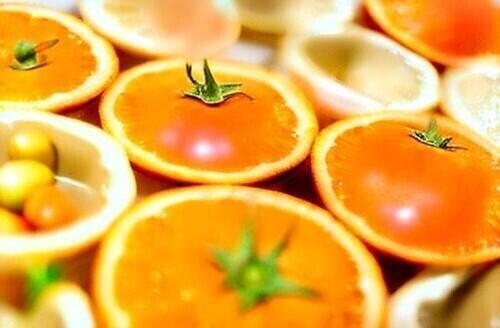} \\
\includegraphics[width=\xx,height=\zz,frame]{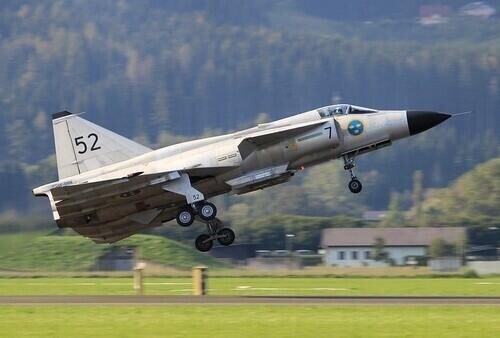} & \includegraphics[width=\xx,height=\zz,frame]{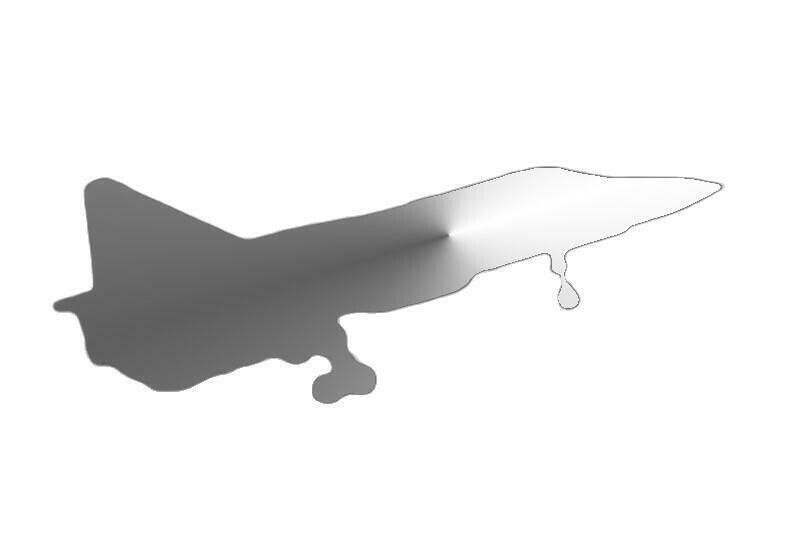} & \includegraphics[width=\xx,height=\zz,frame]{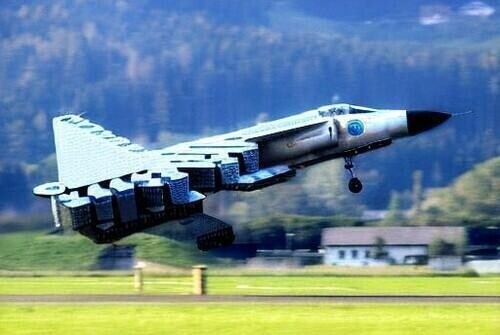} & &
\includegraphics[width=\xx,height=\zz,frame]{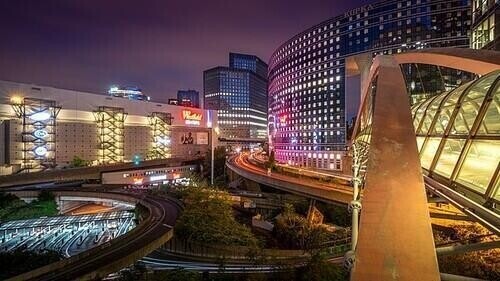} & \includegraphics[width=\xx,height=\zz,frame]{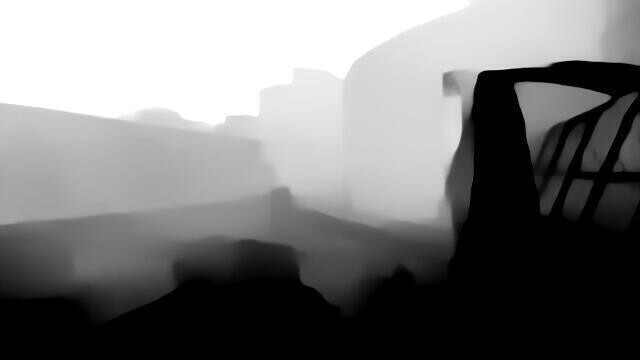} & \includegraphics[width=\xx,height=\zz,frame]{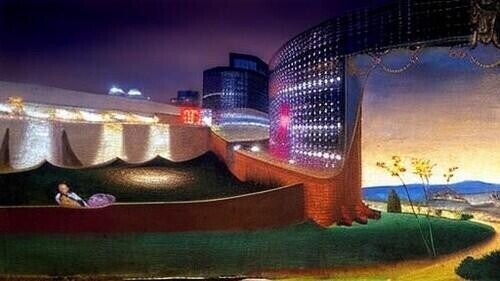} \\

\includegraphics[width=\xx,height=\zz,frame]{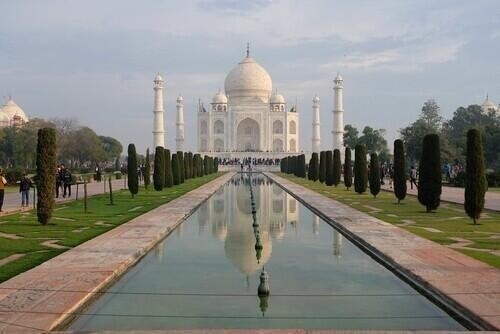} & \includegraphics[width=\xx,height=\zz,frame]{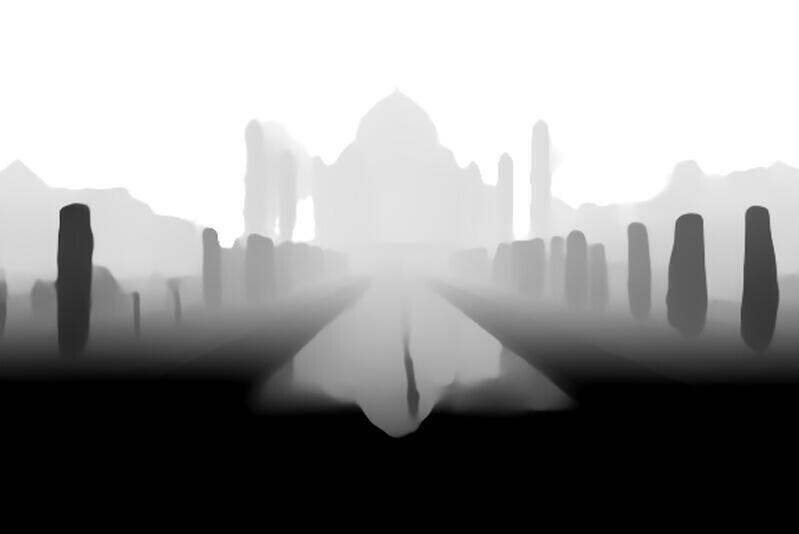} & \includegraphics[width=\xx,height=\zz,frame]{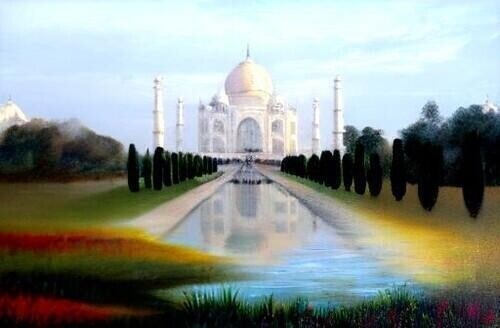} & &
\includegraphics[width=\xx,height=\zz,frame]{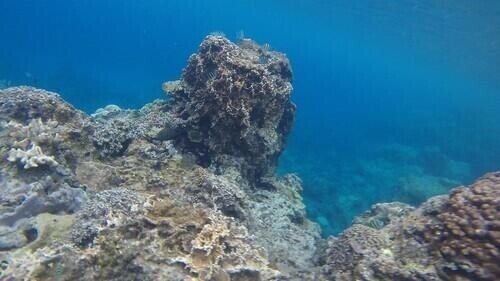} & \includegraphics[width=\xx,height=\zz,frame]{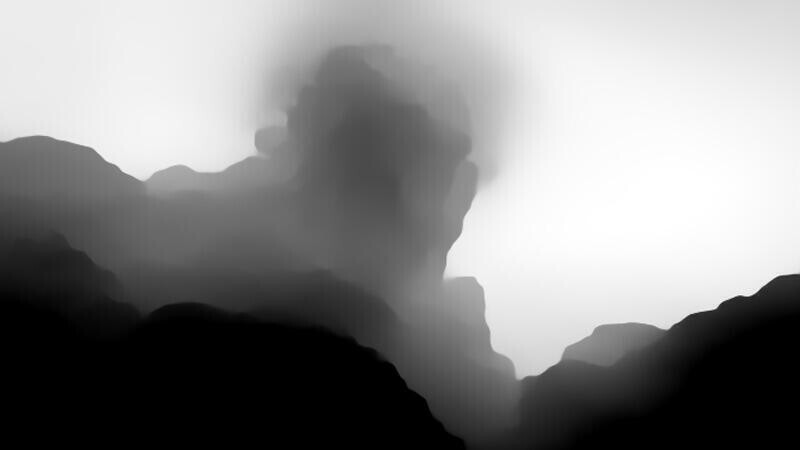} & \includegraphics[width=\xx,height=\zz,frame]{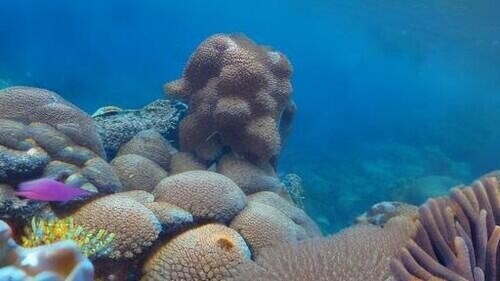} \\
\includegraphics[width=\xx,height=\zz,frame]{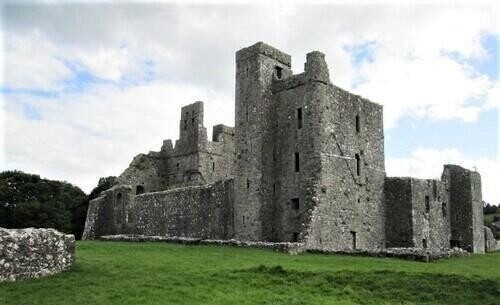} & \includegraphics[width=\xx,height=\zz,frame]{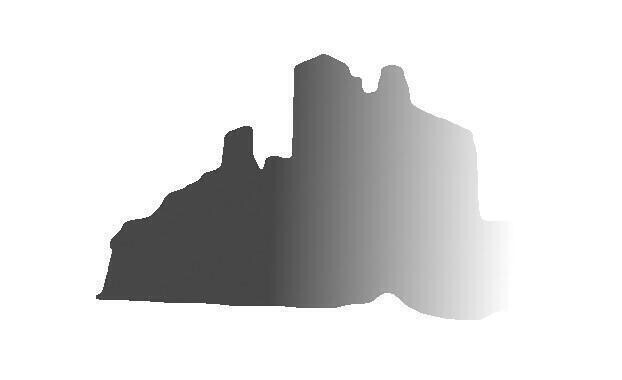} & \includegraphics[width=\xx,height=\zz,frame]{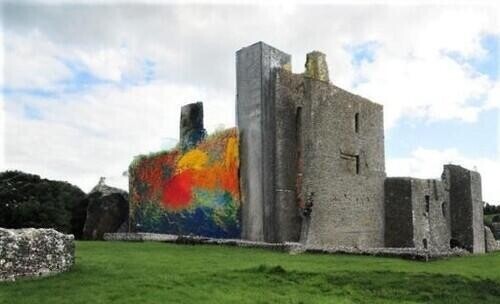} & &
\includegraphics[width=\xx,height=\zz,frame]{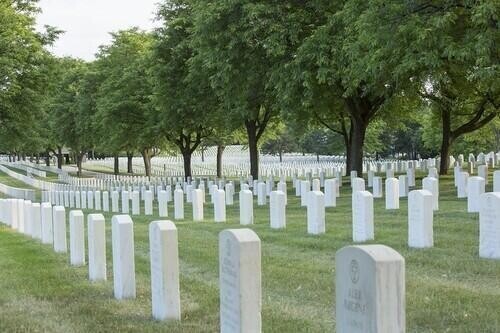} & \includegraphics[width=\xx,height=\zz,frame]{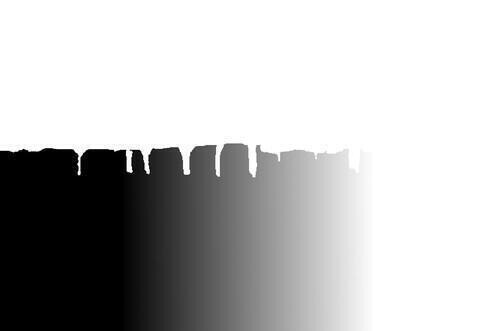} & \includegraphics[width=\xx,height=\zz,frame]{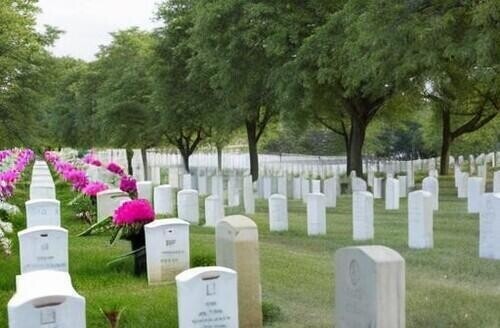} \\
\includegraphics[width=\xx,height=\zz,frame]{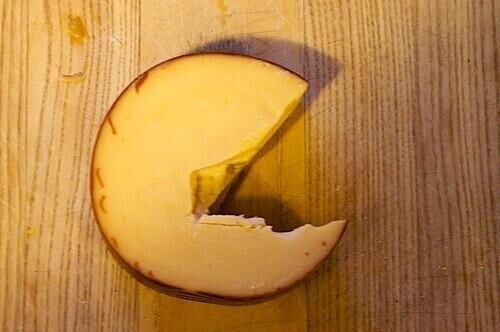} & \includegraphics[width=\xx,height=\zz,frame]{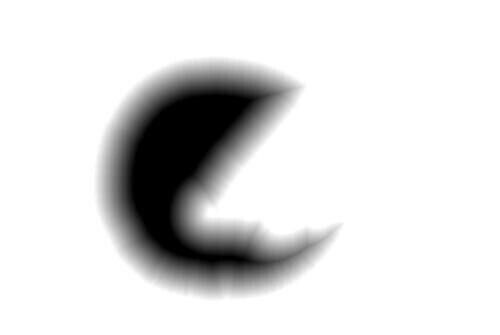} & \includegraphics[width=\xx,height=\zz,frame]{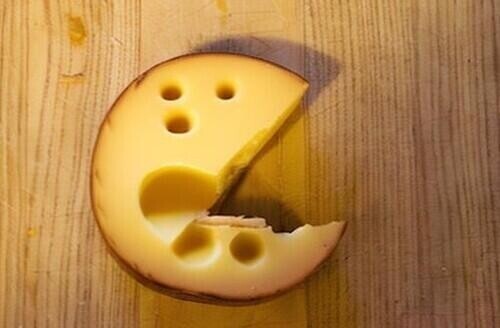} & &
\includegraphics[width=\xx,height=\zz,frame]{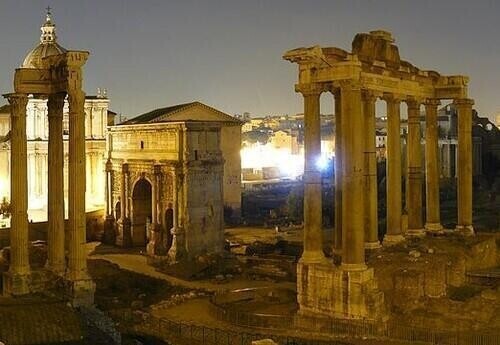} & \includegraphics[width=\xx,height=\zz,frame]{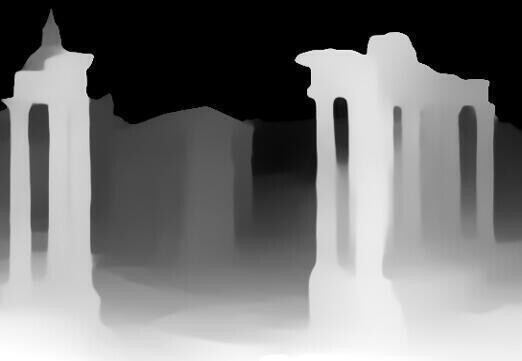} & \includegraphics[width=\xx,height=\zz,frame]{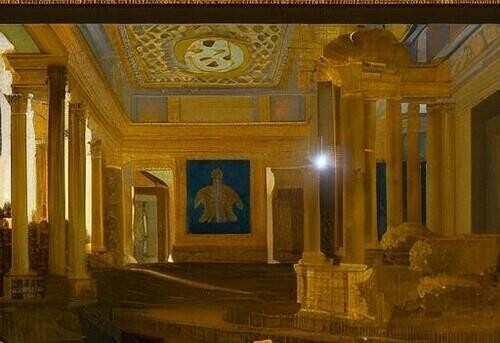} \\

\includegraphics[width=\xx,frame]{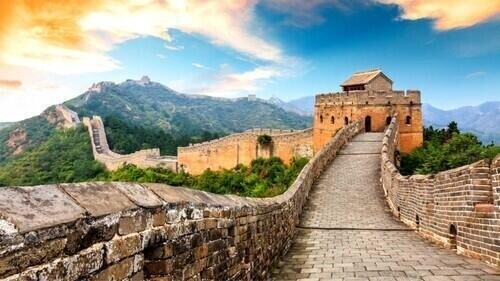} & \includegraphics[width=\xx,frame]{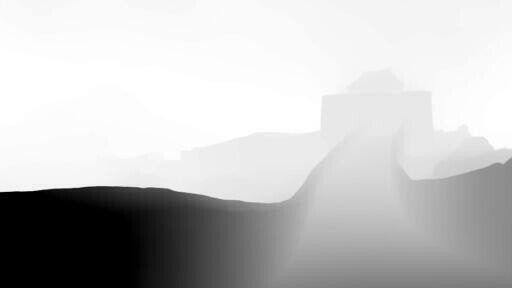} & \includegraphics[width=\xx,frame]{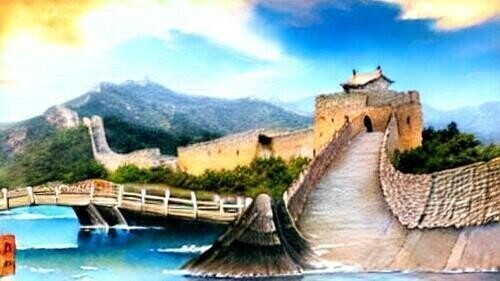} & &

\includegraphics[width=\xx,frame]{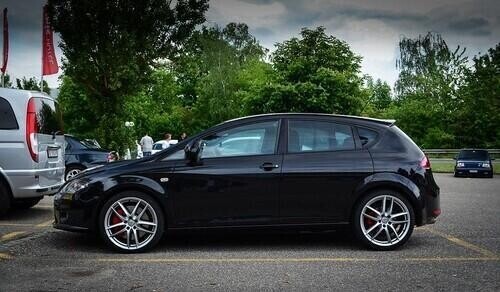} & \includegraphics[width=\xx,frame]{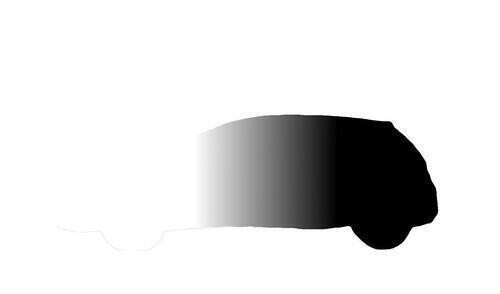} & \includegraphics[width=\xx,frame]{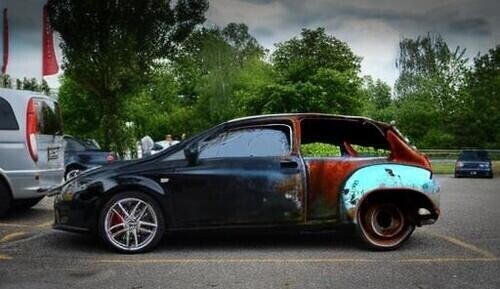}  
\\
\includegraphics[width=\xx,frame]{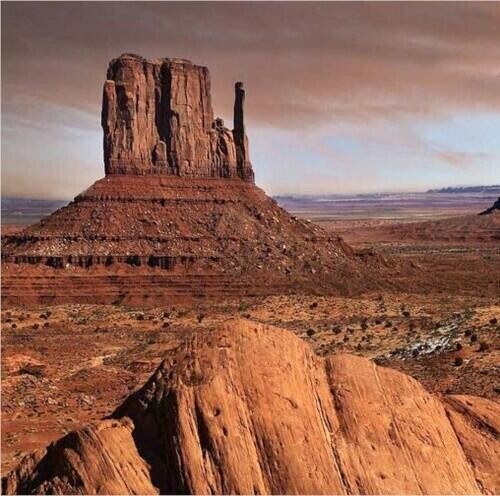} & \includegraphics[width=\xx,frame]{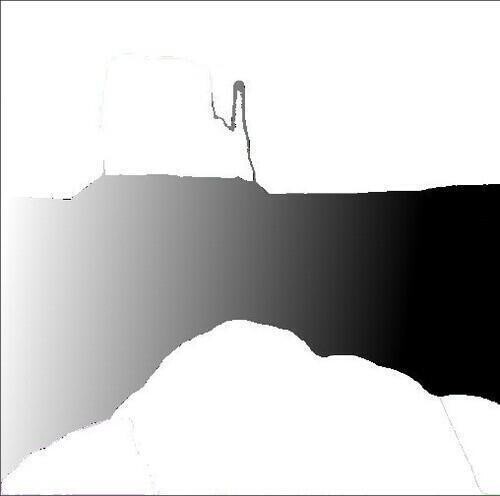} & \includegraphics[width=\xx,frame]{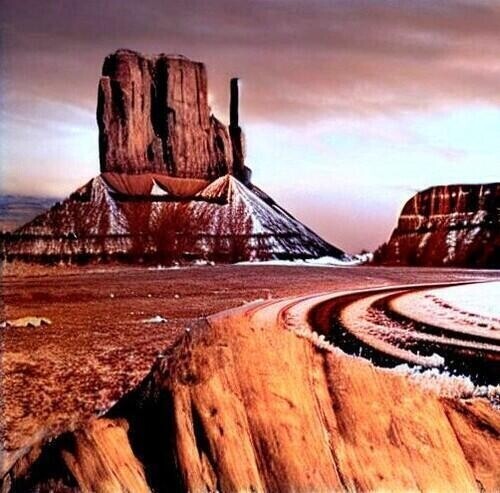} & &
\includegraphics[width=\xx,frame]{imgs/comparison/video_game/origin.jpg} & \includegraphics[width=\xx,frame]{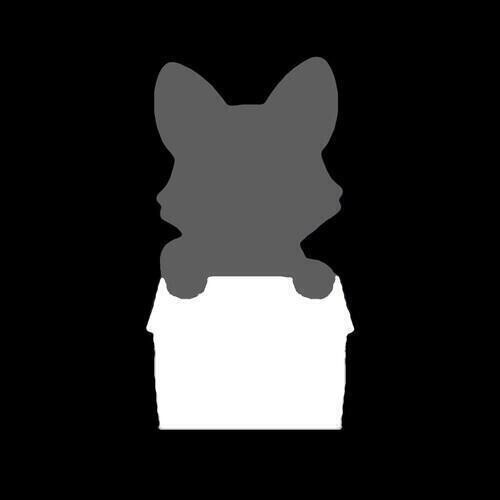} & \includegraphics[width=\xx,frame]{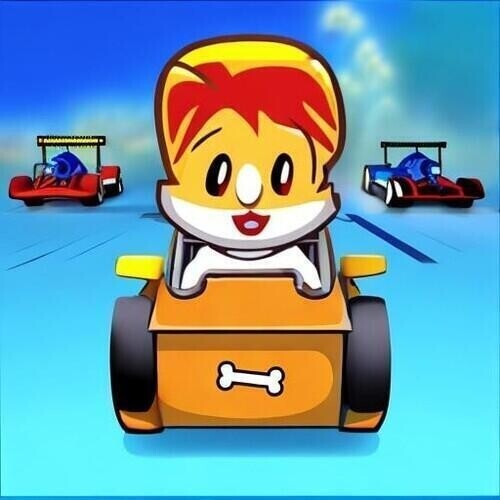} 
\\
\includegraphics[width=\xx,height=\zz,frame]{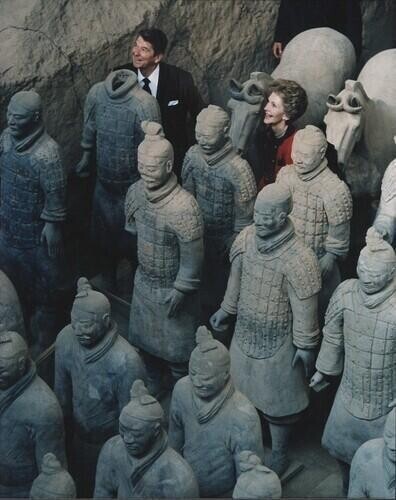} &
\includegraphics[width=\xx,height=\zz,frame]{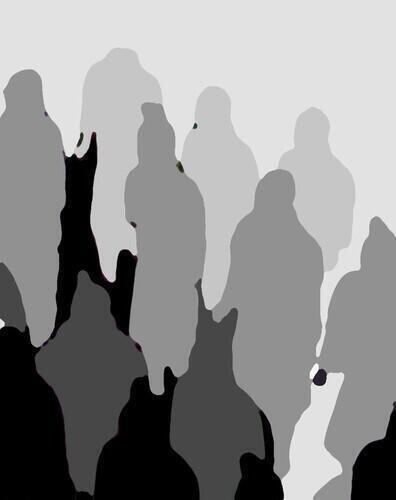} &
\includegraphics[width=\xx,height=\zz,frame]{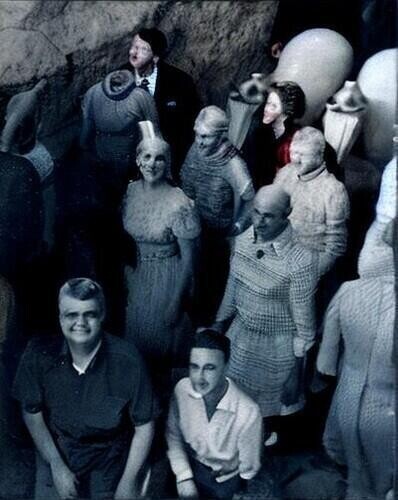} & &
\includegraphics[width=\xx,height=\zz,frame]{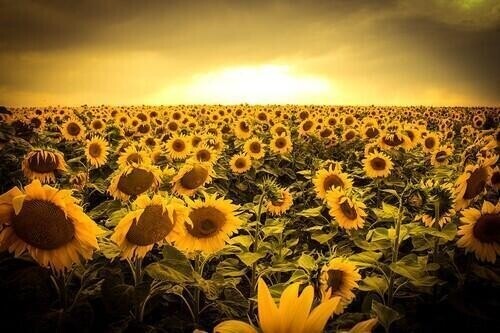} & \includegraphics[width=\xx,height=\zz,frame]{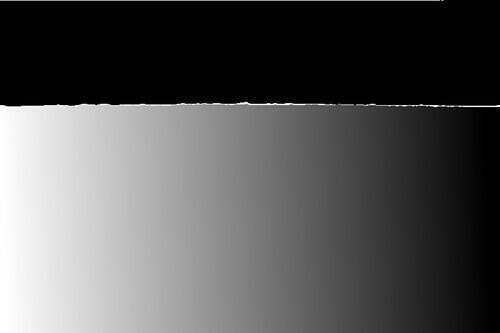} & \includegraphics[width=\xx,height=\zz,frame]{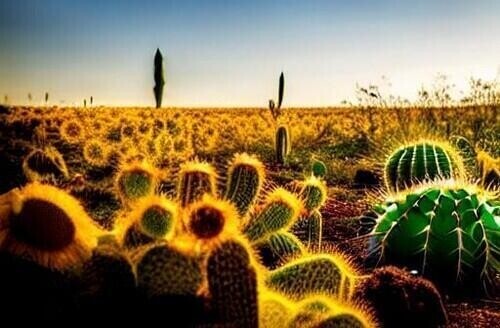} 
\\
\includegraphics[width=\xx,height=\zz,frame]{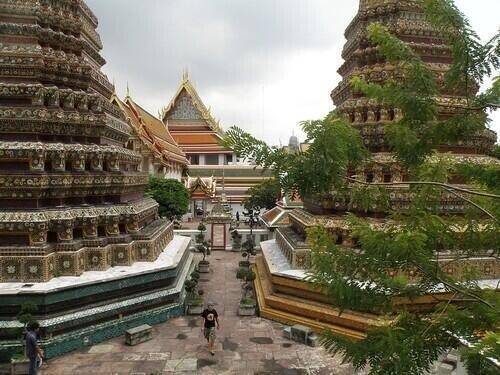} &
\includegraphics[width=\xx,height=\zz,frame]{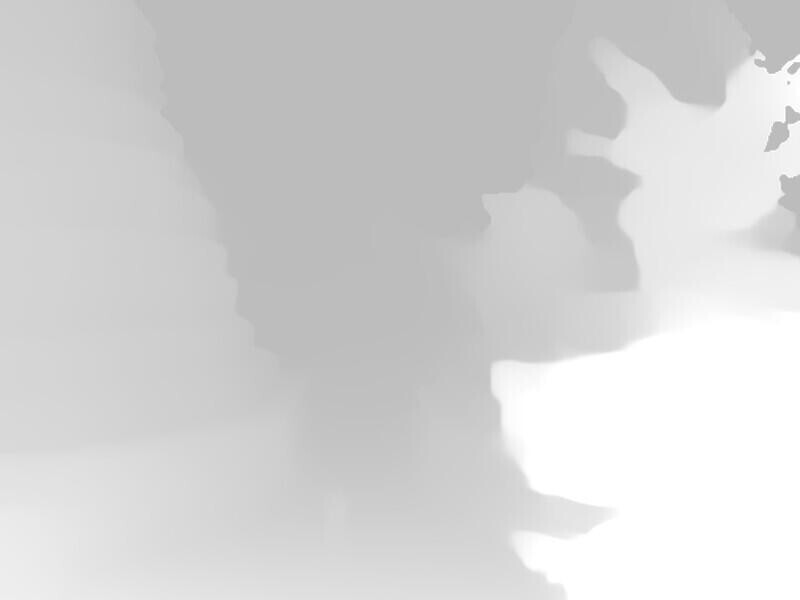} &
\includegraphics[width=\xx,height=\zz,frame]{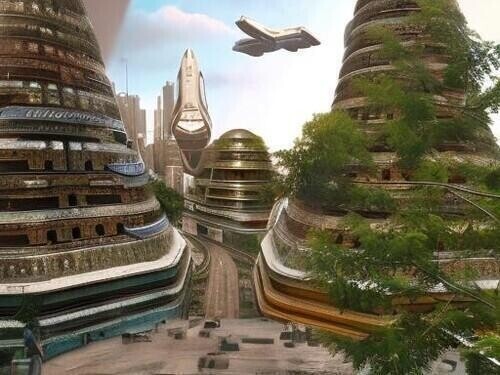} & &
\includegraphics[width=\xx,height=\zz,frame]{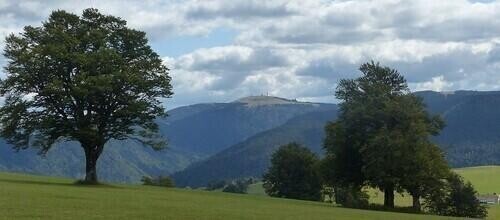} &
\includegraphics[width=\xx,height=\zz,frame]{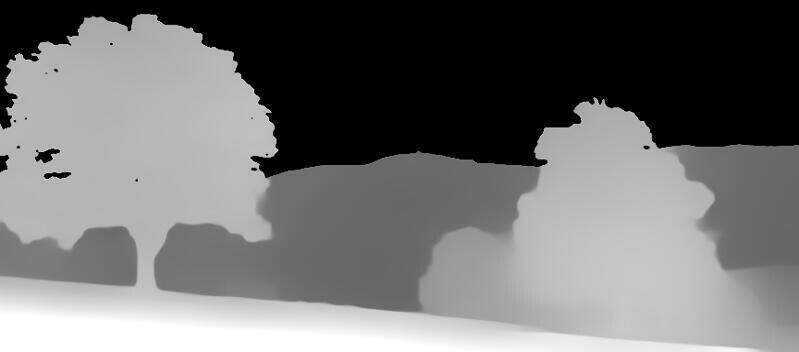} &
\includegraphics[width=\xx,height=\zz,frame]{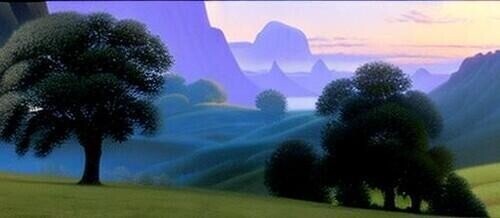}
\\
\includegraphics[width=\xx,height=\bb,frame]{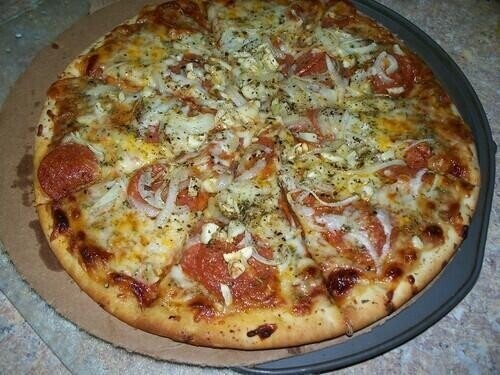} & \includegraphics[width=\xx,height=\bb,frame]{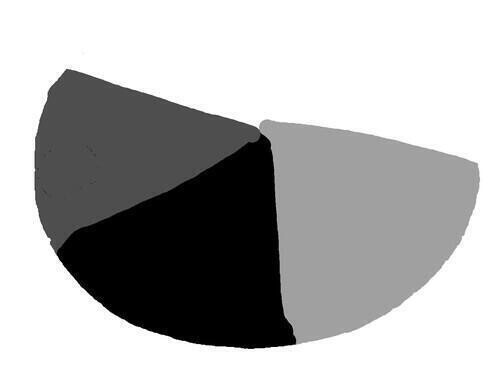} & \includegraphics[width=\xx,height=\bb,frame]{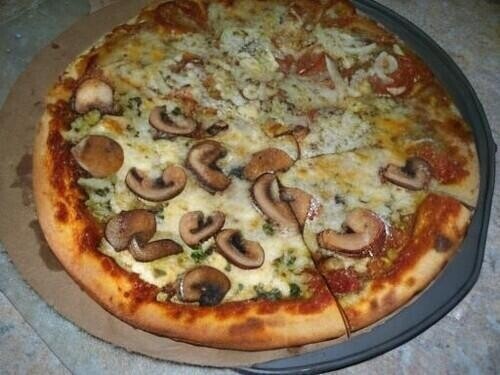} & & 
\includegraphics[width=\xx,height=\bb,frame]{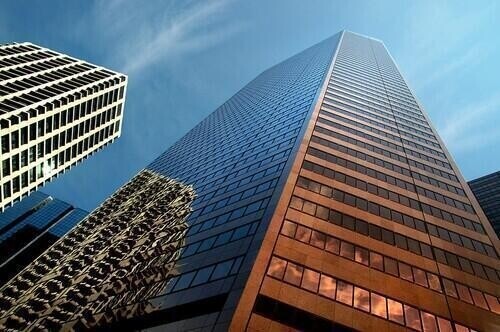} &
\includegraphics[width=\xx,height=\bb,frame]{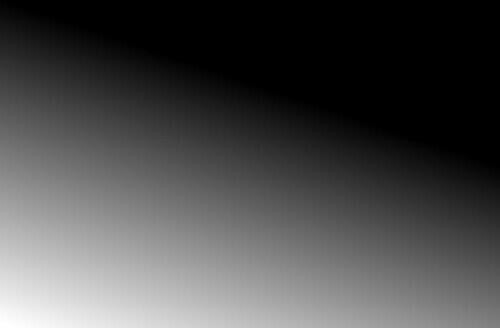} &
\includegraphics[width=\xx,height=\bb,frame]{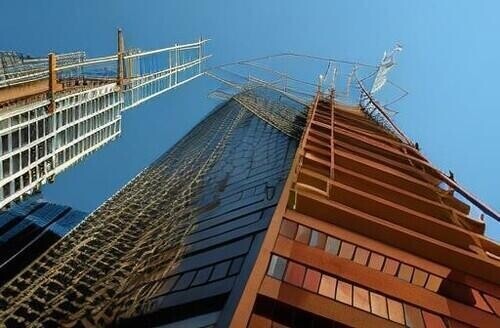}
\end{tabular}

  \caption{\textbf{Various edits with our method.}
  Prompts (row-major order):
  ``zombie'', 
  ``a pile of tomatoes with green stems on them in a market place'',
  ``colorful lego blocks'', 
  ``a detailed painting'',
  ``an oil on canvas painting, metaphysical painting'', 
  ``Coral reef'',
  ``a painting with lots of paint splattered'',
  ``ouquets of flowers are placed in graves'',
  ``Swiss cheese'', 
  ``a mosaic'',
  ``a watercolor painting'',
  ``rusted car'',
  ``snowy surface'',
  ``race car video game'',
  ``group of people are posing for a picture together'',``a cactus with fruit growing on it in a field', “a futuristic city'', ``a detailed matte painting'', ``a bunch of mushrooms on a pizza'',``scaffolding''.
  We used the technique described in Section 3.3.2 in the main paper to expand the prompt for the 2\textsuperscript{nd}, 3\textsuperscript{rd}, 4\textsuperscript{th}, 5\textsuperscript{th}, 6\textsuperscript{th}, 8\textsuperscript{th}, 11\textsuperscript{th}, 13\textsuperscript{th}, 16\textsuperscript{th}, 17\textsuperscript{th}, 18\textsuperscript{th}, 19\textsuperscript{th}, 20\textsuperscript{th} and 21\textsuperscript{st} examples.}
  \label{lego_airplane}
  \label{zombie-bush}
\end{figure*}

\begin{figure*}[t]
  \centering
  \includegraphics[width=0.35\linewidth, margin=5pt]{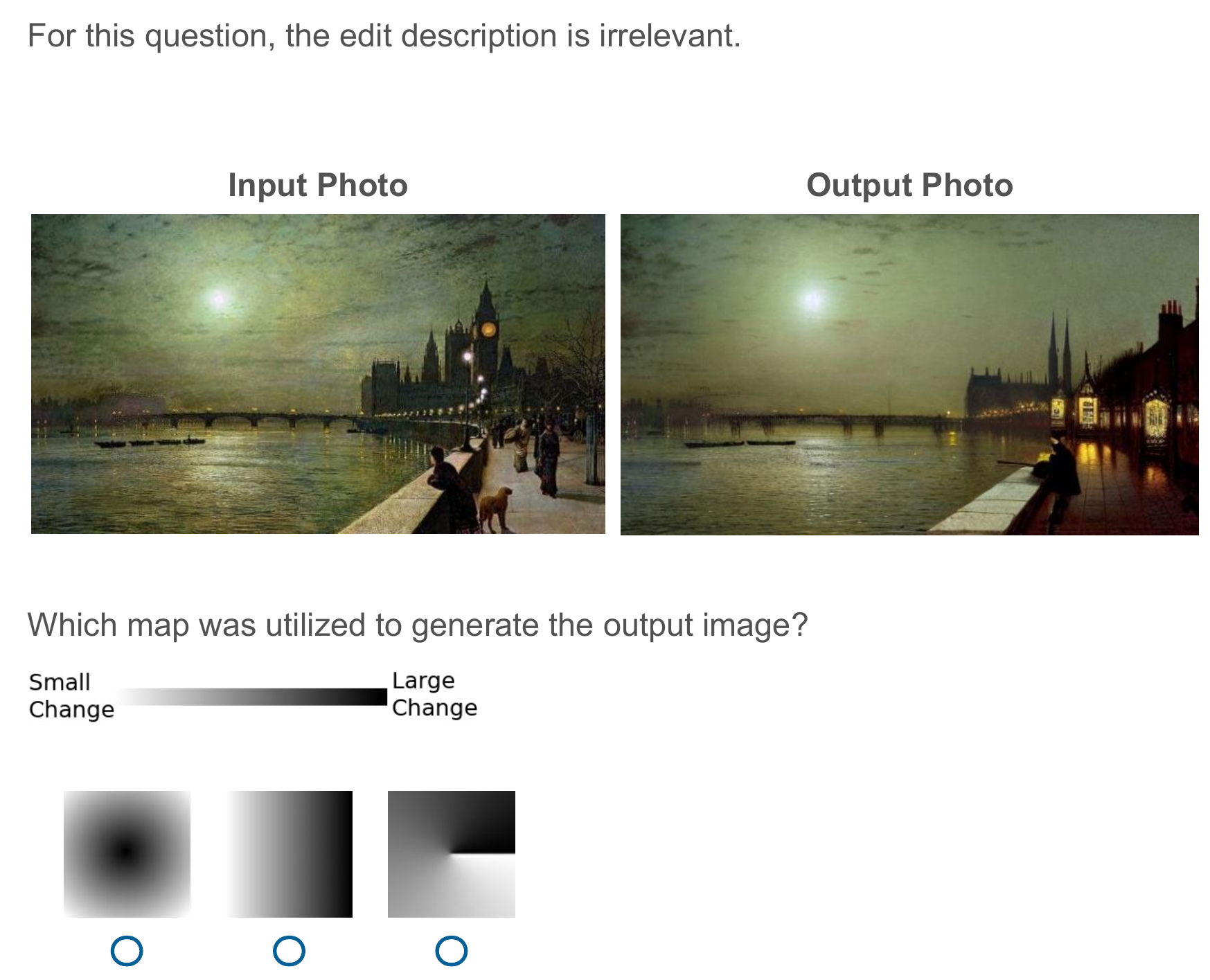}
  \rule{\linewidth}{0.36pt} %
  \\
    \includegraphics[width=0.35\linewidth, margin=5pt]{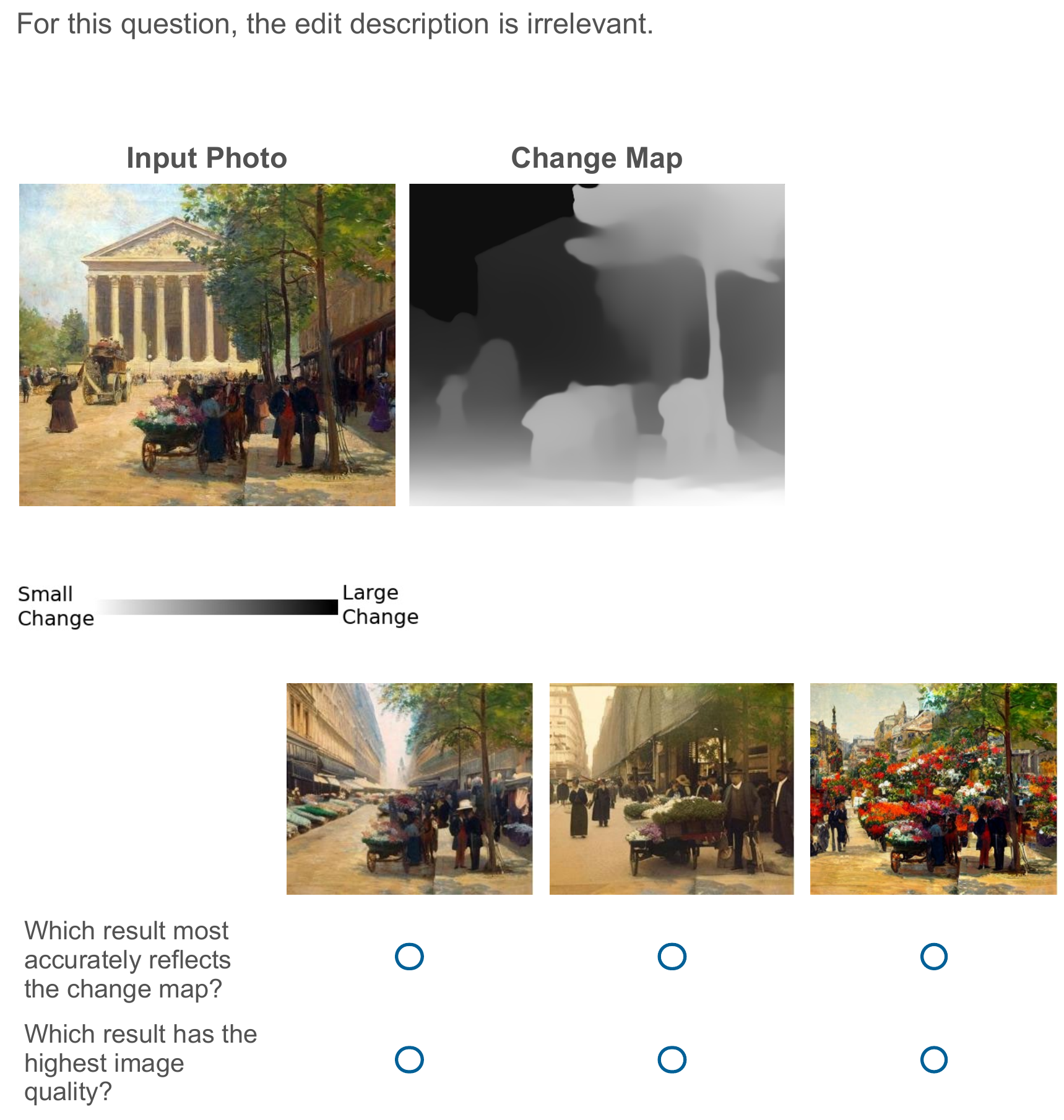}

  \rule{\linewidth}{0.36pt} %
  \\
      \includegraphics[width=0.35\linewidth, margin=5pt]{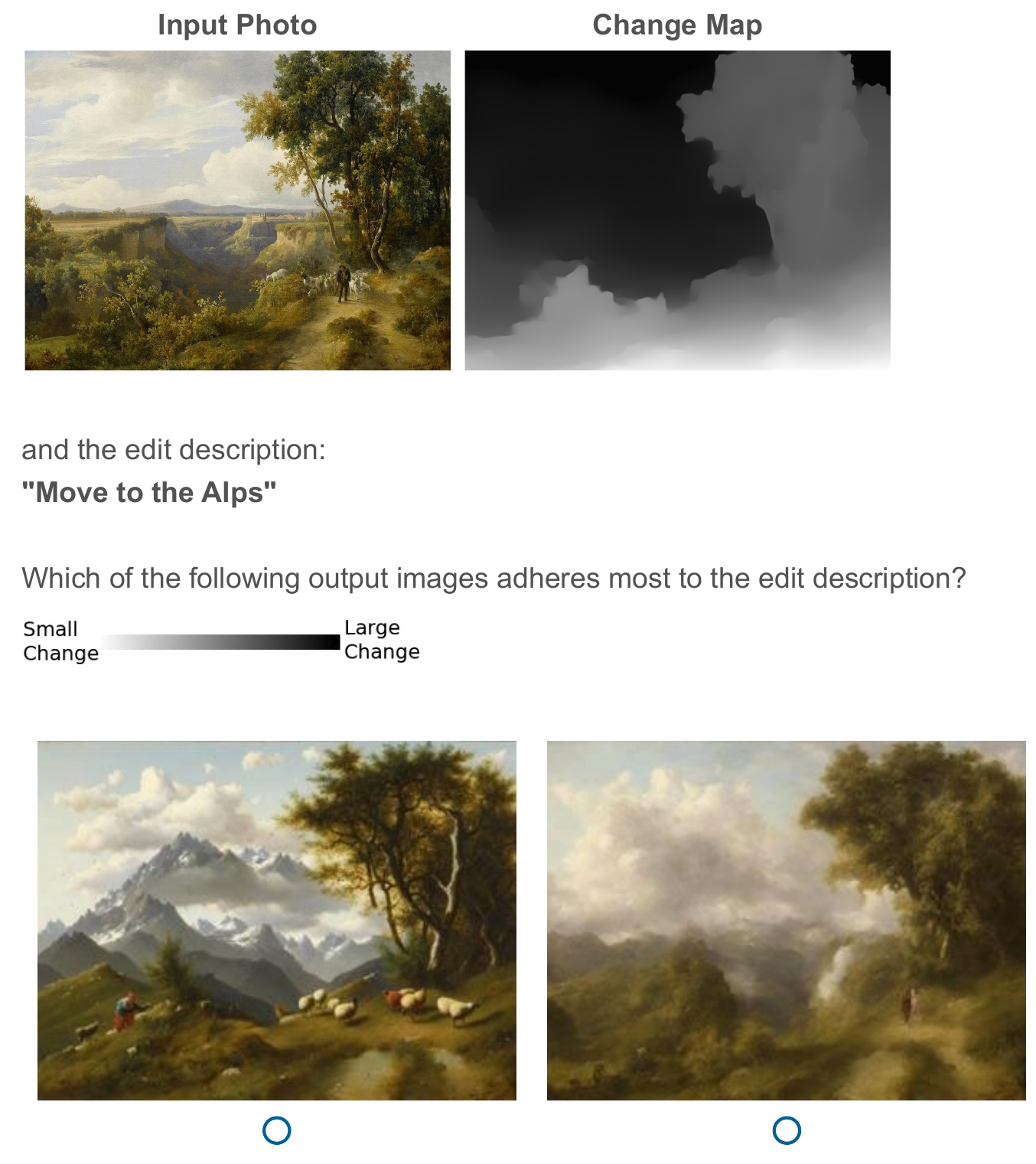}
 \caption{
 \textbf{User study questions.} Users were asked to match maps, judge different methods to produce images, and distinguish guided and unguided inference processes.}
   \label{fig:user-study-screenshots}

\end{figure*}

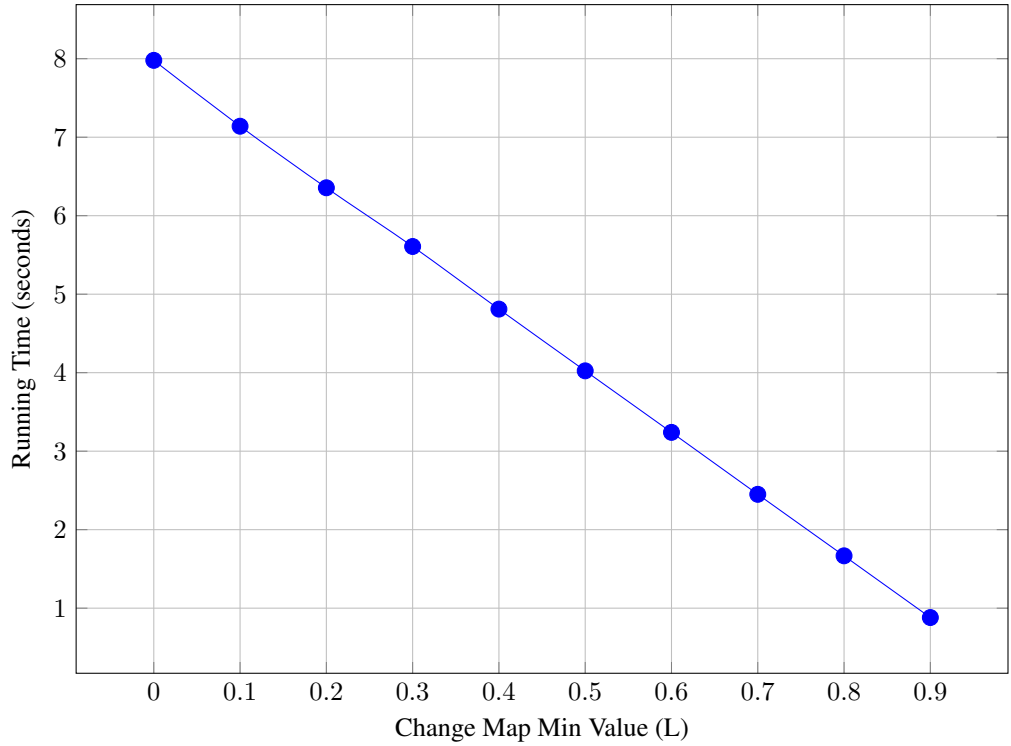
\begin{figure*}
    \centering
    \begin{tikzpicture}
        \begin{axis}[
            xlabel={Change Map Min Value (L)},
            ylabel={Running Time  (seconds)},
            xtick=data,
            legend style={at={(0.5,-0.2)}, anchor=north},
            width=0.8\textwidth,
            height=0.6\textwidth,
            grid=both,
        ]
        
        \addplot[mark=*, mark size=3pt, blue, smooth] table {
            0.0 7.9792
            0.1 7.1399
            0.2 6.3557
            0.3 5.6077
            0.4 4.8100
            0.5 4.0235
            0.6 3.2397
            0.7 2.4504
            0.8 1.6668
            0.9 0.8805
        };
        
        \end{axis}
    \end{tikzpicture}
    \caption{\textbf{Running time with skipping.} The running time of the algorithm with skipping versus the min value of the change-map. As can be seen, the relation is almost linear.}
    \label{fig:runtime-min-value}
\end{figure*}

\newcommand{\jj}{0.18\textwidth}
\newcommand{\kk}{5pt}

\begin{figure*}[t]
\centering
\begin{tabular}{c@{\hskip 0.5cm}cccccc} 
\vspace{-0.1cm}  \includegraphics[width=\jj,height=\kk,frame]{imgs/str_fan/fans/null.jpg}  &
\includegraphics[width=\jj,height=\kk,frame]{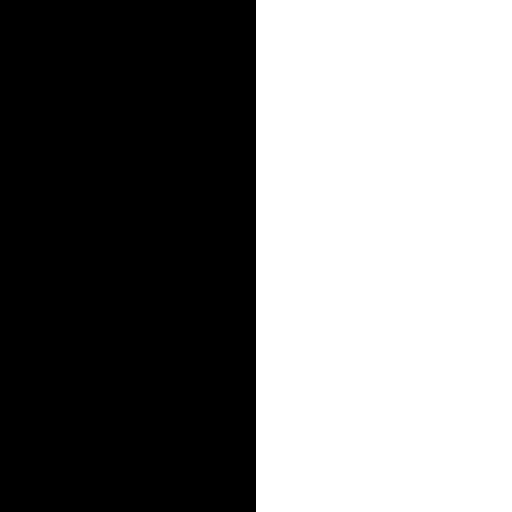} &
\includegraphics[width=\jj,height=\kk,frame]{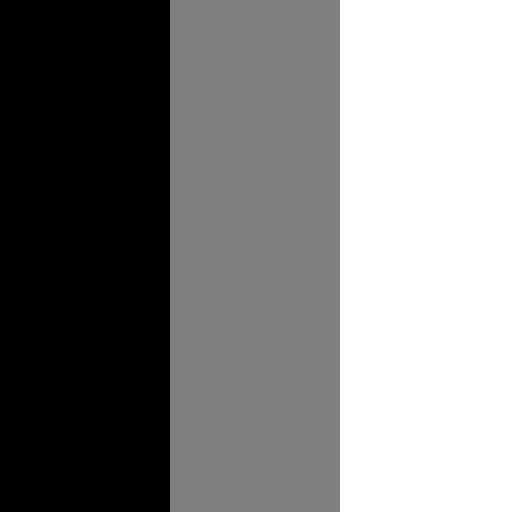} &
\includegraphics[width=\jj,height=\kk,frame]{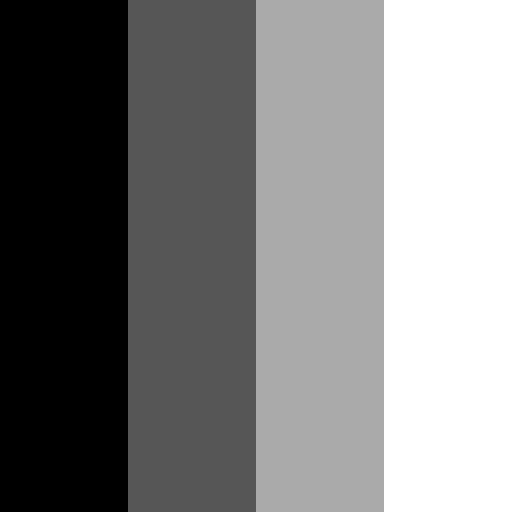} &
\includegraphics[width=\jj,height=\kk,frame]{imgs/str_fan/fans/stripe_5.jpg} \\
\includegraphics[width=\jj,frame]{imgs/str_fan/chinese_house/18778853150_446d91010b_c.jpg} &
\includegraphics[width=\jj,frame]{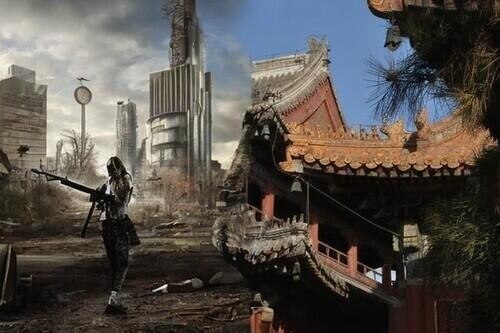} &
\includegraphics[width=\jj,frame]{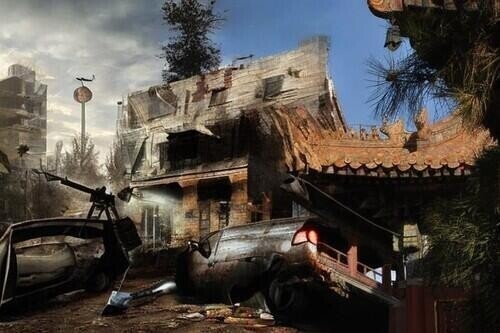} &
\includegraphics[width=\jj,frame]{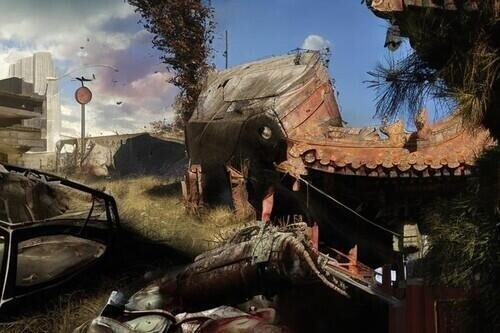} &
\includegraphics[width=\jj,frame]{imgs/str_fan/chinese_house/augmented_53_of_18778853150_446d91010b_c_5.jpg} \\
\multicolumn{5}{c}{post-apocalyptic} \\[0.5ex]
\vspace{-0.1cm}  \includegraphics[width=\jj,height=\kk,frame]{imgs/str_fan/fans/null.jpg}  &
\includegraphics[width=\jj,height=\kk,frame]{imgs/str_fan/fans/stripe_2.jpg} &
\includegraphics[width=\jj,height=\kk,frame]{imgs/str_fan/fans/stripe_3.jpg} &
\includegraphics[width=\jj,height=\kk,frame]{imgs/str_fan/fans/stripe_4.jpg} &
\includegraphics[width=\jj,height=\kk,frame]{imgs/str_fan/fans/stripe_5.jpg} \\
\includegraphics[width=\jj,frame]{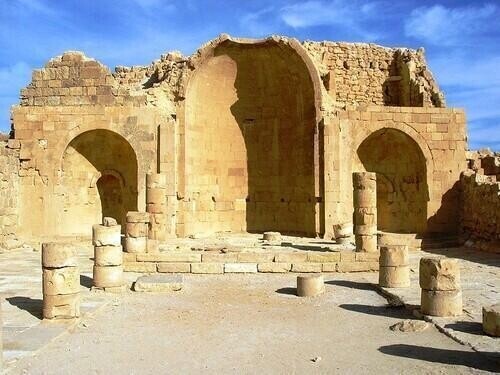} &
\includegraphics[width=\jj,frame]{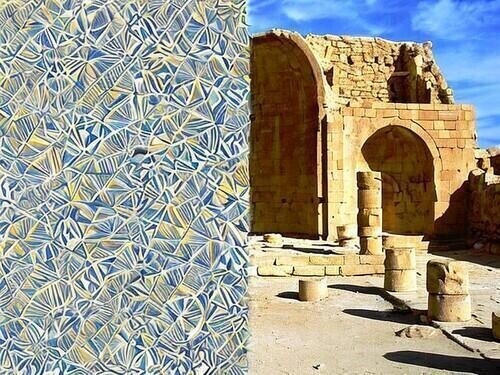} &
\includegraphics[width=\jj,frame]{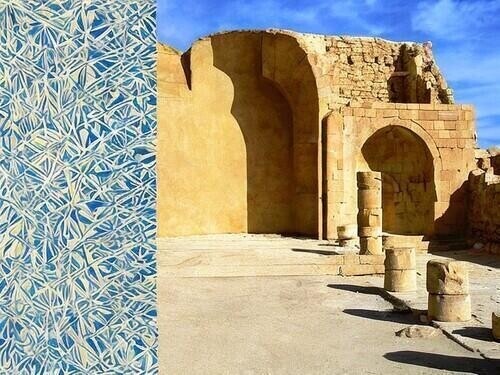} &
\includegraphics[width=\jj,frame]{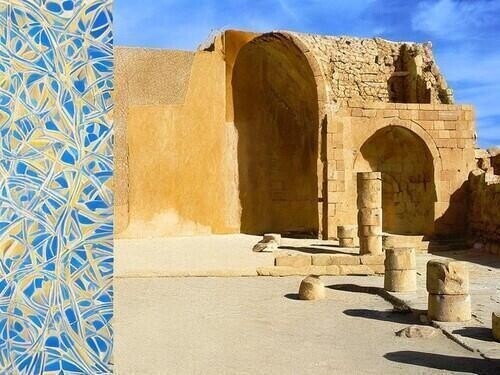} &
\includegraphics[width=\jj,frame]{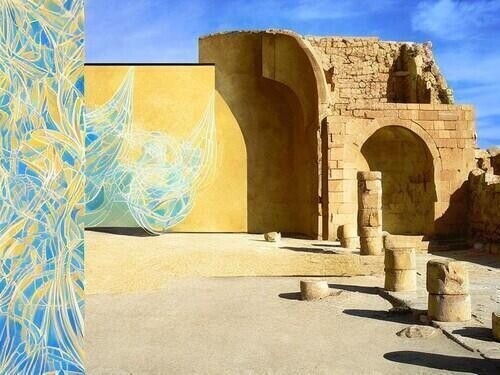} \\
\multicolumn{5}{c}{Enigmatic abstract patterns} \\[0.5ex]
\vspace{-0.1cm}  \includegraphics[width=\jj,height=\kk,frame]{imgs/str_fan/fans/null.jpg}  &
\includegraphics[width=\jj,height=\kk,frame]{imgs/str_fan/fans/stripe_2.jpg} &
\includegraphics[width=\jj,height=\kk,frame]{imgs/str_fan/fans/stripe_3.jpg} &
\includegraphics[width=\jj,height=\kk,frame]{imgs/str_fan/fans/stripe_4.jpg} &
\includegraphics[width=\jj,height=\kk,frame]{imgs/str_fan/fans/stripe_5.jpg} \\
\includegraphics[width=\jj,frame]{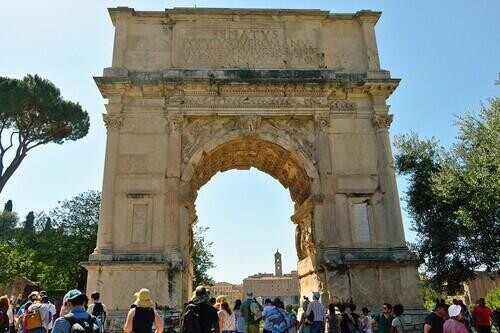} &
\includegraphics[width=\jj,frame]{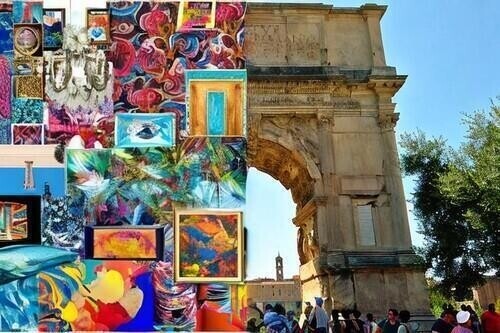} &
\includegraphics[width=\jj,frame]{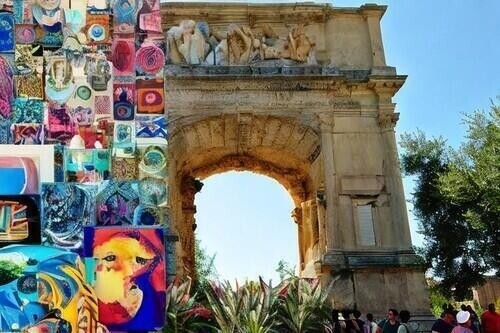} &
\includegraphics[width=\jj,frame]{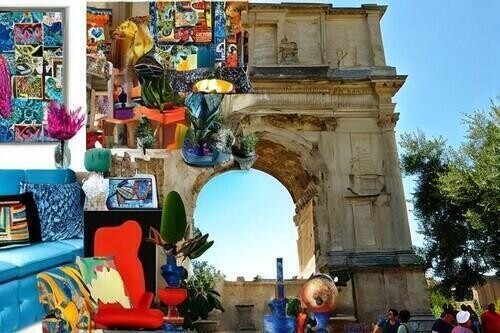} &
\includegraphics[width=\jj,frame]{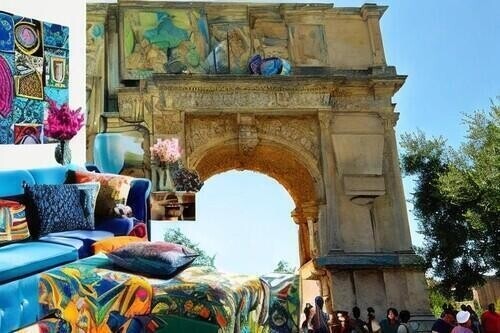} \\
\multicolumn{5}{c}{maximalism} \\[0.5ex]
\end{tabular}
\caption{\textbf{Strength fans.}
For each input image (left), we show strength fans ranging
from two to five strength values. Each strength position and value is represented by the upper bar. Prompts are written below each row. The strength fans allow users to explore and compare the effect of different strength values, and help them find the desired strength they wish to apply. 
}
\label{fig:str_fan_general}
\end{figure*}

\newcommand{\oo}{0.18\textwidth}
\newcommand{\nn}{5pt}

\begin{figure*}[t]
\centering
\begin{tabular}{c@{\hskip 0.5cm}cccc} 
\vspace{-0.1cm}  \includegraphics[width=\oo,height=\nn,frame]{imgs/str_fan/fans/null.jpg}  &
\includegraphics[width=\oo,height=\nn,frame]{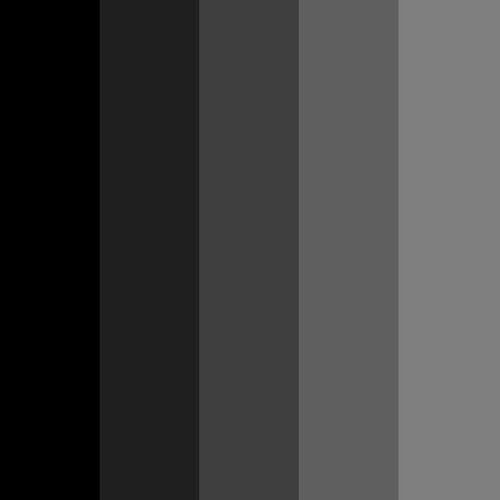} &
\includegraphics[width=\oo,height=\nn,frame]{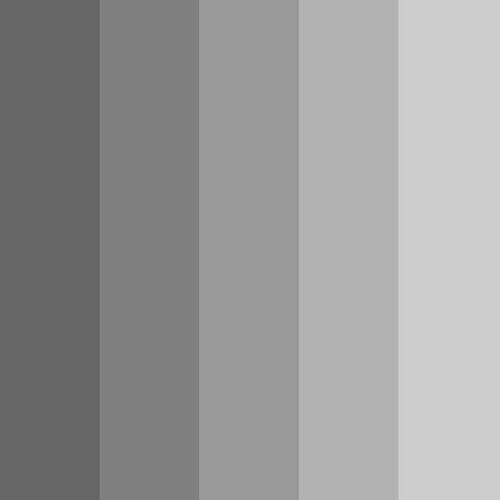} &
\includegraphics[width=\oo,height=\nn,frame]{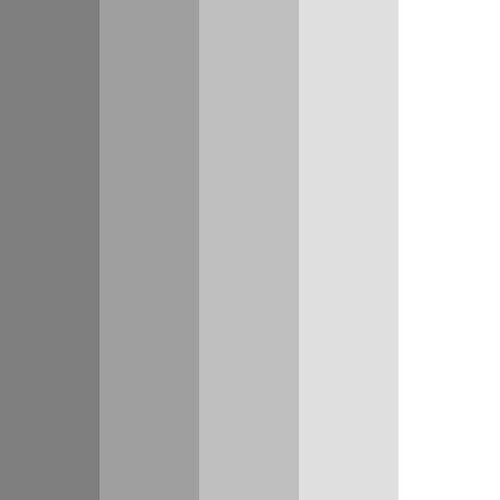}  \\
\includegraphics[width=\oo,frame]{imgs/str_fan/chinese_house/18778853150_446d91010b_c.jpg} &
\includegraphics[width=\oo,frame]{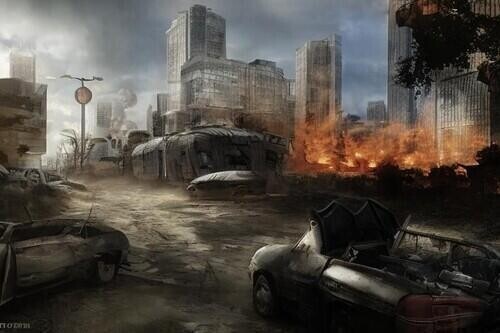} &
\includegraphics[width=\oo,frame]{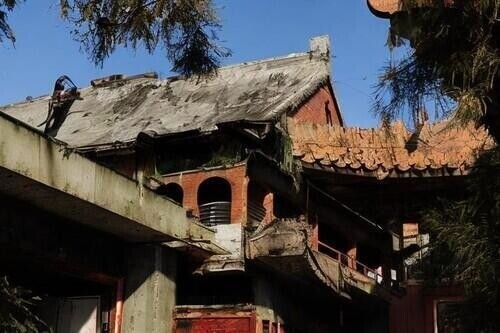} &
\includegraphics[width=\oo,frame]{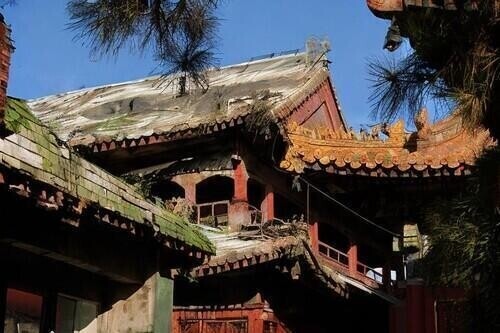}  \\
\vspace{-0.1cm}
\includegraphics[width=\oo,height=\nn,frame]{imgs/str_fan/fans/null.jpg}  &
\includegraphics[width=\oo,height=\nn,frame]{imgs/str_fan/fans/dark.jpg} &
\includegraphics[width=\oo,height=\nn,frame]{imgs/str_fan/fans/centeral.jpg} &
\includegraphics[width=\oo,height=\nn,frame]{imgs/str_fan/fans/light.jpg}  \\
\includegraphics[width=\oo,frame]{imgs/str_fan/shivta/5611370207_16aa56b60f_c.jpg} &
\includegraphics[width=\oo,frame]{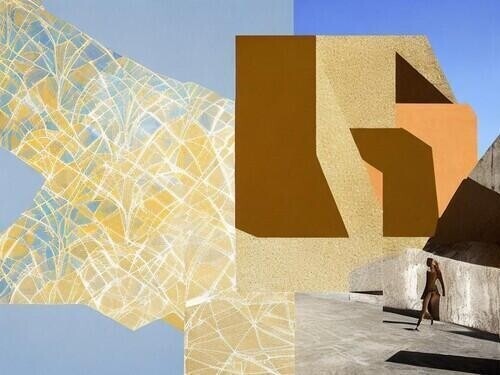} &
\includegraphics[width=\oo,frame]{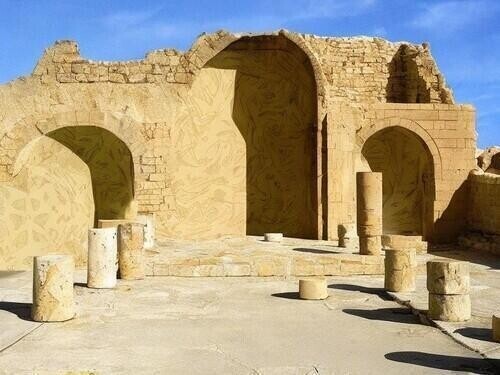} &
\includegraphics[width=\oo,frame]{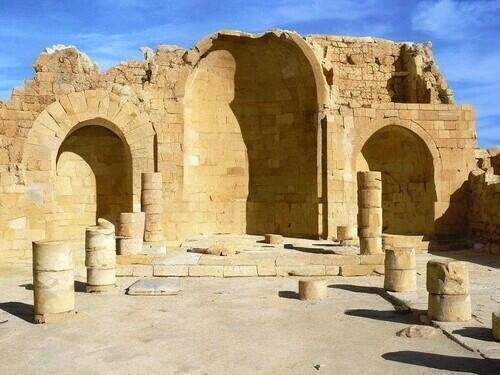} 
\\
\vspace{-0.1cm}
\includegraphics[width=\oo,height=\nn,frame]{imgs/str_fan/fans/null.jpg}  &
\includegraphics[width=\oo,height=\nn,frame]{imgs/str_fan/fans/dark.jpg} &
\includegraphics[width=\oo,height=\nn,frame]{imgs/str_fan/fans/centeral.jpg} &
\includegraphics[width=\oo,height=\nn,frame]{imgs/str_fan/fans/light.jpg}  \\
\includegraphics[width=\oo,frame]{imgs/str_fan/titus/27705711433_f695e2446a_c.jpg} &
\includegraphics[width=\oo,frame]{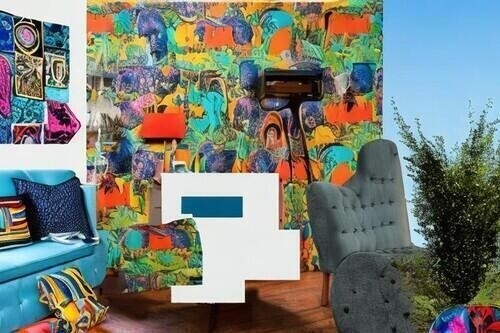} &
\includegraphics[width=\oo,frame]{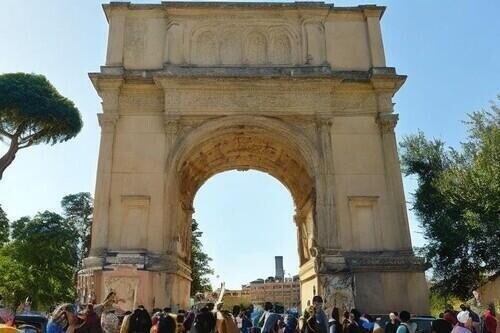} &
\includegraphics[width=\oo,frame]{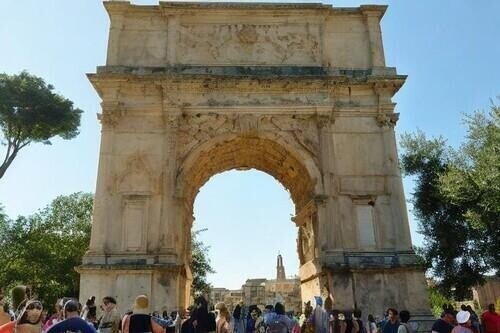} 
\end{tabular}
\caption{\textbf{Strength fans in various magnitudes and offsets.} Users can accurately pick the strength values they wish to investigate. 
For example, a user can start by investigating the full spectrum of strength values, and then zoom in on a specific region to precisely choose a strength parameter.
The prompts and seeds are the same as in \Cref{fig:str_fan_general}.
}
\label{fig:str_fan_tuning}
\end{figure*}

\newcommand{\mm}{0.16\textwidth}
\begin{figure*}
\begin{subfigure}{0.49\columnwidth}
\includegraphics[width=\linewidth,frame]{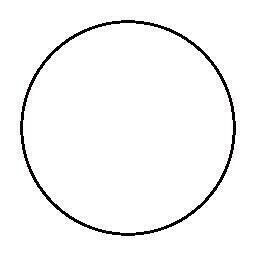}
\caption{Map 1}
\end{subfigure}
\hfill
\begin{subfigure}{0.49\columnwidth}
\includegraphics[width=\linewidth,frame]{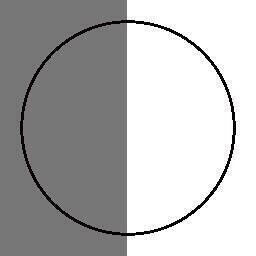}
\caption{\textbf{CAM: 0.43}, DAM: 39.37
}
\end{subfigure}
\hfill
\begin{subfigure}{0.49\columnwidth}
\includegraphics[width=\linewidth,frame]{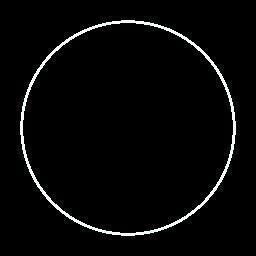}
\caption{CAM: -1, \textbf{DAM: 0}
}
\end{subfigure}
\hfill
\begin{subfigure}{0.49\columnwidth}
\includegraphics[width=\linewidth,frame]{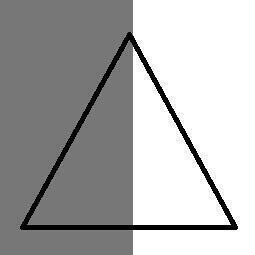}
\caption{CAM: 0.01, DAM: 43.58
}
\end{subfigure}
  \caption{\textbf{CAM \& DAM properties demonstration.} Comparing CAM and DAM scores for Map 1 with three alternative options for the second map.
  Mind that these metrics operate in opposite directions.
  The comparing reveals intriguing insights. CAM selects 'b' as the most similar to 'a' out of the three, whereas DAM indicates maximal similarity to 'c'. Also note that CAM indicates minimal similarity to the pair. Because DAM is sensitive to regional shade changes, despite 'b' pixels changing less dramatically, DAM indicates less similarity compared to 'c'. Despite 'b' and 'd' differing only in a stroke, their CAM scores exhibit a significant disparity, while their DAM scores remain relatively consistent. This discrepancy is particularly interesting as it demonstrates CAM's intolerance to shape variations.
  This underscores the importance of considering both CAM and DAM metrics for a comprehensive assessment of map similarity.}
\label{cam-dam-couterexample}
\end{figure*}

\newcommand{\saf}{0.115\textwidth}
\newcommand{\saftwo}{0.244\textwidth}
\begin{figure*}
\centering
\begin{tabular}{cccc@{\hspace{0.5cm}}cccc}
\textbf{Image} & \textbf{Mask} & \textbf{Blurred Mask} & \textbf{Output} &
\textbf{Image} & \textbf{Mask} & \textbf{Blurred Mask} & \textbf{Output} \\[0.1cm]
\includegraphics[width=\saf,frame]{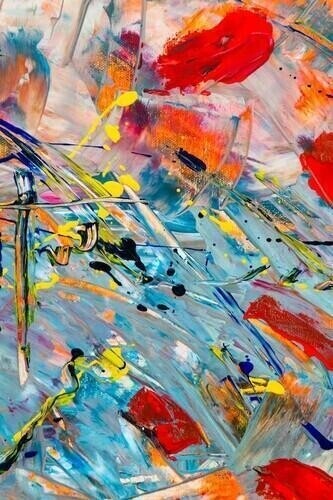} &
\includegraphics[width=\saf,frame]{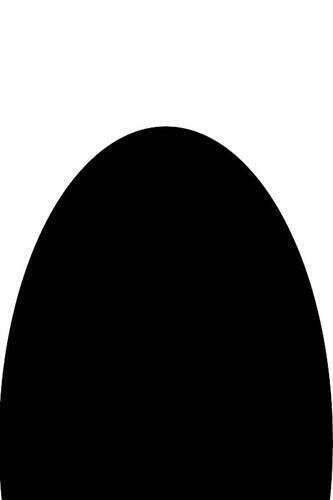} &
\includegraphics[width=\saf,frame]{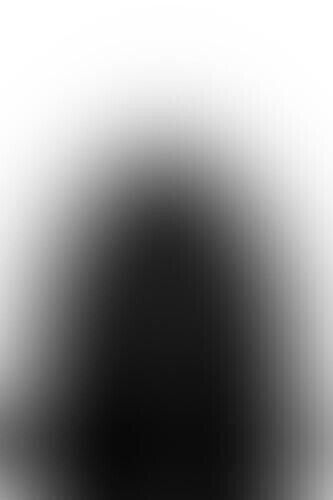} &
\includegraphics[width=\saf,frame]{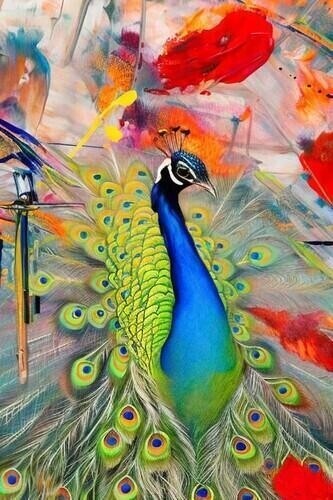} &
\includegraphics[width=\saf,frame]{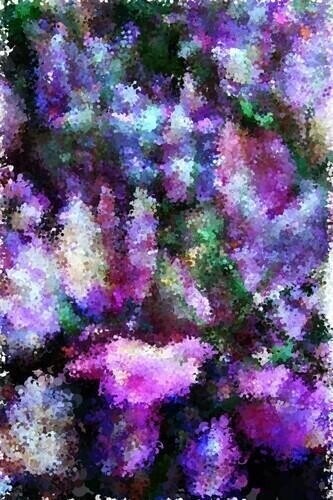} &
\includegraphics[width=\saf,frame]{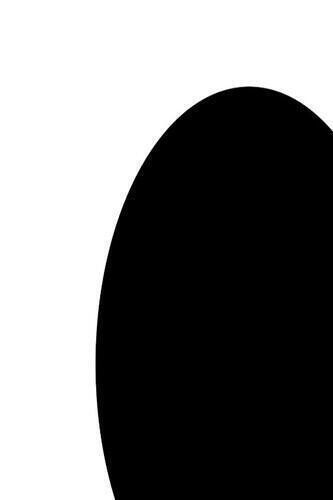} &
\includegraphics[width=\saf,frame]{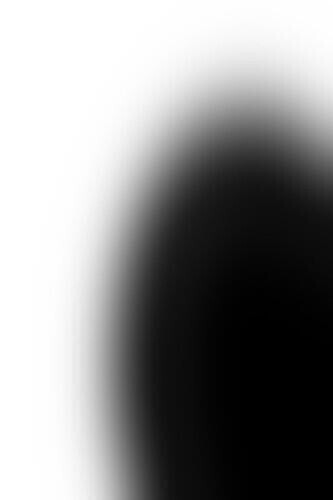} &
\includegraphics[width=\saf,frame]{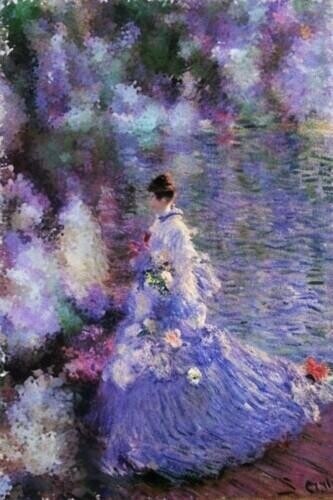} 
\end{tabular}
\vspace{0.2cm}

\centering
\begin{tabular}{cccc}
\textbf{Image} & \textbf{Mask} & \textbf{Blurred Mask} & \textbf{Output} \\ [0.1cm]
\includegraphics[width=\saftwo,frame]{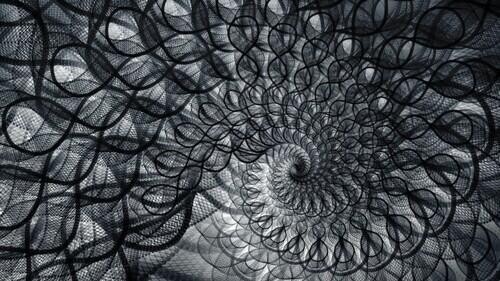} &
\includegraphics[width=\saftwo,frame]{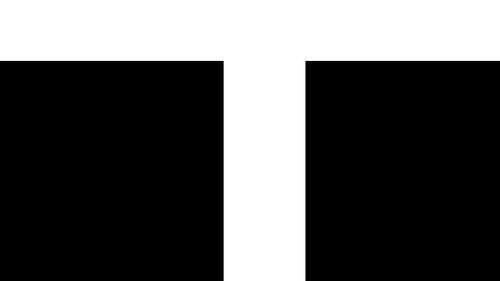} &
\includegraphics[width=\saftwo,frame]{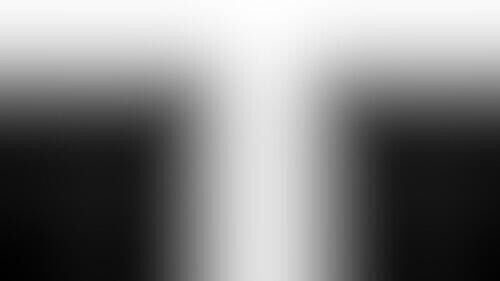} &
\includegraphics[width=\saftwo,frame]{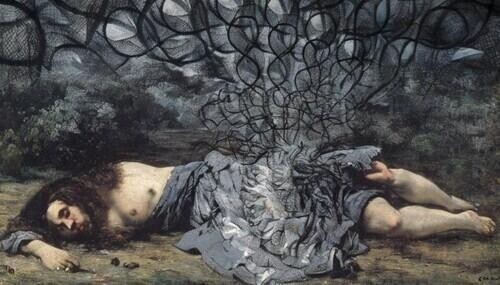} 

\end{tabular}
\caption{\textbf{Additional soft-inpainting results.} Our soft-inpainting results blend smoothly with the backgrounds. Prompts: ``peacock, realism'', ``Camille Monet'', ``Gustave Courbet''. Blurring radii: 64px.  
}
\label{fig:peacock}
\end{figure*}

\newcommand{\salgrule}[1][.2pt]{\par\vskip.5\baselineskip\hrule height #1\par\vskip.5\baselineskip}

\begin{algorithm}[t]
\begin{flushleft}
\caption{Differential Image to Image Diffusion With Skipping}
\label{s_algo-with-skipping}

\hspace*{\algorithmicindent} \textbf{Input} $x$ (image to edit), $k$ (number of steps), $\mu$ (change map with values between 0 and 1), $p$ (prompt) \\
\hspace*{\algorithmicindent} \textbf{Output} $\xhat$
\begin{algorithmic}[1]
\Procedure{inference}{$x$, $k$, $\mu$, $p$}

        \State $z_{init}$ = ldm\_encode($x$)
        \label{s_algo-encode}
        \State $\mu_s$ = down\_sample($\mu$)
        \label{s_algo-downsample}
                \State ${L=\left\lfloor \left(1-\min(\mu_{s})\right)\cdot k\right\rfloor }$
        \State $z'_L$ = add\_noise($z_{init}$, $L$) \label{s_algo-add-noise}

        \State $z_L$ = denoise($z'_L$, $p$, $L$)
        \For{t = $L-1$ to $0$} \label{s_algo-loop-begins}
            
                \State $z_t'$ = add\_noise($z_{init}$, $t$)
                \State $mask$ = $\mu_s \ewlteq \frac{k-t}{k}$ \label{s_no-future-hinting-change}
                \State $z_t^{mix}$ = $z_{t+1} \odot mask +  z_t' \odot (1 - mask)$ \label{s_injection}
                \State $z_t$ = denoise($z_t^{mix}$, $p$, $t$)

        \EndFor
    \State $\xhat$ = ldm\_decode($z_0$)
     \State return $\xhat$

\EndProcedure
\salgrule

\noindent We denote $\ewlteq$, $\odot$ as element-wise less-than and element-wise multiplication, respectively.
$\ewlteq$ returns a tensor of 1s and 0s.
\end{algorithmic}
\end{flushleft}
\end{algorithm}

\begin{table}[htbp]
    \centering
    \setlength{\tabcolsep}{3.4pt}
\begin{tabular}{cccc}
        \toprule
        Min Value & Running Time~(sec) & Delta (sec) & Delta (\%) \\
        \midrule
        no skipping & 8.0 & --- & ---\\ 
        0.1 & 7.14 & -0.86 & -10.74\% \\
        0.2 & 6.36 & -1.64 & -20.55\% \\
        0.3 & 5.61 & -2.39 & -29.90\% \\
        0.4 & 4.81 & -3.19 & -39.87\% \\
        0.5 & 4.02 & -3.98 & -49.70\% \\
        0.6 & 3.24 & -4.76 & -59.51\% \\
        0.7 & 2.45 & -5.55 & -69.37\% \\
        0.8 & 1.67 & -6.33 & -79.17\% \\
        0.9 & 0.88 & -7.12 & -88.99\% \\
        \bottomrule
    \end{tabular}
    \caption{The averaged running time as a function of the minimum value of the map, with delta compared to no skipping. As can be seen, skipping can boost the algorithm's performance dramatically.}
    \label{tbl:skipping-runtime}
\end{table}

\clearpage
\subsection{Results With Automatic Change Maps}
\subsubsection{Constant Change Map}

\begin{figure}[h]

  \centering
  \begin{tabularx}{\columnwidth}{p{1.01\xx}p{1.01\xx}@{\hskip 0.2cm}p{1.01\xx}}

\multicolumn{2}{c}{\textbf{Input Image + Change Map}} & \multicolumn{1}{c}{\textbf{Output Image}} \\[0.2cm] %

\includegraphics[width=\xx,frame]{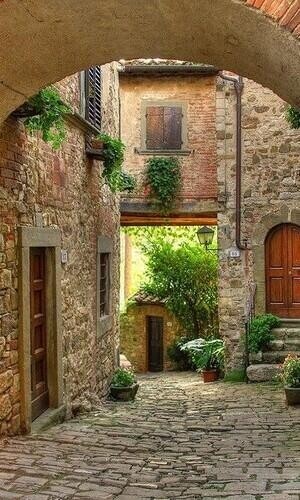} &
\includegraphics[width=\xx,frame]{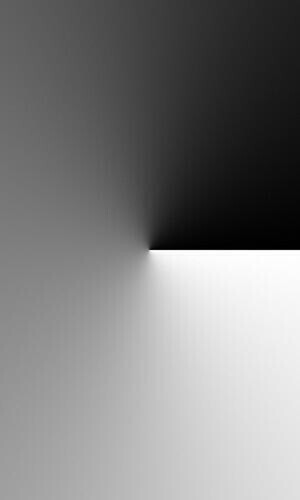} &
\includegraphics[width=\xx,frame]{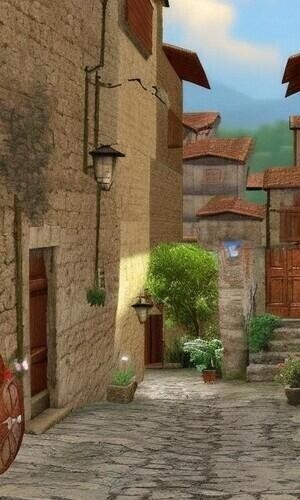} \\[-0.1cm] \multicolumn{3}{X}{\scriptsize{``Medieval village of Montefioralle in Nintendogs game''}} \\[0.2cm] %
\includegraphics[width=\xx,frame]{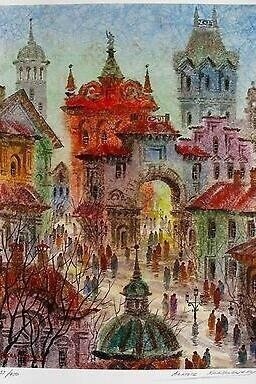} & 
\includegraphics[width=\xx,frame]{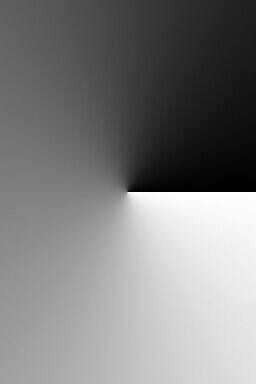} &

\includegraphics[width=\xx,frame]{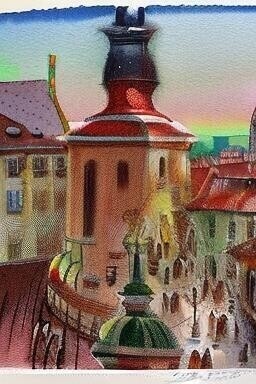} \\[-0.1cm] \multicolumn{3}{X}{\scriptsize{``"""ANATOLE KRASNYANSKY """"MEMORIES OF PRAGUE"""" Watercolor Painting"""''}} \\[0.5cm]

\includegraphics[width=\xx,frame]{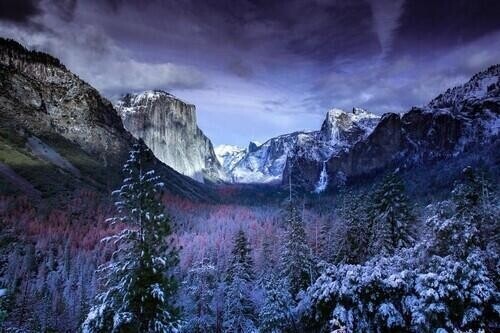} &
\includegraphics[width=\xx,frame]{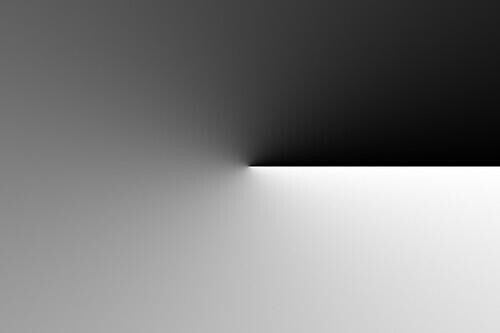} &
\includegraphics[width=\xx,frame]{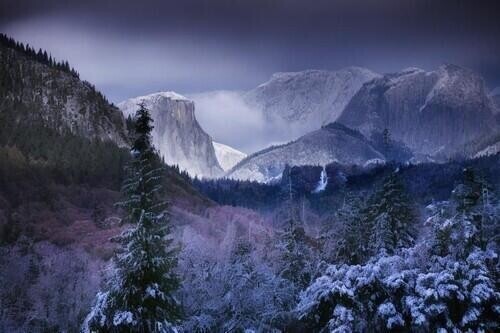} \\[-0.1cm] \multicolumn{3}{X}{\scriptsize{``photo of mountains and trees, with a castle''}} \\[0.2cm] %
\includegraphics[width=\xx,frame]{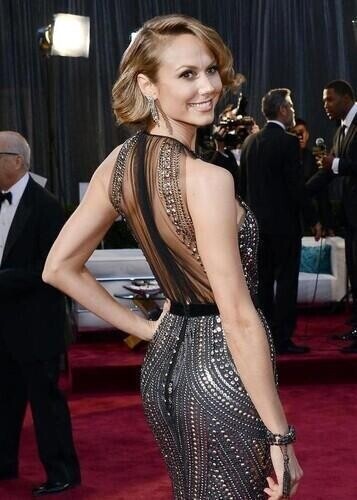} & 
\includegraphics[width=\xx,frame]{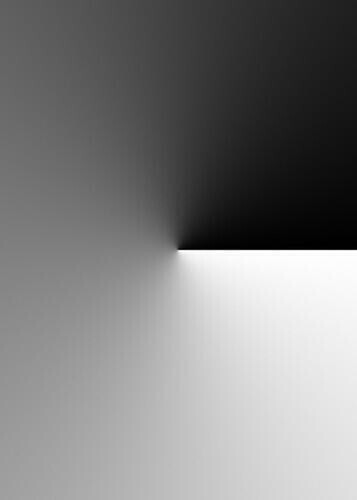} &
\includegraphics[width=\xx,frame]{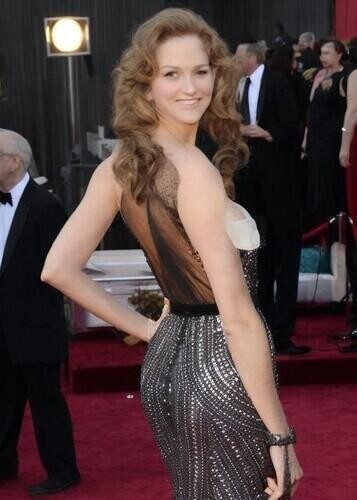} \\
\end{tabularx}
\label{sec:auto-change-maps}
\end{figure}

\begin{figure}[t]

  \centering
  \begin{tabularx}{\columnwidth}{p{1.01\xx}p{1.01\xx}@{\hskip 0.2cm}p{1.01\xx}}

\multicolumn{2}{c}{\textbf{Input Image + Change Map}} & \multicolumn{1}{c}{\textbf{Output Image}} \\[0.2cm] %

\includegraphics[width=\xx,frame]{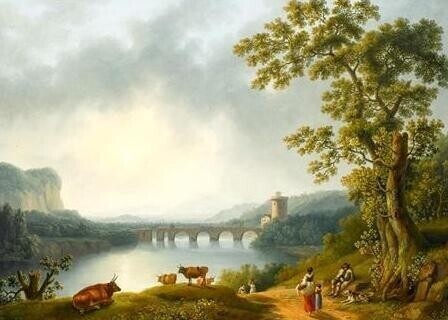} &
\includegraphics[width=\xx,frame]{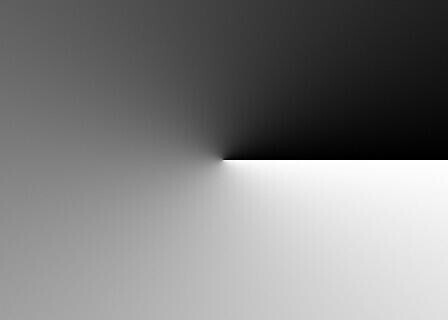} & 

\includegraphics[width=\xx,frame]{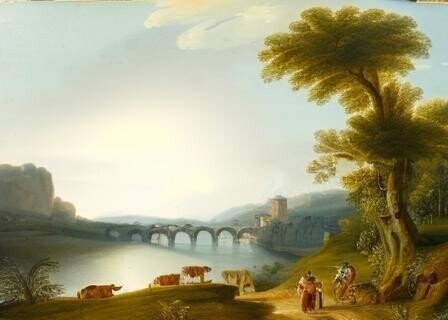} \\[-0.1cm] \multicolumn{3}{X}{\scriptsize{``Jacob Philipp Hackert The Volturno with the Ponte Margherita, near Caserta, with a Herdsman Resting and Peasants on a Path and a Dragon''}} \\[0.5cm] %
\includegraphics[width=\xx,frame]{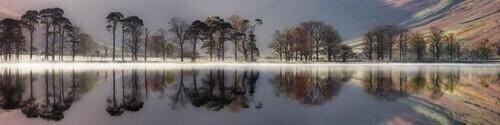} & 
\includegraphics[width=\xx,frame]{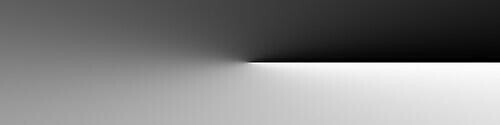} & 

\includegraphics[width=\xx,frame]{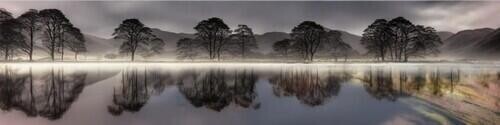} \\[-0.1cm] \multicolumn{3}{X}{\scriptsize{``Buttermere Trees Silhouette with a purple filter - Canvas Prints''}} \\[0.2cm]

\includegraphics[width=\xx,frame]{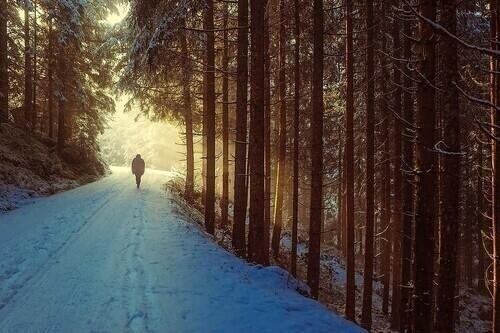} & 
\includegraphics[width=\xx,frame]{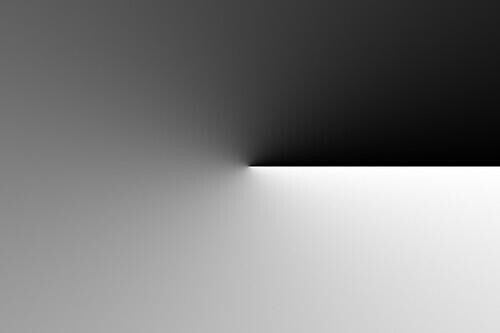} & 

\includegraphics[width=\xx,frame]{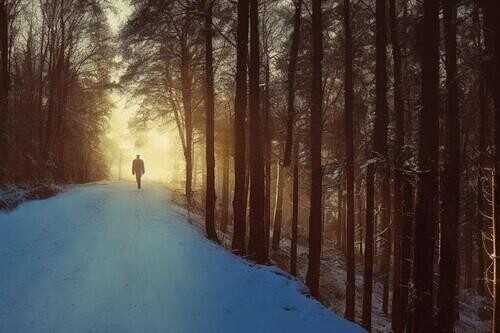} \\[-0.1cm] \multicolumn{3}{X}{\scriptsize{``A person wearing a suit through a snowy forest in the winter.''}} \\[0.2cm]
\includegraphics[width=\xx,frame]{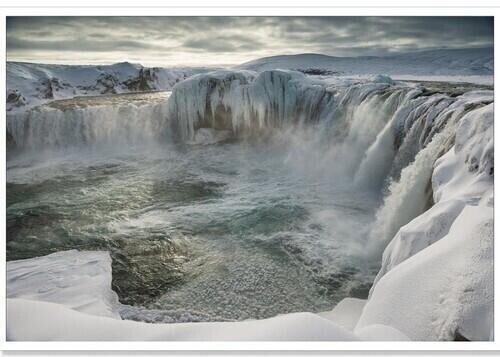} &
\includegraphics[width=\xx,frame]{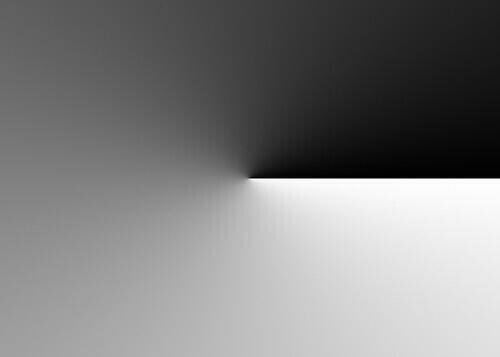} & 

\includegraphics[width=\xx,frame]{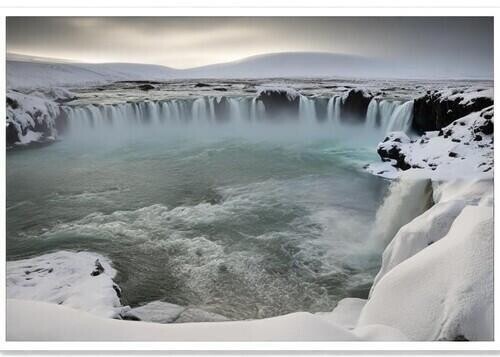} \\
\end{tabularx}

\end{figure}

\clearpage

\begin{figure}[t]

  \centering
  \begin{tabularx}{\columnwidth}{p{1.01\xx}p{1.01\xx}@{\hskip 0.2cm}p{1.01\xx}}

\multicolumn{2}{c}{\textbf{Input Image + Change Map}} & \multicolumn{1}{c}{\textbf{Output Image}} \\[0.2cm] %

\includegraphics[width=\xx,frame]{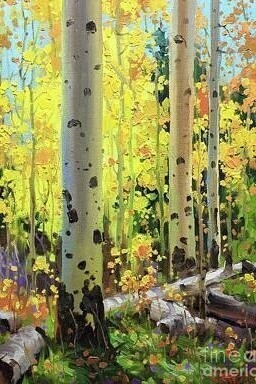} &
\includegraphics[width=\xx,frame]{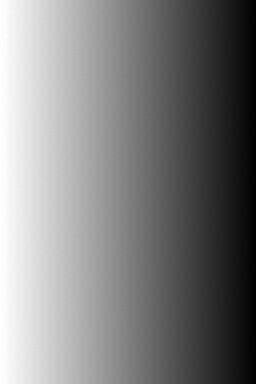} &
\includegraphics[width=\xx,frame]{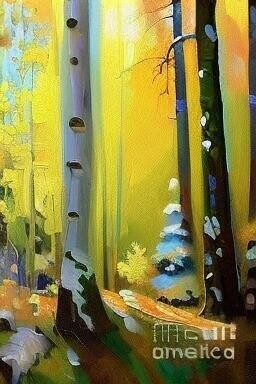} \\[-0.1cm] \multicolumn{3}{X}{\scriptsize{``Winter Forest Symphony II Art Print by Gary Kim''}} \\[0.2cm] %
\includegraphics[width=\xx,frame]{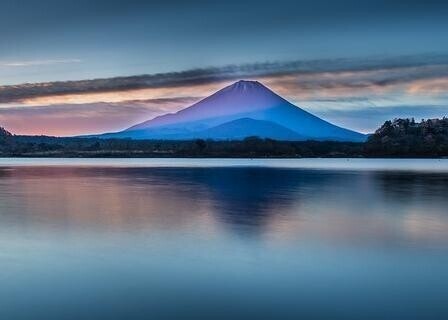} & 
\includegraphics[width=\xx,frame]{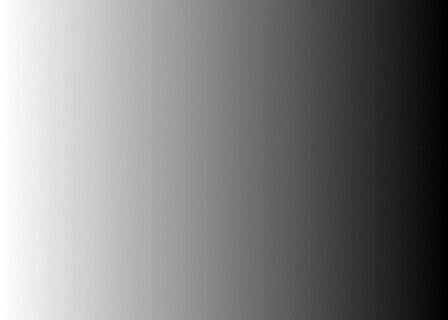} & 
\includegraphics[width=\xx,frame]{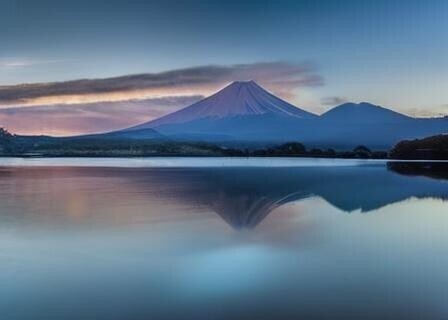} \\[-0.1cm] \multicolumn{3}{X}{\scriptsize{``lake filled with lava morning lake reflection japan sunrise canon glow  1635mm lakeshoji redfuji 5dmarkiii''}} \\[0.5cm]

\includegraphics[width=\xx,frame]{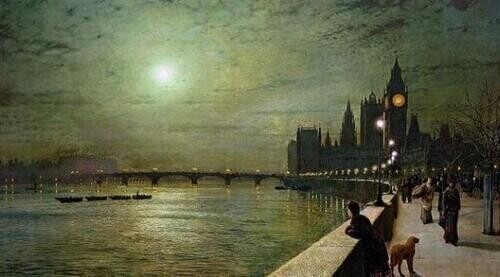} &
\includegraphics[width=\xx,frame]{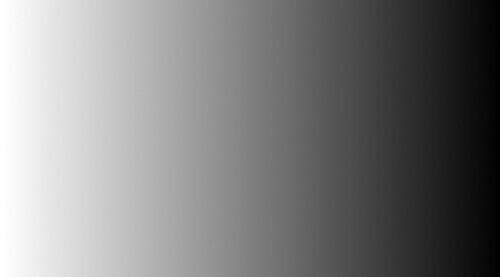} & 
\includegraphics[width=\xx,frame]{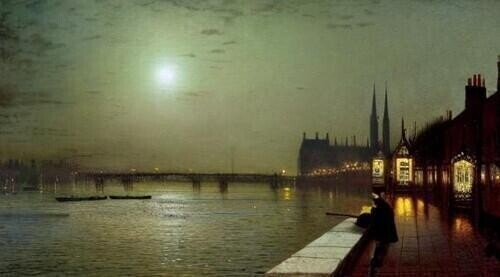} \\[-0.1cm] \multicolumn{3}{X}{\scriptsize{``John Atkinson Grimshaw, Painting on the Thames, 1880''}} \\[0.2cm] %
\includegraphics[width=\xx,frame]{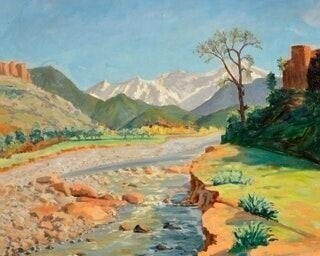} & 
\includegraphics[width=\xx,frame]{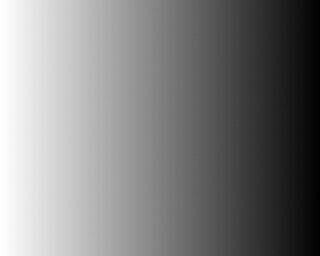} & 
\includegraphics[width=\xx,frame]{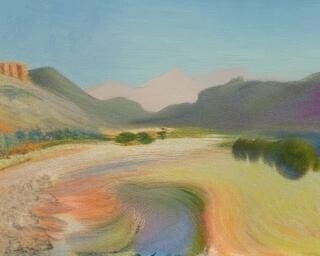} \\
\end{tabularx}

\end{figure}

\begin{figure}[t]

  \centering
  \begin{tabularx}{\columnwidth}{p{1.01\xx}p{1.01\xx}@{\hskip 0.2cm}p{1.01\xx}}

\multicolumn{2}{c}{\textbf{Input Image + Change Map}} & \multicolumn{1}{c}{\textbf{Output Image}} \\[0.2cm] %
\includegraphics[width=\xx,frame]{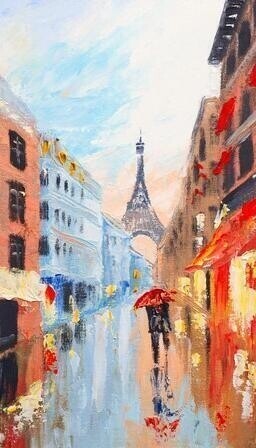} & 
\includegraphics[width=\xx,frame]{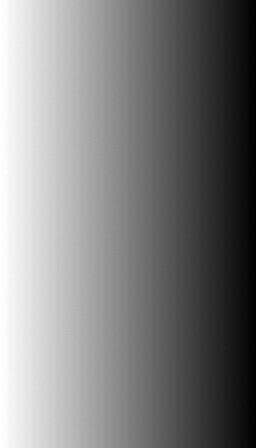} & 
\includegraphics[width=\xx,frame]{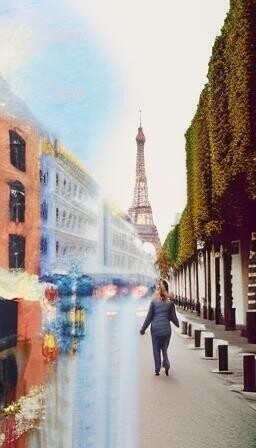} \\[-0.1cm] \multicolumn{3}{X}{\scriptsize{``photograph: couple walking on the streets of Paris against the backdrop of the Eiffel Tower,''}} \\[0.5cm]

\includegraphics[width=\xx,frame]{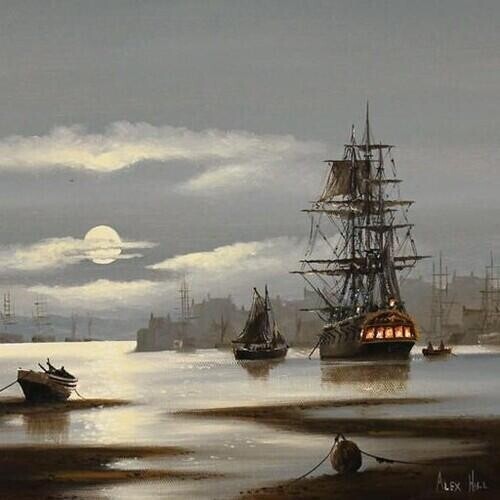} &
\includegraphics[width=\xx,frame]{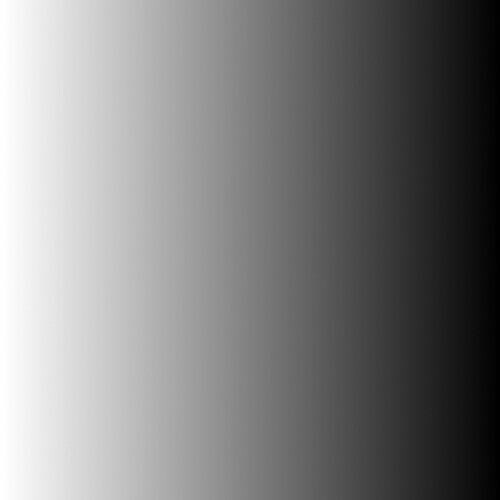} & 

\includegraphics[width=\xx,frame]{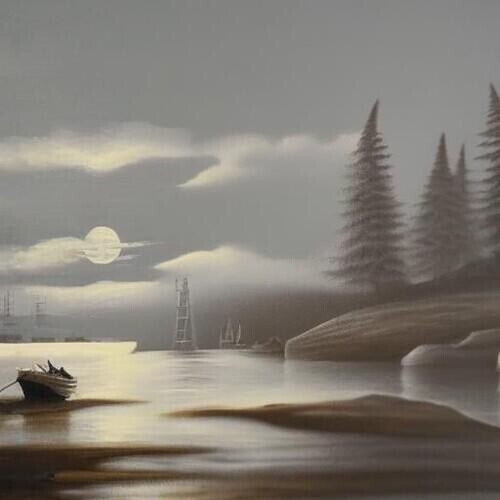} \\[-0.1cm] \multicolumn{3}{X}{\scriptsize{``Alex Hill, Original coloring book illustration, Moonlight Bay''}} \\[0.2cm] %
\includegraphics[width=\xx,frame]{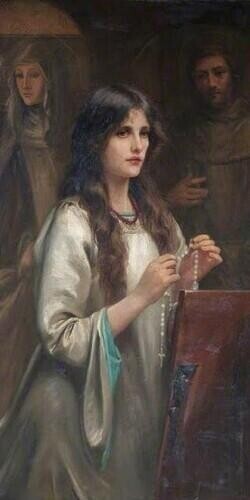} & 
\includegraphics[width=\xx,frame]{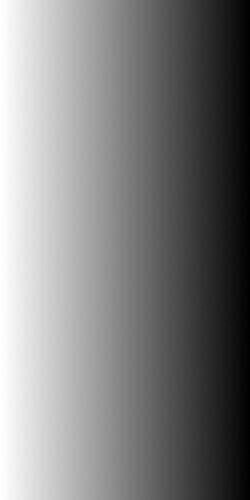} & 
\includegraphics[width=\xx,frame]{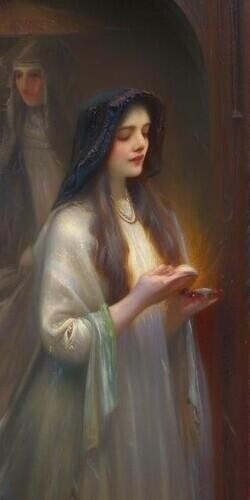} \\[-0.1cm] \multicolumn{3}{X}{\scriptsize{``The Rosary in a blizzard ~  Beatrice Offor (1864-1920)''}} \\[0.2cm]
    \includegraphics[width=\xx,frame]{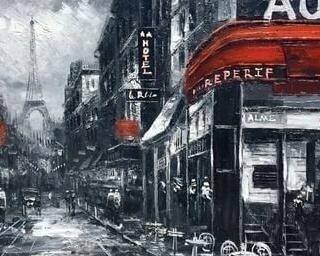} &
    \includegraphics[width=\xx,frame]{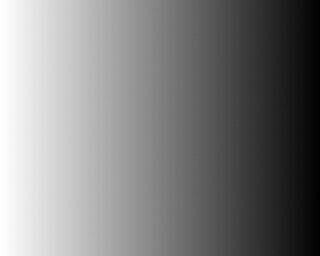} & 

    \includegraphics[width=\xx,frame]{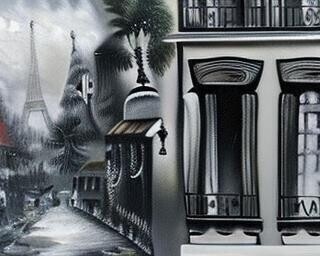} \\
\end{tabularx}

\end{figure}

\clearpage

\begin{figure}[b]

  \centering
  \begin{tabularx}{\columnwidth}{p{1.01\xx}p{1.01\xx}@{\hskip 0.2cm}p{1.01\xx}}

\multicolumn{2}{c}{\textbf{Input Image + Change Map}} & \multicolumn{1}{c}{\textbf{Output Image}} \\[0.2cm] %
    \includegraphics[width=\xx,frame]{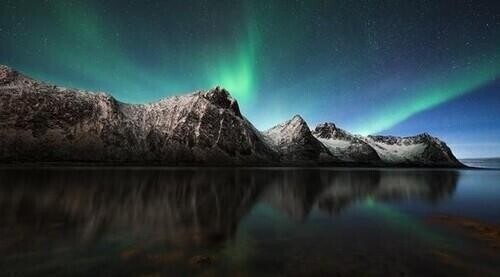} & 
    \includegraphics[width=\xx,frame]{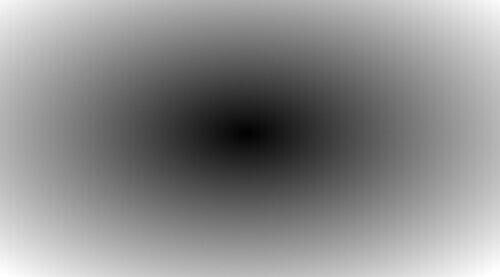} & 
\includegraphics[width=\xx,frame]{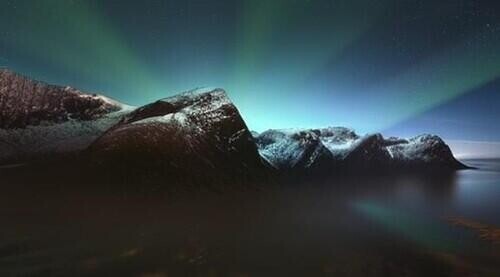} \\[-0.1cm] \multicolumn{3}{X}{\scriptsize{``Picture the sky, sun, mountains, sunset, Norway, North''}} \\[0.2cm]

    \includegraphics[width=\xx,frame]{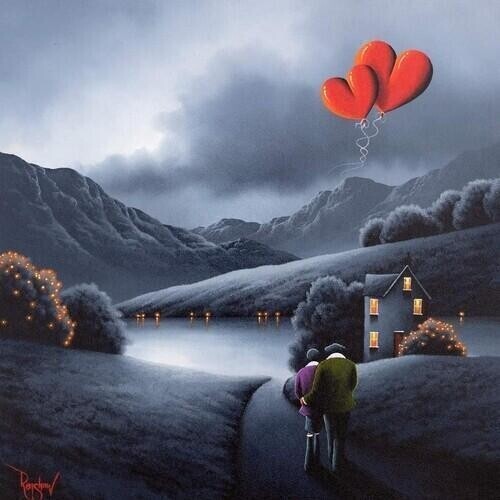} & 
    \includegraphics[width=\xx,frame]{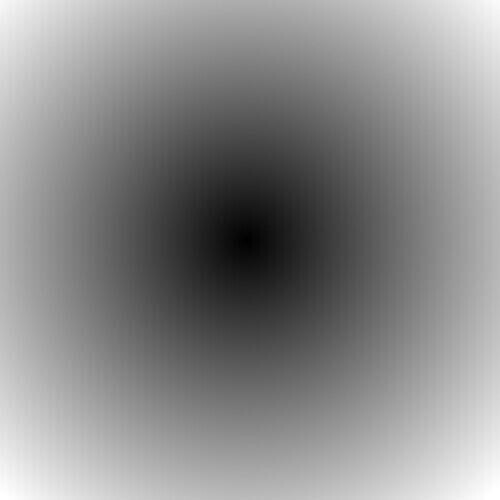} & 
\includegraphics[width=\xx,frame]{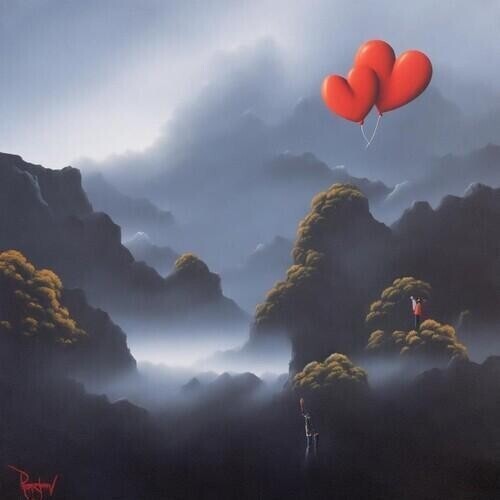} \\[-0.1cm] \multicolumn{3}{X}{\scriptsize{``Someone Like You - Orignal David Renshaw With Mohawk Framed''}} \\[0.2cm] %

    \includegraphics[width=\xx,frame]{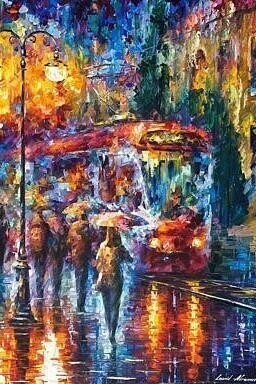} & 
    \includegraphics[width=\xx,frame]{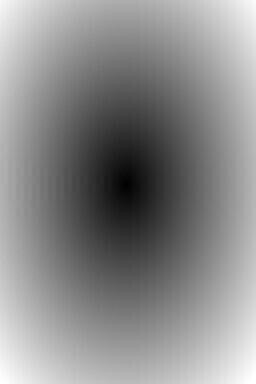} & 
\includegraphics[width=\xx,frame]{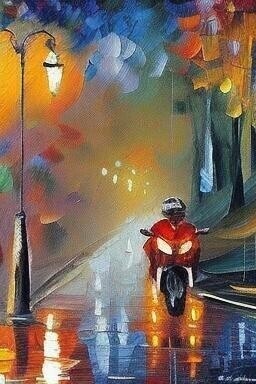} \\[-0.1cm] \multicolumn{3}{X}{\scriptsize{``Rainy Motorbike - Palette Knife Oil Painting On Canvas By Leonid Afremov Original by Leonid Afremov''}} \\[0.5cm]
        \includegraphics[width=\xx,frame]{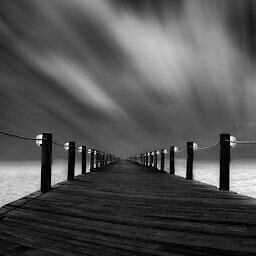} & 
        \includegraphics[width=\xx,frame]{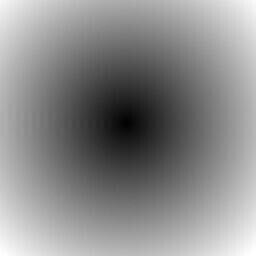} & 
\includegraphics[width=\xx,frame]{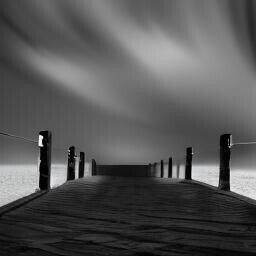} \\
\end{tabularx}
\end{figure}

\begin{figure}

  \centering
  \begin{tabularx}{\columnwidth}{p{1.01\xx}p{1.01\xx}@{\hskip 0.2cm}p{1.01\xx}}

\multicolumn{2}{c}{\textbf{Input Image + Change Map}} & \multicolumn{1}{c}{\textbf{Output Image}} \\[0.2cm] %

    \includegraphics[width=\xx,frame]{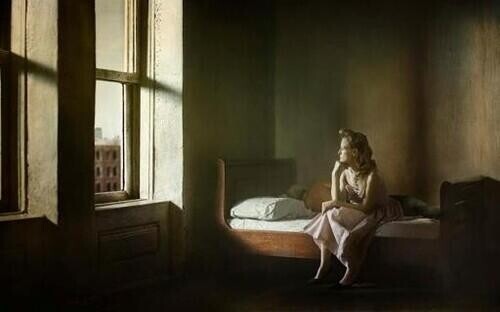} & 
    \includegraphics[width=\xx,frame]{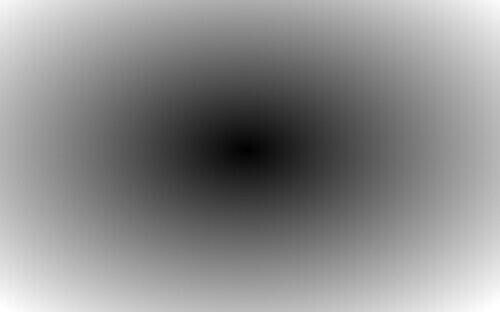} & 
\includegraphics[width=\xx,frame]{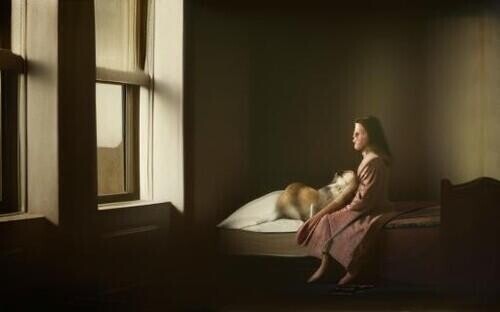} \\[-0.1cm] \multicolumn{3}{X}{\scriptsize{``Richard Tuschman \_Hopper\_Meditations\_Woman\_and\_Dog\_on\_a\_Bed\_20\_x\_29\_in (1)''}} \\[0.5cm]

    \includegraphics[width=\xx,frame]{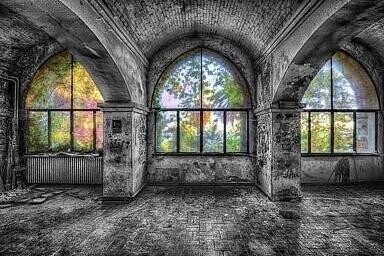} & 
    \includegraphics[width=\xx,frame]{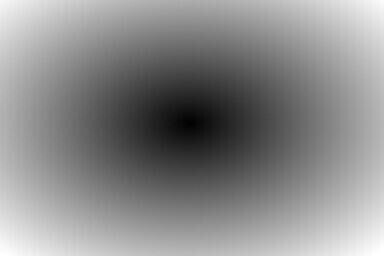} & 

    \includegraphics[width=\xx,frame]{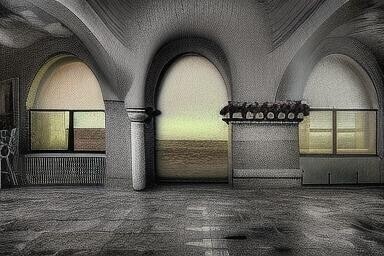} \\[-0.1cm] \multicolumn{3}{X}{\scriptsize{``Photograph - Villa Of Doorways On The Sea - Villa Delle Porte Sul Mare I by Enrico Pelos''}} \\[0.5cm] %

    \includegraphics[width=\xx,frame]{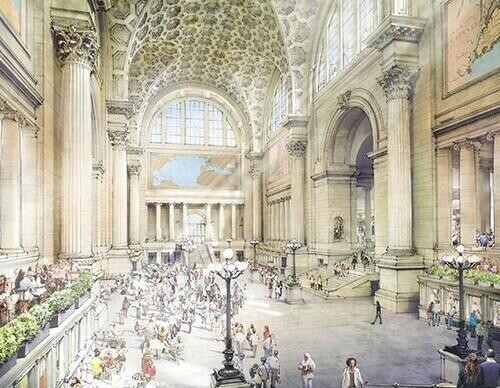} & 
    \includegraphics[width=\xx,frame]{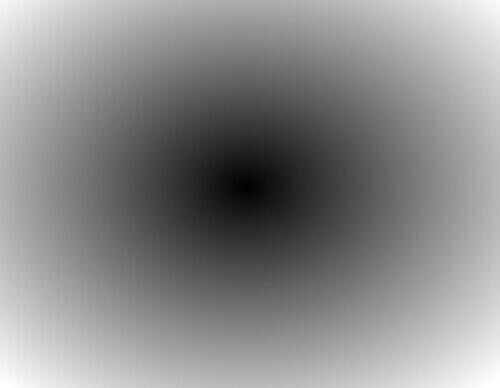} & 

    \includegraphics[width=\xx,frame]{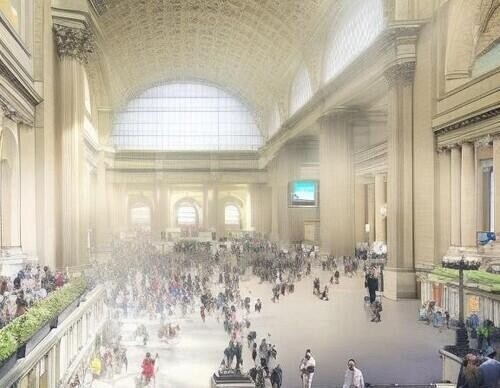} \\[-0.1cm] \multicolumn{3}{X}{\scriptsize{``Rebuild Penn Station's vision for the rebuilt main hall (to life)''}} \\[0.2cm]

    \includegraphics[width=\xx,frame]{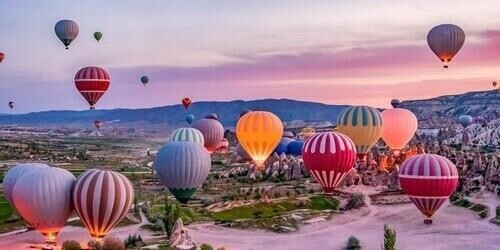} & 
    \includegraphics[width=\xx,frame]{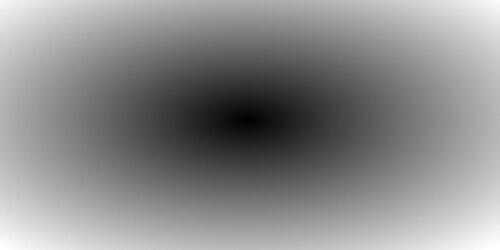} & 
    \includegraphics[width=\xx,frame]{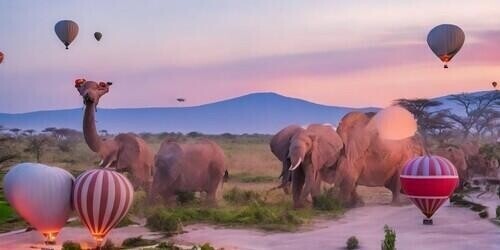} \\
\end{tabularx}
\end{figure}

\clearpage
\subsubsection{Unaltered Depth Map as a Change Map}

\begin{figure}[h]
  \centering
  \begin{tabularx}{\columnwidth}{p{1.01\xx}p{1.01\xx}@{\hskip 0.2cm}p{1.01\xx}}

\multicolumn{2}{c}{\textbf{Input Image + Change Map}} & \multicolumn{1}{c}{\textbf{Output Image}} \\[0.2cm] %
\includegraphics[width=\xx,frame]{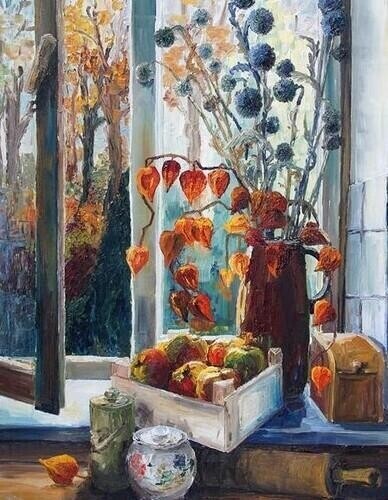} &
\includegraphics[width=\xx,frame]{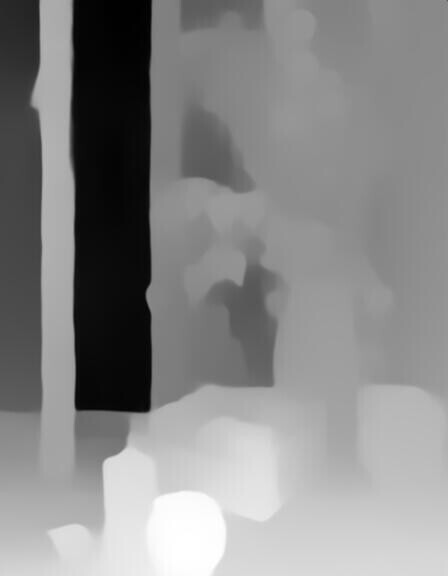} &
\includegraphics[width=\xx,frame]{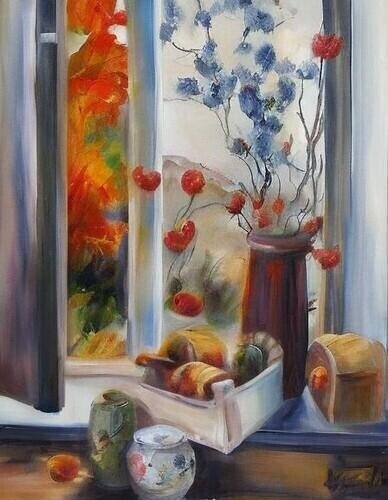}  \\[-0.1cm] \multicolumn{3}{X}{\scriptsize{``Kitchen Window in the Middle of the Desert painting ideas autumn at the kitchen window painting by barbara pommerenke''}} \\[0.5cm] %
\includegraphics[width=\xx,frame]{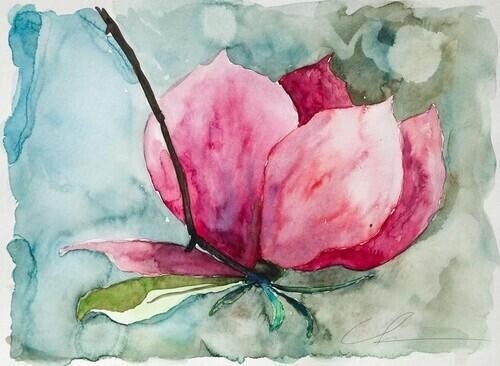} & \includegraphics[width=\xx,frame]{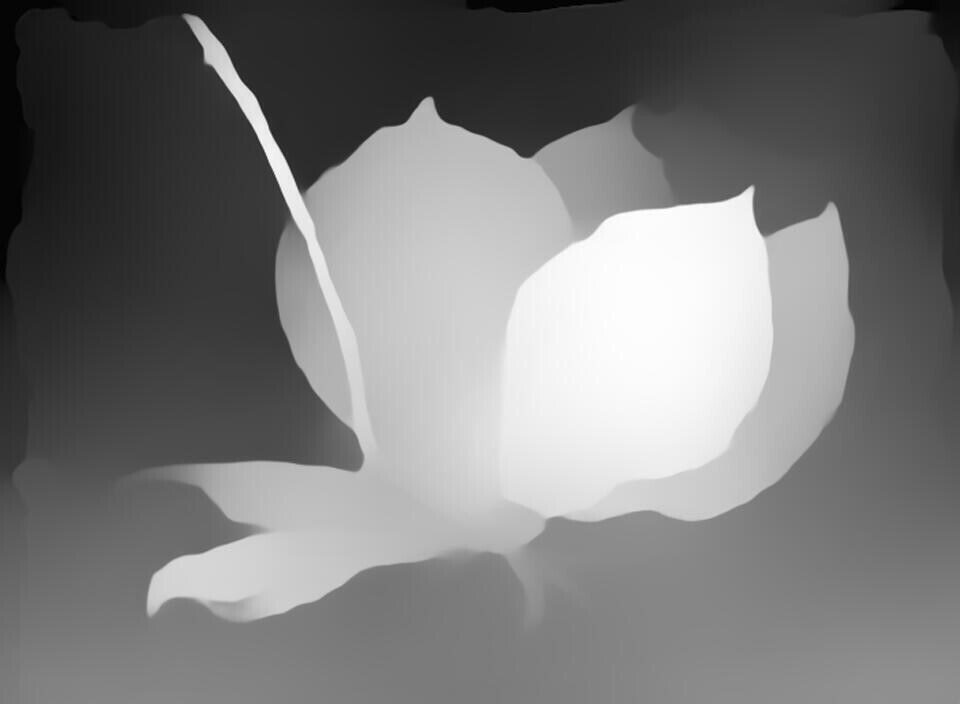} & \includegraphics[width=\xx,frame]{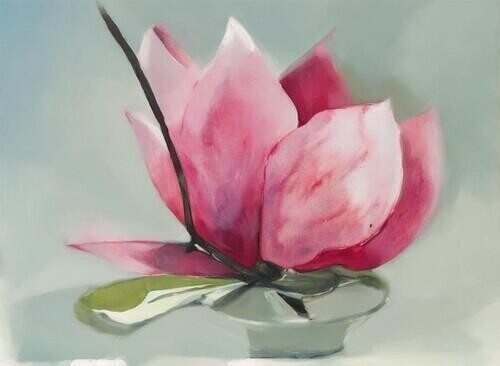} \\[-0.1cm] \multicolumn{3}{X}{\scriptsize{``"""Triangular Saucer Magnolia"" original fine art by Clair Hartmann"''}} \\[0.2cm]

\includegraphics[width=\xx,frame]{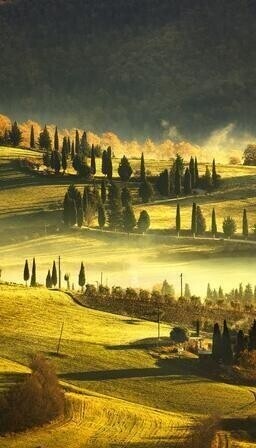} &
\includegraphics[width=\xx,frame]{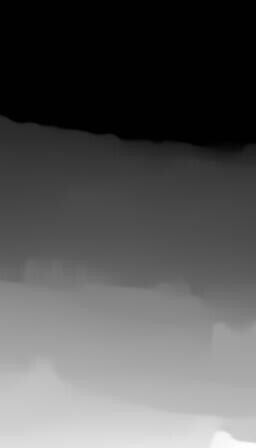} &
\includegraphics[width=\xx,frame]{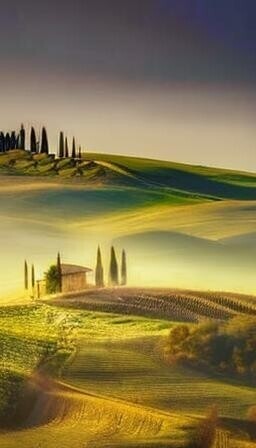}  \\[-0.1cm] \multicolumn{3}{X}{\scriptsize{``Tuscany foggy morning, farmland and palm trees country landscape. Italy, Europe. Stock fotó''}} \\[0.5cm] %
\includegraphics[width=\xx,frame]{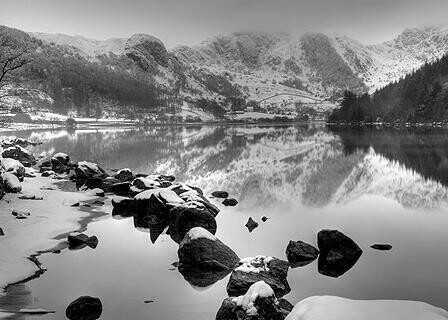} & \includegraphics[width=\xx,frame]{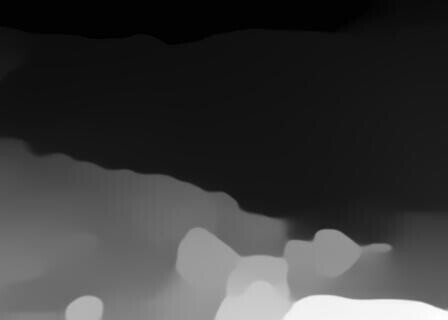} & \includegraphics[width=\xx,frame]{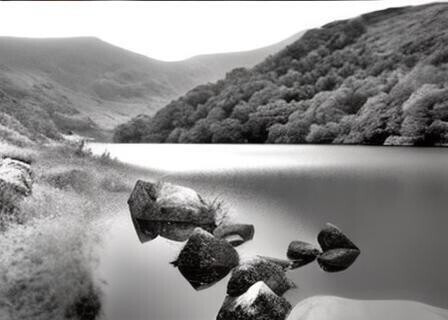} \\
\end{tabularx}
\label{midas-no-intervention}
\end{figure}

\begin{figure}[t]

  \centering
  \begin{tabularx}{\columnwidth}{p{1.01\xx}p{1.01\xx}@{\hskip 0.2cm}p{1.01\xx}}

\multicolumn{2}{c}{\textbf{Input Image + Change Map}} & \multicolumn{1}{c}{\textbf{Output Image}} \\[0.2cm] %
\includegraphics[width=\xx,frame]{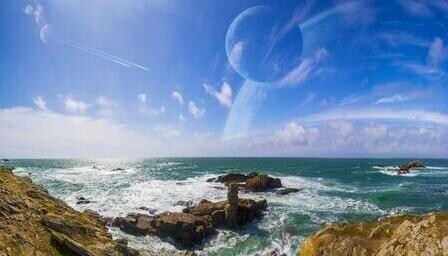} &
\includegraphics[width=\xx,frame]{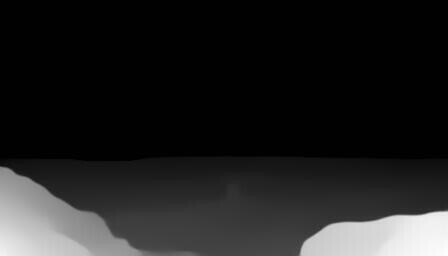} &
\includegraphics[width=\xx,frame]{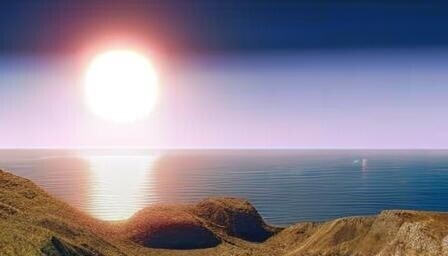} \\[-0.1cm] \multicolumn{3}{X}{\scriptsize{``Movie poster for distant planet system view from cliffs and ocean Stock Photo''}} \\[0.2cm] %
\includegraphics[width=\xx,frame]{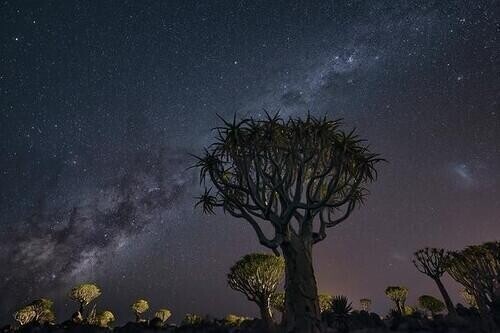} & \includegraphics[width=\xx,frame]{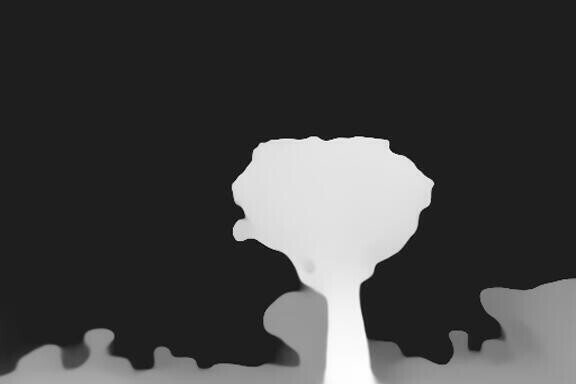} & \includegraphics[width=\xx,frame]{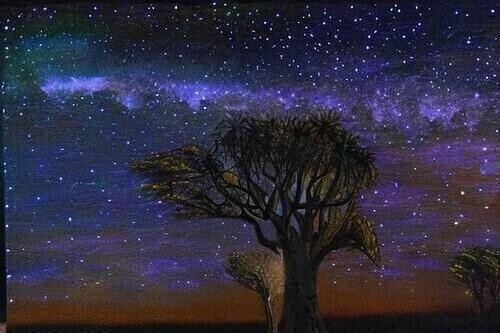} \\[-0.1cm] \multicolumn{3}{X}{\scriptsize{``Painting: The Invasion - Kokerbooms lit up under the Milky Way in Keetmanshoop, Southern Namibia.''}} \\[0.5cm]

\includegraphics[width=\xx,frame]{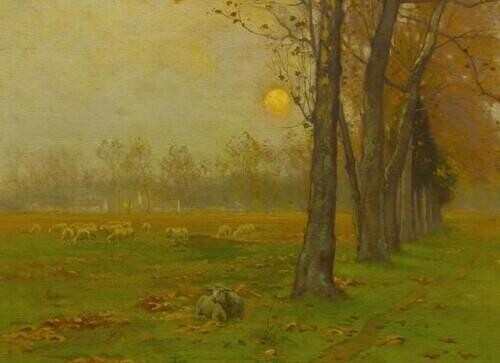} &
\includegraphics[width=\xx,frame]{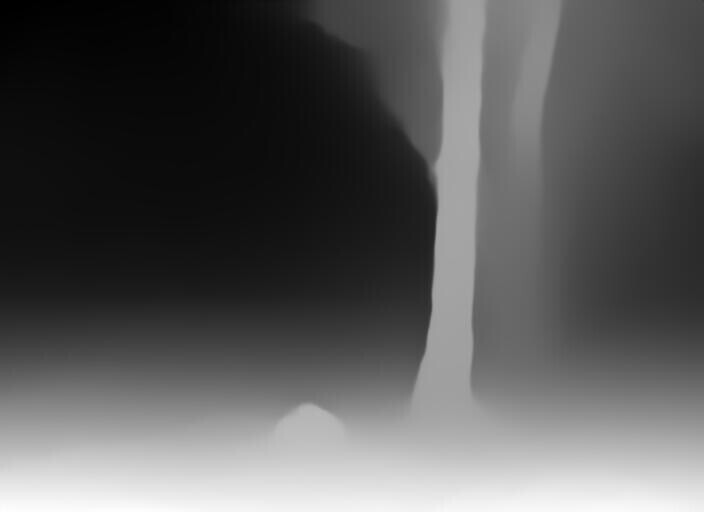} &
\includegraphics[width=\xx,frame]{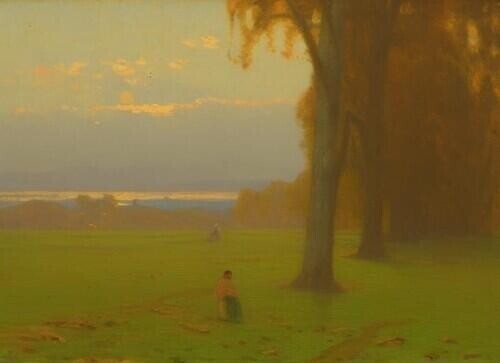} \\[-0.1cm] \multicolumn{3}{X}{\scriptsize{``JOHN CARLETON WIGGINS | End of Day ('The Dying Years') - Bright Day''}} \\[0.2cm] %
\includegraphics[width=\xx,frame]{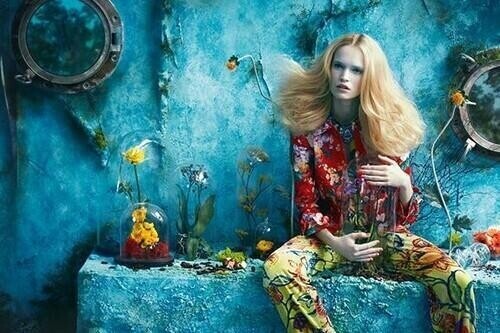} & \includegraphics[width=\xx,frame]{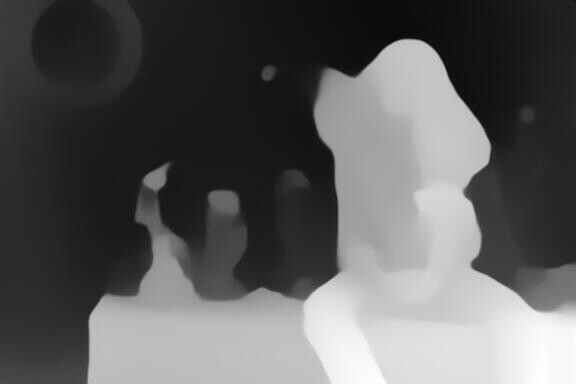} & \includegraphics[width=\xx,frame]{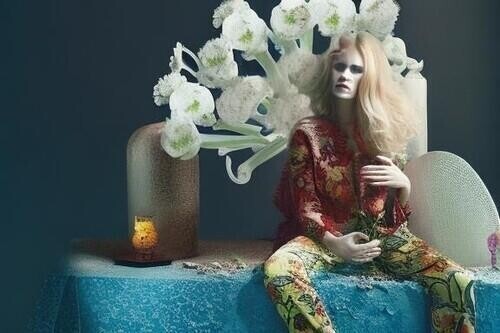} \\
\end{tabularx}

\end{figure}

\begin{figure}[t]

  \centering
  \begin{tabularx}{\columnwidth}{p{1.01\xx}p{1.01\xx}@{\hskip 0.2cm}p{1.01\xx}}

\multicolumn{2}{c}{\textbf{Input Image + Change Map}} & \multicolumn{1}{c}{\textbf{Output Image}} \\[0.2cm] %
\includegraphics[width=\xx,frame]{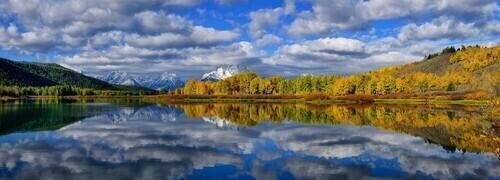} &
\includegraphics[width=\xx,frame]{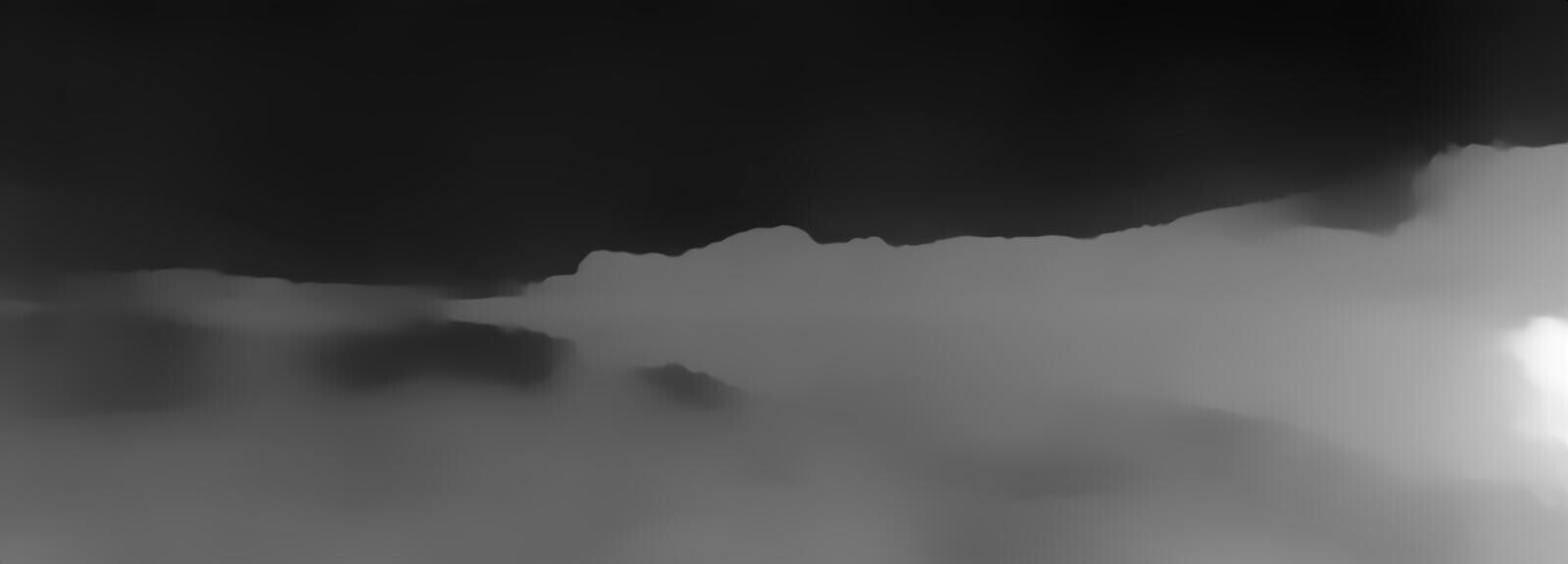} &
\includegraphics[width=\xx,frame]{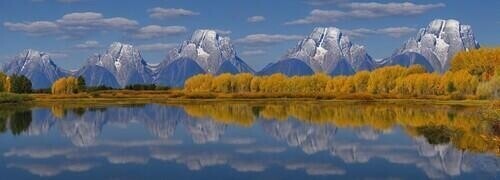} \\[-0.1cm] \multicolumn{3}{X}{\scriptsize{``Oxbow Bend Peak Autumn Panorama Lithograph''}} \\[0.2cm] %
\includegraphics[width=\xx,frame]{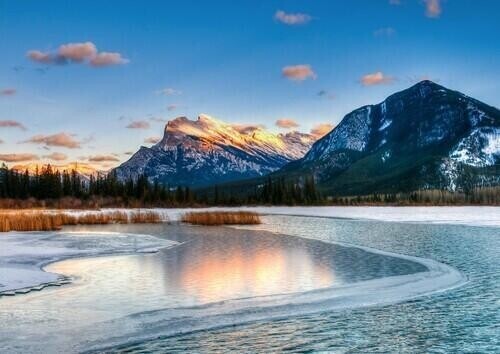} & \includegraphics[width=\xx,frame]{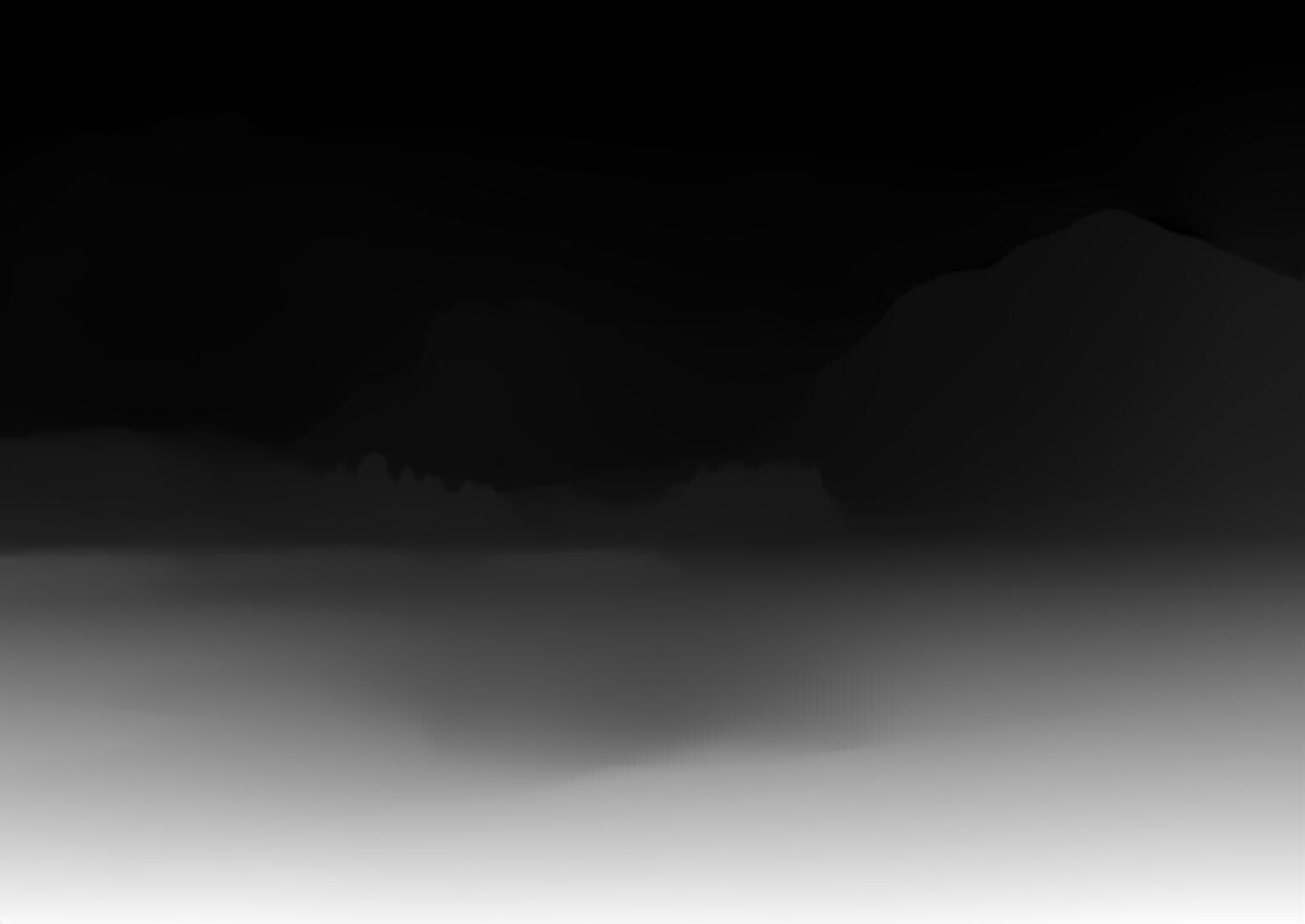} & \includegraphics[width=\xx,frame]{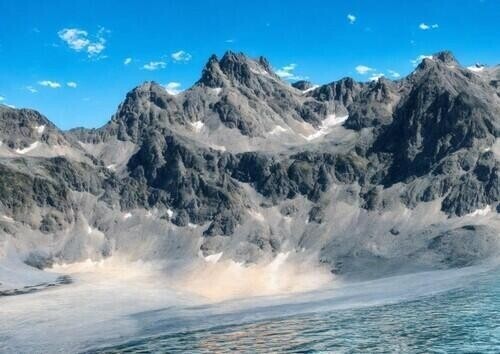} \\[-0.1cm] \multicolumn{3}{X}{\scriptsize{``Fabulous Summer Hikes in the US and Canada''}} \\[0.2cm]

\includegraphics[width=\xx,frame]{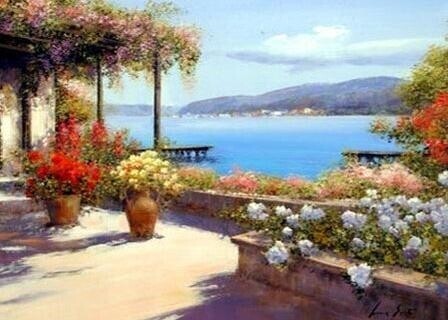} &
\includegraphics[width=\xx,frame]{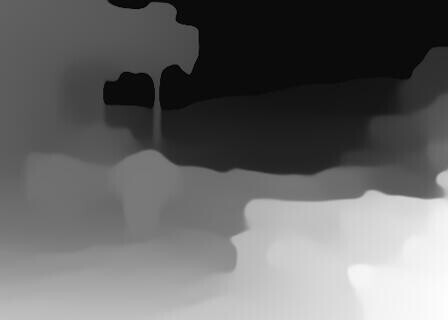} &
\includegraphics[width=\xx,frame]{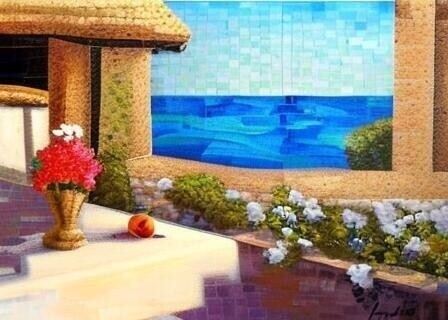} \\[-0.1cm] \multicolumn{3}{X}{\scriptsize{``Mosaic, Wall Art, Large Mosaic, Mediterranean Sea Mosaic, Canvas Mosaic, Kitchen Wall Art, Oil Mosaic, Canvas Art, Seascape, France Summer Resort-Paintingforhome''}} \\[0.5cm] %
\includegraphics[width=\xx,frame]{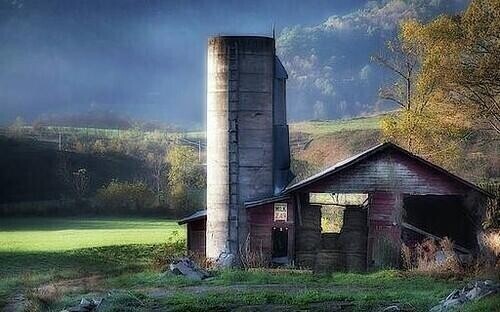} & \includegraphics[width=\xx,frame]{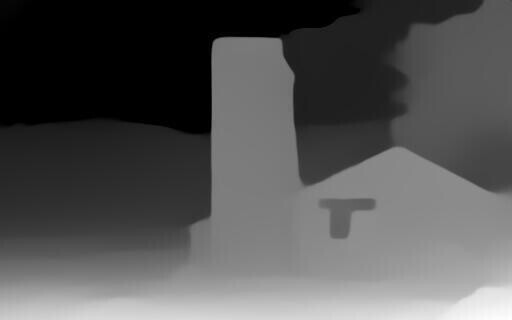} & \includegraphics[width=\xx,frame]{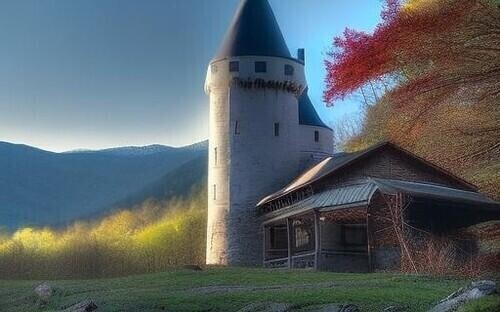} \\
\end{tabularx}

\end{figure}

\begin{figure}[t]

  \centering
  \begin{tabularx}{\columnwidth}{p{1.01\xx}p{1.01\xx}@{\hskip 0.2cm}p{1.01\xx}}

\multicolumn{2}{c}{\textbf{Input Image + Change Map}} & \multicolumn{1}{c}{\textbf{Output Image}} \\[0.2cm] %

\includegraphics[width=\xx,frame]{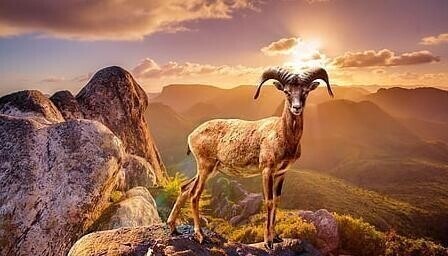} &
\includegraphics[width=\xx,frame]{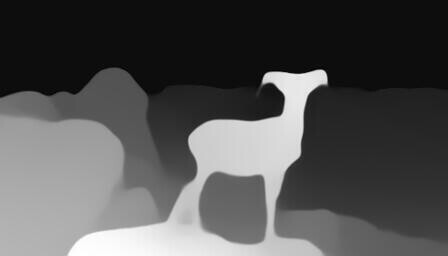} &
\includegraphics[width=\xx,frame]{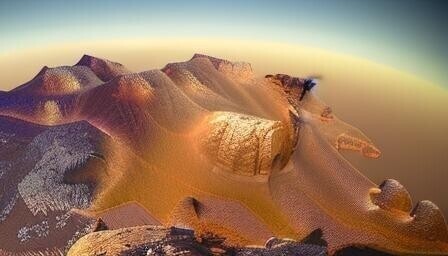} \\[-0.1cm] \multicolumn{3}{X}{\scriptsize{``Mount Ramon on the top of planet''}} \\[0.2cm] %
\includegraphics[width=\xx,frame]{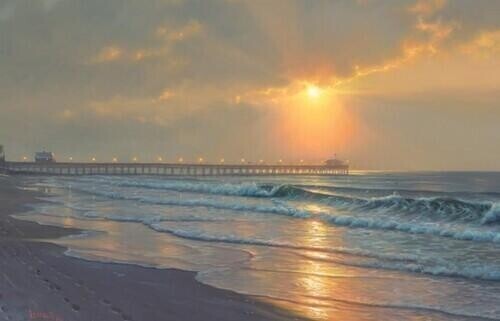} & \includegraphics[width=\xx,frame]{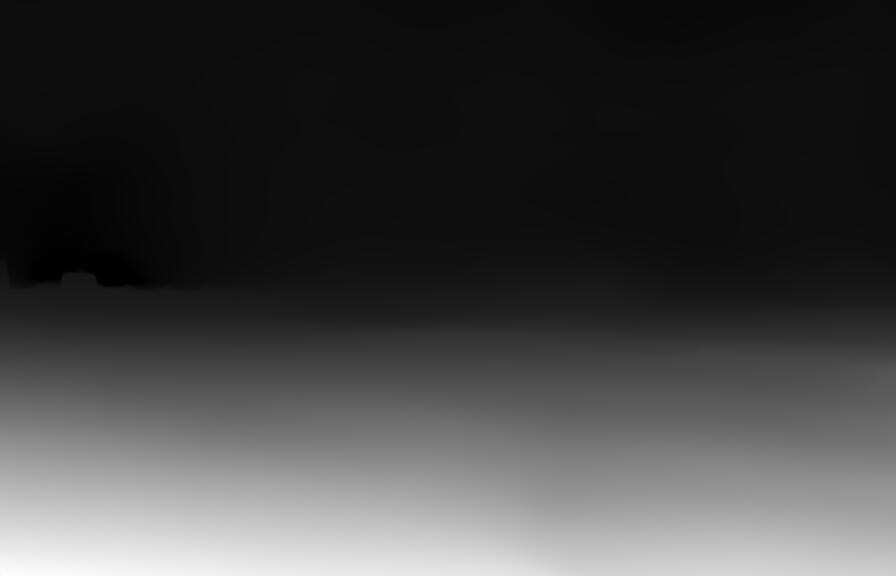} & \includegraphics[width=\xx,frame]{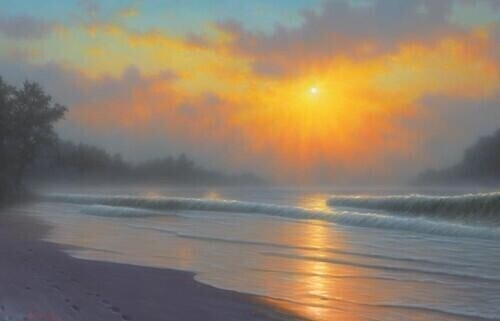} \\[-0.1cm] \multicolumn{3}{X}{\scriptsize{``Intense Radiant Sunrise by Mark Keathley''}} \\[0.2cm]

\includegraphics[width=\xx,frame]{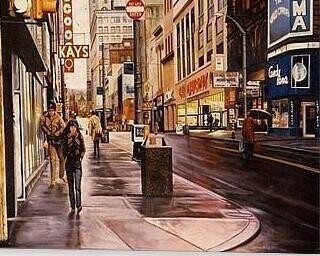} &
\includegraphics[width=\xx,frame]{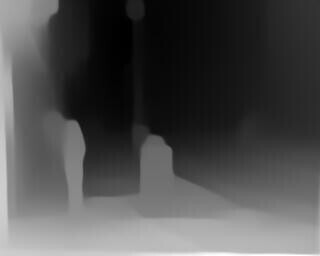} &
\includegraphics[width=\xx,frame]{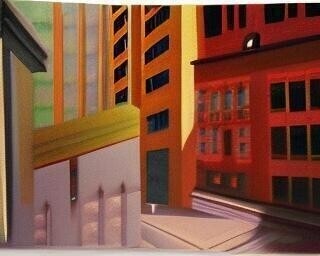} \\[-0.1cm] \multicolumn{3}{X}{\scriptsize{``Fifth Avenue Street View In The 80s Canvas Print''}} \\[0.2cm] %
\includegraphics[width=\xx,frame]{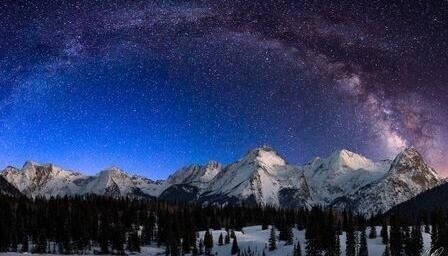} & \includegraphics[width=\xx,frame]{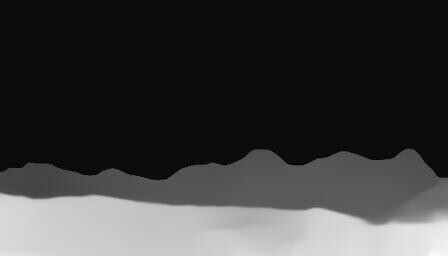} & \includegraphics[width=\xx,frame]{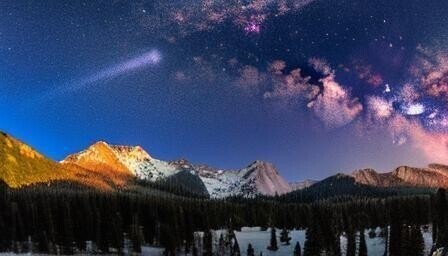} \\
\end{tabularx}

\end{figure}

\begin{figure}[t]

  \centering
  \begin{tabularx}{\columnwidth}{p{1.01\xx}p{1.01\xx}@{\hskip 0.2cm}p{1.01\xx}}

\multicolumn{2}{c}{\textbf{Input Image + Change Map}} & \multicolumn{1}{c}{\textbf{Output Image}} \\[0.2cm] %

\includegraphics[width=\xx,frame]{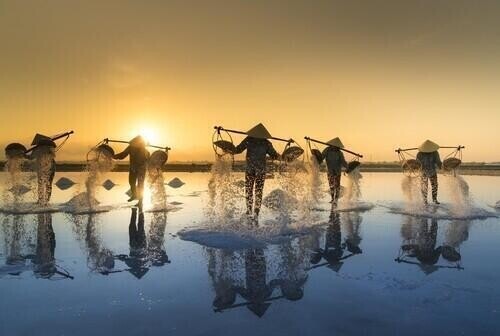} &
\includegraphics[width=\xx,frame]{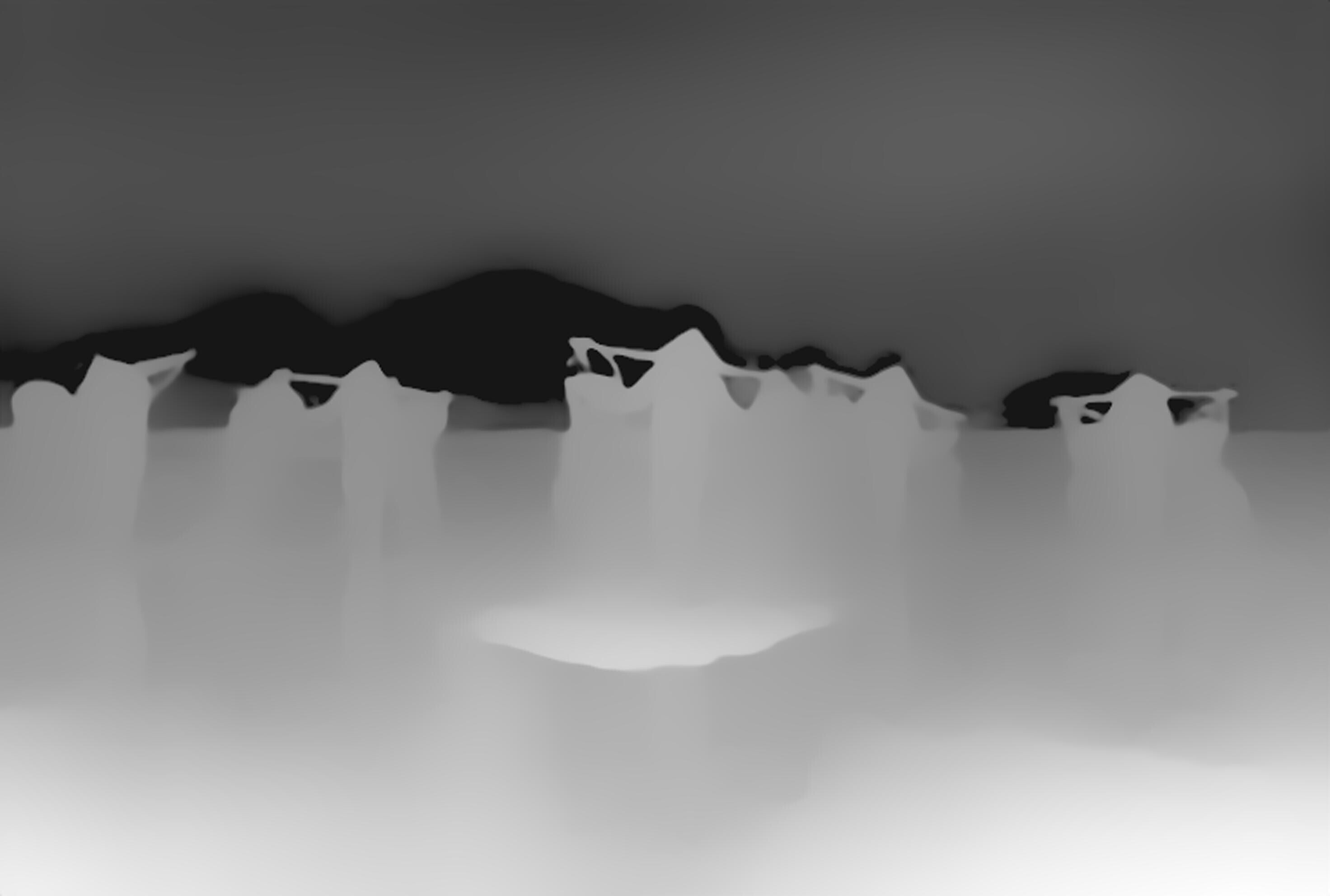} &
\includegraphics[width=\xx,frame]{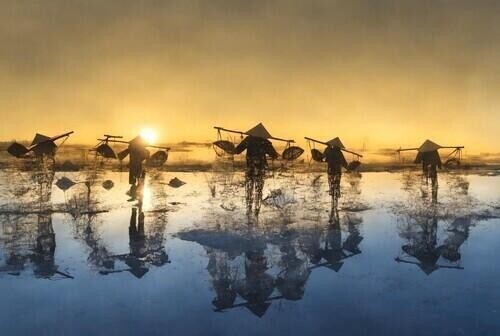} \\[-0.1cm] \multicolumn{3}{X}{\scriptsize{``Hon Khoi Salt Fields, Vietnam, painting''}} \\[0.2cm] %
\includegraphics[width=\xx,frame]{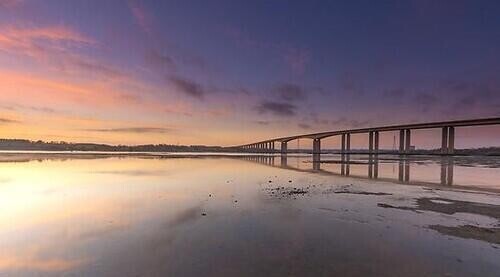} & \includegraphics[width=\xx,frame]{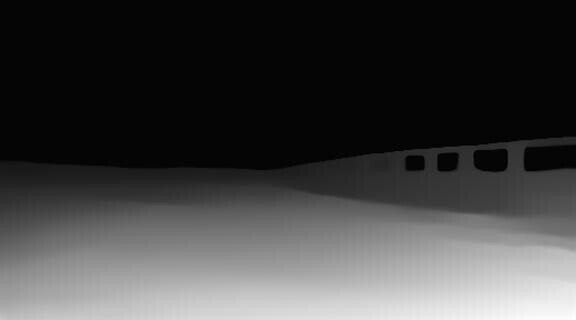} & \includegraphics[width=\xx,frame]{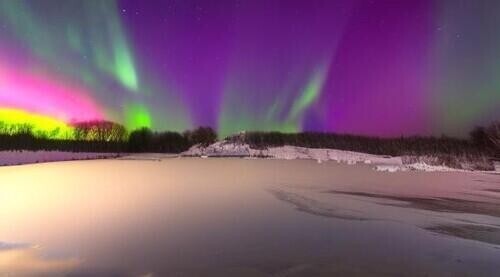} \\[-0.1cm] \multicolumn{3}{X}{\scriptsize{``norwell-bridge--winter-aurora-borealis.jpg''}} \\[0.2cm]

\includegraphics[width=\xx,frame]{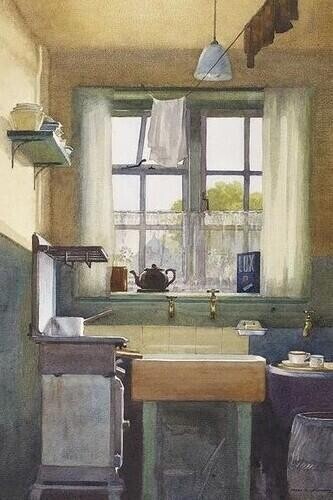} &
\includegraphics[width=\xx,frame]{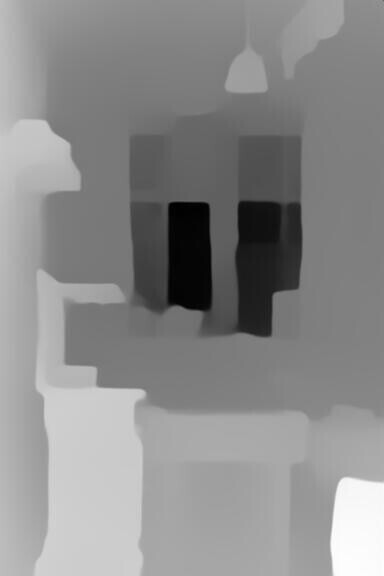} &
\includegraphics[width=\xx,frame]{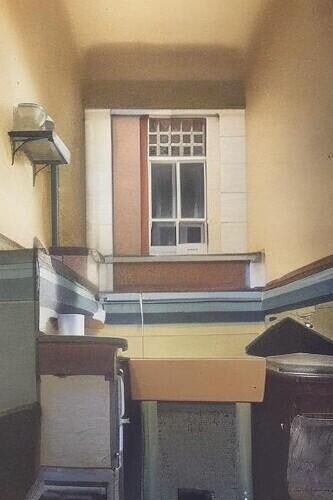} \\[-0.1cm] \multicolumn{3}{X}{\scriptsize{``The Scullery Apartment Building, 15 Dalston Road, Acocks Green, Birmingham by Frank Taylor Lockwood 1944''}} \\[0.5cm] %
\includegraphics[width=\xx,frame]{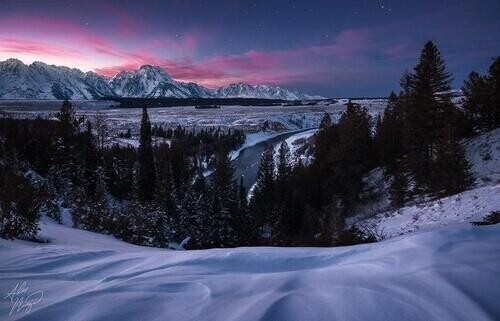} & \includegraphics[width=\xx,frame]{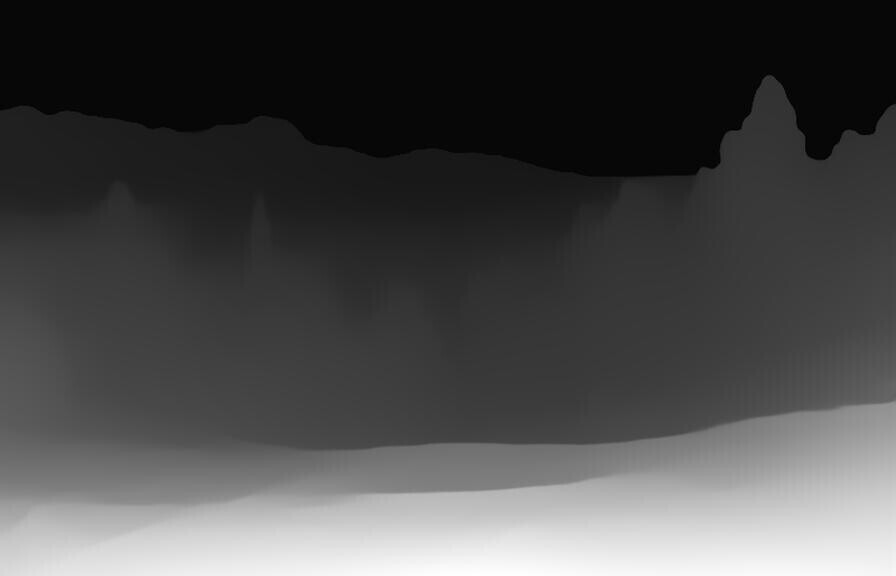} & \includegraphics[width=\xx,frame]{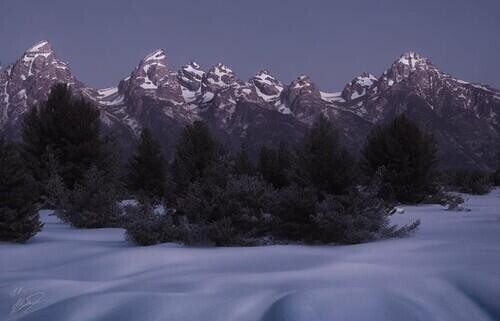} \\
\end{tabularx}

\end{figure}

\begin{figure}[t]

  \centering
  \begin{tabularx}{\columnwidth}{p{1.01\xx}p{1.01\xx}@{\hskip 0.2cm}p{1.01\xx}}

\multicolumn{2}{c}{\textbf{Input Image + Change Map}} & \multicolumn{1}{c}{\textbf{Output Image}} \\[0.2cm] %

\includegraphics[width=\xx,frame]{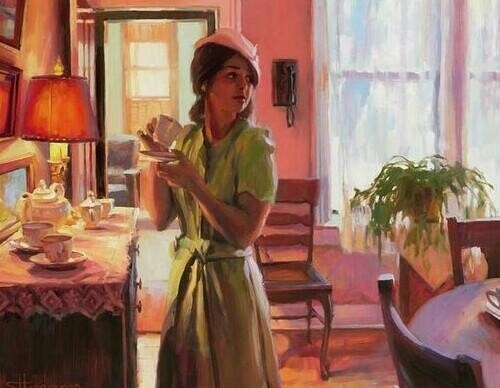} &
\includegraphics[width=\xx,frame]{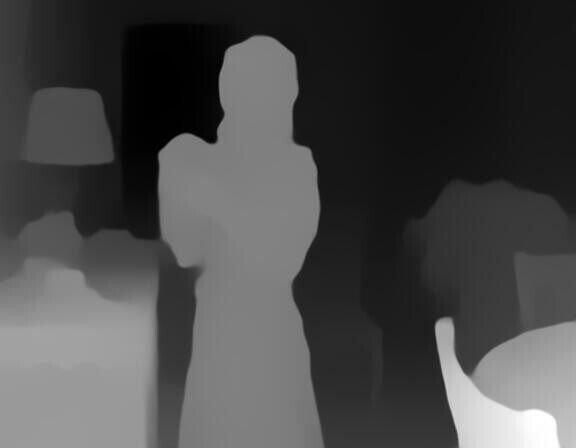} &
\includegraphics[width=\xx,frame]{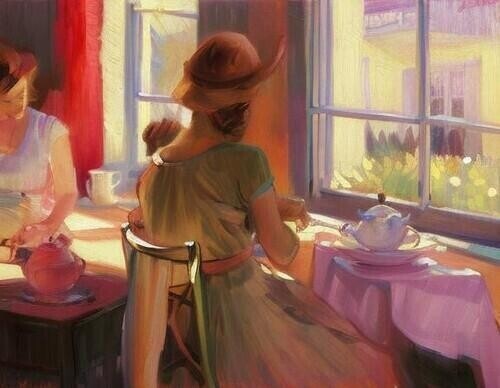} \\[-0.1cm] \multicolumn{3}{X}{\scriptsize{``Vincent van Gogh Painting - Midday Tea by Steve Henderson''}} \\[0.2cm] %
\includegraphics[width=\xx,frame]{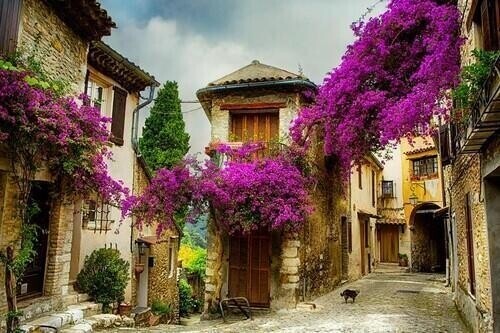} & \includegraphics[width=\xx,frame]{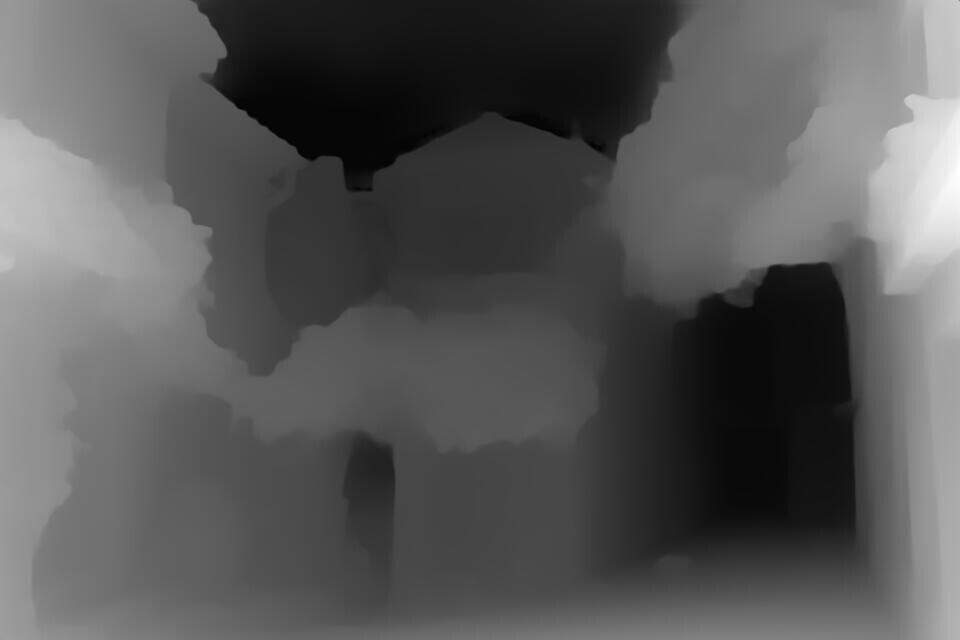} & \includegraphics[width=\xx,frame]{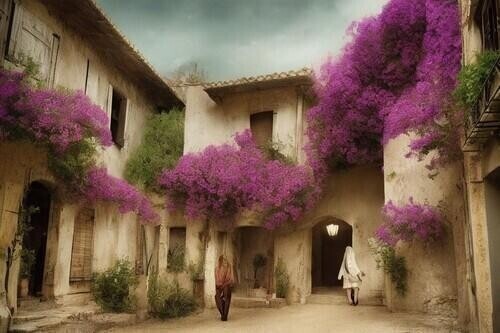} \\[-0.1cm] \multicolumn{3}{X}{\scriptsize{``Provence Horror Movie''}} \\[0.2cm]

\includegraphics[width=\xx,frame]{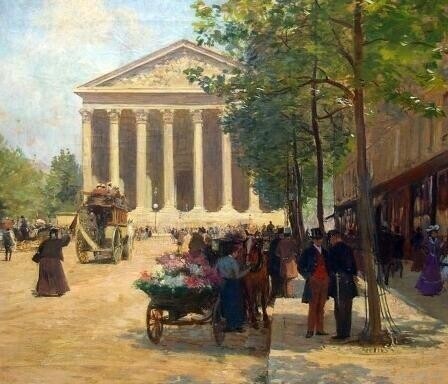} &
\includegraphics[width=\xx,frame]{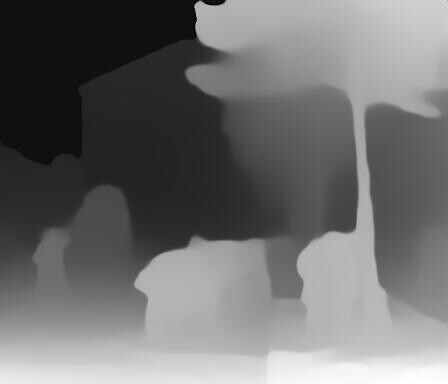} &
\includegraphics[width=\xx,frame]{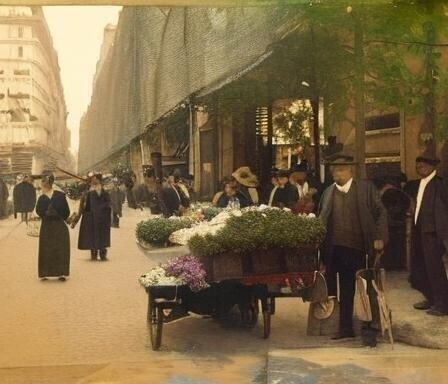} \\[-0.1cm] \multicolumn{3}{X}{\scriptsize{``Victor Gilbert (1847-1933) - Flower Market, rue Royale in Paris, as a photo''}} \\[0.2cm] %
\includegraphics[width=\xx,frame]{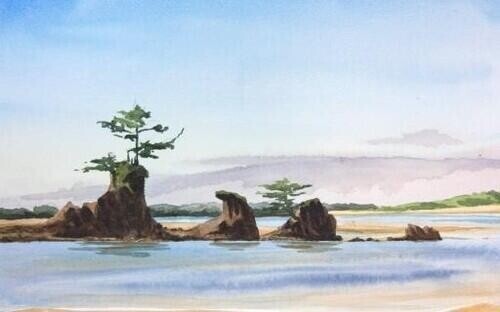} & \includegraphics[width=\xx,frame]{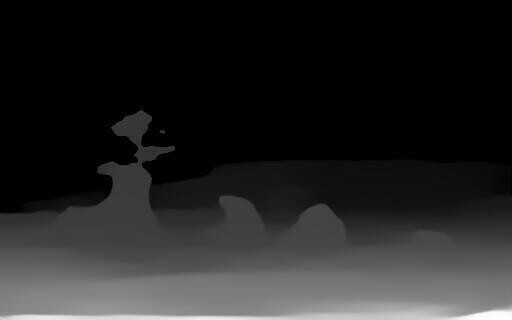} & \includegraphics[width=\xx,frame]{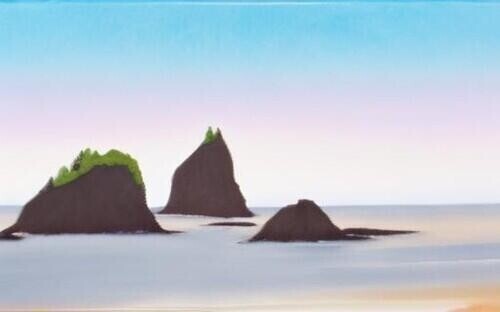} \\
\end{tabularx}

\end{figure}

\begin{figure}[t]

  \centering
  \begin{tabularx}{\columnwidth}{p{1.01\xx}p{1.01\xx}@{\hskip 0.2cm}p{1.01\xx}}

\multicolumn{2}{c}{\textbf{Input Image + Change Map}} & \multicolumn{1}{c}{\textbf{Output Image}} \\[0.2cm] %

\includegraphics[width=\xx,frame]{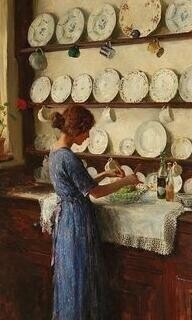} & \includegraphics[width=\xx,frame]{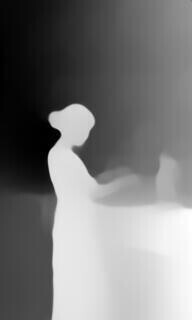} & \includegraphics[width=\xx,frame]{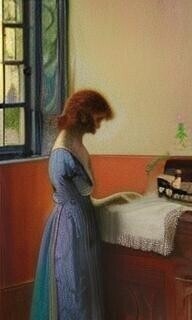} \\[-0.1cm] \multicolumn{3}{X}{\scriptsize{``William Henry Margetson - The Old Lady of the House''}} \\[0.2cm]

\includegraphics[width=\xx,frame]{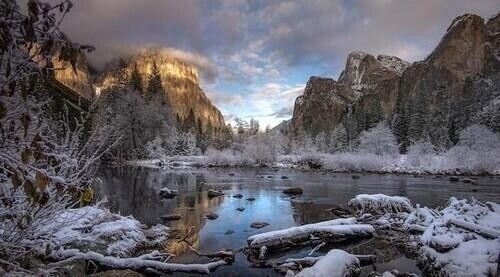} &
\includegraphics[width=\xx,frame]{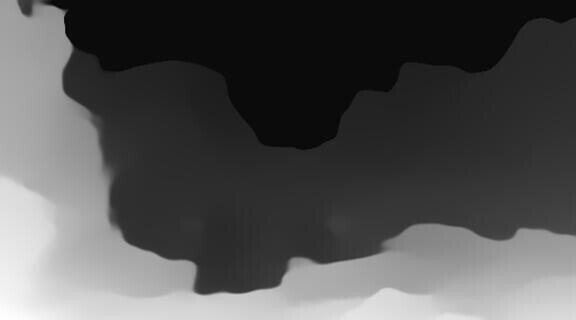} &
\includegraphics[width=\xx,frame]{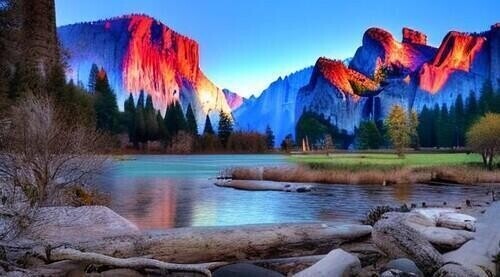} \\[-0.1cm] \multicolumn{3}{X}{\scriptsize{``Breathtaking evening in the summer at valley view yosemite national park''}} \\[0.2cm] %
\includegraphics[width=\xx,frame]{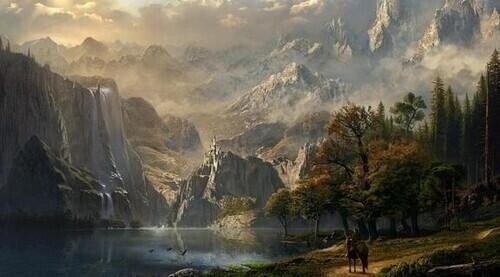} & \includegraphics[width=\xx,frame]{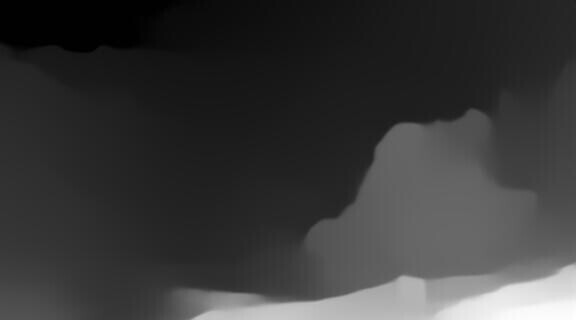} & \includegraphics[width=\xx,frame]{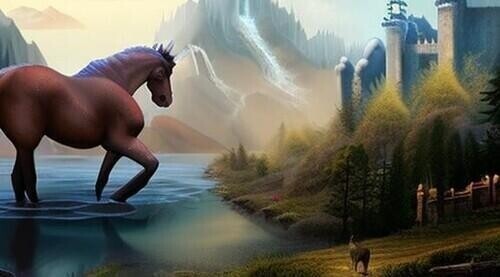} \\[-0.1cm] \multicolumn{3}{X}{\scriptsize{``Picture forest, life-sized puppet, mountains, lake, castle, horse, elf, waterfall, rider, fantasy, art, fantasy, CG wallpapers, Idyll's …''}} \\[0.5cm]
\includegraphics[width=\xx,frame]{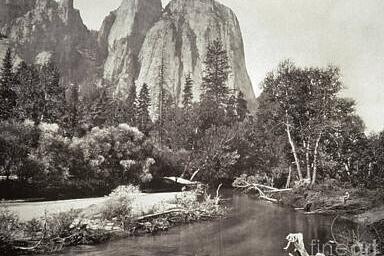} &
\includegraphics[width=\xx,frame]{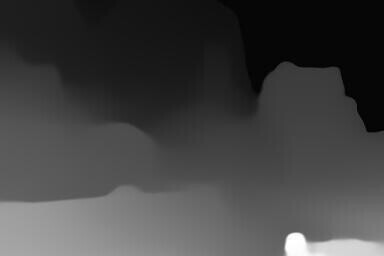} &
\includegraphics[width=\xx,frame]{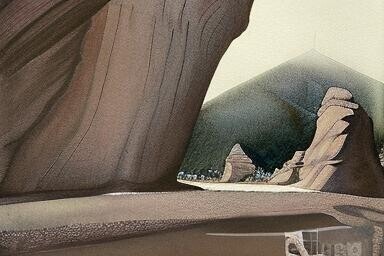} \\
\end{tabularx}

\end{figure}

\begin{figure}[t]

  \centering
  \begin{tabularx}{\columnwidth}{p{1.01\xx}p{1.01\xx}@{\hskip 0.2cm}p{1.01\xx}}

\multicolumn{2}{c}{\textbf{Input Image + Change Map}} & \multicolumn{1}{c}{\textbf{Output Image}} \\[0.2cm] %

\includegraphics[width=\xx,frame]{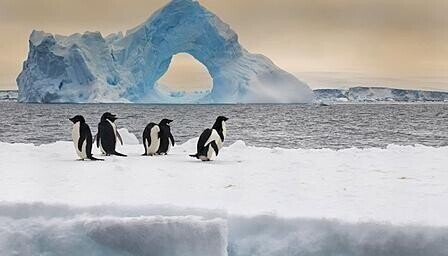} &
\includegraphics[width=\xx,frame]{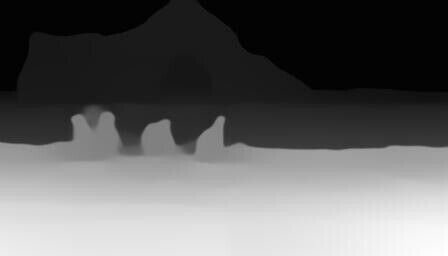} &
\includegraphics[width=\xx,frame]{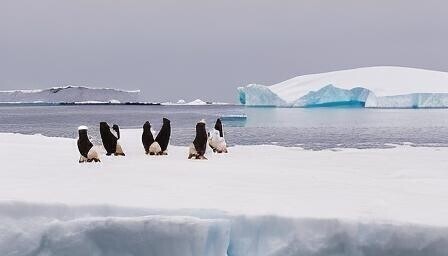} \\[-0.1cm] \multicolumn{3}{X}{\scriptsize{``Group of American bald eagles (Haliaeetus leucocephalus) on an iceberg, natural arch iceberg in the back, Paulet Island, Erebus and Terror Gulf, Antarctic Peninsula, Antarctica''}} \\[0.8cm] %
\includegraphics[width=\xx,frame]{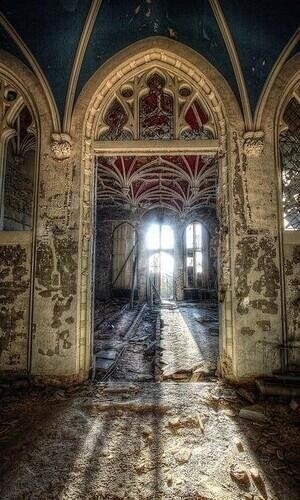} & \includegraphics[width=\xx,frame]{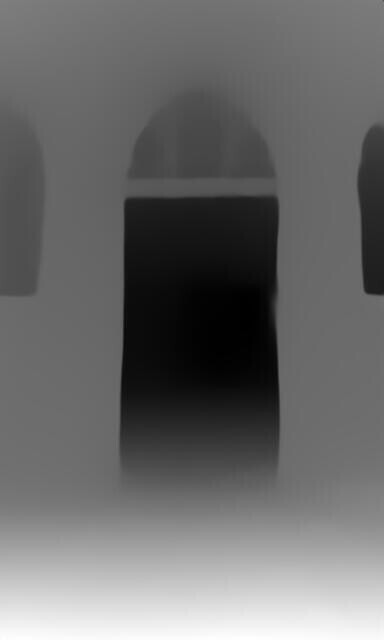} & \includegraphics[width=\xx,frame]{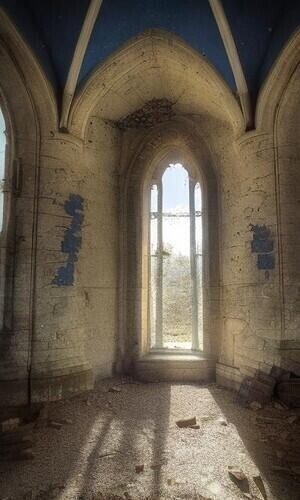} \\[-0.1cm] \multicolumn{3}{X}{\scriptsize{``Sunlight amid ruins at the Chateau de Noisy in the UK.''}} \\[0.2cm]

\includegraphics[width=\xx,frame]{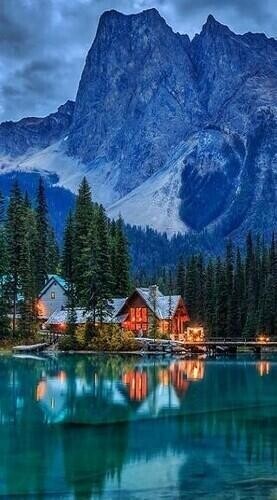} &
\includegraphics[width=\xx,frame]{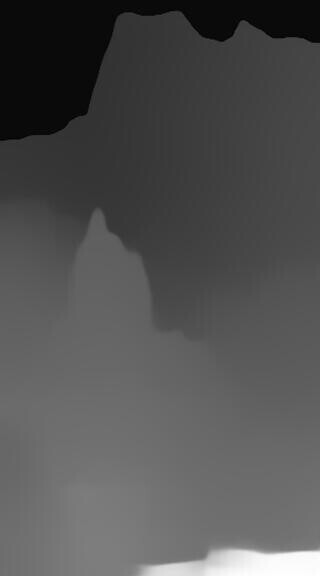} &
\includegraphics[width=\xx,frame]{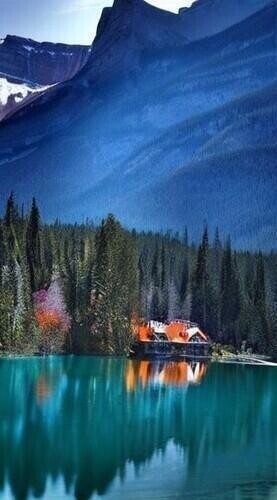} \\[-0.1cm] \multicolumn{3}{X}{\scriptsize{``Emerald Lake, Canada. From the movie "Signature Bride". Follow us @SIGNATUREBRIDE on Twitter and on FACEBOOK @ SIGNATURE BRIDE MAGAZINE''}} \\[0.2cm] %

\end{tabularx}

\end{figure}

\begin{figure}[t]

  \centering
  \begin{tabularx}{\columnwidth}{p{1.01\xx}p{1.01\xx}@{\hskip 0.2cm}p{1.01\xx}}

\multicolumn{2}{c}{\textbf{Input Image + Change Map}} & \multicolumn{1}{c}{\textbf{Output Image}} \\[0.2cm] %

\includegraphics[width=\xx,frame]{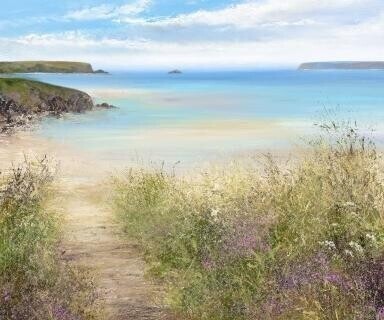} &
\includegraphics[width=\xx,frame]{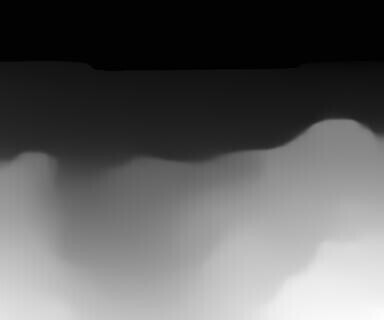} &
\includegraphics[width=\xx,frame]{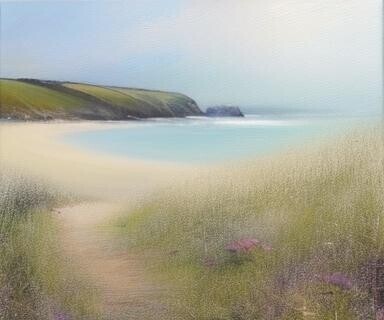} \\[-0.1cm] \multicolumn{3}{X}{\scriptsize{``English Artist Amanda HOSKIN - Summer Path, Daymer Bay Raining''}} \\[0.2cm] %

\includegraphics[width=\xx,frame]{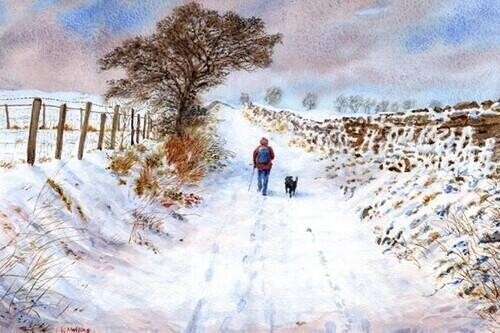} & \includegraphics[width=\xx,frame]{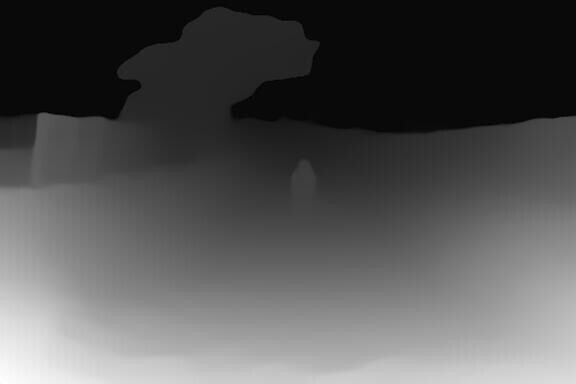} & \includegraphics[width=\xx,frame]{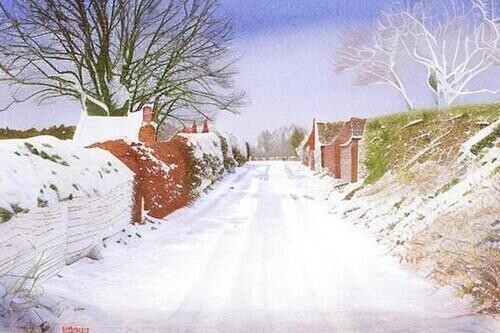} \\[-0.1cm] \multicolumn{3}{X}{\scriptsize{``Sandy Hall Lane, Snow. Poster by Keith Melling''}} \\[0.2cm]
\includegraphics[width=\xx,frame]{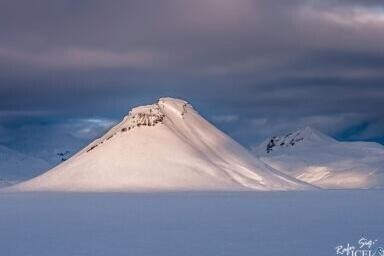} &
\includegraphics[width=\xx,frame]{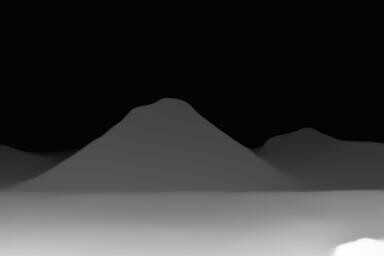} &
\includegraphics[width=\xx,frame]{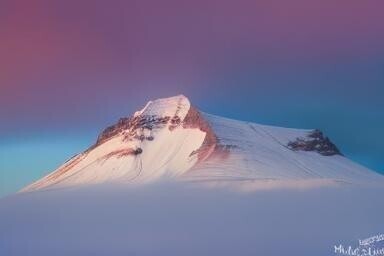} \\[-0.1cm] \multicolumn{3}{X}{\scriptsize{``Mt. Mælifell with Unicorn in the Highlands| Iceland Landscape Photography''}} \\[0.2cm] %
\includegraphics[width=\xx,frame]{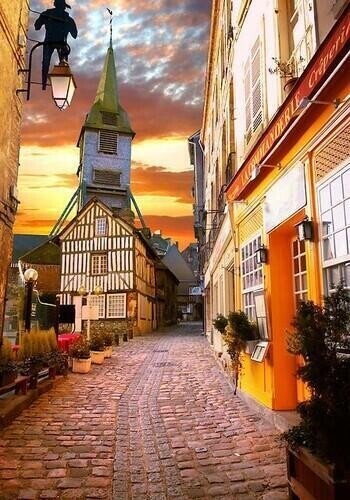} & \includegraphics[width=\xx,frame]{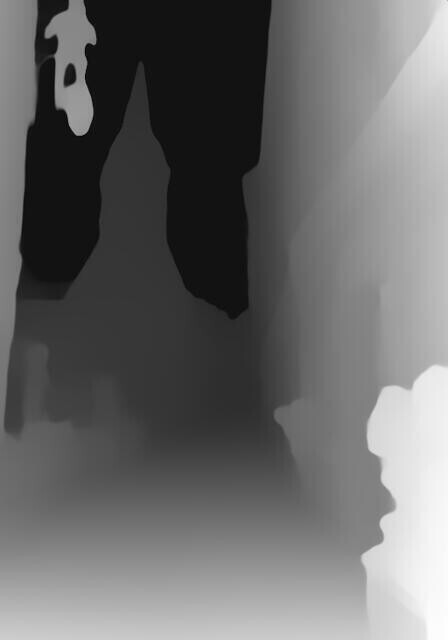} & \includegraphics[width=\xx,frame]{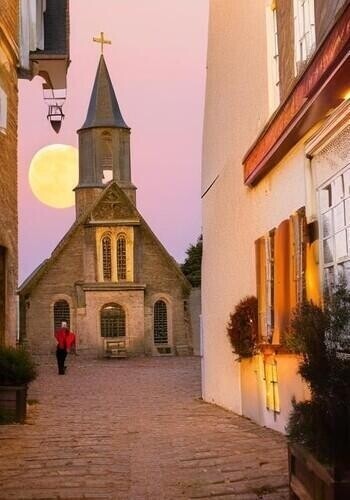} \\
\end{tabularx}

\end{figure}

\begin{figure}[t]

  \centering
  \begin{tabularx}{\columnwidth}{p{1.01\xx}p{1.01\xx}@{\hskip 0.2cm}p{1.01\xx}}

\multicolumn{2}{c}{\textbf{Input Image + Change Map}} & \multicolumn{1}{c}{\textbf{Output Image}} \\[0.2cm] %

\includegraphics[width=\xx,frame]{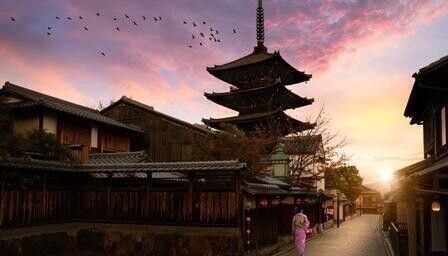} &
\includegraphics[width=\xx,frame]{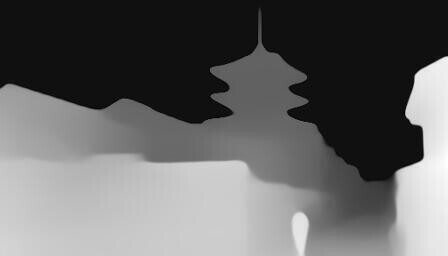} &
\includegraphics[width=\xx,frame]{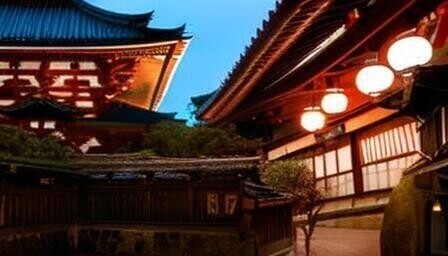} \\[-0.1cm] \multicolumn{3}{X}{\scriptsize{``Yasaka Pagoda and Sannen Zaka Street in the Morning, Kyoto, Japan, Yasaka Pagoda in the morning, Kyoto Japan painting, Yasaka Pagoda in the morning, Kyoto Japan  - 64937280''}} \\[0.8cm] %
\includegraphics[width=\xx,frame]{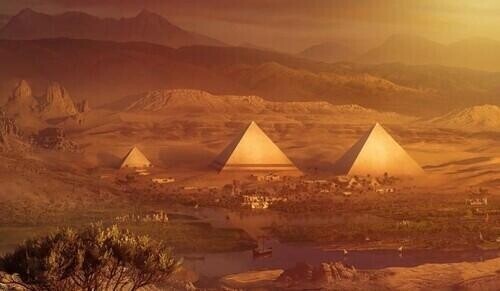} & \includegraphics[width=\xx,frame]{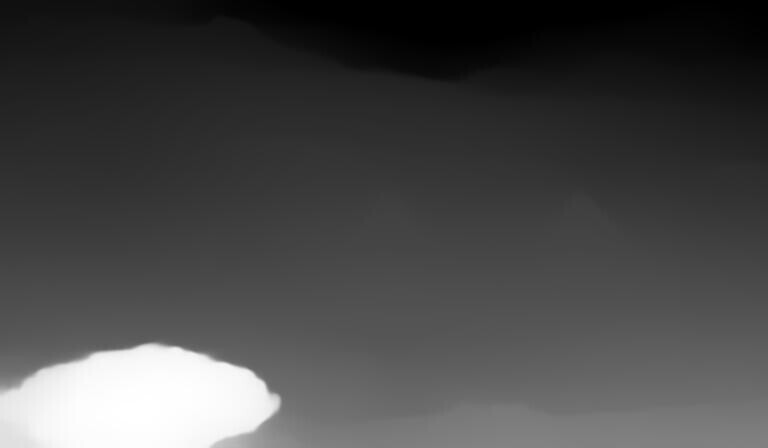} & \includegraphics[width=\xx,frame]{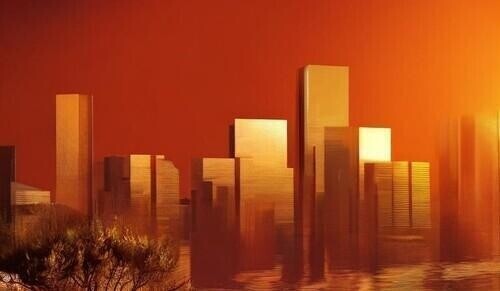} \\[-0.1cm] \multicolumn{3}{X}{\scriptsize{``Artists rendering of modern skyscrapers. Shutterstock.''}} \\[0.2cm]

\includegraphics[width=\xx,frame]{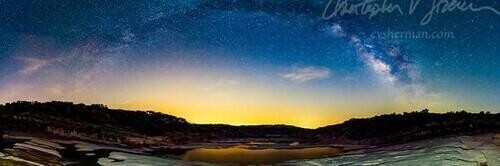} &
\includegraphics[width=\xx,frame]{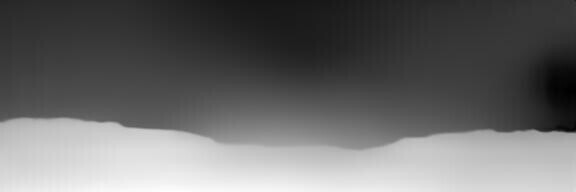} &
\includegraphics[width=\xx,frame]{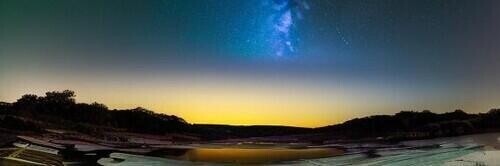} \\[-0.1cm] \multicolumn{3}{X}{\scriptsize{``Bright Green Milky Way Over Austin''}} \\[0.2cm] %
\includegraphics[width=\xx,frame]{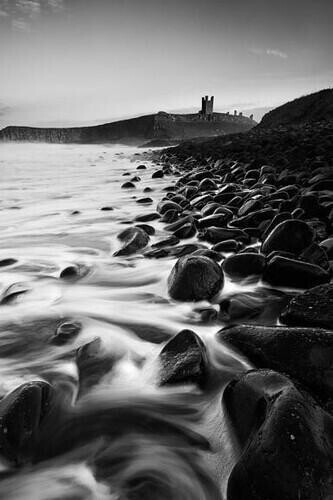} & \includegraphics[width=\xx,frame]{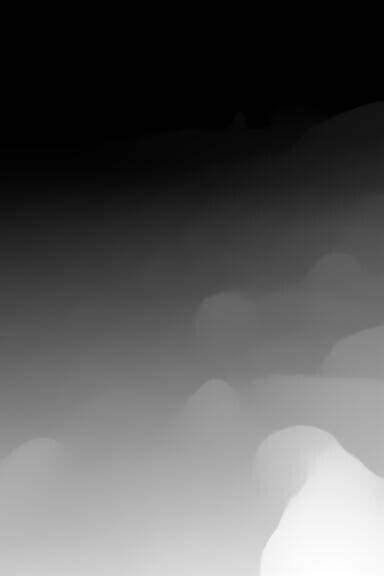} & \includegraphics[width=\xx,frame]{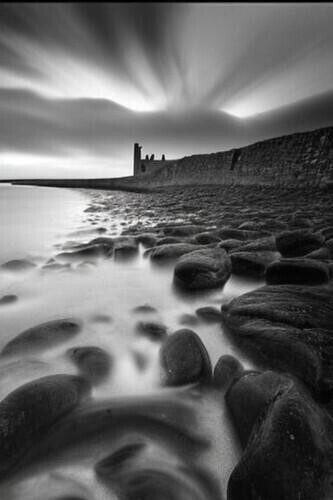} \\
\end{tabularx}

\end{figure}

\clearpage

\begin{figure}

  \centering
  \begin{tabularx}{\columnwidth}{p{1.01\xx}p{1.01\xx}@{\hskip 0.2cm}p{1.01\xx}}

\multicolumn{2}{c}{\textbf{Input Image + Change Map}} & \multicolumn{1}{c}{\textbf{Output Image}} \\[0.2cm]

\includegraphics[width=\xx,frame]{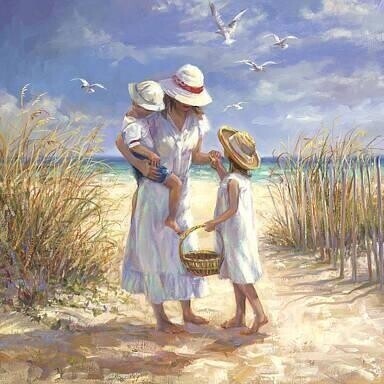} &
\includegraphics[width=\xx,frame]{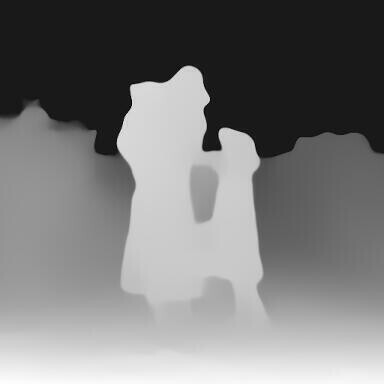} &
\includegraphics[width=\xx,frame]{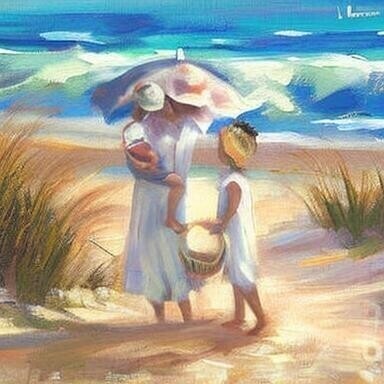}  \\[-0.1cm]
\multicolumn{3}{X}{\scriptsize{``Mothers Day Red Beach Original by Laurie Hein''}} \\[0.2cm]

\includegraphics[width=\xx,frame]{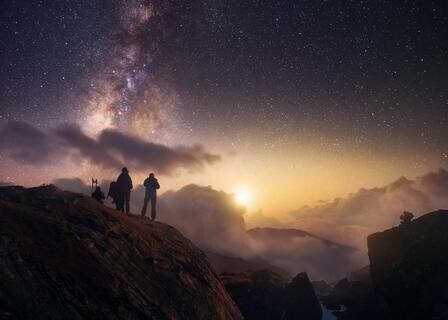} & \includegraphics[width=\xx,frame]{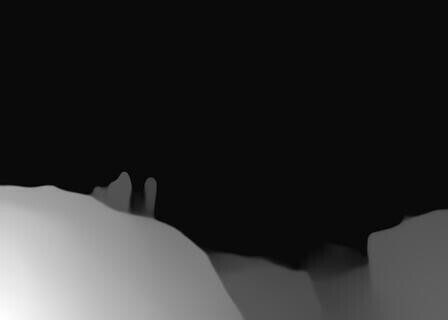} & \includegraphics[width=\xx,frame]{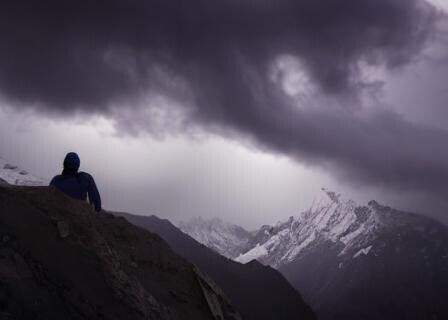} \\[-0.1cm]
\multicolumn{3}{X}{\scriptsize{``Group of hikers looking at the stormy sky in mountains of the Himalayas''}} \\[0.2cm]

\includegraphics[width=\xx,frame]{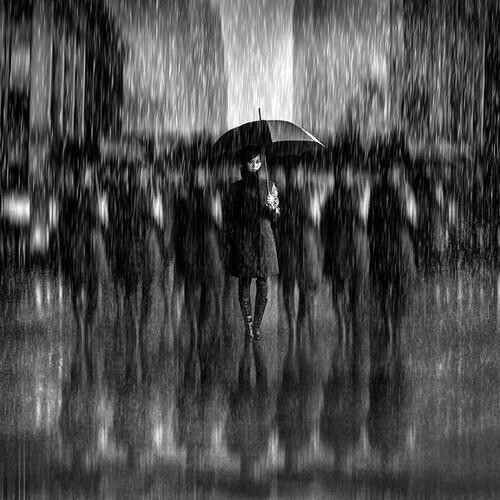} &
\includegraphics[width=\xx,frame]{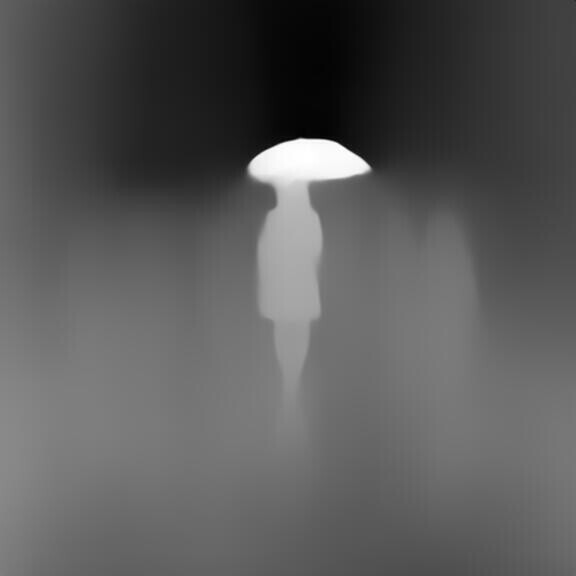} &
\includegraphics[width=\xx,frame]{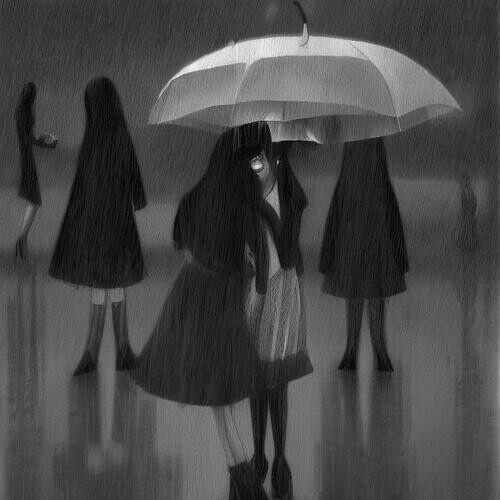}  \\[-0.1cm]
\multicolumn{3}{X}{\scriptsize{``Cartoon - Girls In The Rain by Antonyus Bunjamin (abe)''}} \\[0.2cm]

\includegraphics[width=\xx,frame]{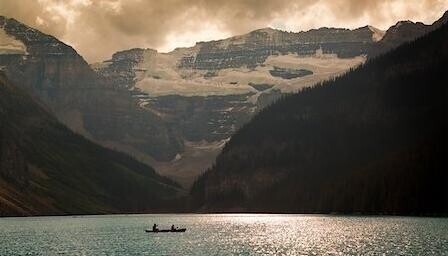} & \includegraphics[width=\xx,frame]{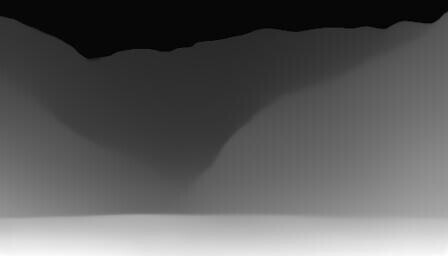} & \includegraphics[width=\xx,frame]{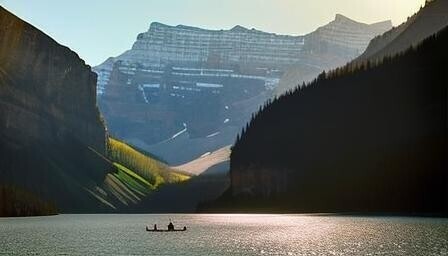} \\
\end{tabularx}

\end{figure}
\begin{figure}

  \centering
  \begin{tabularx}{\columnwidth}{p{1.01\xx}p{1.01\xx}@{\hskip 0.2cm}p{1.01\xx}}

\multicolumn{2}{c}{\textbf{Input Image + Change Map}} & \multicolumn{1}{c}{\textbf{Output Image}} \\[0.2cm] %
\includegraphics[width=\xx,frame]{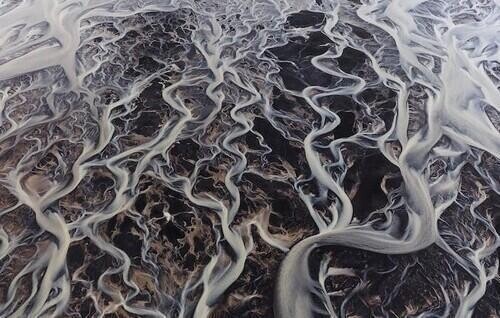} & \includegraphics[width=\xx,frame]{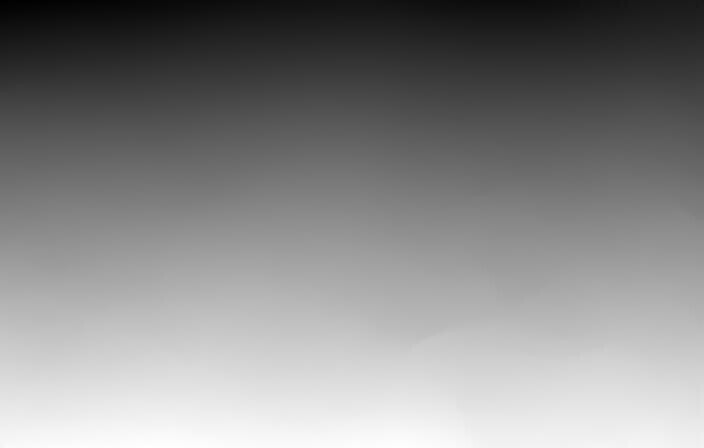} & \includegraphics[width=\xx,frame]{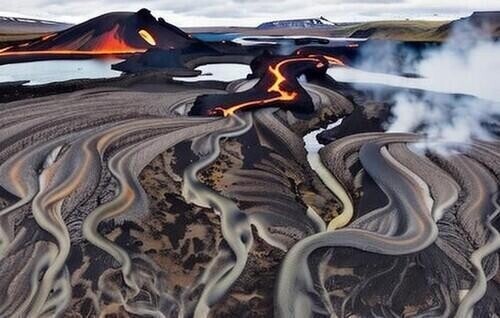} \\[-0.1cm] \multicolumn{3}{X}{\scriptsize{``A river of lava winding its way through the Icelandic Highlands.''}} \\[0.2cm]
\includegraphics[width=\xx,frame]{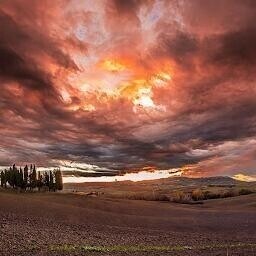} &
\includegraphics[width=\xx,frame]{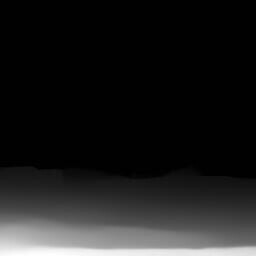} &
\includegraphics[width=\xx,frame]{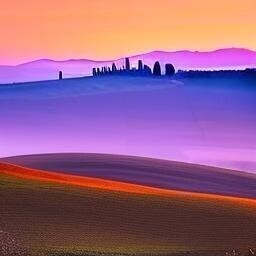}   \\[-0.1cm] \multicolumn{3}{X}{\scriptsize{``Dramatic Tramonto in Val d'Orcia by Pasquale Bimonte - Landscapes Sunsets \& Sunrises ''}} \\[0.5cm] %

\includegraphics[width=\xx,frame]{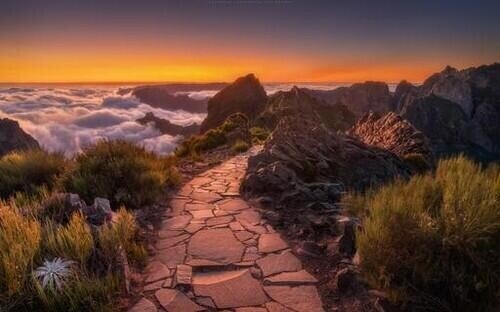} & \includegraphics[width=\xx,frame]{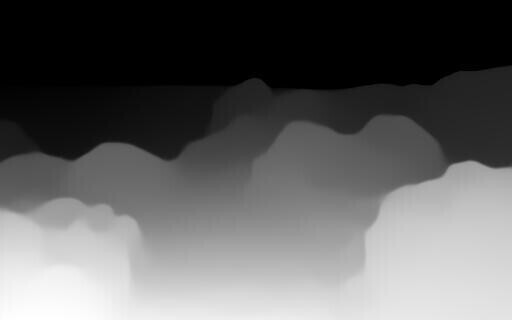} & \includegraphics[width=\xx,frame]{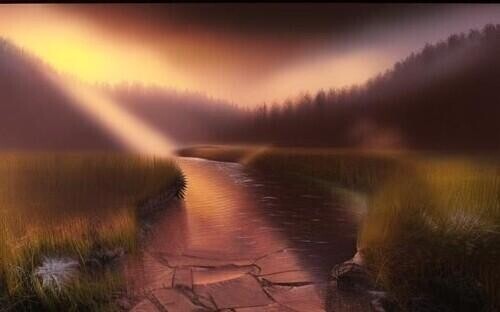} \\[-0.1cm] \multicolumn{3}{X}{\scriptsize{``River into the Light by Alexander Lauterbach''}} \\[0.2cm]

\includegraphics[width=\xx,frame]{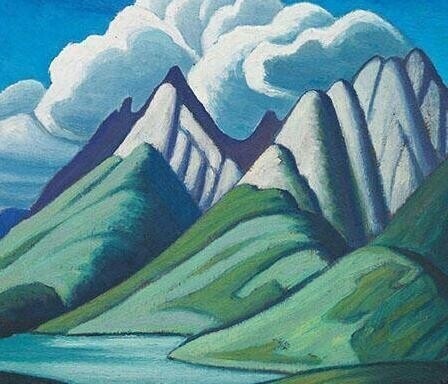} &
\includegraphics[width=\xx,frame]{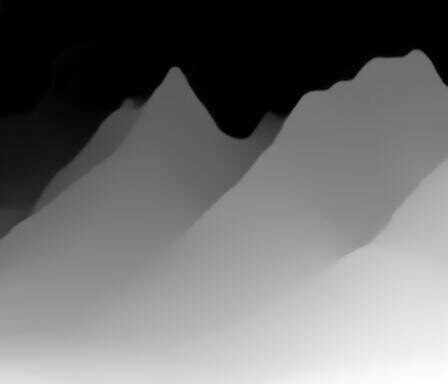} &
\includegraphics[width=\xx,frame]{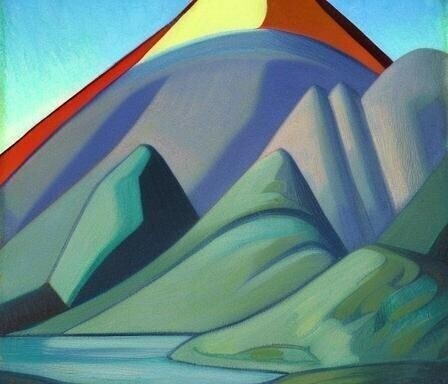}  \\
\end{tabularx}

\end{figure}

\begin{figure}

  \centering
  \begin{tabularx}{\columnwidth}{p{1.01\xx}p{1.01\xx}@{\hskip 0.2cm}p{1.01\xx}}

\multicolumn{2}{c}{\textbf{Input Image + Change Map}} & \multicolumn{1}{c}{\textbf{Output Image}} \\[0.2cm] %

\includegraphics[width=\xx,frame]{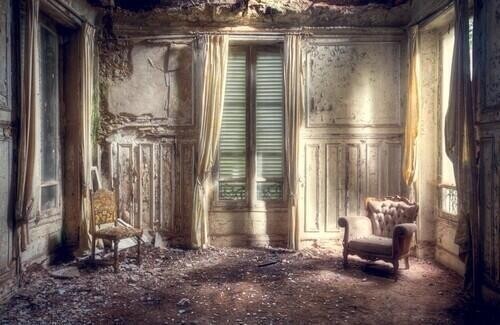} &
\includegraphics[width=\xx,frame]{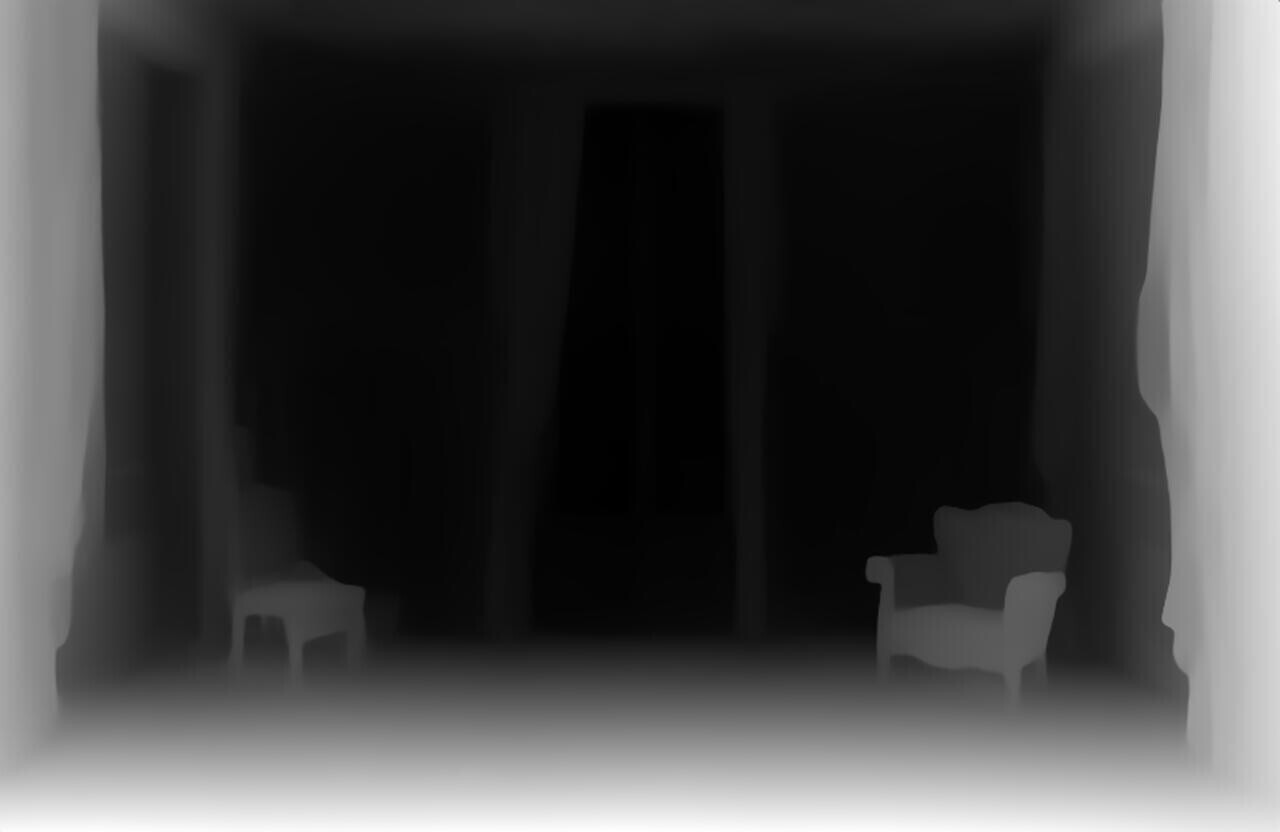} &
\includegraphics[width=\xx,frame]{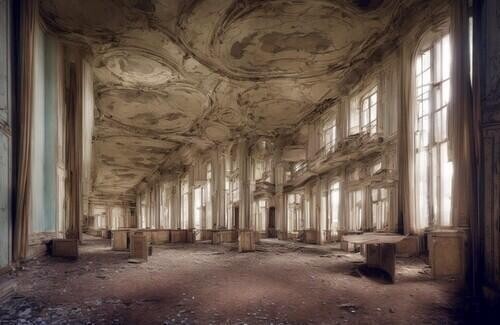}  \\[-0.1cm] \multicolumn{3}{X}{\scriptsize{``15 Photos of Abandoned Libraries in Decay - Urban ...''}} \\[0.2cm] %
\includegraphics[width=\xx,frame]{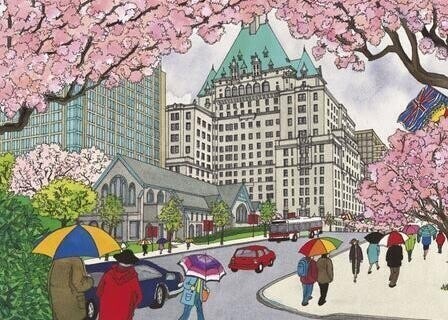} & \includegraphics[width=\xx,frame]{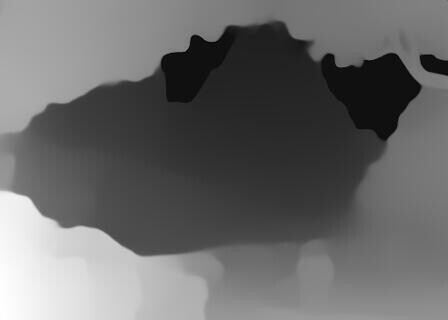} & \includegraphics[width=\xx,frame]{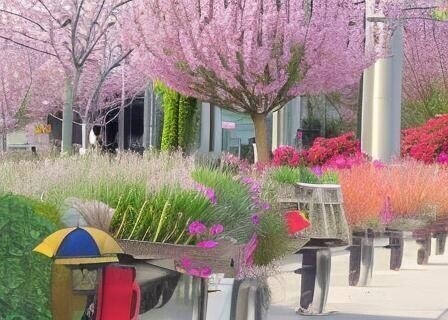} \\[-0.1cm] \multicolumn{3}{X}{\scriptsize{``Spring Plants on Burrard Street''}} \\[0.2cm]

\includegraphics[width=\xx,frame]{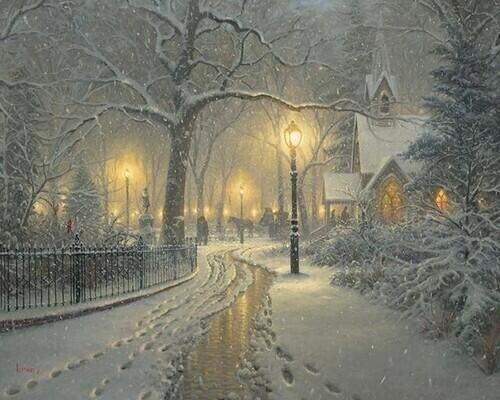} &
\includegraphics[width=\xx,frame]{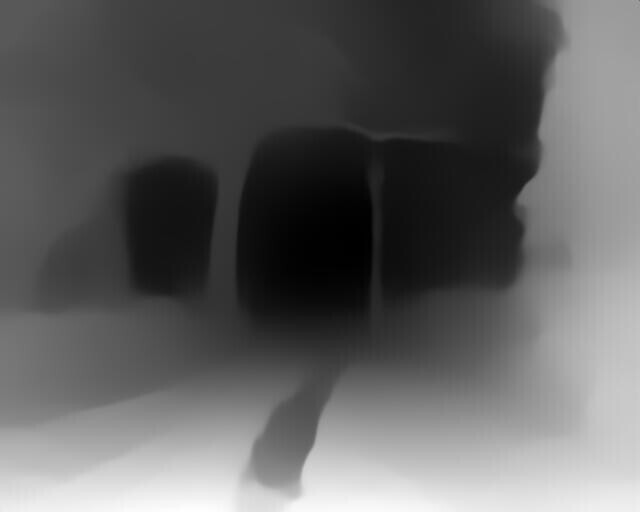} &
\includegraphics[width=\xx,frame]{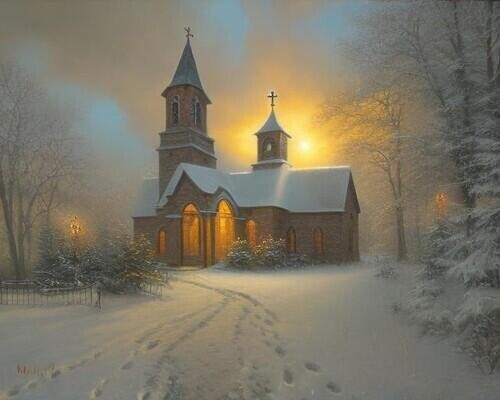}  \\[-0.1cm] \multicolumn{3}{X}{\scriptsize{``SWinter Chapel with Tornado by Mark Keathley''}} \\[0.2cm] %
\includegraphics[width=\xx,frame]{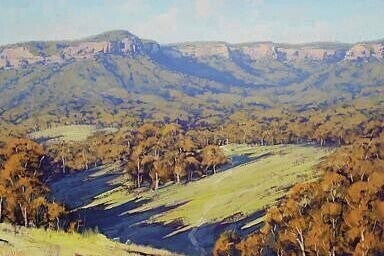} & \includegraphics[width=\xx,frame]{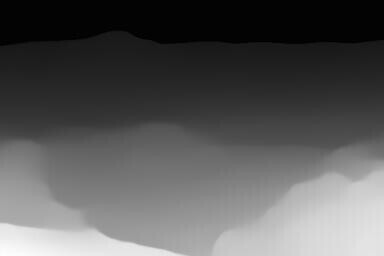} & \includegraphics[width=\xx,frame]{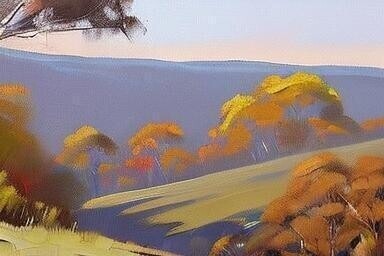} \\
\end{tabularx}

\end{figure}

\begin{figure}

  \centering
  \begin{tabularx}{\columnwidth}{p{1.01\xx}p{1.01\xx}@{\hskip 0.2cm}p{1.01\xx}}

\multicolumn{2}{c}{\textbf{Input Image + Change Map}} & \multicolumn{1}{c}{\textbf{Output Image}} \\[0.2cm] %
\includegraphics[width=\xx,frame]{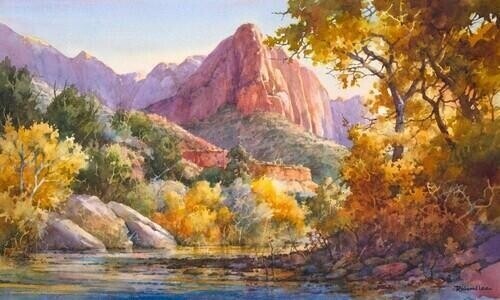} & \includegraphics[width=\xx,frame]{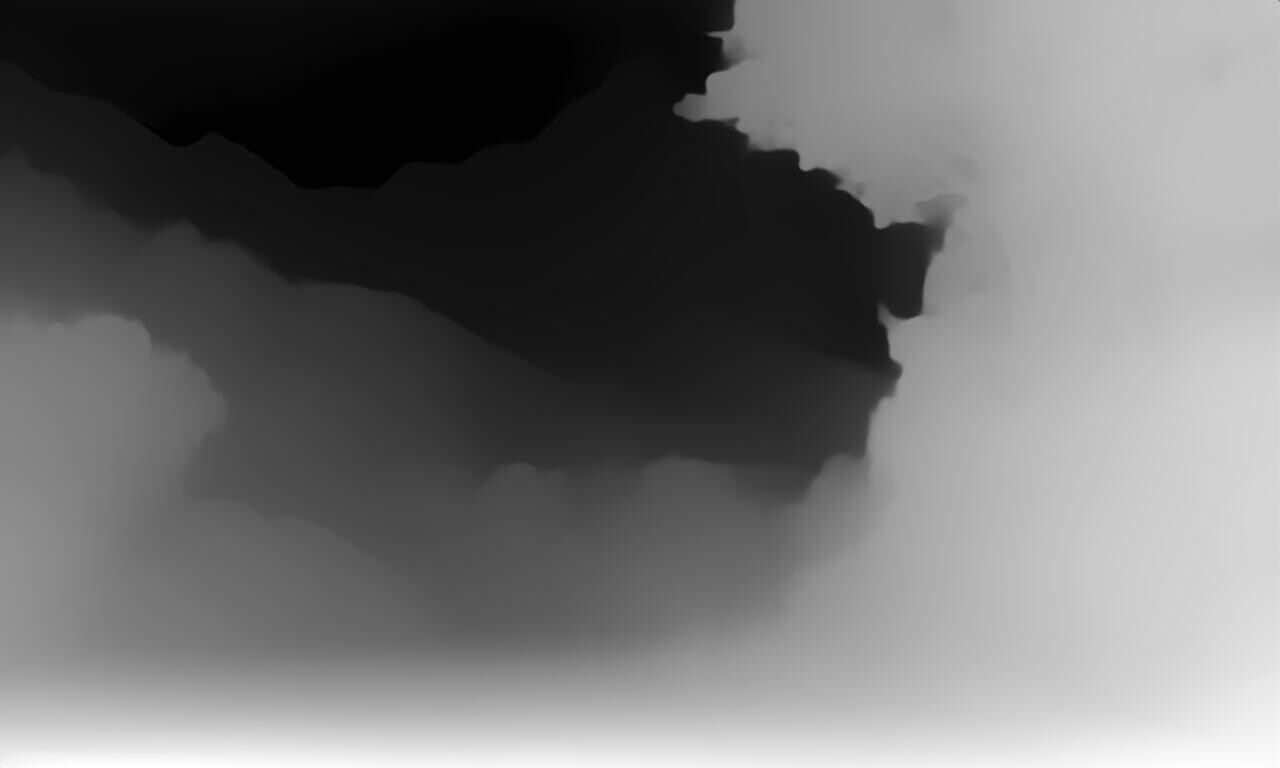} & \includegraphics[width=\xx,frame]{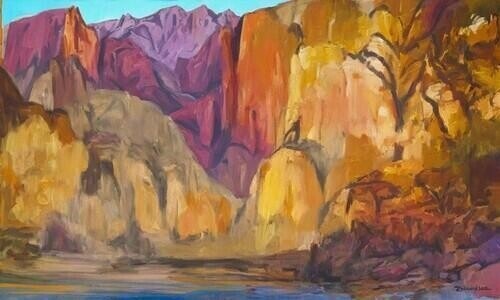} \\[-0.1cm] \multicolumn{3}{X}{\scriptsize{``Andy Warhol's Zion Canyon Oil Painting''}} \\[0.2cm]

\includegraphics[width=\xx,frame]{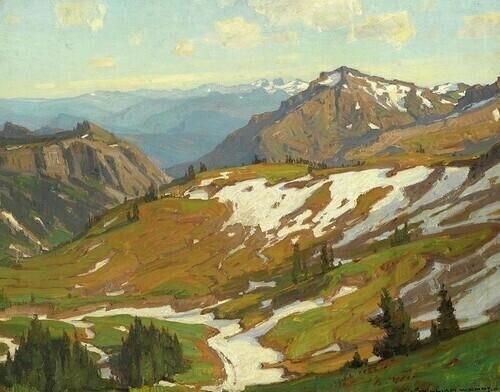} &
\includegraphics[width=\xx,frame]{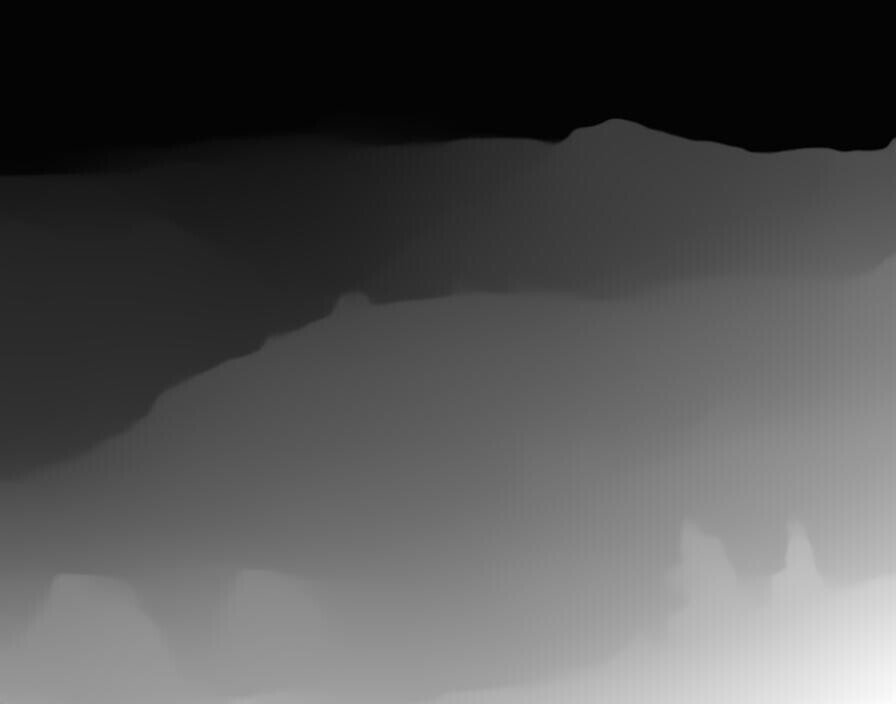} &
\includegraphics[width=\xx,frame]{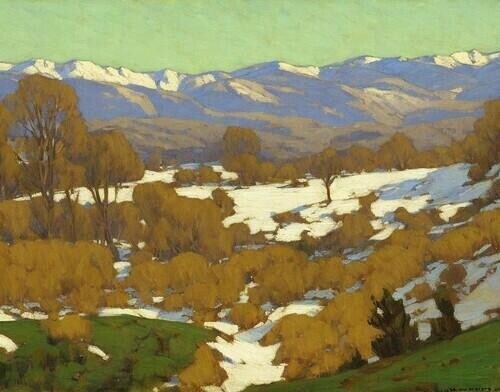}  \\[-0.1cm] \multicolumn{3}{X}{\scriptsize{``William Wendt Painting - Winter Thaw by William Wendt''}} \\[0.2cm] %
\includegraphics[width=\xx,frame]{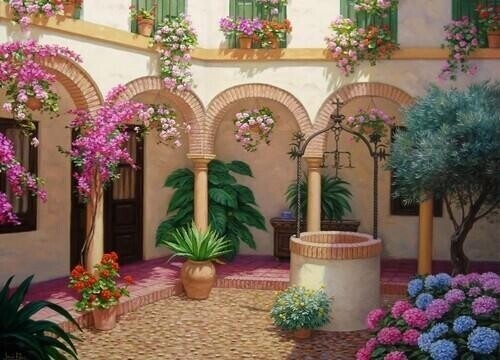} & \includegraphics[width=\xx,frame]{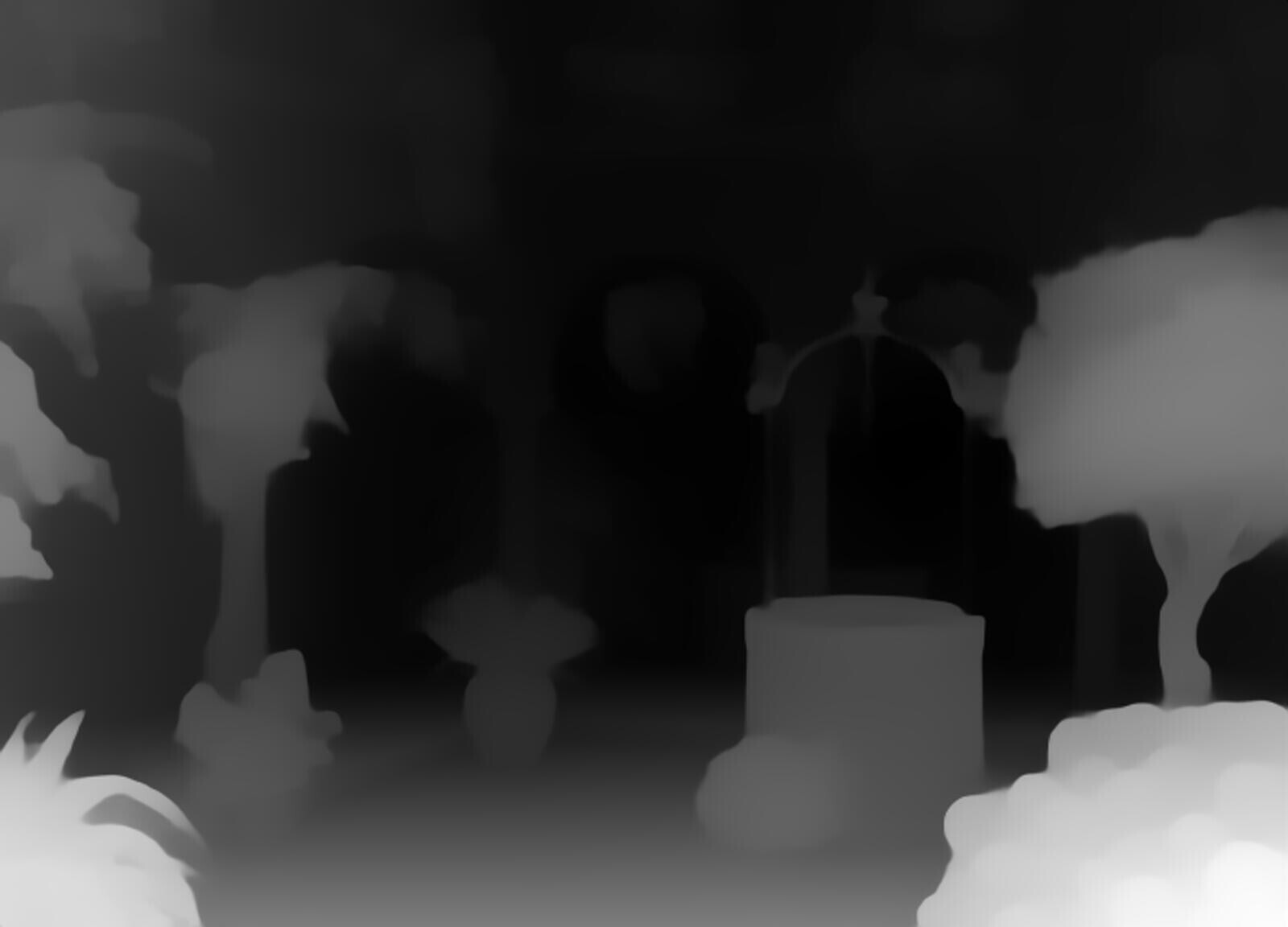} & \includegraphics[width=\xx,frame]{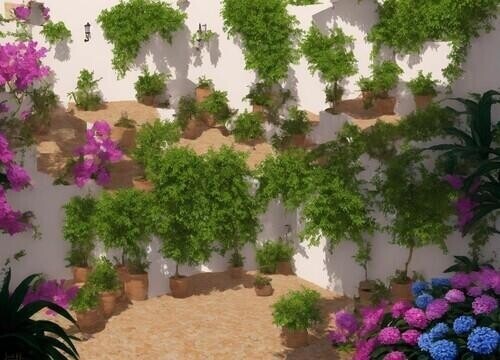} \\[-0.1cm] \multicolumn{3}{X}{\scriptsize{``Jes s fern ndez 3D rendering patio andaluz con pozo y olivo for Patios andaluces decoracion''}} \\[0.5cm]

\includegraphics[width=\xx,frame]{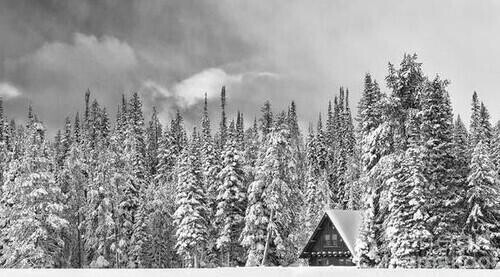} &
\includegraphics[width=\xx,frame]{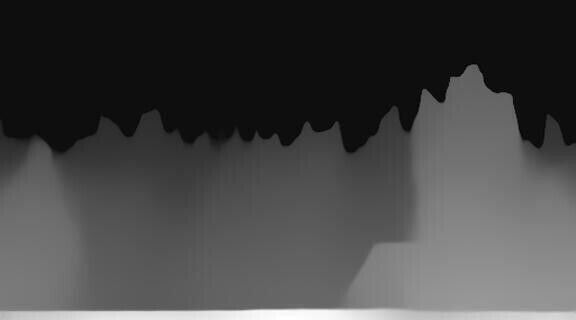} &
\includegraphics[width=\xx,frame]{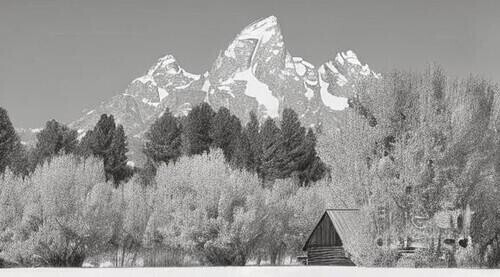}  \\
\end{tabularx}

\end{figure}

\begin{figure}

  \centering
  \begin{tabularx}{\columnwidth}{p{1.01\xx}p{1.01\xx}@{\hskip 0.2cm}p{1.01\xx}}

\multicolumn{2}{c}{\textbf{Input Image + Change Map}} & \multicolumn{1}{c}{\textbf{Output Image}} \\[0.2cm] %
\includegraphics[width=\xx,frame]{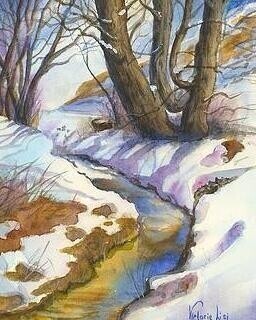} & \includegraphics[width=\xx,frame]{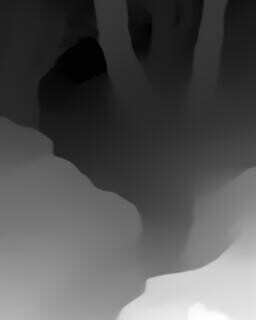} & \includegraphics[width=\xx,frame]{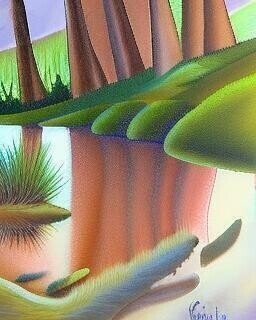} \\[-0.1cm] \multicolumn{3}{X}{\scriptsize{``Creek at Bobcat Ridge with kudzu by Victoria Lisi''}} \\[0.2cm]

\includegraphics[width=\xx,frame]{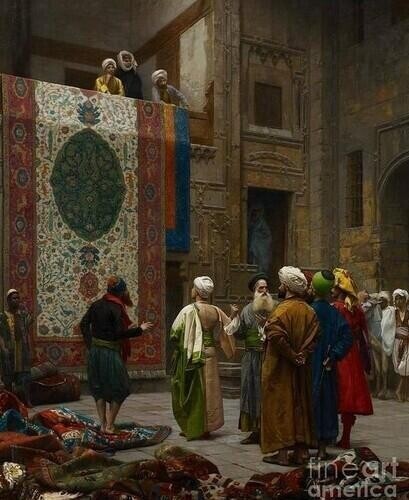} &
\includegraphics[width=\xx,frame]{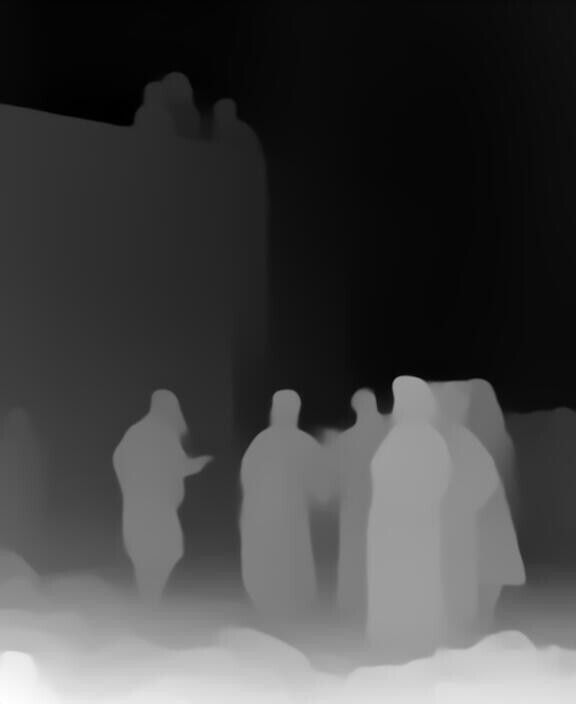} &
\includegraphics[width=\xx,frame]{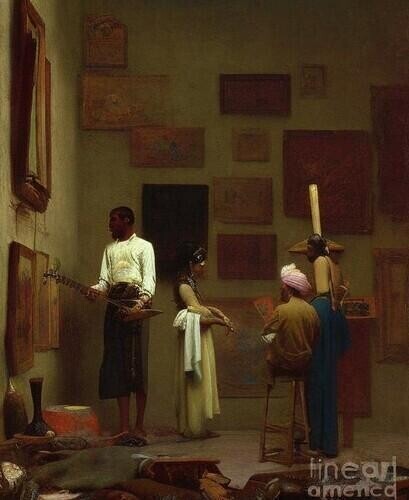}  \\[-0.1cm] \multicolumn{3}{X}{\scriptsize{``Gerome Art Print featuring the painting The Guitar Merchant by Jean Leon Gerome''}} \\[0.2cm] %

\includegraphics[width=\xx,frame]{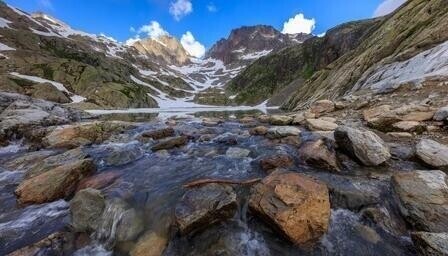} & \includegraphics[width=\xx,frame]{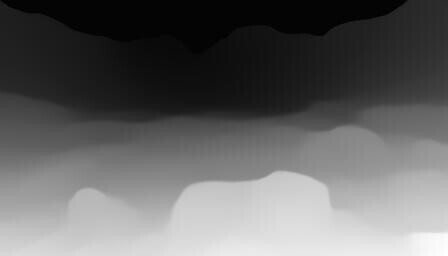} & \includegraphics[width=\xx,frame]{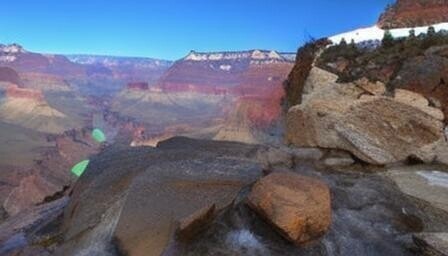} \\[-0.1cm] \multicolumn{3}{X}{\scriptsize{``lac: Lac Blanc in Grand Canyon, USA''}} \\[0.2cm]

\includegraphics[width=\xx,frame]{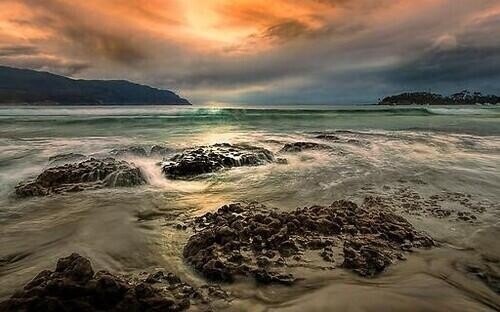} &
\includegraphics[width=\xx,frame]{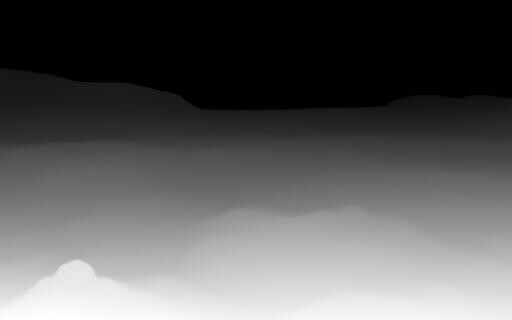} &
\includegraphics[width=\xx,frame]{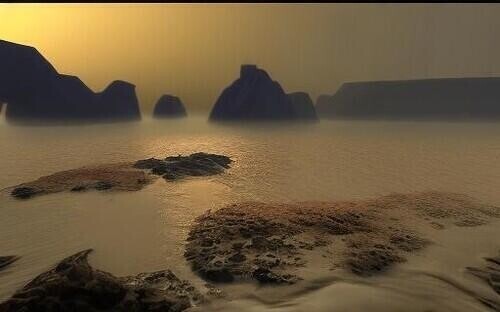}  \\
\end{tabularx}

\end{figure}
\begin{figure}

  \centering
  \begin{tabularx}{\columnwidth}{p{1.01\xx}p{1.01\xx}@{\hskip 0.2cm}p{1.01\xx}}

\multicolumn{2}{c}{\textbf{Input Image + Change Map}} & \multicolumn{1}{c}{\textbf{Output Image}} \\[0.2cm] %
\includegraphics[width=\xx,frame]{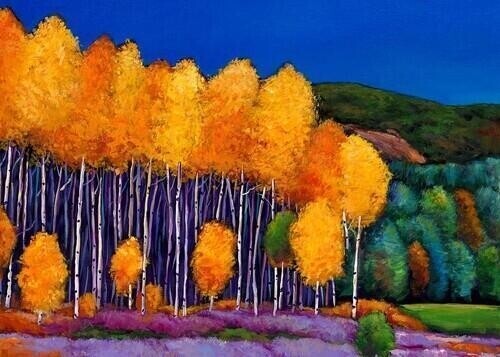} &
\includegraphics[width=\xx,frame]{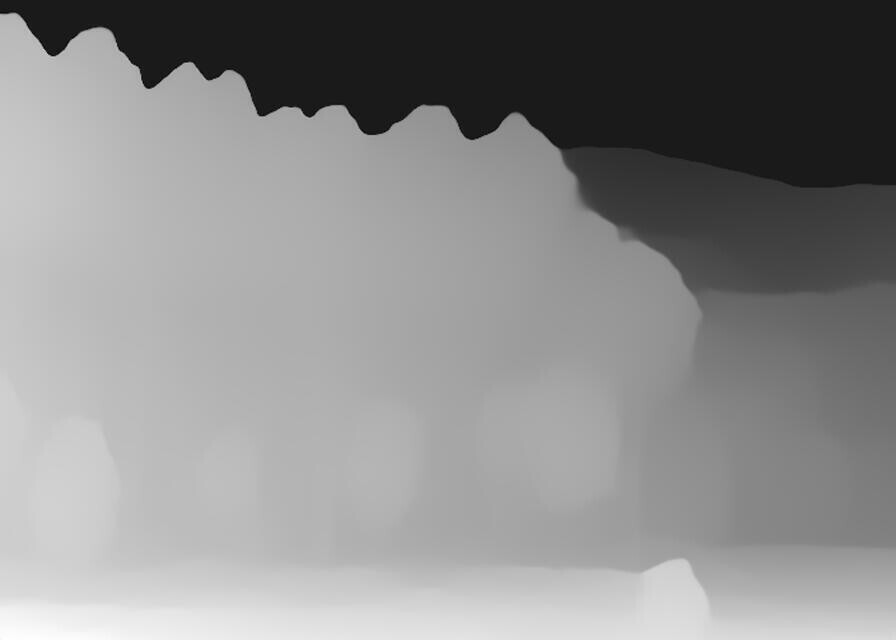} &
\includegraphics[width=\xx,frame]{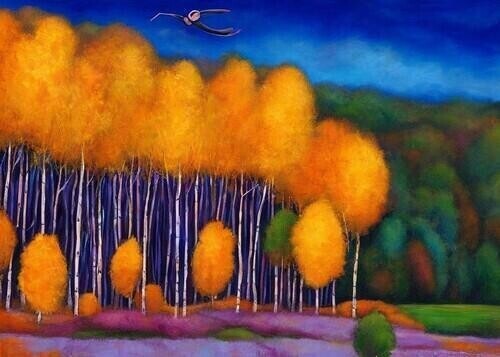}  \\[-0.1cm] \multicolumn{3}{X}{\scriptsize{``Statue - A Moment In Time by Johnathan Harris''}} \\[0.2cm] %
\includegraphics[width=\xx,frame]{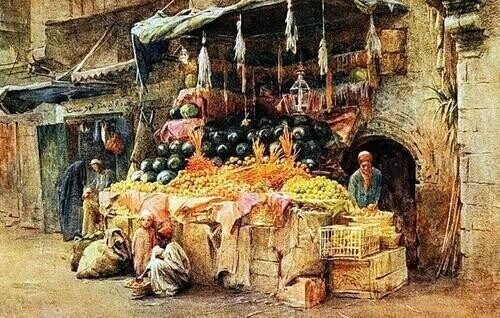} & \includegraphics[width=\xx,frame]{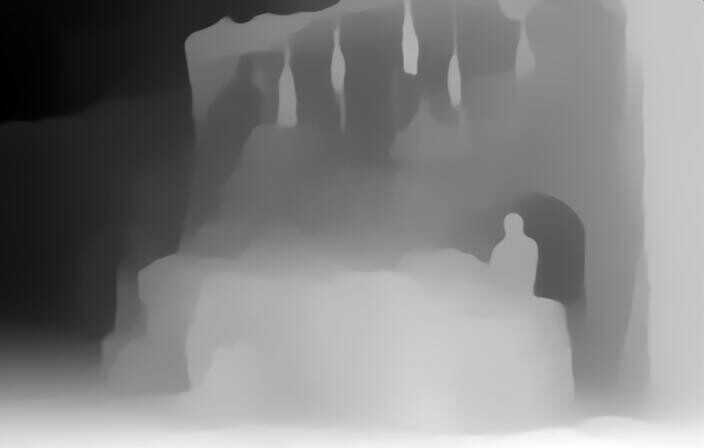} & \includegraphics[width=\xx,frame]{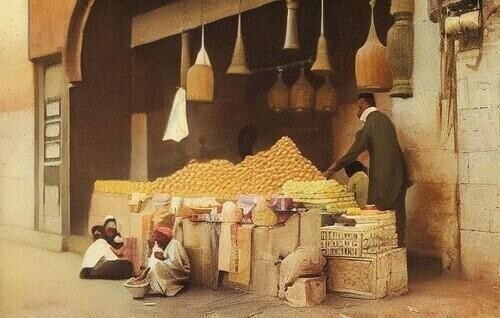} \\[-0.1cm] \multicolumn{3}{X}{\scriptsize{``Chocolatier in Cairo , Egypt 1907 by Walter Tyndale (English , 1855–1943)''}} \\[0.2cm]
\includegraphics[width=\xx,frame]{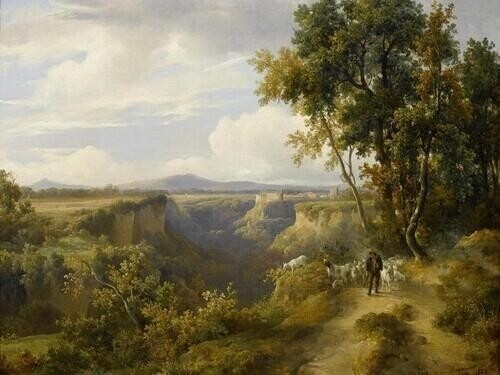} &
\includegraphics[width=\xx,frame]{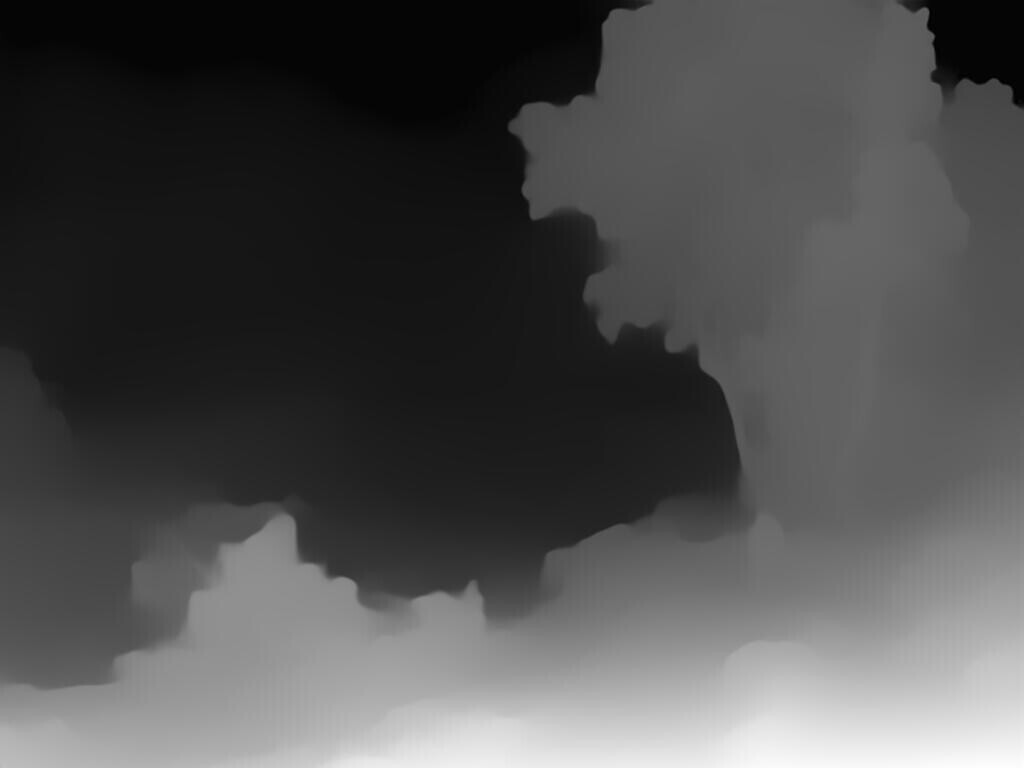} &
\includegraphics[width=\xx,frame]{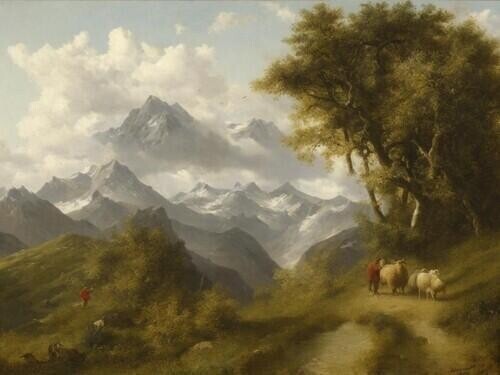}  \\[-0.1cm] \multicolumn{3}{X}{\scriptsize{``Jacques Raymond Brascassat - A goatherder and his flock in an Alpine landscape''}} \\[0.2cm] %
\includegraphics[width=\xx,frame]{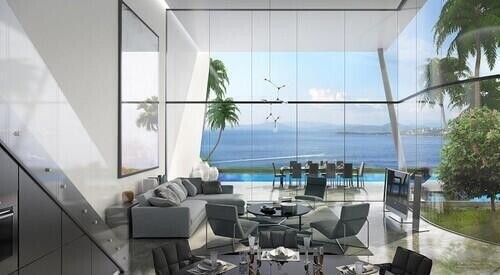} & \includegraphics[width=\xx,frame]{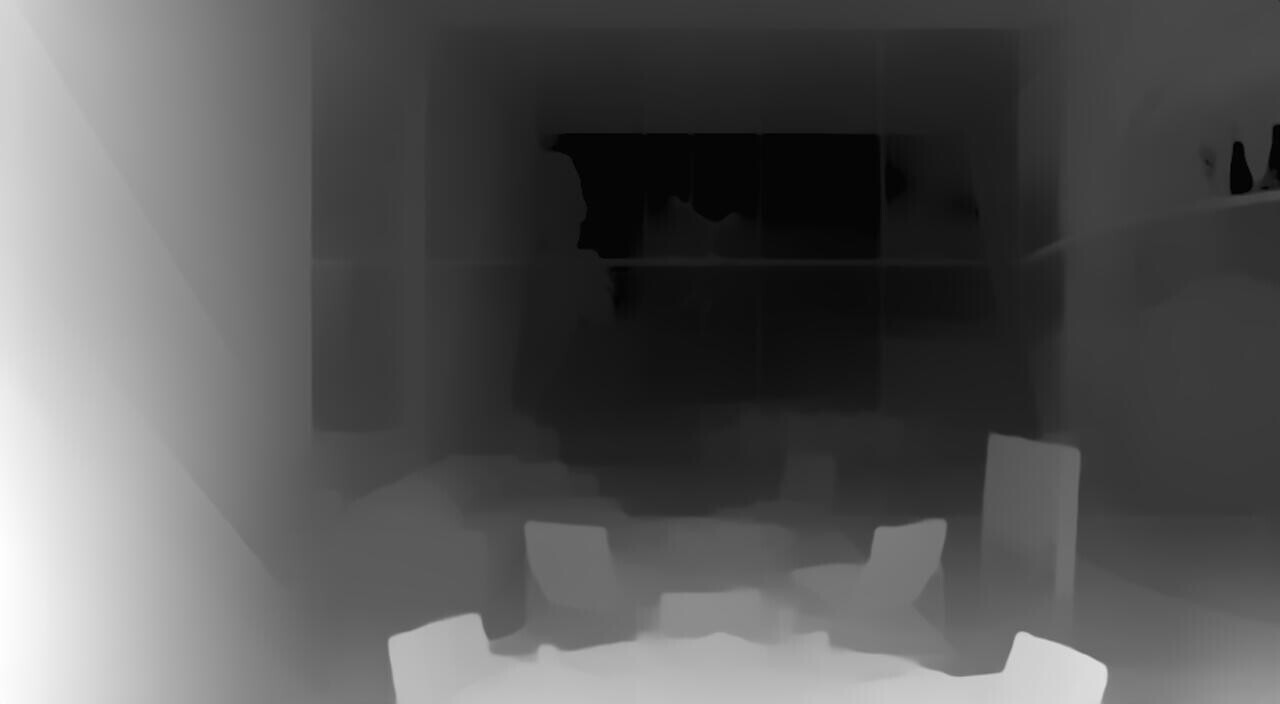} & \includegraphics[width=\xx,frame]{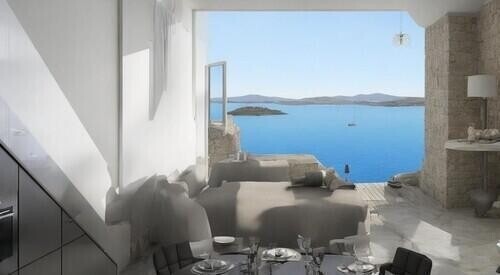} \\
\end{tabularx}

\end{figure}

\begin{figure}

  \centering
  \begin{tabularx}{\columnwidth}{p{1.01\xx}p{1.01\xx}@{\hskip 0.2cm}p{1.01\xx}}

\multicolumn{2}{c}{\textbf{Input Image + Change Map}} & \multicolumn{1}{c}{\textbf{Output Image}} \\[0.2cm] %

\includegraphics[width=\xx,frame]{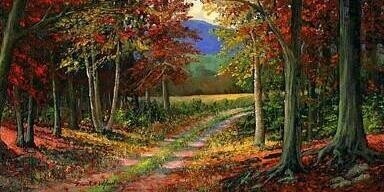} &
\includegraphics[width=\xx,frame]{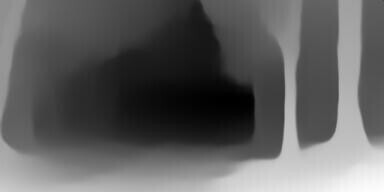} &
\includegraphics[width=\xx,frame]{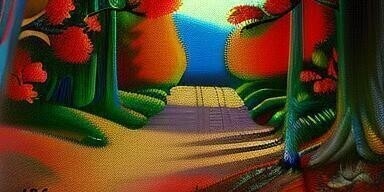}  \\[-0.1cm] \multicolumn{3}{X}{\scriptsize{``Foliage Painting movie poster - Forgotten Road by Frank Wilson''}} \\[0.2cm] %
\includegraphics[width=\xx,frame]{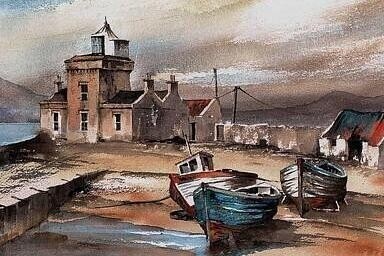} & \includegraphics[width=\xx,frame]{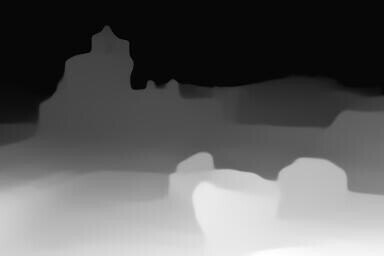} & \includegraphics[width=\xx,frame]{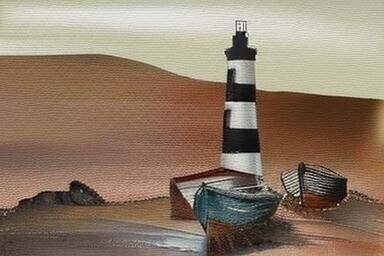} \\[-0.1cm] \multicolumn{3}{X}{\scriptsize{``Mayo Painting - Belmullet Lighthouse  Mayo by Val Byrne, broken lighthouse''}} \\[0.2cm]
\includegraphics[width=\xx,frame]{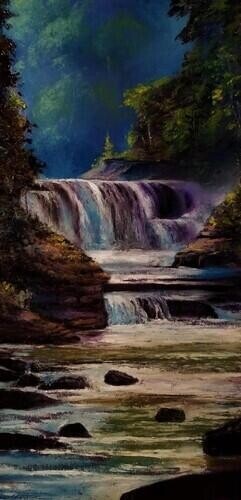} &
\includegraphics[width=\xx,frame]{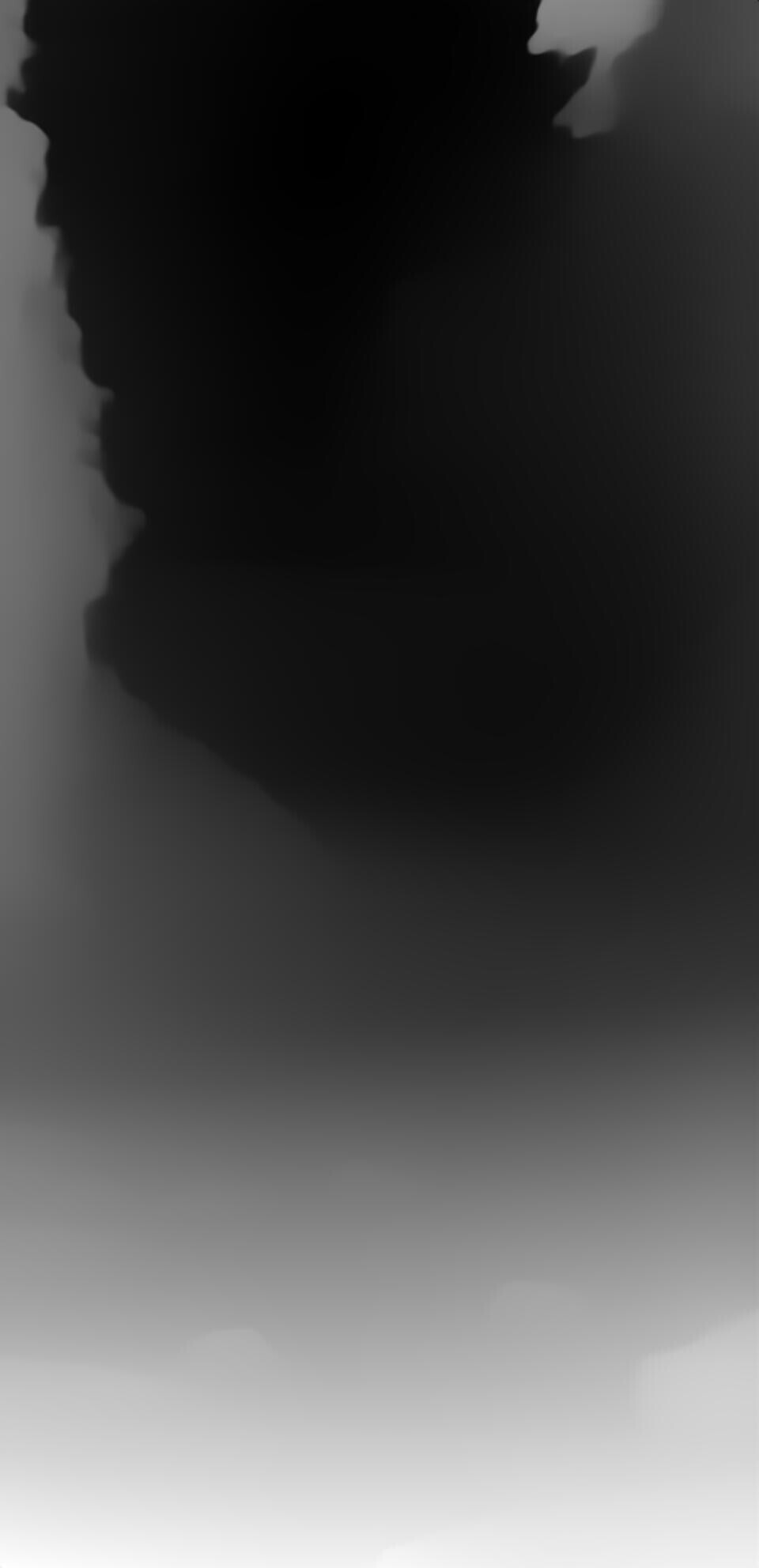} &
\includegraphics[width=\xx,frame]{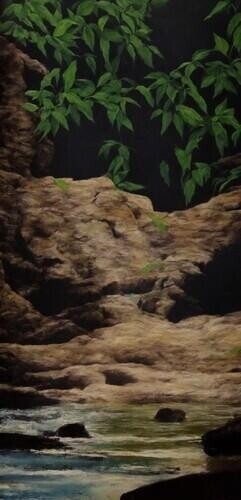}  \\[-0.1cm] \multicolumn{3}{X}{\scriptsize{``Add Artwork | Wallhanging by Ray Easton | Artists for Conservation, Jungle Setting |''}} \\[0.5cm] %
\includegraphics[width=\xx,frame]{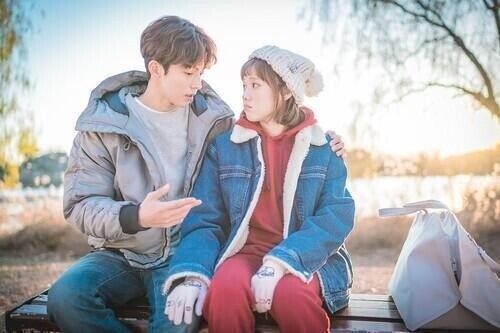} & \includegraphics[width=\xx,frame]{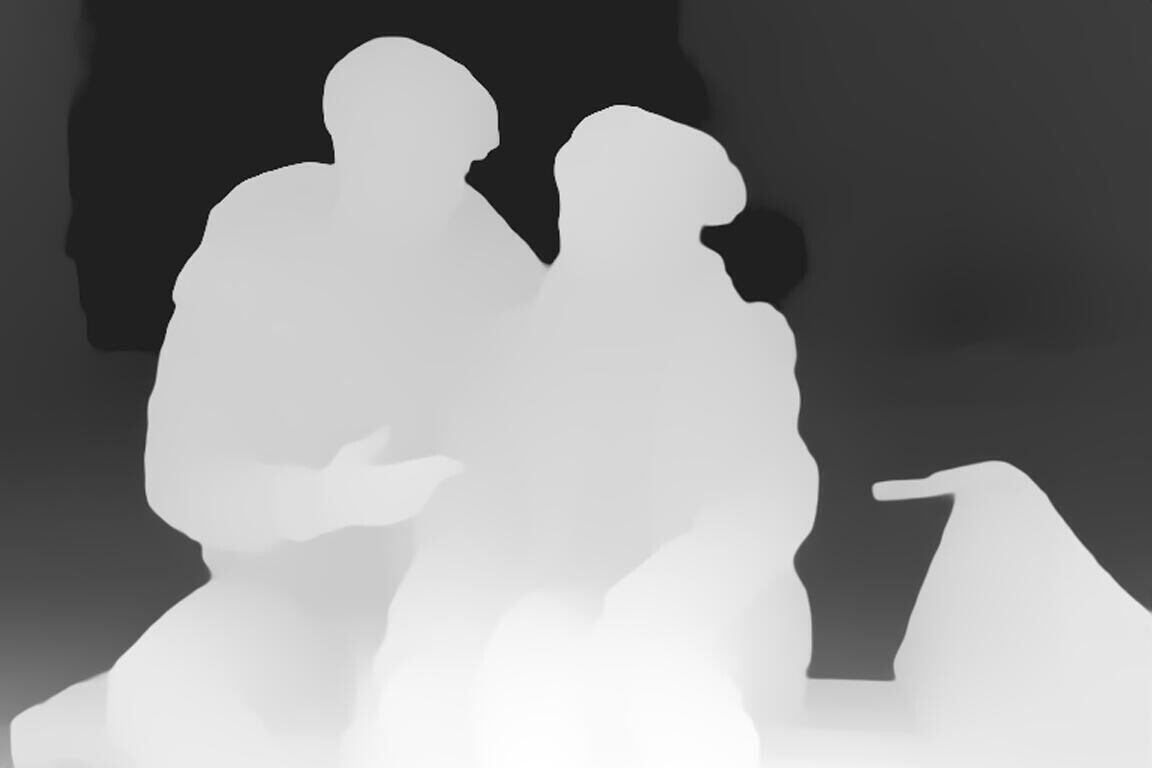} & \includegraphics[width=\xx,frame]{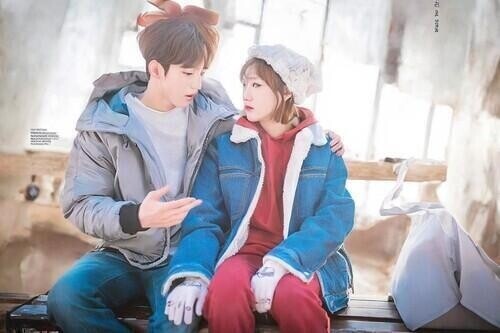} \\
\end{tabularx}

\end{figure}

\begin{figure}

  \centering
  \begin{tabularx}{\columnwidth}{p{1.01\xx}p{1.01\xx}@{\hskip 0.2cm}p{1.01\xx}}

\multicolumn{2}{c}{\textbf{Input Image + Change Map}} & \multicolumn{1}{c}{\textbf{Output Image}} \\[0.2cm] %

\includegraphics[width=\xx,frame]{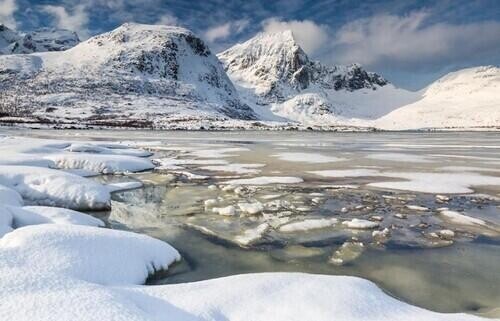} &
\includegraphics[width=\xx,frame]{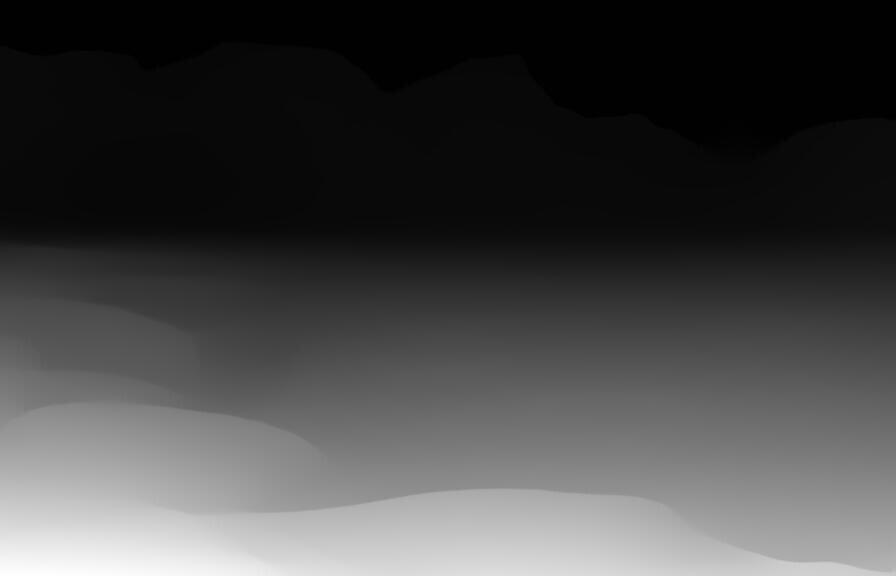} &
\includegraphics[width=\xx,frame]{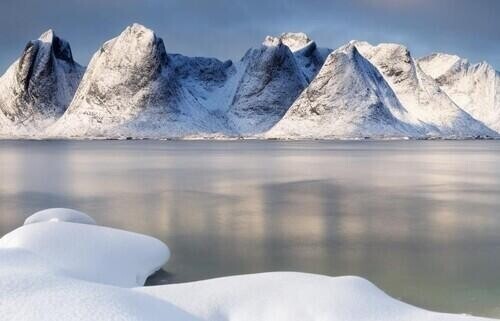}  \\[-0.1cm] \multicolumn{3}{X}{\scriptsize{``Winter warm winters of the Lofoten Islands 10''}} \\[0.2cm] %
\includegraphics[width=\xx,frame]{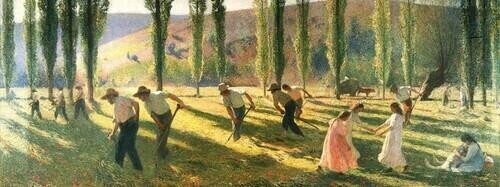} & \includegraphics[width=\xx,frame]{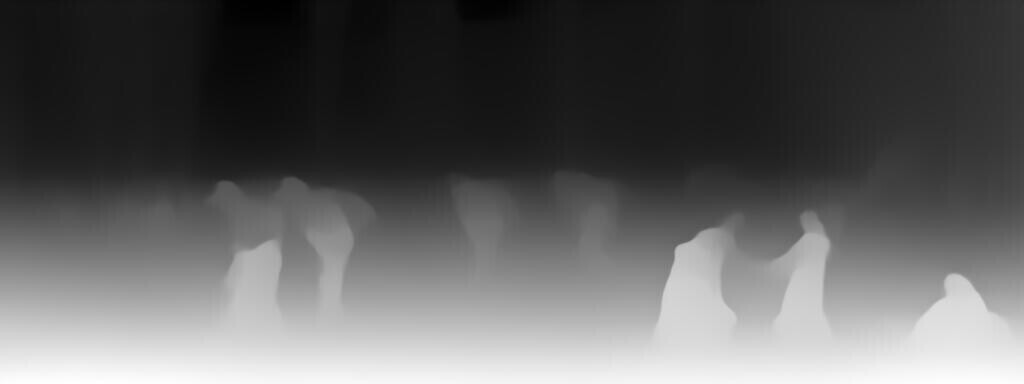} & \includegraphics[width=\xx,frame]{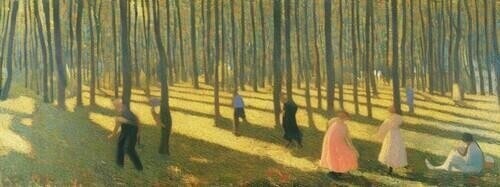} \\[-0.1cm] \multicolumn{3}{X}{\scriptsize{``Image result for henri martin in a forest''}} \\[0.2cm]
\includegraphics[width=\xx,frame]{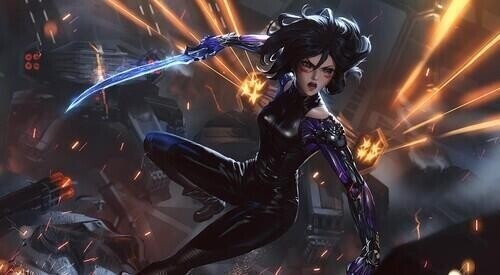} &
\includegraphics[width=\xx,frame]{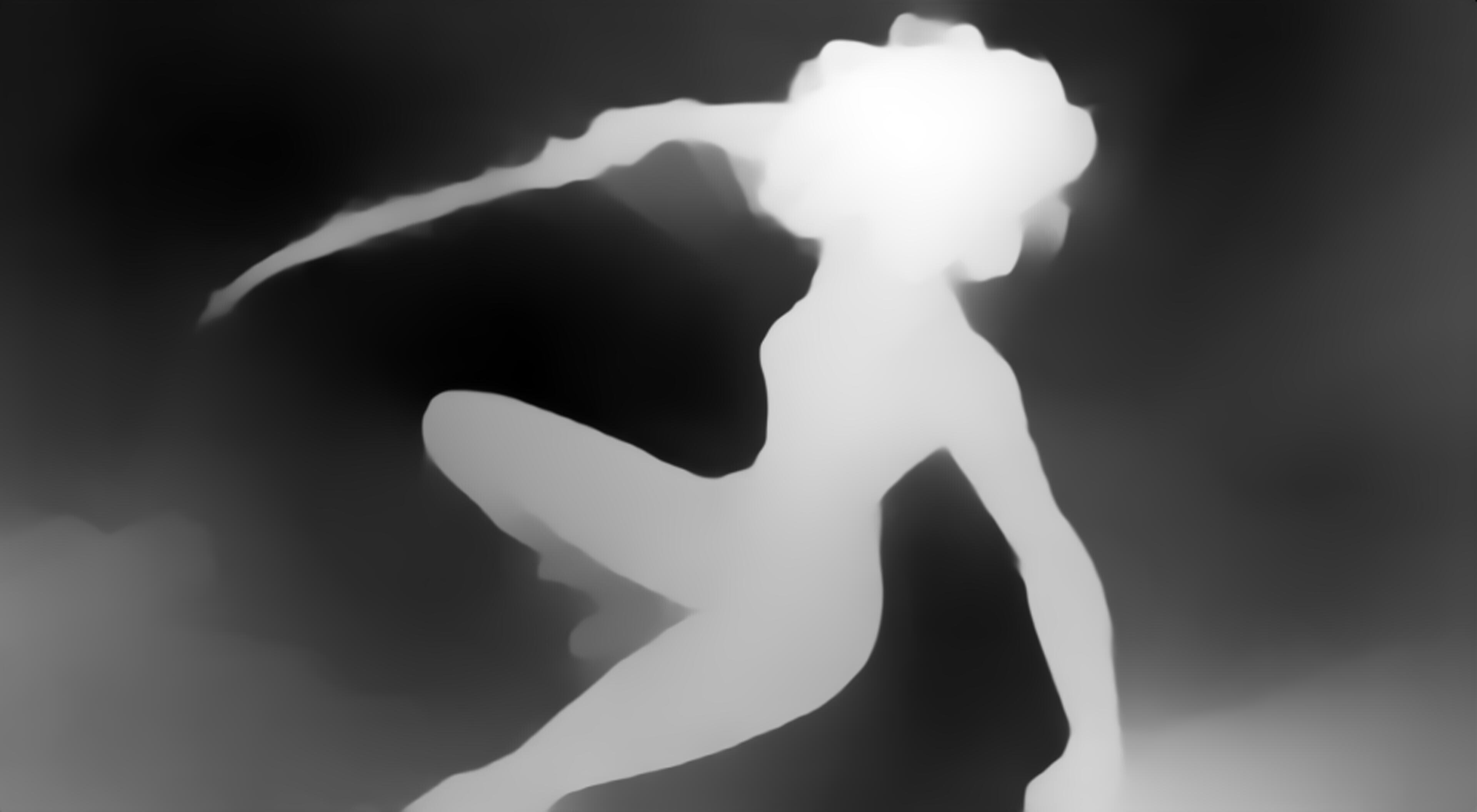} &
\includegraphics[width=\xx,frame]{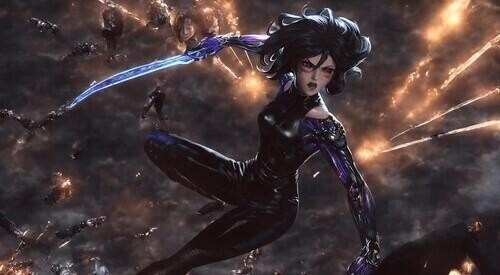}  \\[-0.1cm] \multicolumn{3}{X}{\scriptsize{``Alita Battle Angel Sword 4k Stage Prop 34''}} \\[0.2cm] %
\includegraphics[width=\xx,frame]{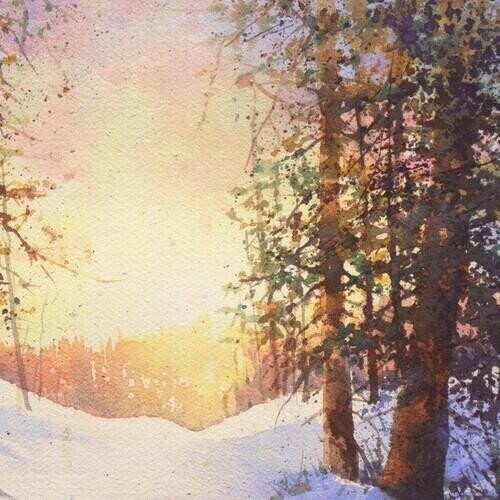} & \includegraphics[width=\xx,frame]{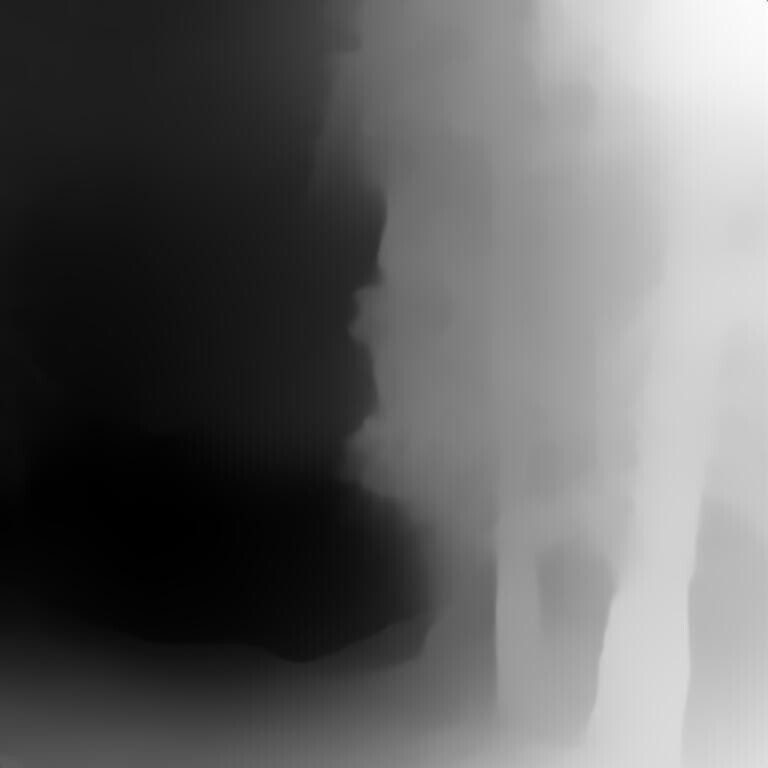} & \includegraphics[width=\xx,frame]{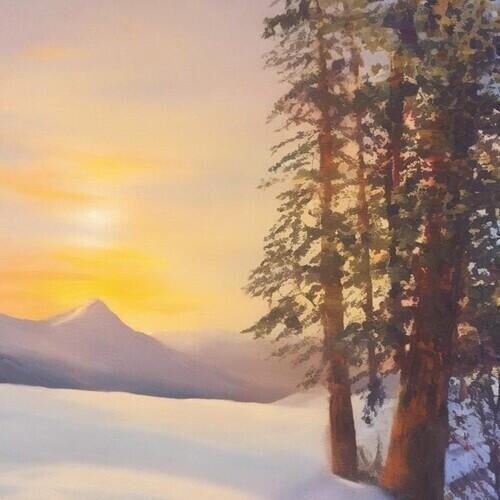} \\
\end{tabularx}

\end{figure}

\begin{figure}

  \centering
  \begin{tabularx}{\columnwidth}{p{1.01\xx}p{1.01\xx}@{\hskip 0.2cm}p{1.01\xx}}

\multicolumn{2}{c}{\textbf{Input Image + Change Map}} & \multicolumn{1}{c}{\textbf{Output Image}} \\[0.2cm] %
\includegraphics[width=\xx,frame]{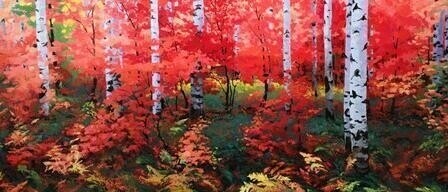} &
\includegraphics[width=\xx,frame]{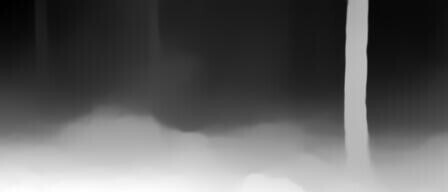} &
\includegraphics[width=\xx,frame]{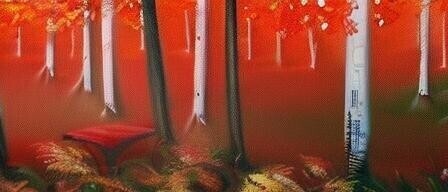}  \\[-0.1cm] \multicolumn{3}{X}{\scriptsize{``""Ruby Red Autumn"" Stanislav Sidorov 30x60" Painting''}} \\[0.2cm] %
\includegraphics[width=\xx,frame]{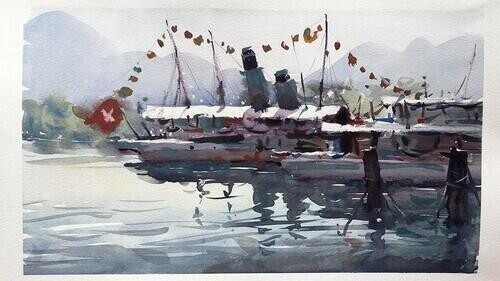} & \includegraphics[width=\xx,frame]{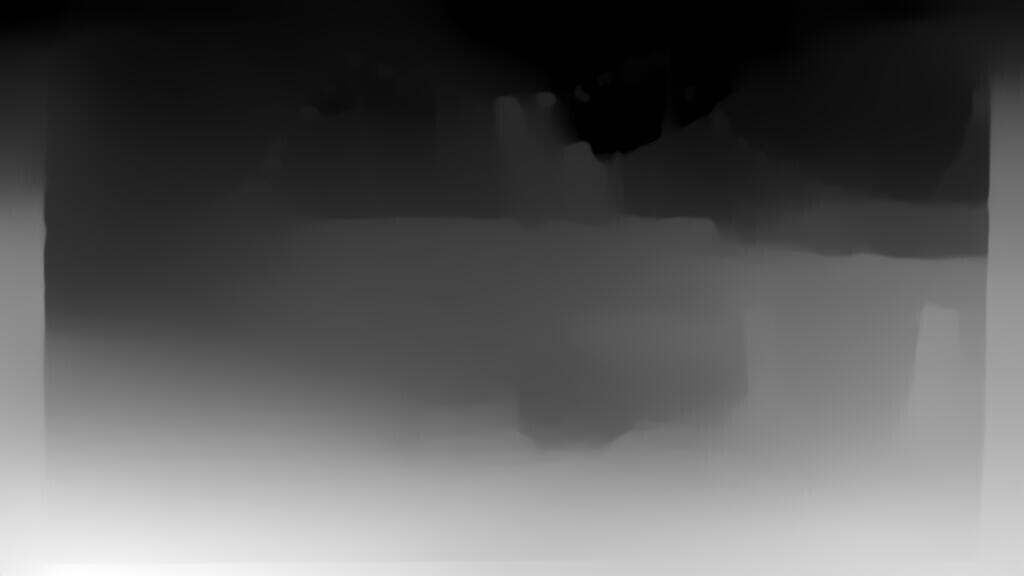} & \includegraphics[width=\xx,frame]{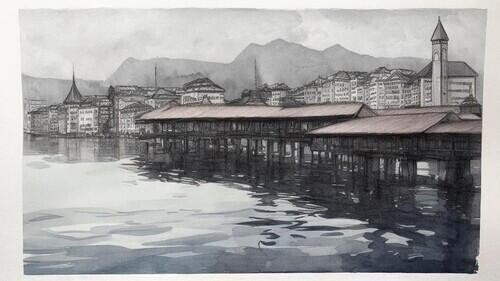} \\[-0.1cm] \multicolumn{3}{X}{\scriptsize{``Lucerne Dock - Drawing - Marco Bucci Art Store''}} \\[0.2cm]

\includegraphics[width=\xx,frame]{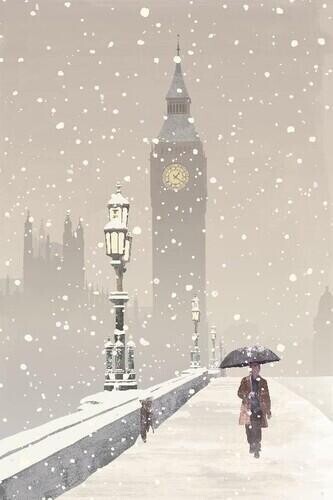} &
\includegraphics[width=\xx,frame]{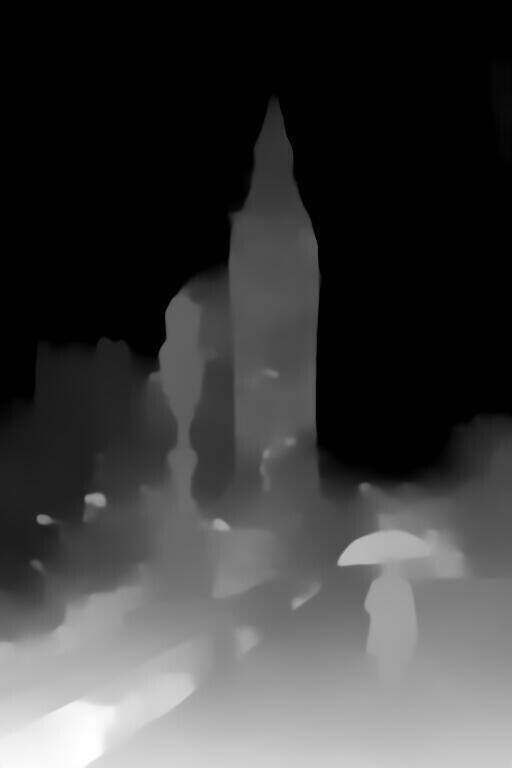} &
\includegraphics[width=\xx,frame]{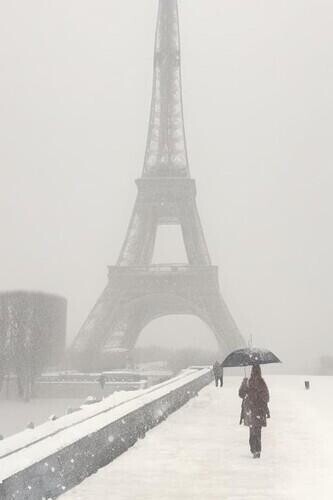}  \\[-0.1cm] \multicolumn{3}{X}{\scriptsize{``Big Eiffel Tower in the Snow''}} \\[0.2cm] %

\includegraphics[width=\xx,frame]{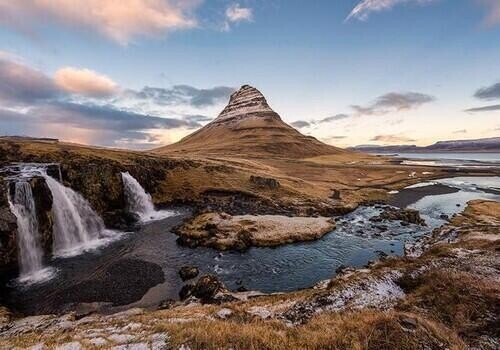} & \includegraphics[width=\xx,frame]{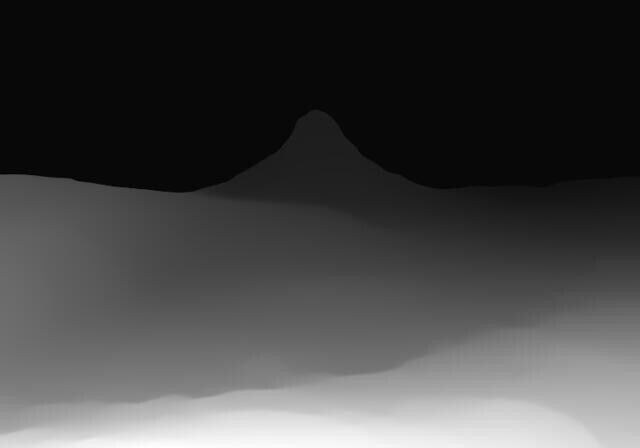} & \includegraphics[width=\xx,frame]{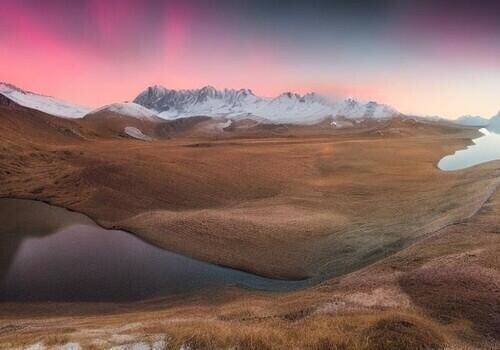} \\
\end{tabularx}

\end{figure}

\begin{figure}

  \centering
  \begin{tabularx}{\columnwidth}{p{1.01\xx}p{1.01\xx}@{\hskip 0.2cm}p{1.01\xx}}

\multicolumn{2}{c}{\textbf{Input Image + Change Map}} & \multicolumn{1}{c}{\textbf{Output Image}} \\[0.2cm] %
\includegraphics[width=\xx,frame]{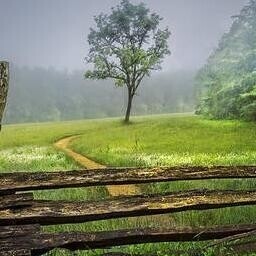} & \includegraphics[width=\xx,frame]{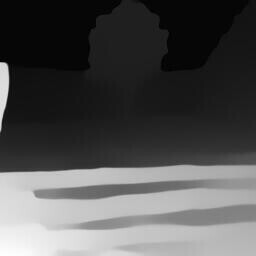} & \includegraphics[width=\xx,frame]{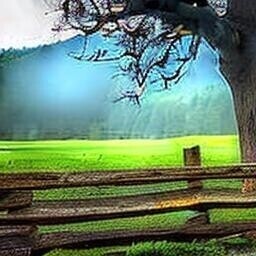} \\[-0.1cm] \multicolumn{3}{X}{\scriptsize{``Debra and Dave Vanderlaan - Huge Cades Cove Misty Tree'''}} \\[0.2cm]

\includegraphics[width=\xx,frame]{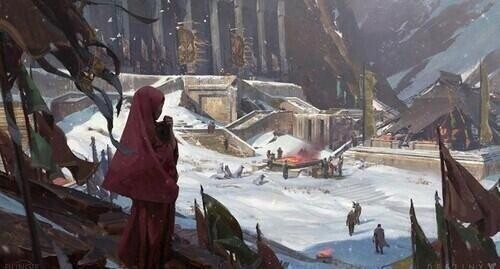} &
\includegraphics[width=\xx,frame]{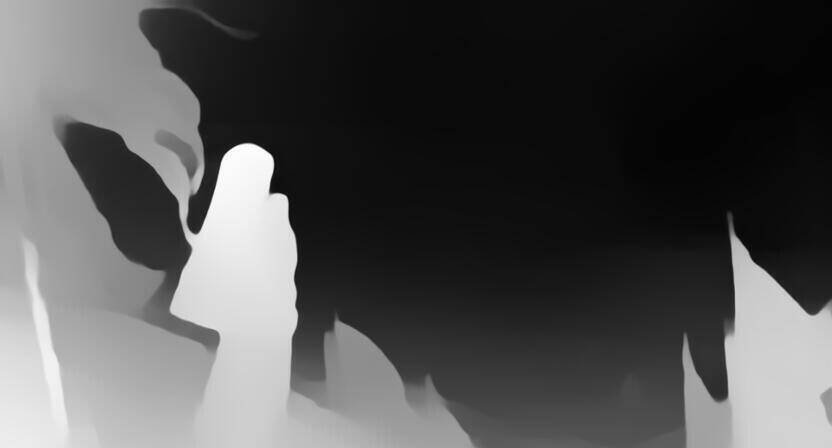} &
\includegraphics[width=\xx,frame]{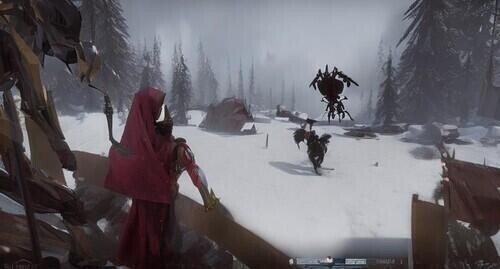}  \\[-0.1cm] \multicolumn{3}{X}{\scriptsize{``Rise of Iron with giant ant (3)''}} \\[0.2cm] %
\includegraphics[width=\xx,frame]{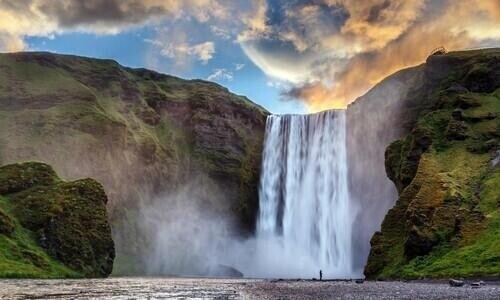} & \includegraphics[width=\xx,frame]{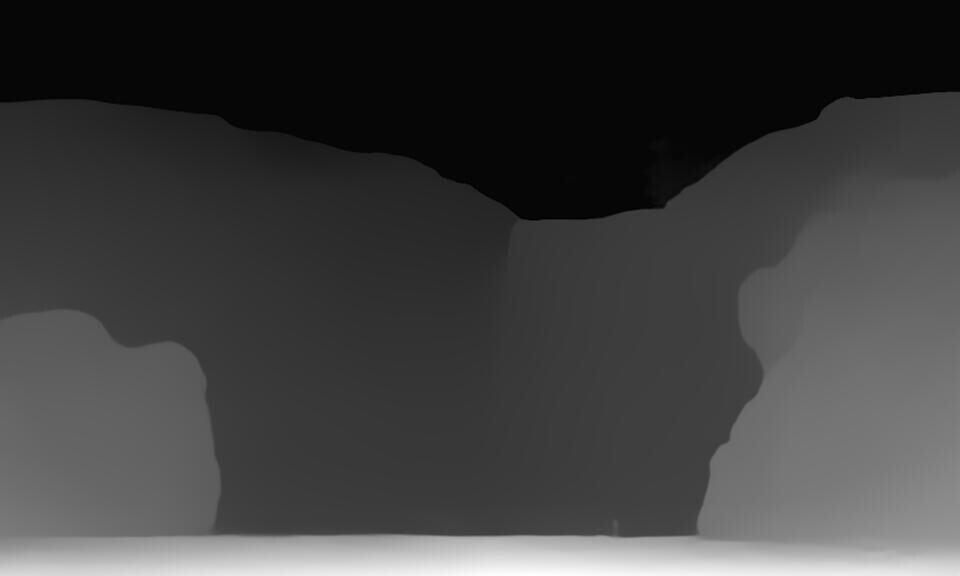} & \includegraphics[width=\xx,frame]{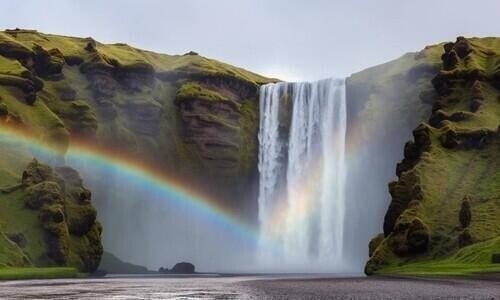} \\[-0.1cm] \multicolumn{3}{X}{\scriptsize{``Skogafoss waterfall and rainbow in the south of Iceland''}} \\[0.2cm]

\includegraphics[width=\xx,frame]{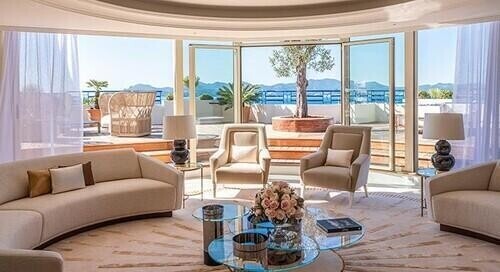} &
\includegraphics[width=\xx,frame]{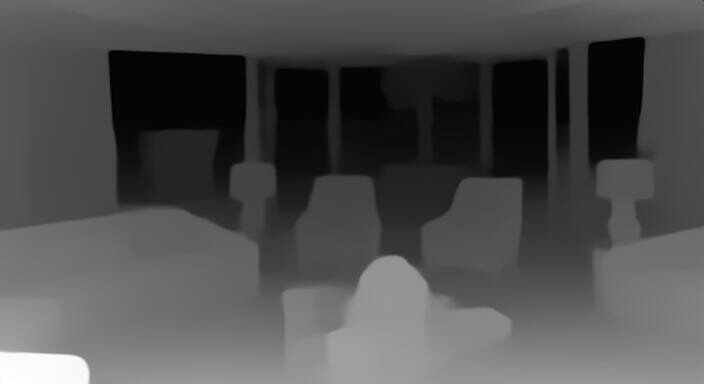} &
\includegraphics[width=\xx,frame]{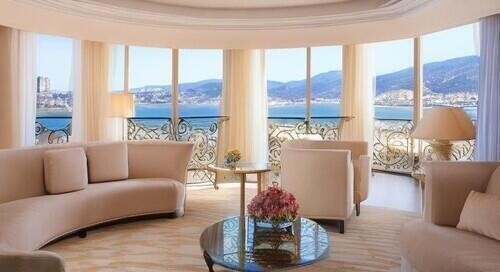}  \\
\end{tabularx}

\end{figure}

\begin{figure}

  \centering
  \begin{tabularx}{\columnwidth}{p{1.01\xx}p{1.01\xx}@{\hskip 0.2cm}p{1.01\xx}}

\multicolumn{2}{c}{\textbf{Input Image + Change Map}} & \multicolumn{1}{c}{\textbf{Output Image}} \\[0.2cm] %

\includegraphics[width=\xx,frame]{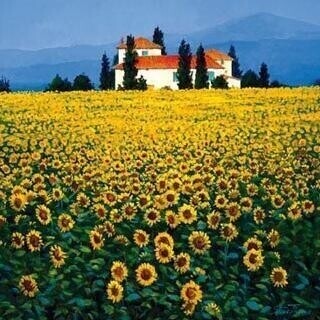} & \includegraphics[width=\xx,frame]{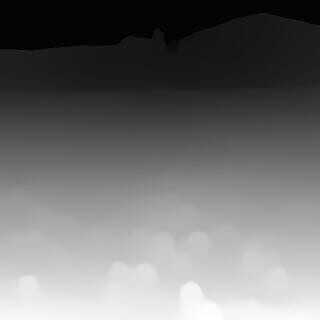} & \includegraphics[width=\xx,frame]{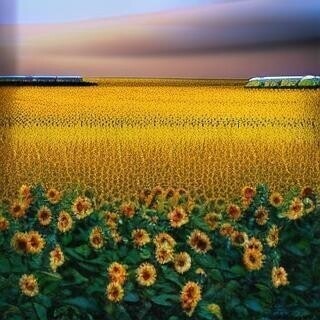} \\[-0.1cm] \multicolumn{3}{X}{\scriptsize{``Sunflowers Field with Snow Cover Art Print''}} \\[0.2cm]

\includegraphics[width=\xx,frame]{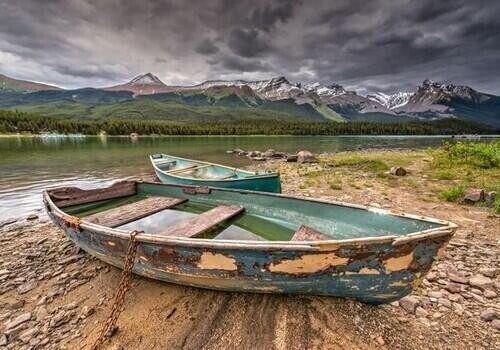} &
\includegraphics[width=\xx,frame]{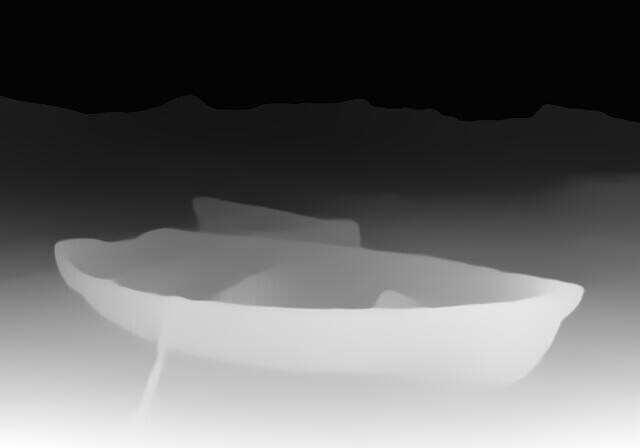} &
\includegraphics[width=\xx,frame]{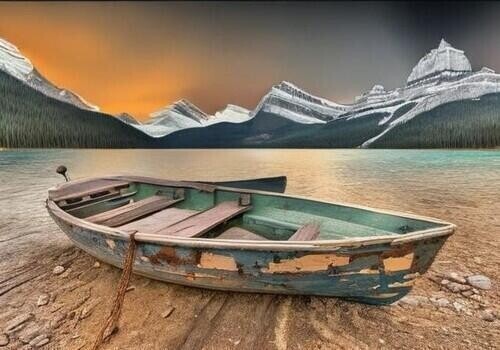}  \\[-0.1cm] \multicolumn{3}{X}{\scriptsize{``Illustration of Maligne Ghosts by Wayne-Stadler-Illustration - Ships And Boats Photo Contest''}} \\[0.5cm] %
\includegraphics[width=\xx,frame]{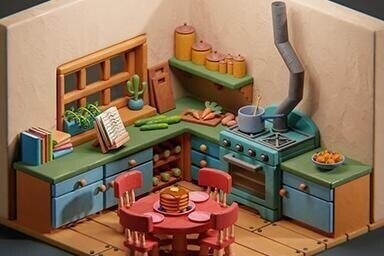} & \includegraphics[width=\xx,frame]{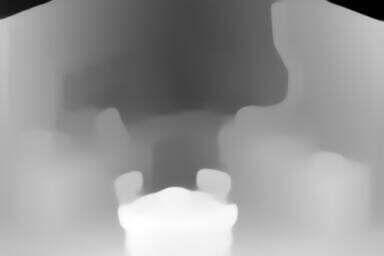} & \includegraphics[width=\xx,frame]{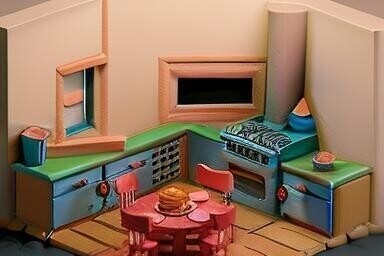} \\[-0.1cm] \multicolumn{3}{X}{\scriptsize{``Clay Kitchen - low poly diorama with huge explosion''}} \\[0.2cm]
\includegraphics[width=\xx,frame]{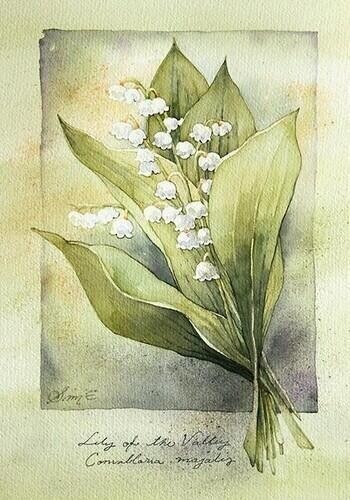} &
\includegraphics[width=\xx,frame]{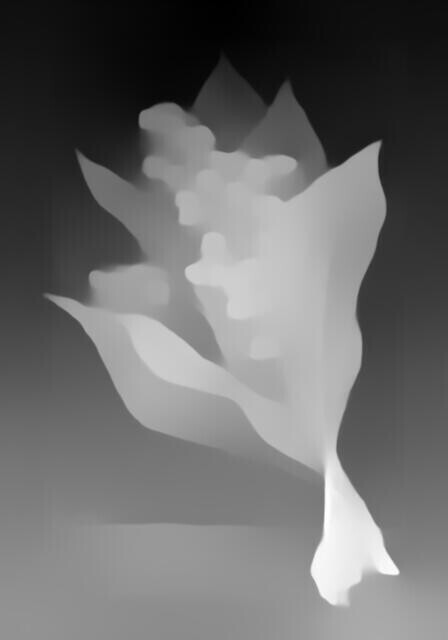} &
\includegraphics[width=\xx,frame]{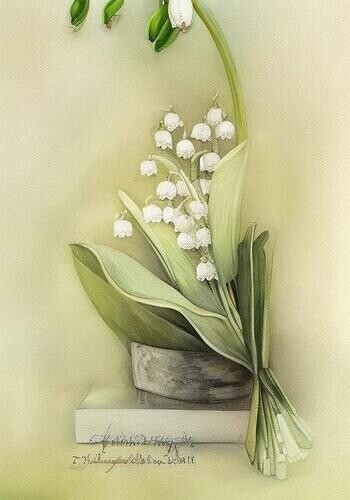}  \\
\end{tabularx}%

\end{figure}

\clearpage
\section{Teaser Settings}
\label{prompt}
In the main paper's teaser, we used various well-known checkpoints of Stable Diffusion. This demonstrates our framework's ability to use community checkpoints.
Most settings have been inspired by publicly published results on Mage~\cite{mage_space}.
\begin{enumerate}
    \item \textbf{Prompt:} ``Tree of life under the sea, ethereal, glittering, lens flares, cinematic lighting, artwork by Anna Dittmann \& Carne Griffiths, 8k, unreal engine 5, hightly detailed, intricate detailed''. 
    \textbf{Negative prompt:} ``bad anatomy, poorly drawn face, out of frame, gibberish, lowres, duplicate, morbid, darkness, maniacal, creepy, fused, blurry background, crosseyed, extra limbs, mutilated, dehydrated, surprised, poor quality, uneven, off-centered, bird illustration, painting, cartoons, sketch, worst quality, low quality, normal quality, lowres, bad anatomy, bad hands, monochrome, grayscale, collapsed eyeshadow, multiple eyeblows, vaginas in breasts, cropped, oversaturated, extra limb, missing limbs, deformed hands, long neck, long body, imperfect, bad hands, signature, watermark, username, artist name, conjoined fingers, deformed fingers, ugly eyes, imperfect eyes, skewed eyes, unnatural face, unnatural body, error, two body, two faces''. 
    \textbf{Checkpoint:} AbsoluteReality.
    \item \textbf{Prompt:} ``yellow-white-blue-purple-golden glowing mountains palace above the clouds, magical reality, high definition, 32K, dynamic lights, cinematic sorrounding, intricate, natural lighting, ray tracing, bloom, extreme hdr, Octane render, unreal engine, 16K hyper realism, character design, hyper detailed, volumetric lighting, hdr, shining, vibrant, photo realism, Canon EOS 7D, Canon EF 70-200mm f/2.8L IS, vibrant colors, beautiful picture quality, breathtaking scene, focused''. 
    \textbf{Negative prompt:} ``Extra limbs, extra fingers, long neck, deformed, More than one nipple per breast, pointy nose, asian, japanese, chinese, lowres, disfigured, ostentatious, ugly, oversaturated, grain, low resolution, disfigured, blurry, bad anatomy, disfigured, poorly drawn face, mutant, mutated, extra limb, ugly, poorly drawn hands, missing limbs, blurred, floating limbs, disjointed limbs, deformed hands, blurred, out of focus, long neck, long body, ugly, disgusting, bad drawing, childish, cut off cropped, distorted, imperfect, surreal, bad hands, text, error, extra digit, fewer digits, cropped , worst quality, low quality, normal quality, jpeg artifacts, signature, watermark, username, blurry, artist name, Lots of hands, extra limbs, extra fingers, conjoined fingers, deformed fingers, old, ugly eyes, imperfect eyes, skewed eyes , unnatural face, stiff face, stiff body, unbalanced body, unnatural body, lacking body, details are not clear, details are sticky, details are low, distorted details, ugly hands, imperfect hands, mutated hands and fingers, long body , mutation, poorly drawn bad hands, fused hand, missing hand, disappearing arms, disappearing thigh, disappearing calf, disappearing legs, ui, missing fingers, text, letters, illustration, painting, cartoons, sketch, worst quality, low quality, normal quality, lowres, bad anatomy, bad hands, monochrome, grayscale, collapsed eyeshadow, multiple eyeblows, vaginas in breasts, cropped, oversaturated, extra limb, missing limbs, deformed hands, long neck, long body, imperfect, bad hands, signature, watermark, username, artist name, conjoined fingers, deformed fingers, ugly eyes, imperfect eyes, skewed eyes, unnatural face, unnatural body, error, two body, two faces, hat, hats, pirate hat, headgear, helmet captains hat''. \textbf{Checkpoint:} DucHaiten GODofSIMP.    
    \item \textbf{Prompt:} ``Fully In frame, 3D, centered, colorful, complete, poofy, single alien, fantasy Surreal Tree, 3d depth outer space nebulae background, 3D Art nouveau curvilinear lines, centered, no circles, no frame, dark atmosphere, curvilinear clouds, matte painting, deep color, fantastical, 3D intricate detail, complementary colors, 8k, cgsociety, artstation, hyperrealistic, cinematic, ultra hd, 4k, 8k, highly detailed cinematic global lighting, octane render, unreal engine 5''. 
    \textbf{Negative prompt:} ``Multiple trees, stacked trees, trees on top of trees, illustrations, split image, 2d, painting, cartoons, sketch, worst quality, low quality, normal quality, low res, monochrome, grayscale, error, bad image, bad photo illustration, 2d, painting, cartoons, sketch, worst quality, low quality, normal quality, monochrome, grayscale, cropped, oversaturated, signature, watermark, username, artist name, error, bad image, bad photo illustration, 2d, painting, cartoons, sketch, worst quality, low quality, normal quality, low res, monochrome, greyscale, signature, watermark, error, bad image, bad photo black and white, monochrome, comic, text, error, cropped, letterbox, jpeg artifacts, signature, watermark, username, artist name, censored, worst quality, low quality, anime, digital illustration, 3d rendering, comic panel, scanlation, multiple views, lowres, ostentatious, ugly, oversaturated, grain, bad drawing, childish, cropped , worst quality, low quality, normal quality, jpeg artifacts, signature, watermark, username, blurry, details are not clear, details are sticky, details are low, distorted details, mutation, poorly drawn 2d, painting, cartoons, sketch, worst quality, low quality, normal quality, low res, monochrome, grayscale, cropped, oversaturated, signature, watermark, username, artist , error, bad image, bad photo deformed, distorted, disfigured, poorly drawn, bad anatomy, wrong anatomy, extra limb, missing limb, floating limbs, mutated hands and fingers, disconnected limbs, mutation, mutated, ugly, disgusting, blurry, amputation illustration, painting, cartoons, sketch, worst quality, low quality, normal quality, lowres, bad anatomy, bad hands, monochrome, grayscale, collapsed eyeshadow, multiple eyeblows, vaginas in breasts, cropped, oversaturated, extra limb, missing limbs, deformed hands, long neck, long body, imperfect, bad hands, signature, watermark, username, artist name, conjoined fingers, deformed fingers, ugly eyes, imperfect eyes, skewed eyes, unnatural face, unnatural body, error, two body, two faces Poorly drawn face, poorly drawn hands, poorly drawn weapons, unnatural pose, blank background, boring background, render, unreal engine''. \textbf{Checkpoint:} ``AbsoluteReality''.
    \item \textbf{Prompt:} ``everything is burning, fire''. 
    \textbf{Negative prompt:} ``bad anatomy, bad proportions, blurry, cloned face, cropped, deformed, dehydrated, disfigured, duplicate, error, fused fingers, worst quality''. 
    \textbf{Checkpoint:} ``Stable-Diffusion v1.5''.
    \item \textbf{Prompt:} ``A highly detailed alien landscape, alien buildings, weird colors, strange plants, xenomorphs, slime, oozing, HDR, 4k, volumetric lights, fantasy art, digital painting, beautiful, colorful, serene, intricate, eldritch, Nvidia ray tracing, Imax, slow shutter speed.''. 
    \textbf{Negative prompt:} ``cartoon, 3d, disfigured,
    bad art, deformed,extra limbs,close up,b\&w, wierd colors, blurry, duplicate, morbid, mutilated, [out of frame], extra fingers, mutated hands, poorly drawn hands, poorly drawn face, mutation, deformed, ugly, blurry, bad anatomy, bad proportions, extra limbs, cloned face, disfigured, out of frame, ugly, extra limbs, bad anatomy, gross proportions, malformed limbs, missing arms, missing legs, extra arms, extra legs, mutated hands, fused fingers, too many fingers, long neck, Photoshop, video game, ugly, tiling, poorly drawn hands, poorly drawn feet, poorly drawn face, out of frame, mutation, mutated, extra limbs, extra legs, extra arms, disfigured, deformed, cross-eye, body out of frame, blurry, bad art, bad anatomy, double face, double face realistic, semi-realistic, cgi, 3d, render, sketch, cartoon, drawing, anime, cropped, worst quality, low quality, jpeg artifacts, ugly, duplicate, morbid, mutilated, out of frame, extra fingers, mutated hands, poorly drawn hands, poorly drawn face, mutation, deformed, blurry, dehydrated, bad anatomy, bad proportions, extra limbs, cloned face, disfigured, gross proportions, malformed limbs, missing arms, missing legs, extra arms, extra legs, fused fingers, too many fingers, long neck''. \textbf{Checkpoint:} Realistic Vision V2.
    \item \textbf{Prompt:} ``no humans,scenery, flower A whimsical illustration of a rainbow connecting two vibrant and cheerful worlds, signifying the connection between people who find humor in life's moments''. 
    \textbf{Negative prompt:} ``blur haze child, loli, paintings, sketches, worst quality, low quality, normal quality, lowres, normal quality, monochrome, grayscale, skin spots, acnes, skin blemishes, age spot, glans, mutated hands, poorly drawn hands, blurry, bad anatomy, extra limbs, lowers, bad hands, missing fingers, extra digit, bad hands, missing fingers worst quality, low quality, sketch, watermark, text copyright signature, cut out, cgi, hands, Two bodies, Two heads, doll, extra nipples, bad anatomy, blurry, fuzzy, extra arms, extra fingers, poorly drawn hands, disfigured, tiling, deformed, mutated, out of frame, cloned face illustration, painting, cartoons, sketch, worst quality, low quality, normal quality, lowres, bad anatomy, bad hands, monochrome, grayscale, collapsed eyeshadow, multiple eyeblows, vaginas in breasts, cropped, oversaturated, extra limb, missing limbs, deformed hands, long neck, long body, imperfect, bad hands, signature, watermark, username, artist name, conjoined fingers, deformed fingers, ugly eyes, imperfect eyes, skewed eyes, unnatural face, unnatural body, error illustration, 3d, 2d, painting, cartoons, sketch, worst quality, low quality, normal quality, lowres, bad anatomy, bad hands, vaginas in breasts, monochrome, grayscale, collapsed eyeshadow, multiple eyeblows, cropped, oversaturated, extra limb, missing limbs, deformed hands, long neck, long body, imperfect, bad hands, signature, watermark, username, artist name, conjoined fingers, deformed fingers, ugly eyes, imperfect eyes, skewed eyes, unnatural face, unnatural body, error, bad image, bad photo worst quality, low quality, normal quality, lowres, bad anatomy, bad hands, vaginas in breasts, monochrome, grayscale, collapsed eyeshadow, multiple eyeblows, cropped, oversaturated, extra limb, missing limbs, deformed hands, long neck, long body, imperfect, bad hands, signature, watermark, username, artist name, conjoined fingers, deformed fingers, ugly eyes, imperfect eyes, skewed eyes, unnatural face, unnatural body, error, painting by bad-artist sketch, worst quality, low quality, normal quality, lowres, bad anatomy, bad hands, monochrome, grayscale, collapsed eyeshadow, multiple eyeblows, vaginas in breasts, cropped, oversaturated, extra limb, missing limbs, deformed hands, long neck, long body, imperfect, bad hands, signature, watermark, username, artist name, conjoined fingers, deformed fingers, ugly eyes, imperfect eyes, skewed eyes, unnatural face, unnatural body, error''. 
    \textbf{Checkpoint:} DucHaiten StyleLikeMe.
    
\end{enumerate}

\end{document}